\documentclass[10pt,twocolumn,letterpaper]{article}

\usepackage[accsupp]{axessibility}  %
\usepackage[pagenumbers]{arxivs} %

\usepackage{booktabs} %
\usepackage{amsmath}
\usepackage{graphicx}
\usepackage{multirow}
\usepackage{booktabs}
\usepackage{array}
\usepackage{subcaption}
\usepackage{tabularx}
\usepackage{xcolor}
\usepackage{colortbl} %
\usepackage{graphics}
\usepackage{xcolor} %

\definecolor{arxivsblue}{rgb}{0.21,0.49,0.74}
\usepackage[pagebackref,breaklinks,colorlinks,allcolors=arxivsblue]{hyperref}

\title{Im2Haircut: Single-view Strand-based Hair Reconstruction for Human Avatars}

\author{
Vanessa Sklyarova$^{1,2}$ \and 
Egor Zakharov$^{2}$ \and
Malte Prinzler$^{1,2}$ \and 
Giorgio Becherini$^{1}$ \and 
Michael J. Black$^{1}$ \and
Justus Thies$^{1,3}$ \and 
\vspace{0.1cm}\\
$^1$Max Planck Institute for Intelligent Systems \ \ $^2$ETH Zürich \ \ $^3$Technical University of Darmstadt
}

\begin{document}

\twocolumn[{%
\renewcommand\twocolumn[1][]{#1}%
\maketitle
\begin{center}
    \centering
    
    \captionsetup{type=figure}
    \vspace{-0.3cm}
    \includegraphics[width=0.99\textwidth]{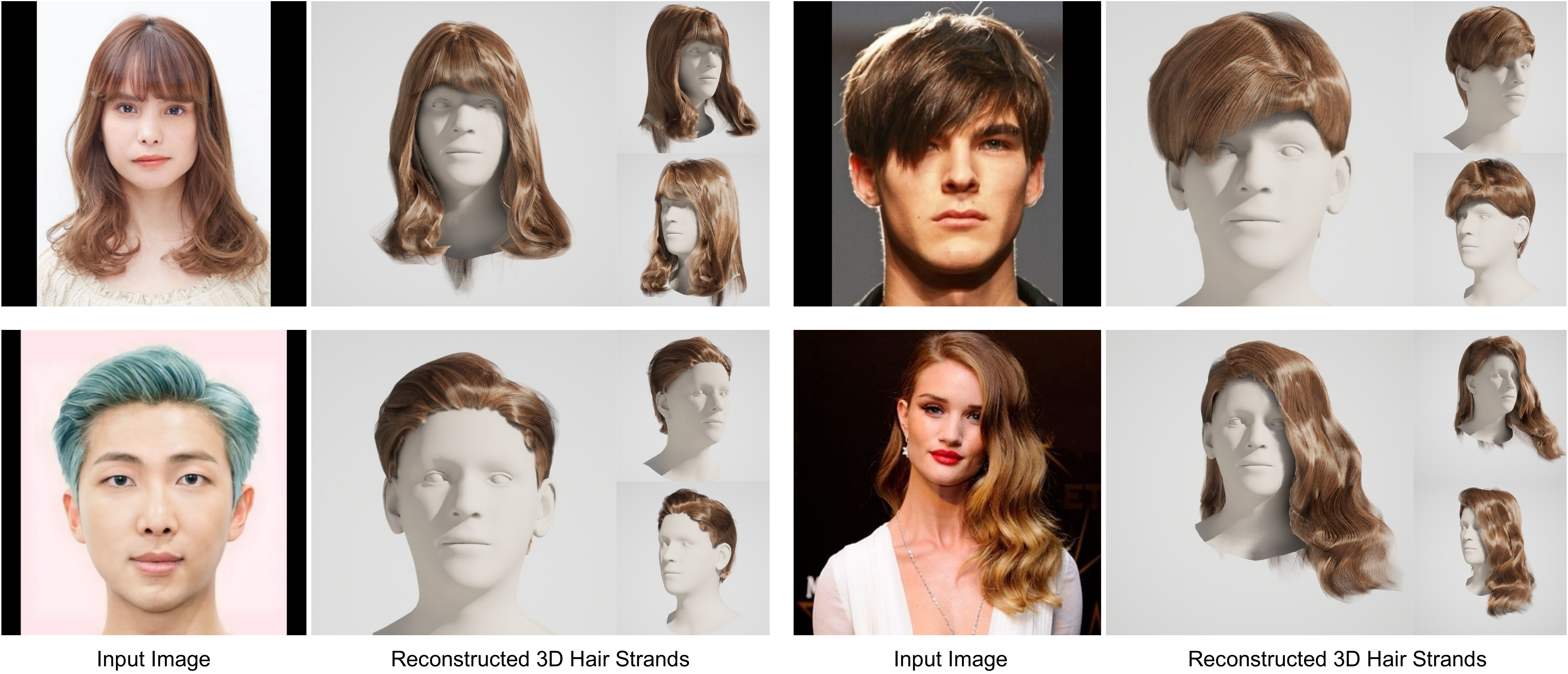}
    \vspace{-0.2cm} 
    \caption{Given a single image, our method, Im2Haircut, generates high-quality, strand-based 3D hair geometry. Im2Haircut consists of a prior hair geometry model trained on a mixture of synthetic and real data that is finetuned using the input image at inference time.}
    \label{fig:teaser}
\end{center}%
}]

\begin{abstract}

We present a novel approach for 3D hair reconstruction from \textit{single} photographs based on a global hair prior combined with local optimization.
Capturing strand-based hair geometry from single photographs is challenging due to the variety and geometric complexity of hairstyles and the lack of ground truth training data.
Classical reconstruction methods like multi-view stereo only reconstruct the visible hair strands, missing the inner structure of hairstyles and hampering realistic hair simulation.
To address this, existing methods leverage hairstyle priors trained on synthetic data.
Such data, however, is limited in both quantity and quality since it requires 
manual work from skilled artists to model the 3D hairstyles and create near-photorealistic renderings.
To address this, we propose a novel approach that uses both, real and synthetic data to learn an effective hairstyle prior.
Specifically, we train a transformer-based prior model on synthetic data to obtain knowledge of the internal hairstyle geometry and introduce real data in the learning process to model the outer structure.
This training scheme is able to model the visible hair strands depicted in an input image, while preserving the general 3D structure of hairstyles.
We exploit this prior to create a Gaussian-splatting-based reconstruction method that creates hairstyles from one or more images.
Qualitative and quantitative comparisons with existing reconstruction pipelines demonstrate the effectiveness and superior performance of our method for capturing detailed hair orientation, overall silhouette, and backside consistency. For additional results and code, please refer to \url{https://im2haircut.is.tue.mpg.de}.
\end{abstract}

\section{Introduction}

Realistic digital representation of a person's appearance, including their hair, is essential for creating highly personalized and realistic 3D head avatars.
However, capturing a digital representation of hair is challenging, as hair strands are very thin structures, with complex geometry.
Varying hair densities, orientations, and textures make accurate reconstruction both computationally intensive and algorithmically complex.
In addition, photographs only capture the visible hair, while the inner structure is occluded.

Numerous approaches have been developed to reconstruct strand-based hair from monocular video or multi-view data  \cite{luo2024gaussianhair, GaussianHaircut, neuralhaircut, neuralstrands, deepmvshair, monohair} using learned hairstyle priors to recover the inner structure.
These methods require extensive processing times, often taking several hours~\cite{GroomCap, GaussianHaircut, deepmvshair, monohair, drhair, luo2024gaussianhair} or even days~\cite{neuralhaircut, neuralstrands}.
Such techniques are impractical for user-friendly applications and are very expensive to deploy.
Reconstruction from a single photograph is much simpler and more user friendly than requiring video.
However, this one-shot reconstruction process, e.g.~from a frontal photo, poses considerable challenges, particularly in modeling the internal geometry and unseen backside geometry.

To address these challenges, strong hair priors need to be learned.
Existing work relies on training models using synthetic datasets and subsequently applying them to real images \cite{hairmony, hairnet, neuralhd, hairstep, perm}.
However, a significant obstacle in these approaches is the domain gap between synthetic and real-world data due to difficulties in rendering hair with realistic textures, as most existing datasets feature coarse hairstyles with relatively few strands.
In contrast to these synthetic hair models, real human hair often comprises over 100,000 strands, each contributing to the overall appearance and realism of the hairstyle.
This domain gap can undermine the model's performance and generalization ability.
To reduce this domain gap, some works~\cite{hairnet, neuralhd, hairstep} employ orientation maps~\cite{Paris2004CaptureOH, Paris2008HairPG} as input signals instead of relying on RGB images.
HairStep~\cite{hairstep} expands on this approach by incorporating depth maps and direction maps, which provide richer geometric and structural information.
These advancements represent promising steps toward achieving accurate and efficient hair modeling, but significant challenges remain in bridging the gap between synthetic training and real-world applications (e.g.~predicting direction maps). 

In this paper, we learn a hair prior using a novel mixture of synthetic and real data to decrease the domain gap and improve reconstruction realism.
For the synthetic data, we assume that we have full control over the training image generation process, including access to the 3D structure of the hairstyles, which can be used as a training objective.
For real ``in-the-wild'' data, this supervision is not available.
However, following an approach proposed in~\cite{GaussianHaircut}, we can learn from the visible structures in the images using differentiable rendering based on 3D Gaussian Splatting~\cite{Kerbl20233DGS}.
After training on this real and synthetic data mixture, our model can predict global hairstyles from single photographs.
To improve the personalization and matching of the hairstyle, we propose an additional optimization step, where local detail is recovered using the same 3D Gaussian-Splatting-based differentiable rendering.

We demonstrate the capabilities of our proposed method in comparison to state-of-the-art single-view reconstruction methods.
From the qualitative and quantitative results we can conclude that our technique improves the hair strand details like orientation, the overall silhouette, and back-side consistency of the hairstyle.

In summary, our contributions are: (1)
A novel training strategy to learn a hairstyle prior that mixes synthetic and real data, improving reconstruction quality and narrowing the domain gap between the training data and real test data.
(2) An inversion approach that reconstructs hairstyles from single or multi-view photographs in the learned hairstyle prior using Gaussian splatting.

\section{Related Work}

\begin{figure*}
    \centering
    \includegraphics[width=\textwidth]{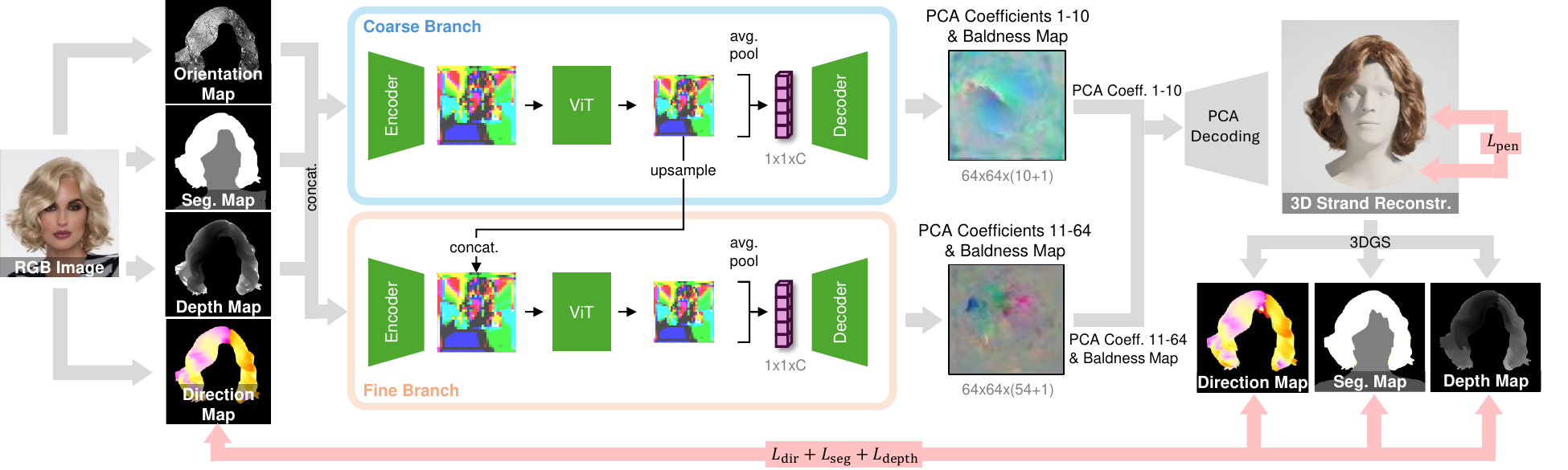}
    \caption{Our inference pipeline consists of two stages: coarse and fine. We first train to predict the first 10 components of the PCA hair representation and then the remaining 54 components for more detail. To mitigate the domain gap, we train on synthetic data using 3D reconstruction losses and then mix in real data. As we do not have ground-truth geometry for real images, we use Gaussian splatting to render from the desired view and supervise with 2D rendering losses.}
    \label{fig:optim}
    \vspace{-0.3cm}
\end{figure*}

\subsection{Strand-based Hair Modeling}

The de-facto hair representation standard which is supported by the majority of computer graphics and simulation pipelines is strand-based hair modeling~\cite{Bergou2008DiscreteER, Marschner2003LightSF, Chiang2015APA, Daviet2023InteractiveHS}.
To create such assets for computer games and VFX requires skilled 3D artists to model the hair strands.
To automatically recover strand-based 3D hairstyles from real subjects, several methods have been proposed that use single photographs~\cite{hairnet, neuralhd, perm, hairstep, Hairgan}, multi-view / video~\cite{neuralstrands, neuralhaircut, GaussianHaircut, drhair, Luo2012MultiviewHC, monohair, Zhang2017ADA, GroomCap, Chai2016AutoHair, Chai2012SingleviewHM, hairsalon, Zhang2018ModelingHF, strandaccurate}, or even CT data~\cite{ct2hair}.
It is worth noting that a key challenge in reconstructing hair from optical data is the unknown inner structure of the hairstyle which is mostly occluded. While CT data addresses this, it is not feasible for consumer  applications.

\subsection{Image-based 3D Hair Reconstruction.}
Most existing approaches for strand-based hair modeling conduct 3D hair volume reconstruction that is followed by either autoregressive decoding of the explicit hair strands, the so-called \emph{hair growing}~\cite{deepmvshair, monohair, GroomCap, hairstep, Hairgan}, or direct reconstruction of the hairstyle in the form of a strand-based hair texture, or \emph{hair map}~\cite{difflocks2025, neuralstrands, neuralhaircut, GaussianHaircut, luo2024gaussianhair, drhair, hairnet, perm, long2025tangledgenerating3dhair}.

Modern hair-growing-based approaches~\cite{deepmvshair, monohair, GroomCap, hairstep} require a complete volumetric reconstruction of the hairstyle that includes not only densities but also 3D-lifted direction maps modeled throughout an entire hair volume.
Hair strands are then extracted as the trajectory of the heuristic-based approaches, which traverse the learned orientation field.
The main limitation of this approach is that it lacks a prior at the level of the strand-based hairstyle, which may result in hairstyles with unrealistic internal structures that are hard to physically simulate.

Alternatively, methods that utilize a hair-map-based representation~\cite{difflocks2025, neuralstrands, neuralhaircut, GaussianHaircut, luo2024gaussianhair, drhair, perm} rely only on the surface-level geometry of hair to reconstruct the underlying hair strands.
Since these structures are visible and, thus, require no prior knowledge, the hair prior is applied directly at the strand level, either per strand~\cite{neuralstrands} or jointly over the entire hairstyle~\cite{difflocks2025, neuralhaircut, GaussianHaircut, luo2024gaussianhair}.
Our approach adopts the strategy of hair-map-based modeling for a single image setting that omits any volumetric hair reconstruction, similar to HairNet~\cite{hairnet}, which allows us to use real-world data to train our method via differentiable hair rendering.
Our single image fine-tuning strategy also favorably compares to retrieval-based methods such as Hairmony~\cite{hairmony}, since we are able to achieve precise pixel-aligned reconstructions.

\section{Background}

\paragraph{Strand-based Hair Modeling.}
Strand-based hairstyles are defined by a hair map $\mathcal{H}$, which is a texture map associated with the scalp region of the head mesh.
Each texel $i = 1 \dots H\cdot W$ of this map contains a strand $S_i$ that originates at the corresponding point on the scalp mesh.
These strands are modeled as polylines with $L$ line segments: $S_i = \{ p_{ij} \}$, where $j = 1 \dots L$ is the index of the point on the strand, which makes $\mathcal{H}$ have dimensions $H \times W \times 3L$.
To reduce the dimensionality, %
we use PCA-based compression in the frequency domain of the hair strands~\cite{perm}:
\begin{equation}
    S_i = \text{iDFT} \big( \bar{\mathcal{S}} + \gamma_i^T X \big),
\end{equation}
where $\gamma_i$ are the PCA coefficients which are used as the parameterization of a hair strand.
$X \in \mathbb{R}^{|\gamma_i| \times 6k}$ are the principal components, where $k = \lfloor L/2 \rfloor + 1$ is a number of frequency bands, $\bar{\mathcal{S}} = \{ \bar{\mathcal{S}}_\text{real}, \bar{\mathcal{S}}_\text{imag} \} \in \mathbb{R}^{k \times 3 \times 2}$ is a mean phase vector of strands in the frequency domain, and $\text{iDFT}(\cdot)$ denotes the inverse Discrete Fourier Transform.
These principal components are estimated using available strand-based hair geometries from synthetic datasets.

In the following, we denote a hair map composed of these coefficients as $Z$, with dimensions of $H\times W\times |\gamma|$, and refer to it as a PCA hair map. Similar to \cite{perm, groomgen} we use $\gamma=64$.
Following~\cite{perm}, we split $Z$ into coarse $Z_{c}$ and fine $Z_{f}$ components, where $Z_c$ contains the coefficients for the first ten principal components and $Z_{f}$ the rest.

\smallskip
\noindent\textbf{3D Gaussian Splatting for Hair Reconstruction.}
Recent work~\cite{GaussianHaircut,luo2024gaussianhair,GroomCap} uses 3D Gaussian Splatting~\cite{kerbl2023gaussian} (3DGS) for hair reconstruction from multi-view data.  
The general idea of these approaches is to use 3DGS to achieve differentiable rendering of the hair strands.
This is done by constraining the degrees of freedom of the rasterized 3D Gaussians to make them lie on the line segments of the strand polylines. 
Specifically, the mean of each Gaussian is defined at the midpoints of hair segments, while the direction of the highest variance corresponds to 3D orientation. 
Such reparameterization allows us to use a 3DGS rasterization method to match strand-based hair renders against given silhouette and orientation maps (for more details, please refer to the supplementary).

\section{Method}

Our proposed method for one-shot hair reconstruction has two major components: (i) a global hairstyle prior, and (ii) an optimization-based local hair strand reconstruction scheme.
The hairstyle prior is learned using a hybrid training scheme with full 3D supervision from synthetic data and self-supervised losses on real data using Gaussian splatting~\cite{kerbl2023gaussian}.
The training on synthetic data enables the inner 3D structure of hairstyles to be learned, while the self-supervised training on real data improves realistic modeling of the outer layer of visible hair.
During test time, we leverage our global hairstyle prior, which is conditioned on the input image, to regress an initial geometry state of the hairstyle.
This initial state is then further refined based on rendering losses to match the intricate details of the hair strands in the input image.

\subsection{Hairstyle Prior}

As detailed above, we learn a hairstyle prior using synthetic and real data.
Specifically, we leverage the PERM~\cite{perm} dataset that includes 21,054 synthetic hairstyles with around 10,000 strands each, and the Hairstep~\cite{hairstep} dataset which contains 1,250 ``in-the-wild'' photographs without any available 3D hair geometry, but with manually annotated directional maps (see Sec.~\ref{sec:data}).

Our hairstyle prior model is conditioned on a single input image which is mapped by a vision transformer to predict a PCA-based hair map $Z$ and a baldness mask $M$.
These maps are defined on the UV space of the FLAME parametric head model \cite{FLAME:SiggraphAsia2017}, covering the skull region.
Leveraging the properties of PCA, we employ a coarse-to-fine architectural scheme, see Fig.~\ref{fig:optim}.
Specifically, we employ a coarse branch that is trained to produce the basic structure of the hairstyle by only predicting the first 10 PCA coefficients $Z_c$, and the baldness map $M$.
In a second branch, another 54 PCA coefficients $Z_f$ are predicted, which contain the higher frequencies of the hair strands, as well as a refined version of the baldness map $M$.
The outputs of both branches are concatenated to produce the final output.

This architecture is trained in three stages:
(1) training the coarse branch considering the synthetic dataset with 3D supervision on the output, (2) training the fine branch with the same losses, and (3) joint training with additional real data using self-supervised rendering losses.%

\medskip \noindent
\textbf{(1) Coarse branch.}
The \textit{coarse branch} of our architecture is trained to retrieve the basic structure of the hairstyle, i.e. the PCA hair map of size $64\times64\times11$ containing the first 10 PCA components of the hair strands $Z_c$ and the coarse version of the baldness map $M_c$ of size $H\times W$, that defines whether there is a hair root or not.
The input to this branch is a feature map of shape $H\times W \times 4$, which is a concatenation of orientation maps $O_i\in\mathbb{R}^{H\times W\times1}$, segmentation maps $S_i\in\mathbb{R}^{H\times W\times2}$, and depth maps $D_i\in\mathbb{R}^{H\times W\times1}$, all of which are extracted from the input image with off-the-shelf models. 
These inputs are processed with a convolutional encoder, followed by a vision transformer (ViT)~\cite{vit}. 
Similar to \cite{li2022mat}, in the transformer's attention layers, we discard tokens from patches that lie outside of the hair region.  
The output of the ViT is converted to the output feature map using a convolutional decoder, where we concatenate learnable positional encodings to the activation maps in each layer.

To train this coarse branch, we employ the following losses based on synthetic data:
\begin{equation}
    \mathcal{L}_\text{coarse} = \mathcal{L}_\text{MAE} + \lambda_\text{PCA} \big\| Z_c - \hat{Z}_c \big\|_2 + \lambda_\text{mask} \big\| M_c - \hat{M} \big\|_1.
\end{equation}
We denote a ground truth PCA-based hair map as $\hat{Z}_c$, its corresponding strand-based map as $\hat{H}_c = \{ \hat{S}_i \} = \{ \hat{p}_{ij} \}$, and a strand-based map decoded from the network's prediction $Z_c$ as $H_c = \{ S_i \} = \{ p_{ij} \} $.
If we also denote point-wise directions as $v_{ij}$, and their normalized versions as $b_{ij}$, then the strand's scalar curvature can be written as:
\begin{equation}
    g_{ij} = \big\| b_{ij}\, \times\, b_{i,j+1} \big\|_2.    
\end{equation}
Our reconstruction loss $\mathcal{L}_\text{MAE}$ is inspired by  %
\cite{neuralhaircut} and is a sum of individual point-wise losses,
$ \mathcal{L}_{ij}^\text{point} =$
\begin{equation}
    \big\| p_{ij} - \hat{p}_{ij} \big\|_2 + \lambda_\text{dir} \big\| v_{ij} - \hat{v}_{ij} \big\|_1 + \lambda_\text{curv} \big\| g_{ij} - \hat{g}_{ij} \big\|_2, \\
\end{equation}
\begin{equation}
    \mathcal{L}_\text{MAE} = \frac{1}{HWL} \sum_{ij} \mathcal{L}_{ij}^\text{point}.
\end{equation}
Moreover, the latent hair maps $Z_c$ and baldness maps $M_c$ are supervised with the ground-truth $\hat{Z}_c$ and $\hat{M}$.

\medskip \noindent
\textbf{(2) Fine branch.}
In the \textit{fine branch}, we predict a feature map of shape $64\times{64}\times{(54+1)}$ containing the remaining PCA components $Z_f$ to bring more high-frequency details into the hairstyle, as well as an updated baldness map $M_f$.
We use a similar architecture to the coarse stage and use its weights to initialize the encoder and decoder. 
However, the resolution of the feature maps on which the fine branch transformer operates is two times higher than the coarse branch.
Thus, to generate the fine branch transformer's input, we first upsample the final features of the coarse branch and concatenate them channel-wise with the features extracted with the fine branch encoder.
For training, we employ a loss similar to $\mathcal{L}_\text{coarse}$.
However, we modify the coarse point loss $\mathcal{L}_\text{point}$ to focus it on matching the hair regions visible on the input view.
This follows HairNet~\cite{hairnet}, where it was shown to improve reconstruction details.
To do that, we define per-view visibility weights $w_{ij}$ for each point in the synthetic hairstyle and use a weighted version of the loss $\mathcal{L}_\text{MAE}$:
\begin{equation}
    \mathcal{L}'_\text{MAE} = \frac{\sum_{ij}  w_{ij} \mathcal{L}_{ij}^\text{point}}{\sum_{ij} w_{ij}} ,
\end{equation}
where $w_{ij} =3$ for visible points and $1$ otherwise.
To supervise the output of the fine branch, we use:
\begin{equation}
    \mathcal{L}_\text{fine} = \mathcal{L}'_\text{MAE} + \lambda_\text{PCA} \big\| Z_f - \hat{Z}_f \big\|_2 + \lambda_\text{mask} \big\| M_f - \hat{M} \big\|_1.
\end{equation}
After initializing the fine branch using only $\mathcal{L}_\text{fine}$, we further optimize both branches together with synthetic data:
\begin{equation}
    \mathcal{L}_\text{synth} = \mathcal{L}_\text{coarse} + \mathcal{L}_\text{fine}.
\end{equation}
Note that we use a fine baldness map $M_f$ as the final output $M$, and the PCA map $Z$ is a concatenation of $Z_c$ and $Z_f$.

\medskip \noindent
\textbf{(3) Hybrid training on mixed data.}
Using the model, which is trained on purely synthetic data, we introduce additional self-supervised losses based on rendering to handle real training images where no 3D data is available.
Specifically, during this training step, we mix synthetic and real training data.
For the synthetic data, we keep the geometry-based losses defined above, while real data is used by comparing renderings of the predicted hairstyle with the ground truth images.
For differentiable rasterization, we use a hair rendering pipeline based on 3D Gaussian Splatting~\cite{GaussianHaircut}.
We utilize a similar set of losses as \cite{GaussianHaircut} with the main difference that we use directed orientation maps and depth instead of Gabor orientation maps and RGB loss.
The rendering-based losses include a segmentation loss, $\mathcal{L}_\text{seg}$, that uses an L1 loss to match the rendered silhouette $s$ with the ground-truth hair segmentation $\hat{s}$, and a direction loss, $\mathcal{L}_\text{dirmap}$:
\begin{equation*}
    \mathcal{L}_\text{seg} = 
    \frac{1}{M}  \sum_p\big| s_p - \hat{s}_p \big|,\quad \mathcal{L}_\text{dirmap} = \frac{1}{M} \sum_p \big\| b_p - \hat{b}_p \big\|_1,
\end{equation*}
where $p$ denotes a pixel index, $M$ is the number of pixels, $b_p \in \mathbb{R}^2$ is the rendered hair growth direction, and $\hat{b}_p$ is the ground-truth hair growth direction that was manually annotated for our training dataset.
Despite not having 3D ground truth for the real data, we can introduce an objective that penalizes hair penetration with the head mesh:
\begin{equation}
    \mathcal{L}_\text{pen} = \frac{1}{HWL}\sum_{ij} \text{ReLU}\big( -\text{sdf}(p_{ij}) \big),
\end{equation}
where $\text{sdf}$ denotes a signed distance function for the template head mesh.
The loss for the real data is:
\begin{equation}
    \mathcal{L}_\text{real} = \lambda_{seg}\mathcal{L}_\text{seg} + \lambda_{dirmap} \mathcal{L}_\text{dirmap} + \lambda_{pen}\mathcal{L}_{pen}.
\end{equation}
We note that there is a tradeoff between alignment with the target view and realism of the hairstyle, especially in non-visible regions.
Therefore, we adopt a hybrid training strategy that trains the network on both synthetic and real data, which is combined in a mini-batch with a mixing rate $r$: 
\begin{equation}
    \mathcal{L}_{hybrid} = \mathcal{L}_{synth} + r\mathcal{L}_{real}.
\end{equation}
To stabilize the training, we propagate the gradient from the real data only into the coarse branch of our architecture, while synthetic data influences both branches (i.e., all 64 components).
We note that the signal stemming from the in-the-wild rendering-based objective is too noisy to train the fine branch of our network and results in the appearance of unrealistic hair details.
However, such fine-tuning allows us to achieve a substantially higher diversity and realism in the reconstructed hairstyles.

We use AdamW~\cite{adamw} for optimization with learning rate 0.0001, batch size 32, and train the whole method for 139 hours on 4 GPUs A100, see supplementary for details.

\subsection{Hairstyle Reconstruction} 
To achieve detailed and aligned hairstyles from single photographs, we perform a single-view fine-tuning of our pre-trained model for 400 steps.
We fine-tune all parts of the model using a coarse-to-fine strategy: first optimizing the initial ten components for 20 steps, followed by training all components together for the remaining steps.
For optimization, we use the $L_{real}$ loss from the hybrid training stage and further regularize the reconstruction process by computing a depth loss between the rasterized Gaussians and depth estimations from Depth Pro~\cite{depth-pro}.
We follow HAAR~\cite{haar} and increase the number of hair strands through interpolation during the optimization. 
The whole reconstruction pipeline takes around 10 minutes on an  A100.

\subsection{Dataset}
\label{sec:data}
We render each hairstyle from the PERM~\cite{perm} dataset from 8 different views using Blender~\cite{blender}, resulting in 168,032 RGB images with silhouette masks for the train and 400 images for the test set.
Similar to \cite{neuralhd, hairnet}, we calculate orientation maps using Gabor filters and depth maps using the Depth Pro~\cite{depth-pro} model.
Thus, our model's input consists of an orientation map, a hair segmentation map, a body segmentation mask, and a depth map. %
For training, we project each synthetic hairstyle onto a regular hair texture grid of the size $64~\times{64}$ using the nearest interpolation, which also contains a baldness map.
During inference, we additionally calculate segmentation maps using Segment-Anything~\cite{SAM}, direction maps using Hairstep~\cite{hairstep}, and cameras using Deep3DFaceRecon~\cite{Deep3dFaceRecon}.

\begin{figure*}
    \setlength{\tabcolsep}{0pt} %
    \renewcommand{\arraystretch}{0.0} %
    \begin{tabular}{@{}ccccccccc}
        \multirow{2}{*}[0.481in]{\includegraphics[width=0.14\textwidth]{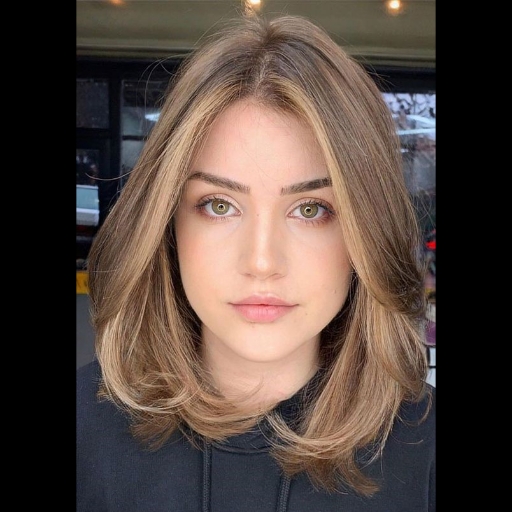}} 
            &
        \multirow{2}{*}[0.481in]{\includegraphics[trim={125 200 125 50},clip,width=0.14\textwidth]{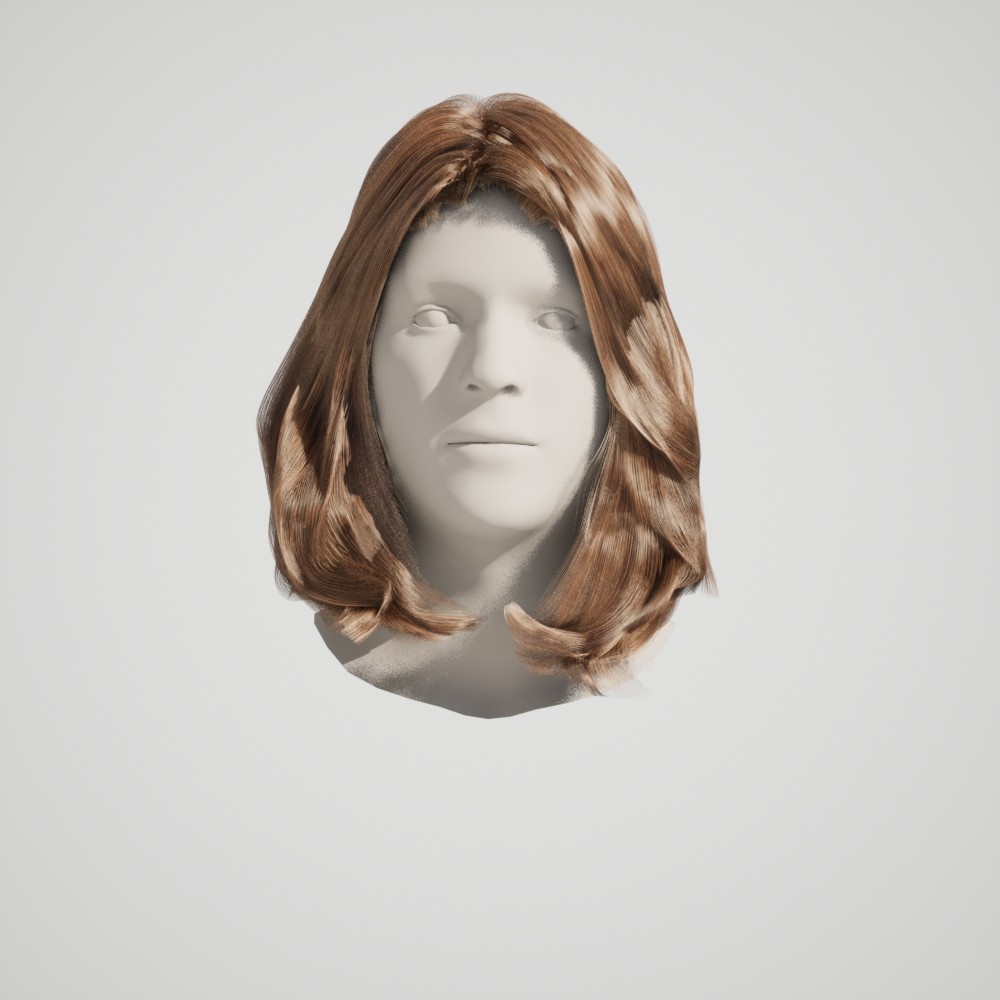}} 
        &
        \includegraphics[width=0.07\textwidth]{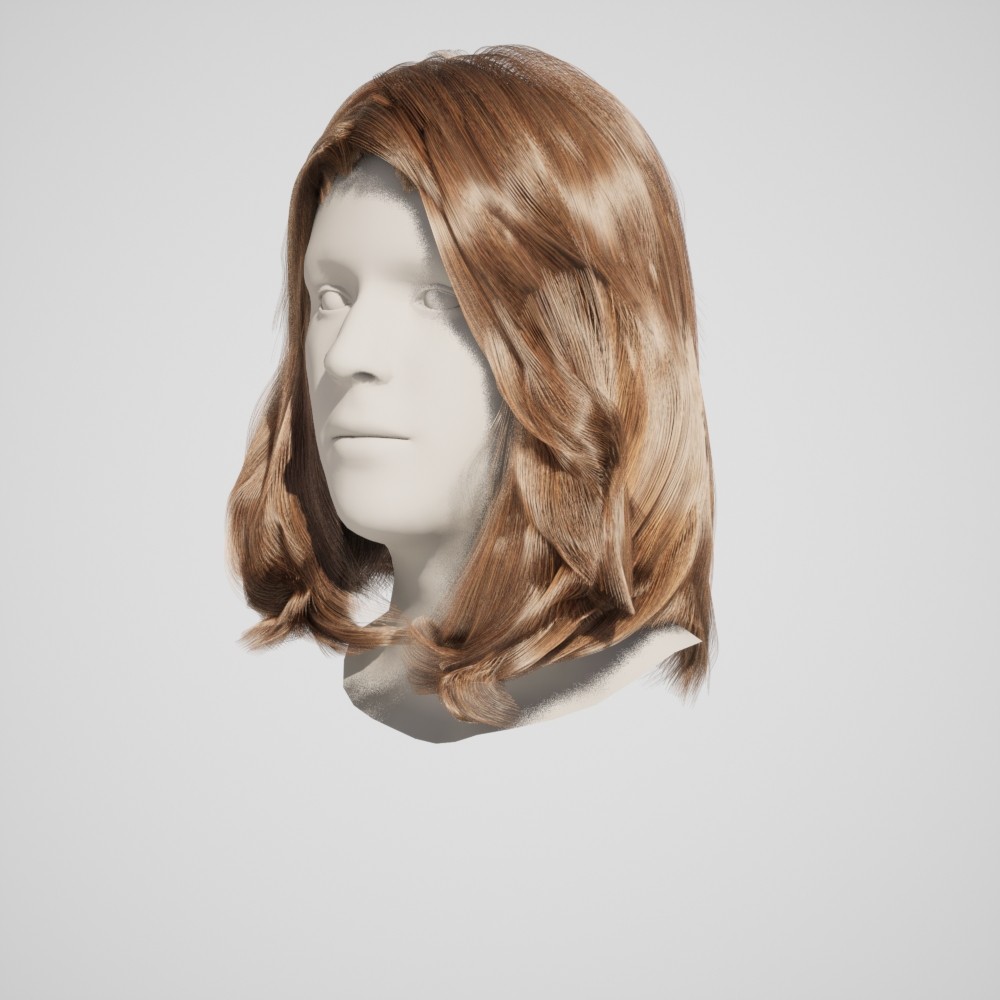} 
        &
        \multirow{2}{*}[0.481in]{\includegraphics[trim={125 200 125 50},clip,width=0.14\textwidth]{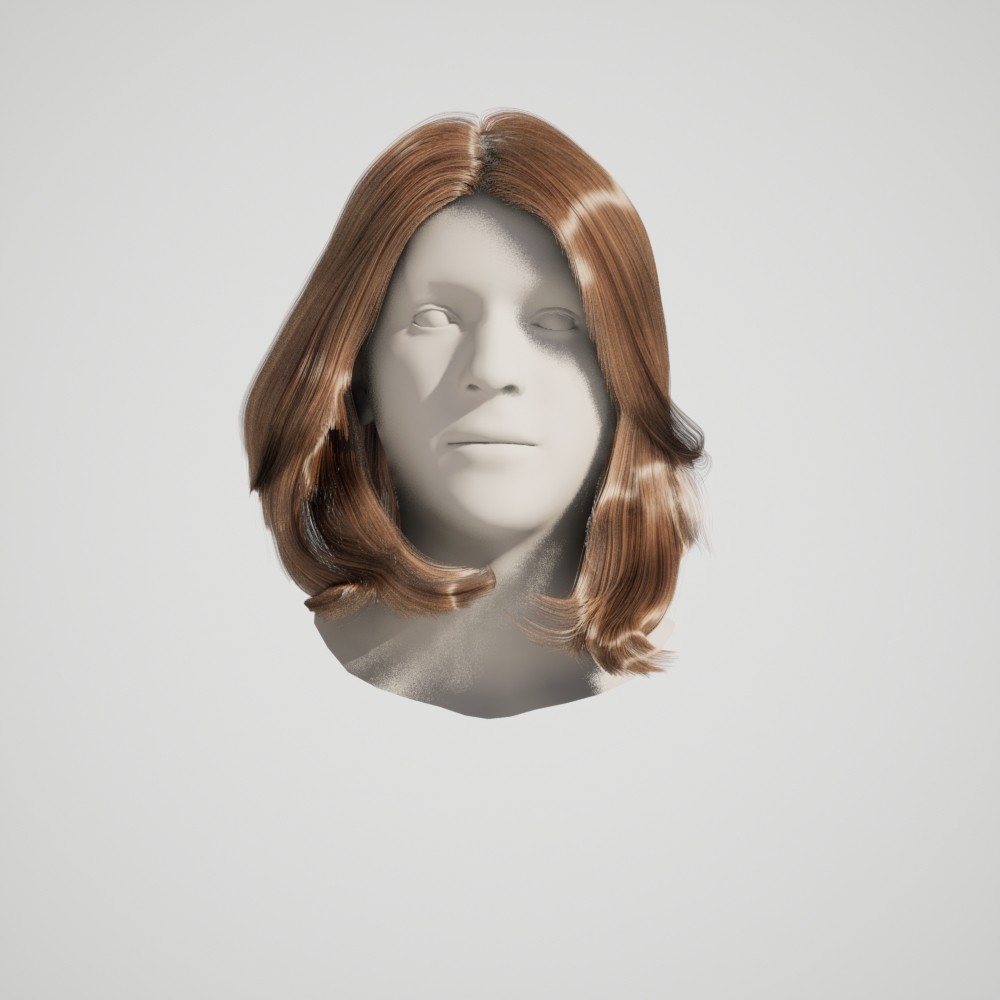}} 
        &
        \includegraphics[width=0.07\textwidth]{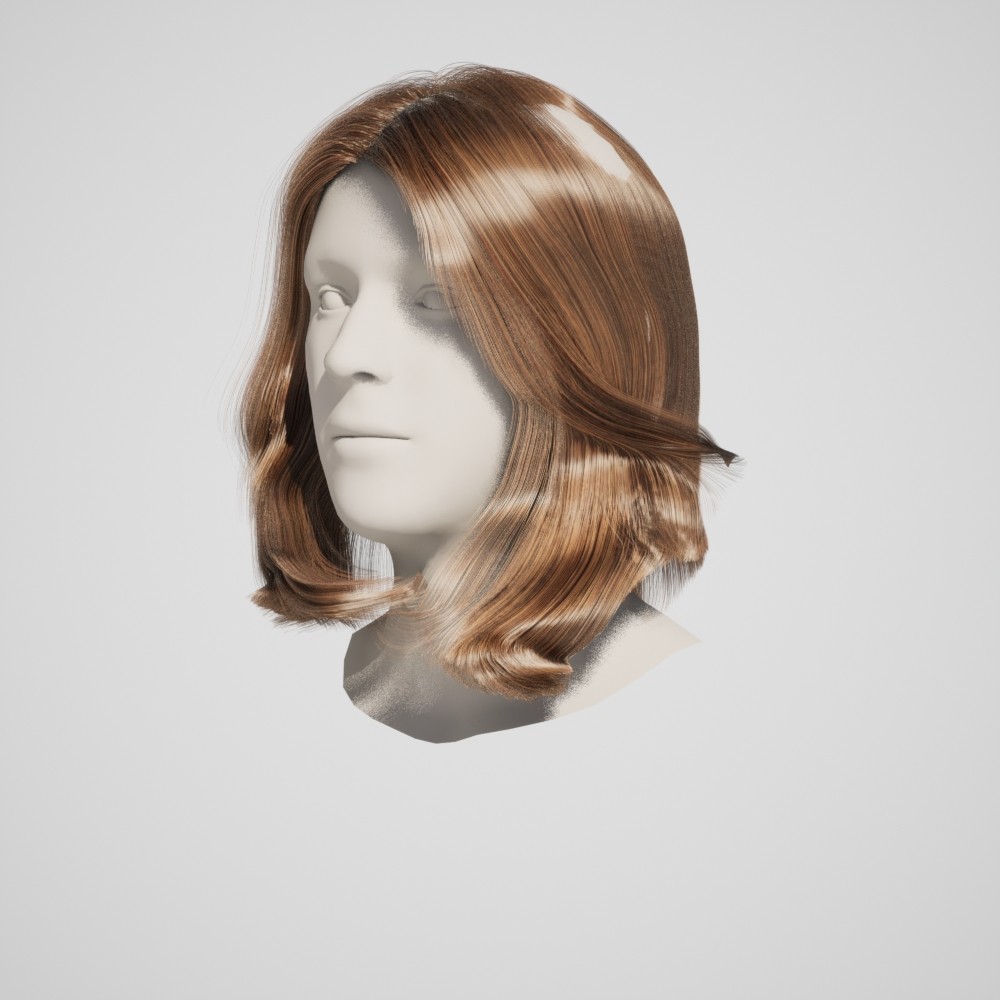} 
        &
        \multirow{2}{*}[0.481in]{\includegraphics[trim={125 200 125 50},clip,width=0.14\textwidth]{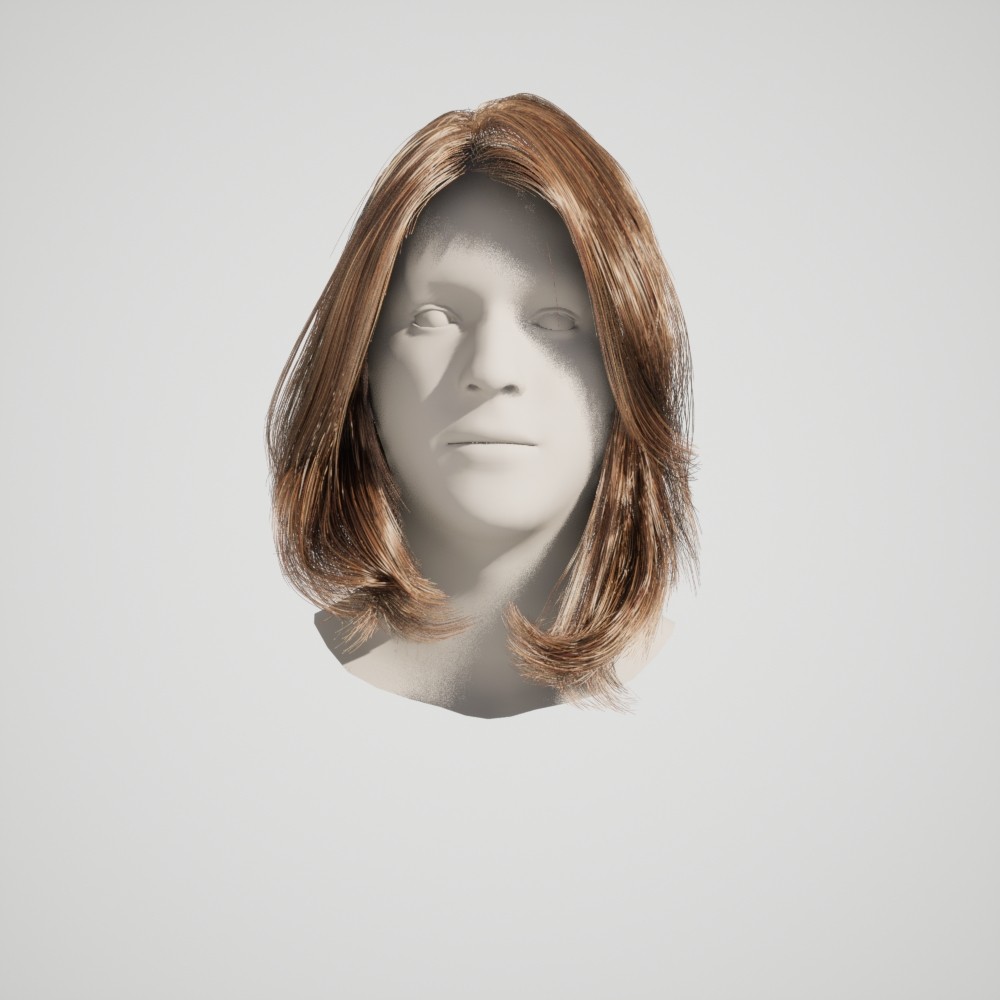}} 
        &
        \includegraphics[width=0.07\textwidth]{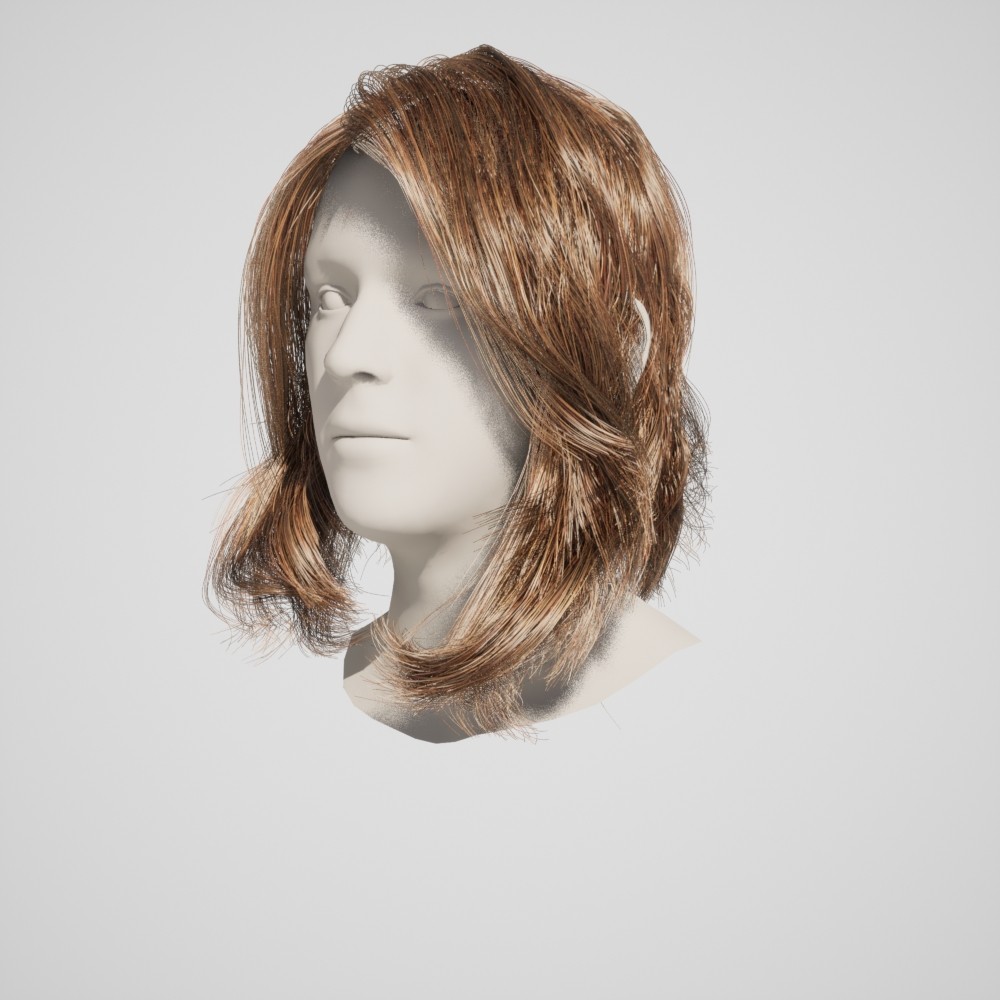} 
        &

        \multirow{2}{*}[0.481in]{\includegraphics[trim={125 200 125 50},clip,width=0.14\textwidth]{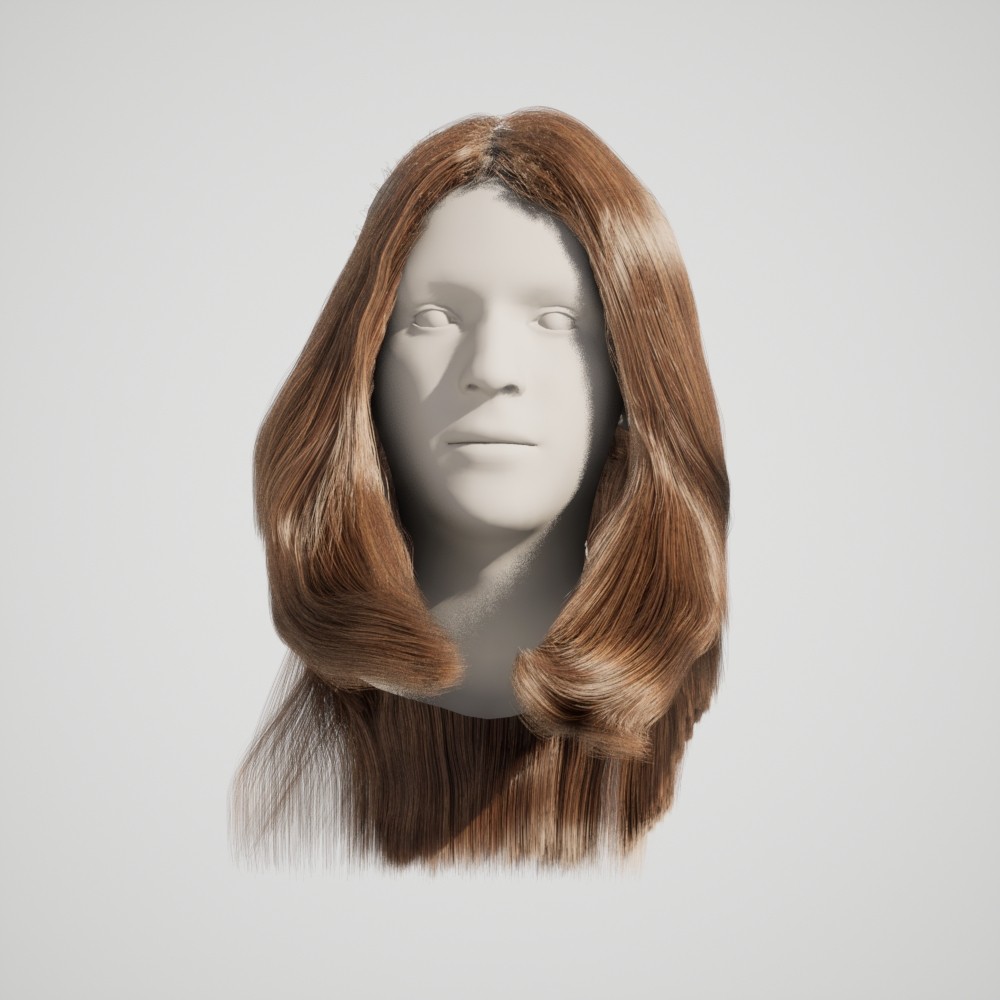}} 
        &
        \includegraphics[width=0.07\textwidth]{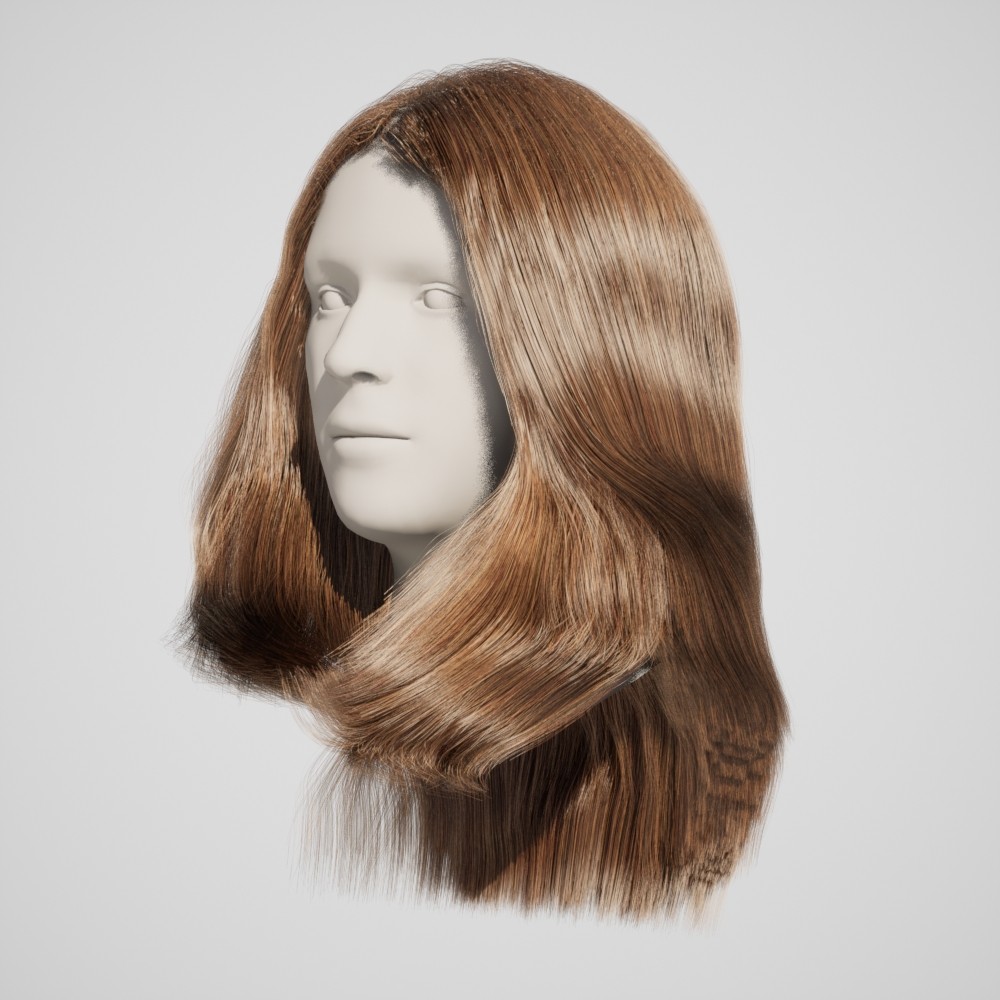}

        \\

        &
        &
        \includegraphics[width=0.07\textwidth]{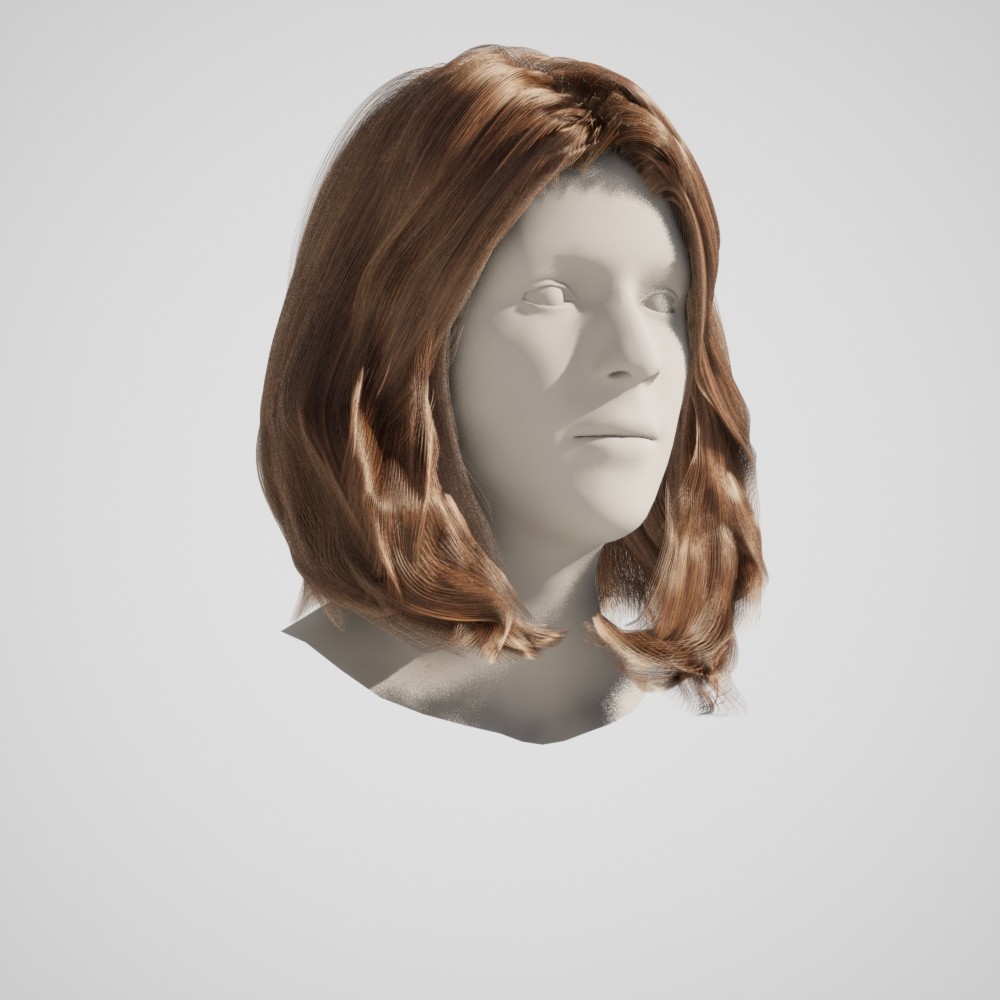}
        &
        &
        \includegraphics[width=0.07\textwidth]{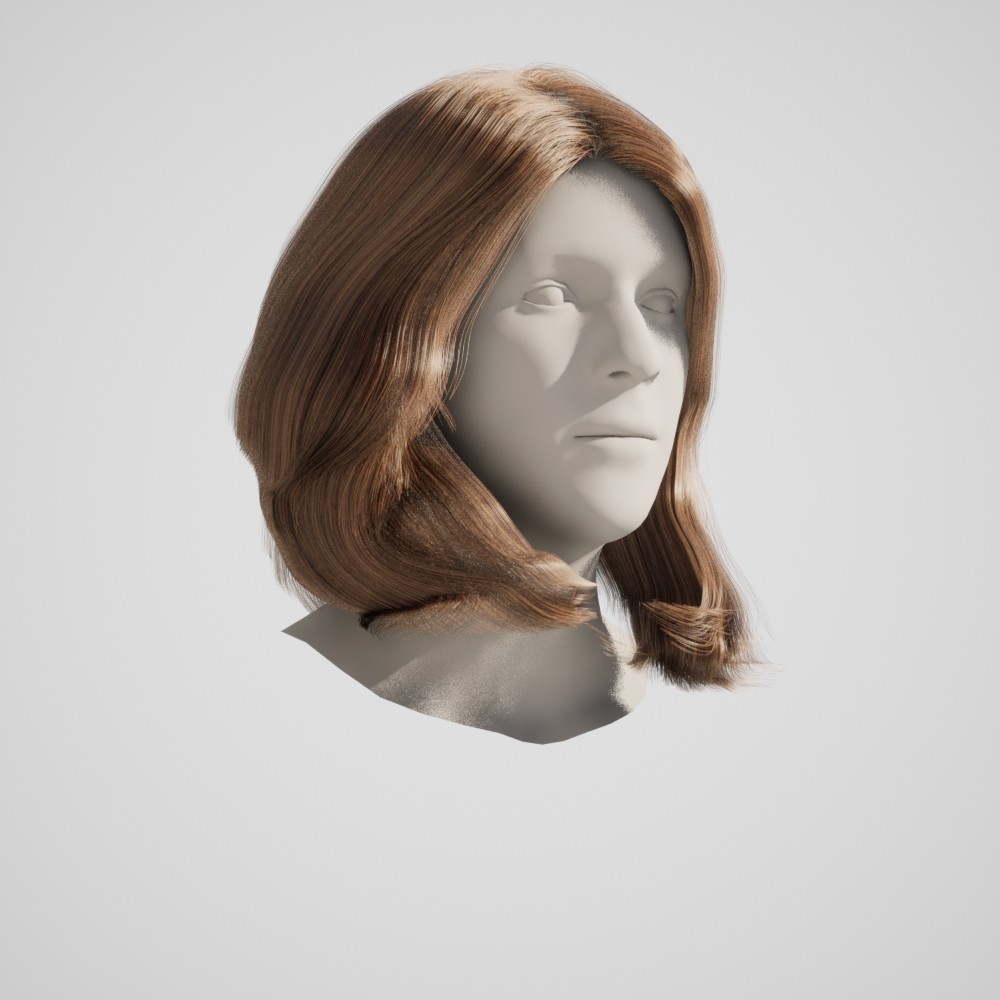}
        &
        &
        \includegraphics[width=0.07\textwidth]{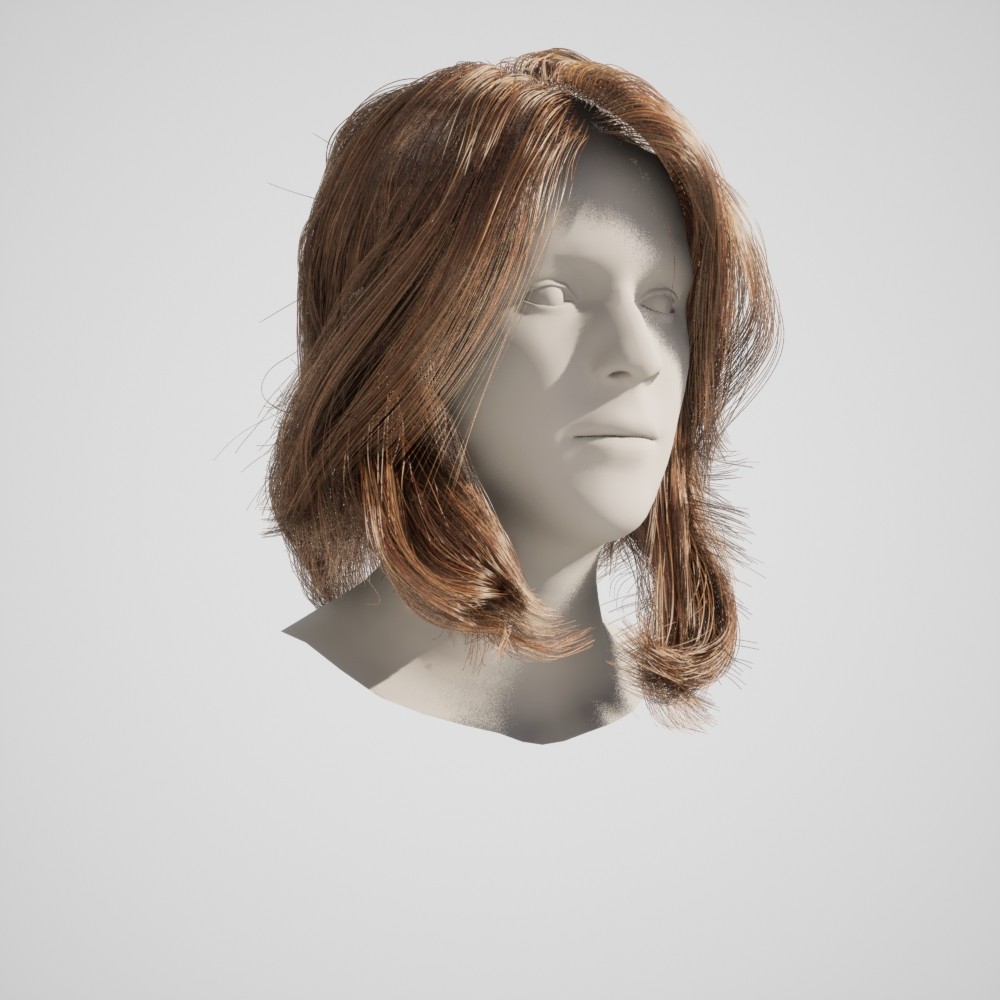}

        & 
        &
        \includegraphics[width=0.07\textwidth]{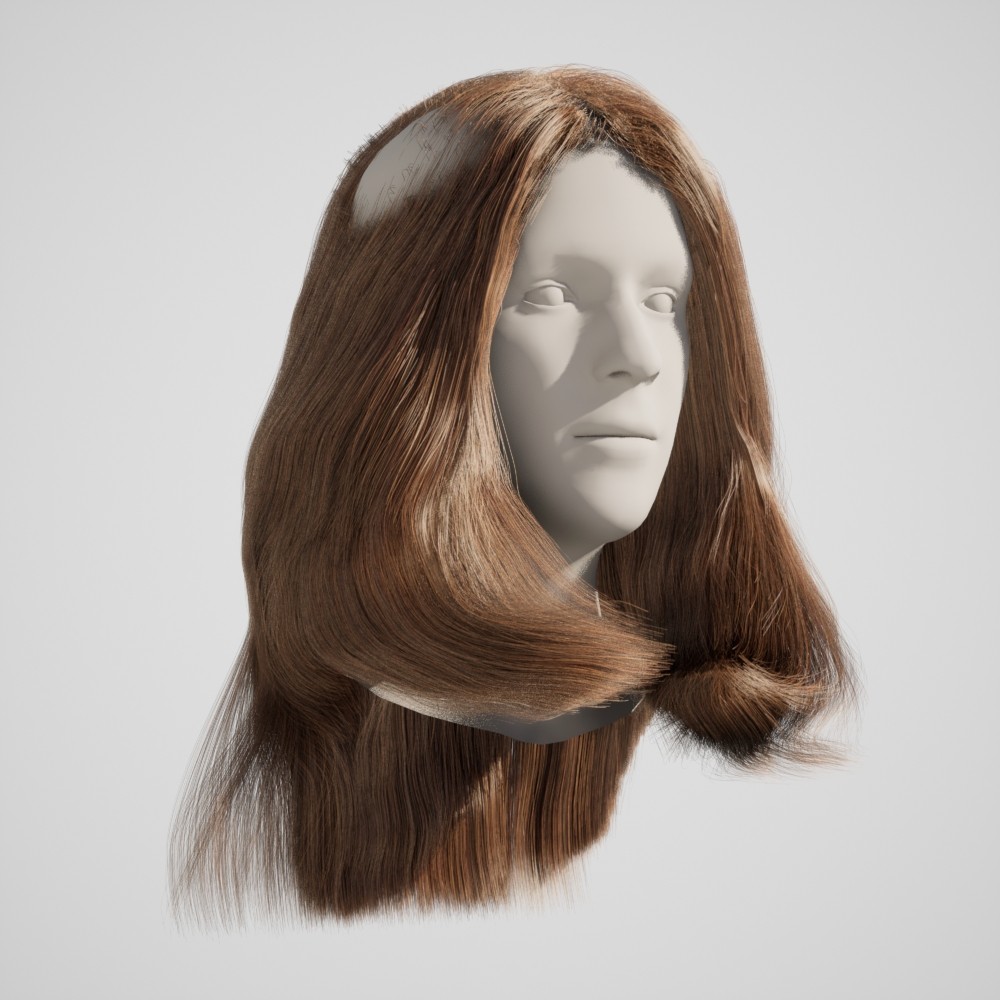} 

        \\

        \multirow{2}{*}[0.481in]{\includegraphics[width=0.14\textwidth]{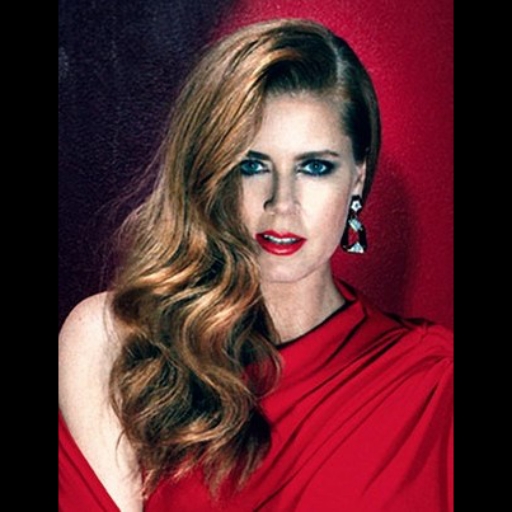}} 
            &
        \multirow{2}{*}[0.481in]{\includegraphics[width=0.14\textwidth]{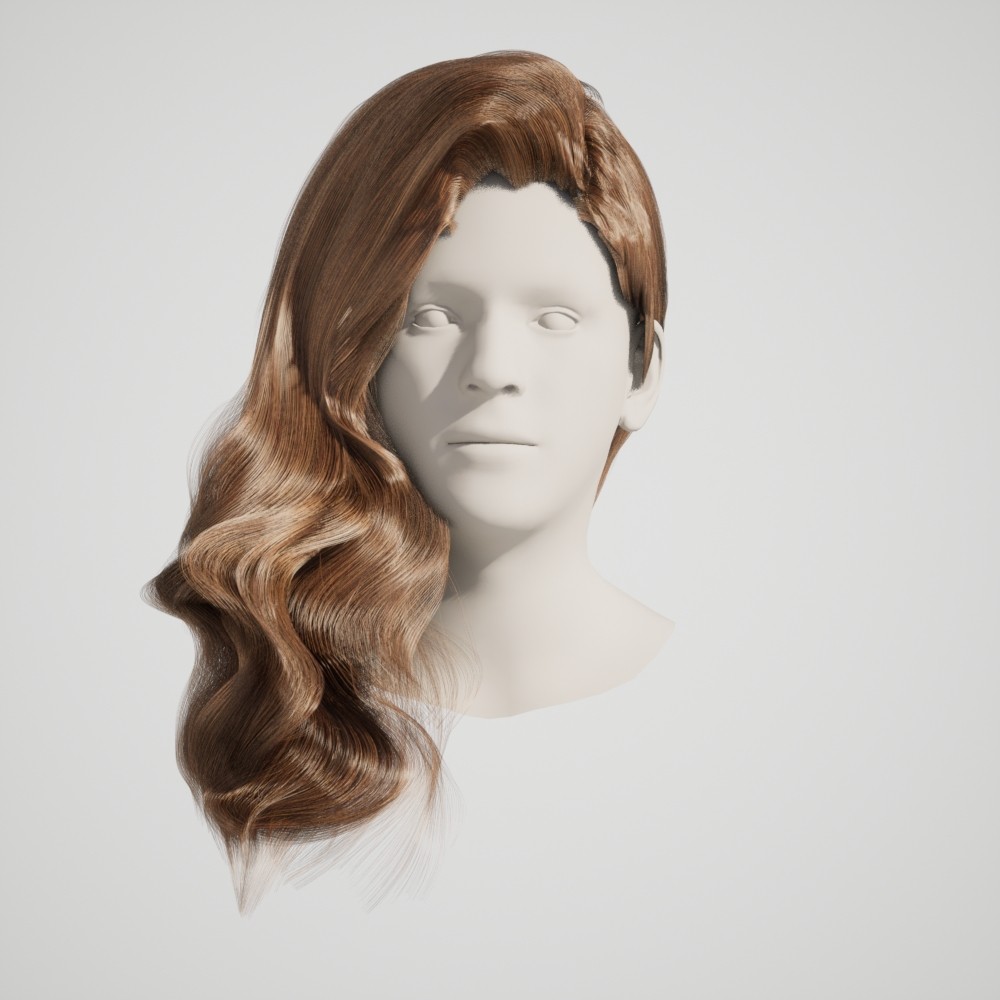}} 
        &
        \includegraphics[width=0.07\textwidth]{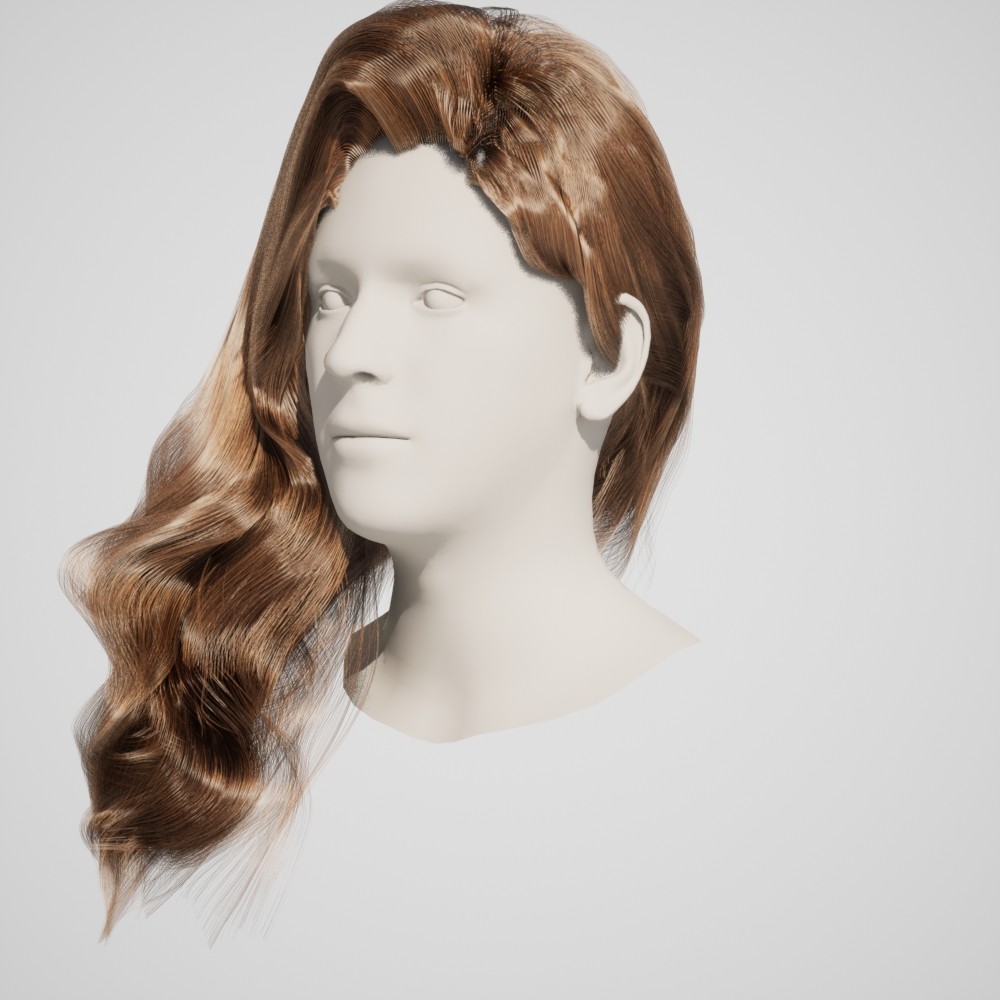} 
        &
        \multirow{2}{*}[0.481in]{\includegraphics[width=0.14\textwidth]{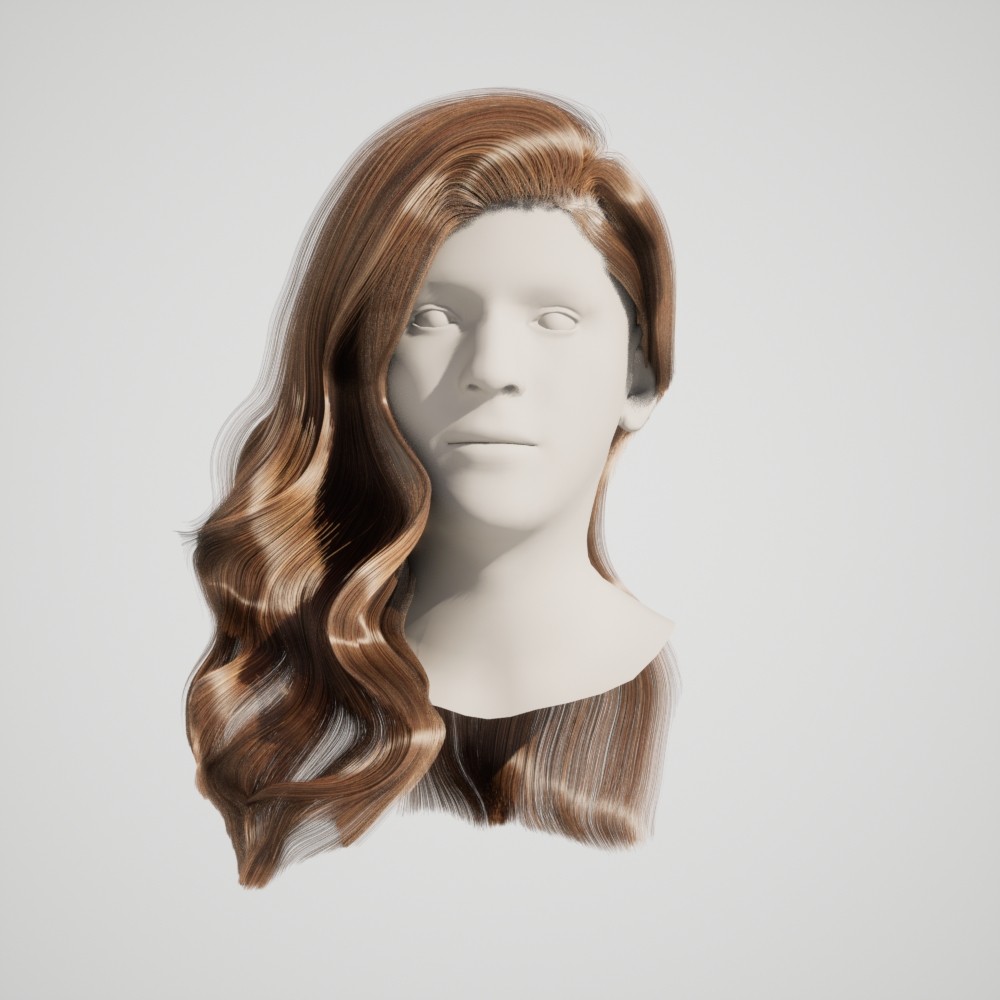}} 
        &
        \includegraphics[width=0.07\textwidth]{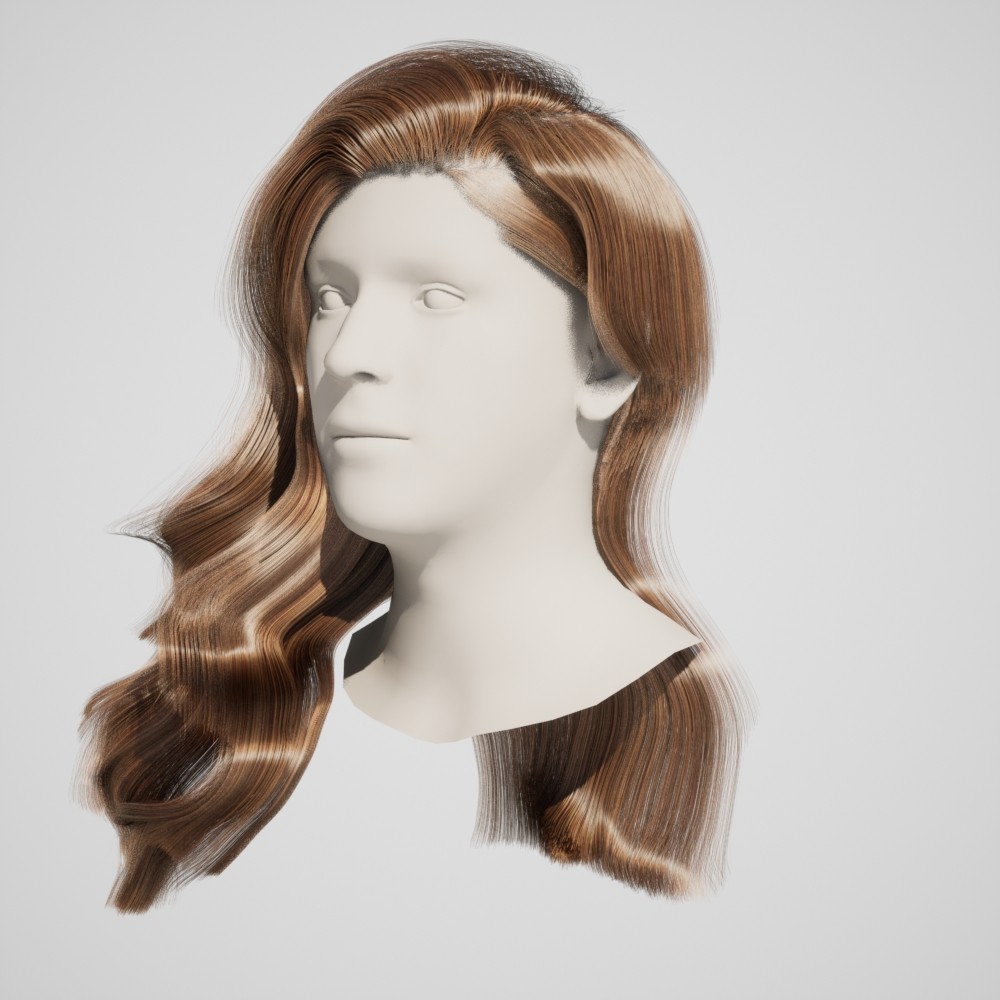} 
        &
        \multirow{2}{*}[0.481in]{\includegraphics[width=0.14\textwidth]{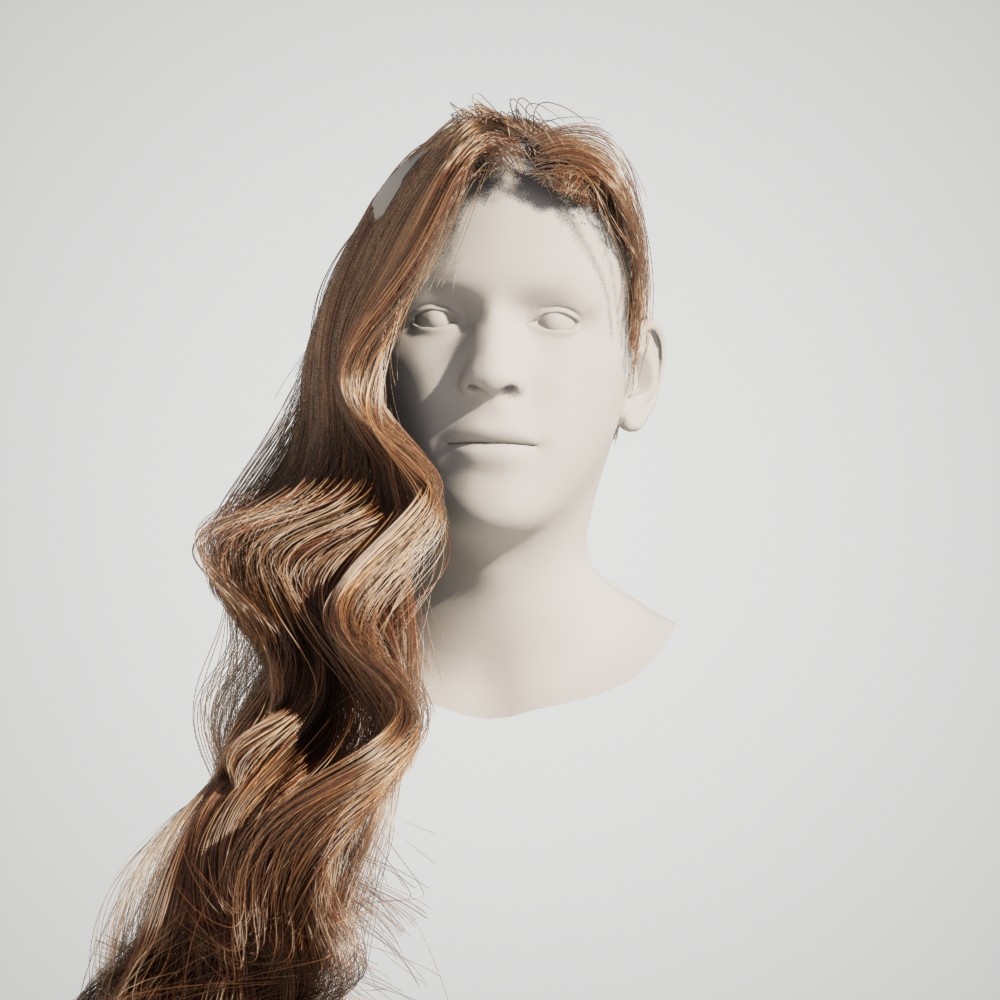}} 
        &
        \includegraphics[width=0.07\textwidth]{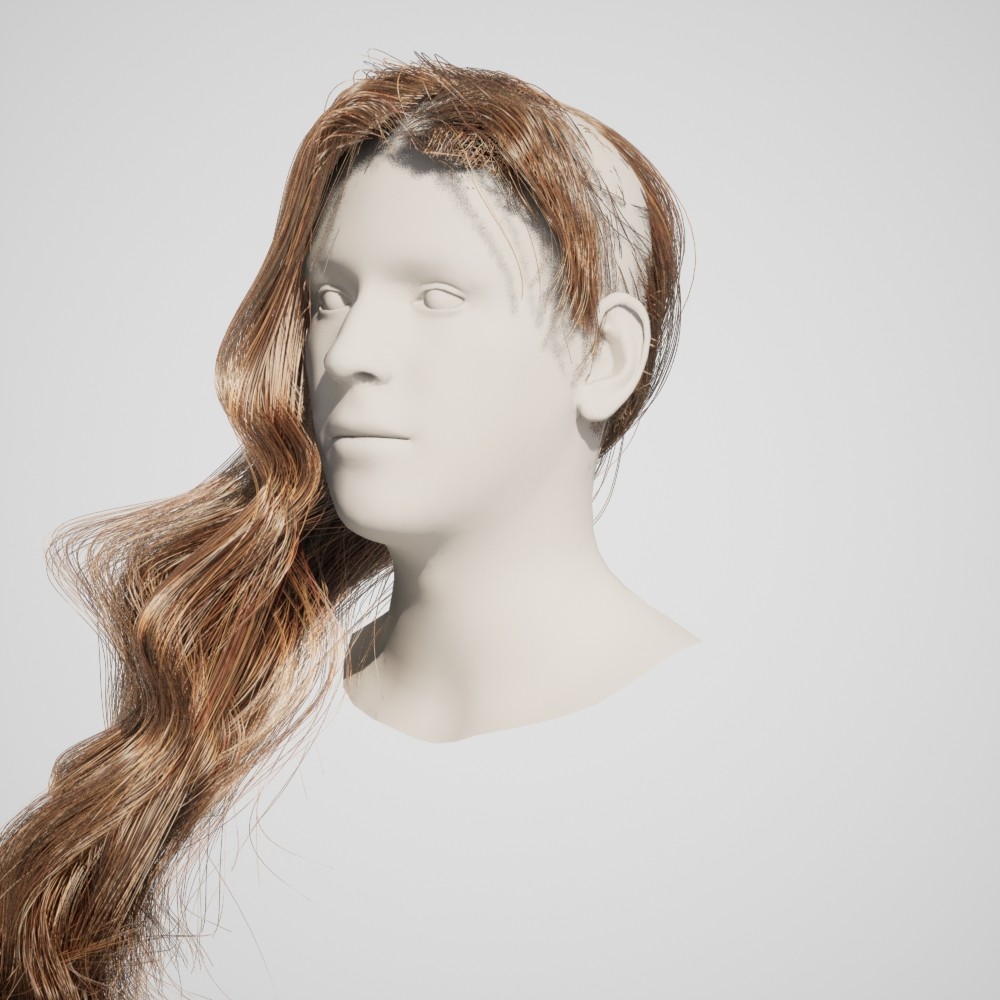} 
        &

        \multirow{2}{*}[0.481in]{\includegraphics[width=0.14\textwidth]{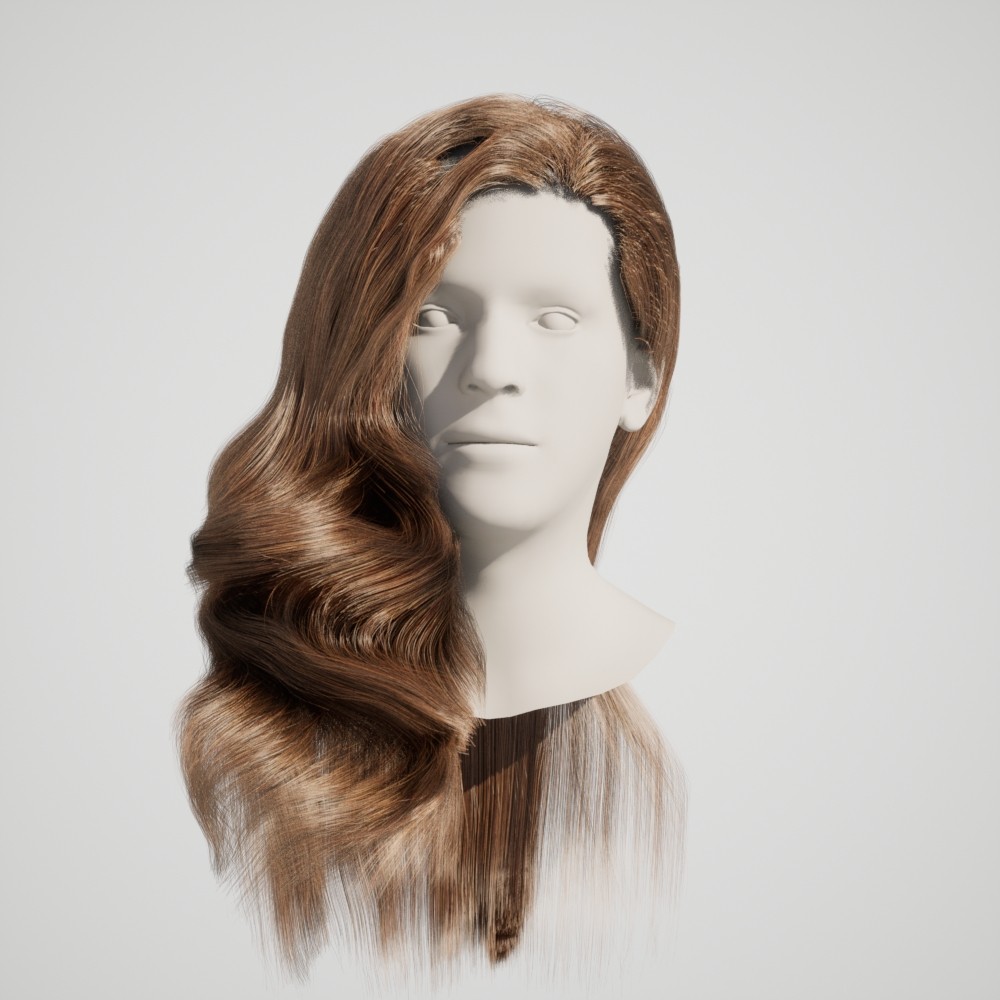}} 
        &
        \includegraphics[width=0.07\textwidth]{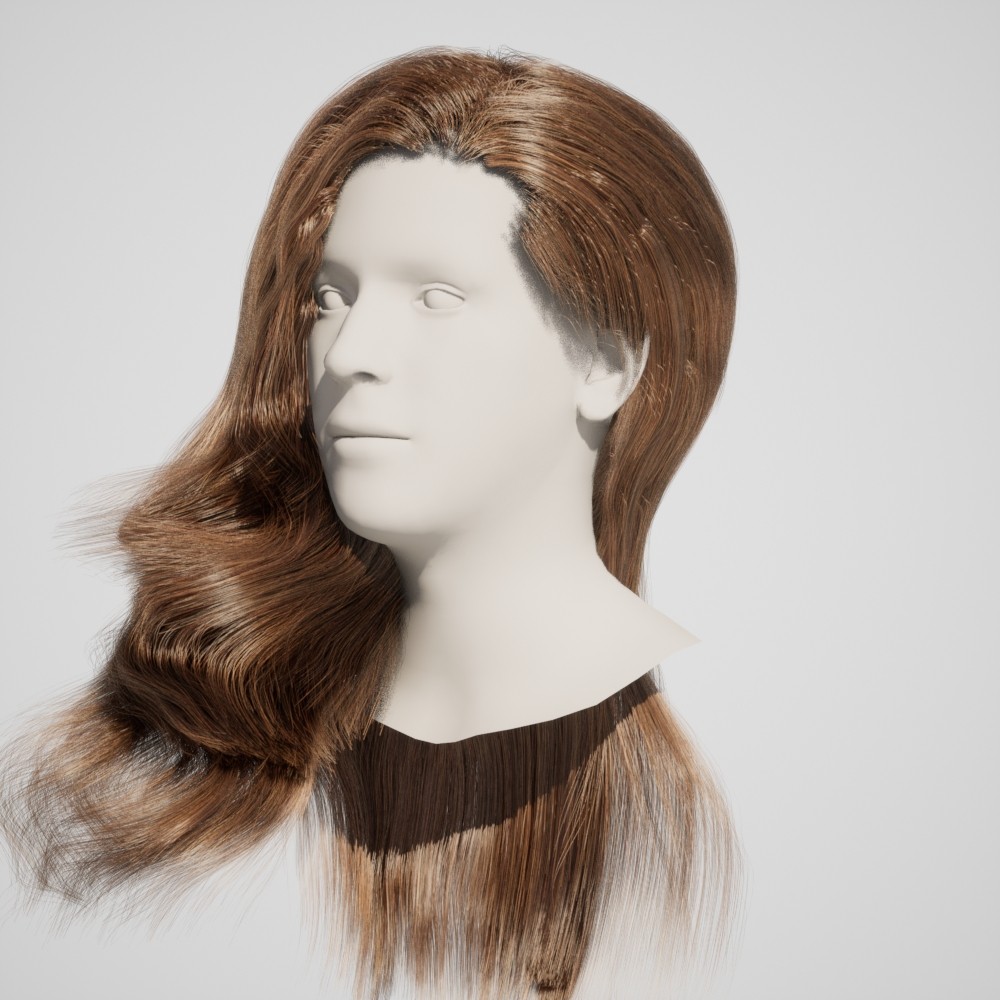}

        \\

        &
        &
        \includegraphics[width=0.07\textwidth]{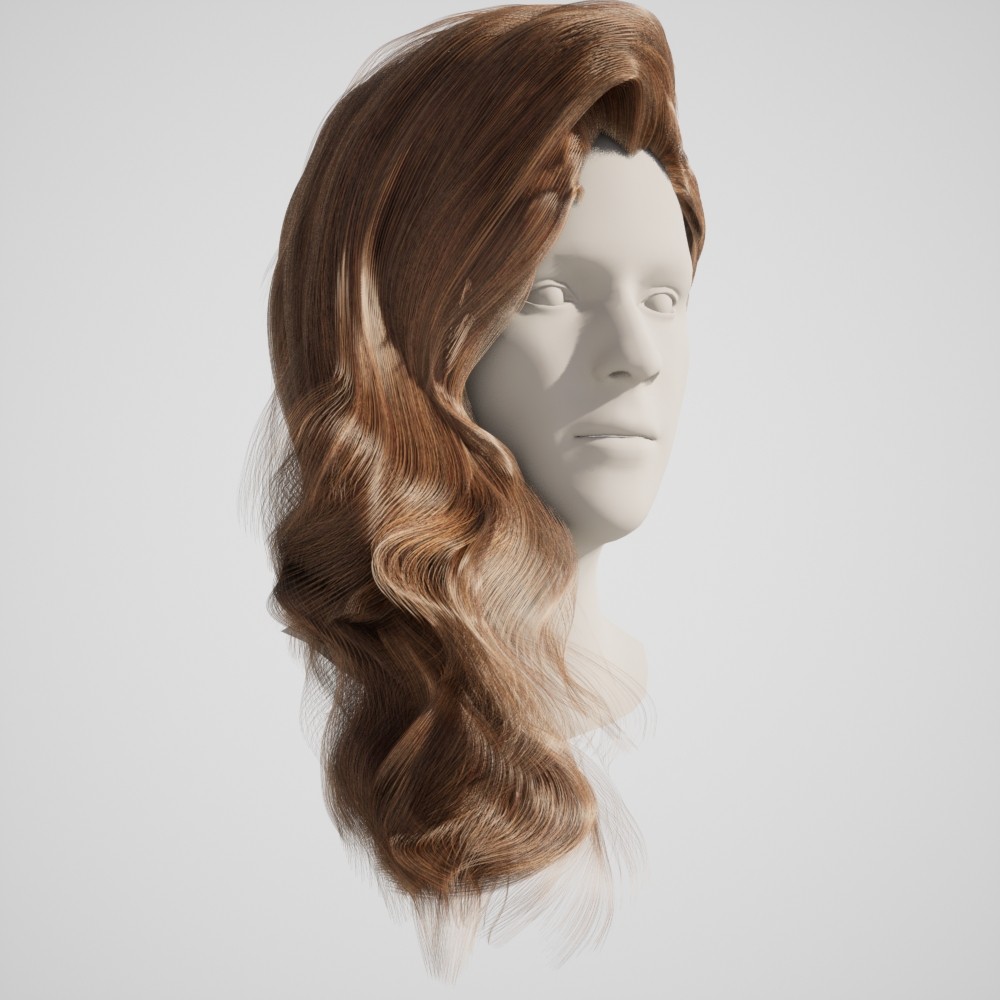}
        &
        &
        \includegraphics[width=0.07\textwidth]{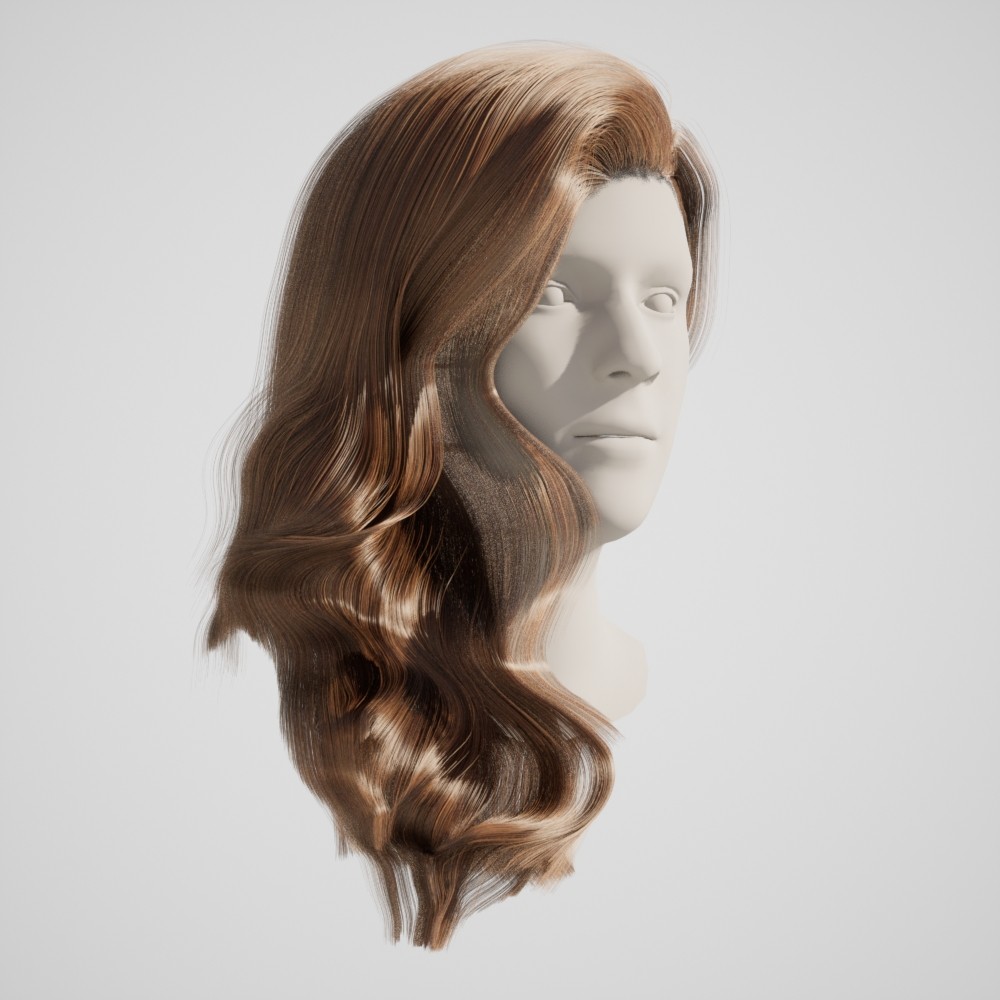}
        &
        &
        \includegraphics[width=0.07\textwidth]{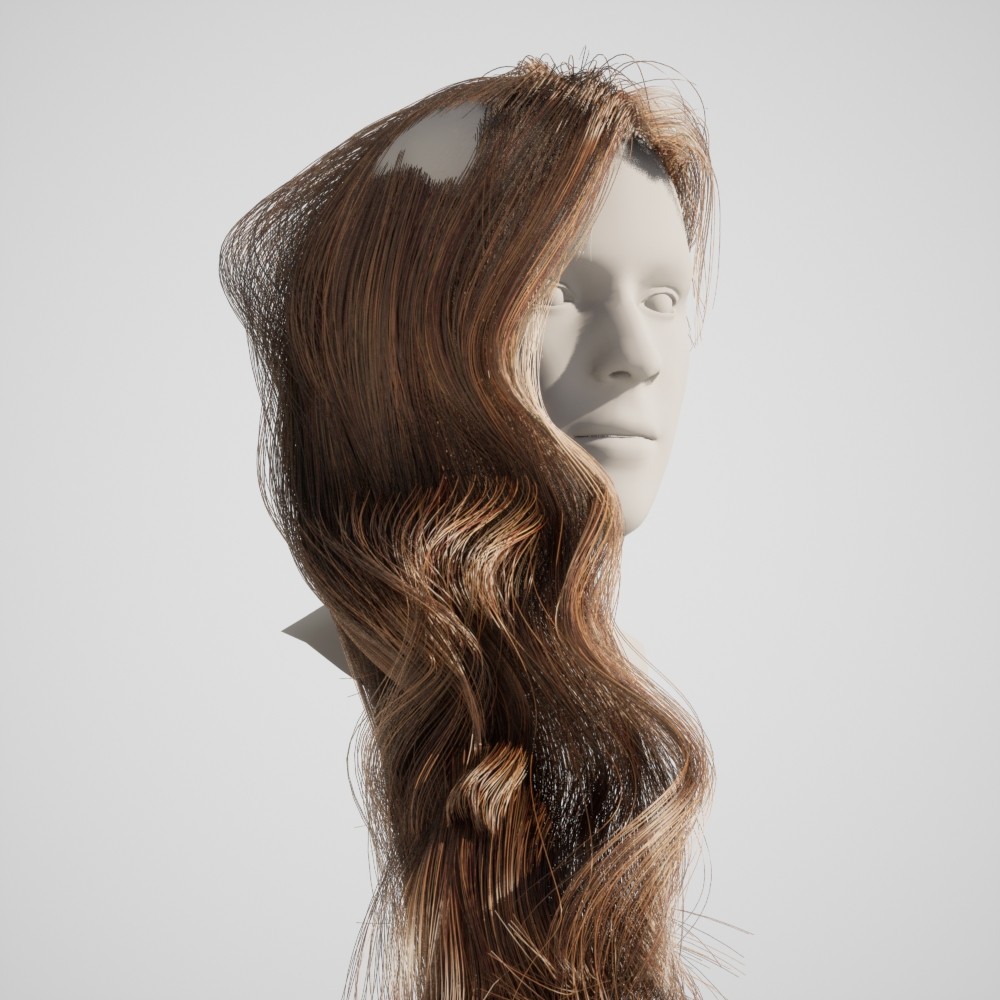}

        & 
        &
        \includegraphics[width=0.07\textwidth]{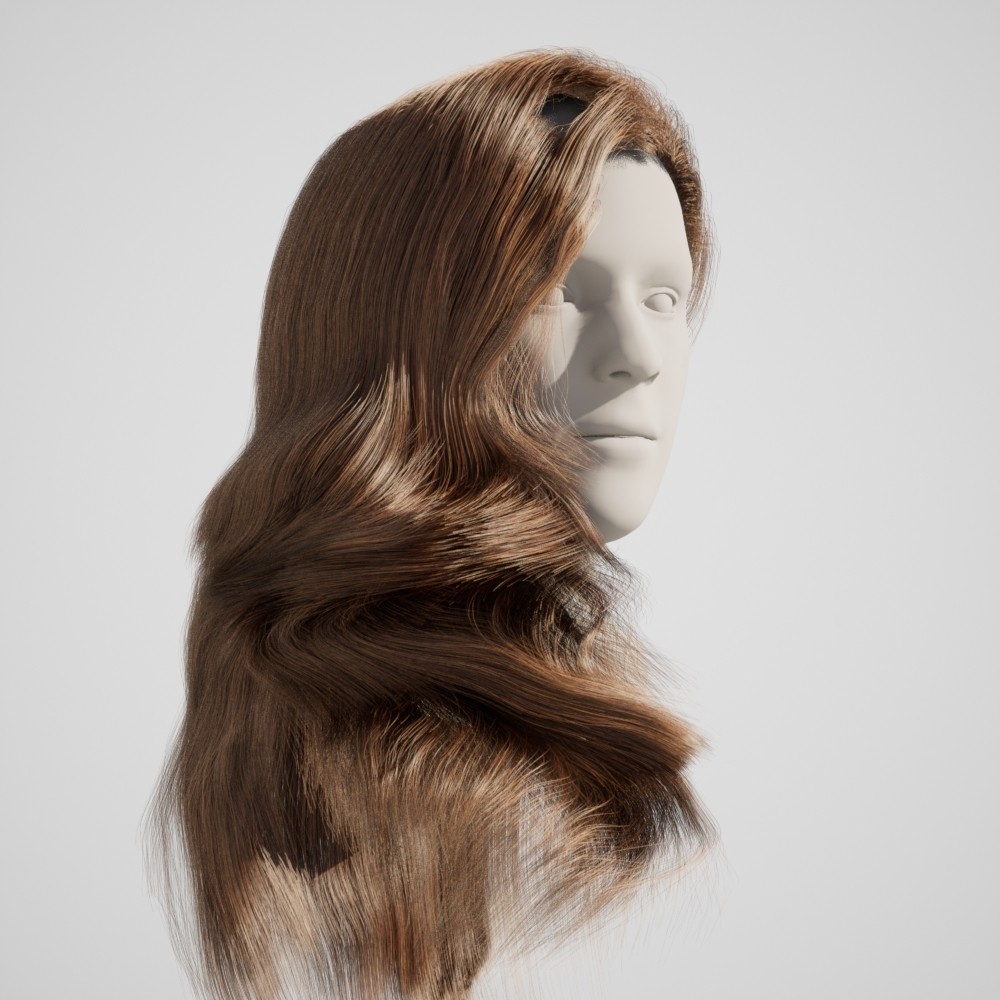} 

        \\

        \multirow{2}{*}[0.481in]{\includegraphics[width=0.14\textwidth]{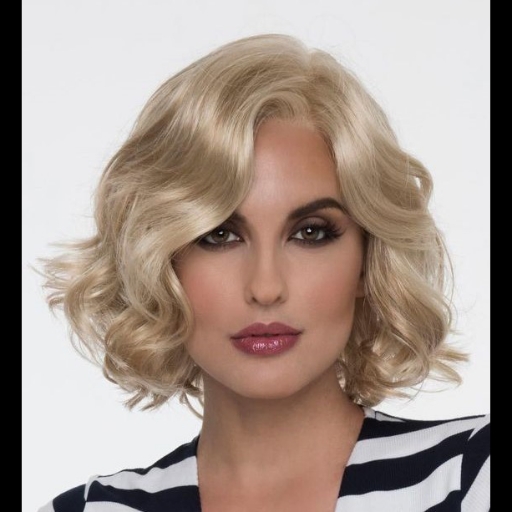}} 
            &
        \multirow{2}{*}[0.481in]{\includegraphics[trim={125 210 125 40},clip,width=0.14\textwidth]{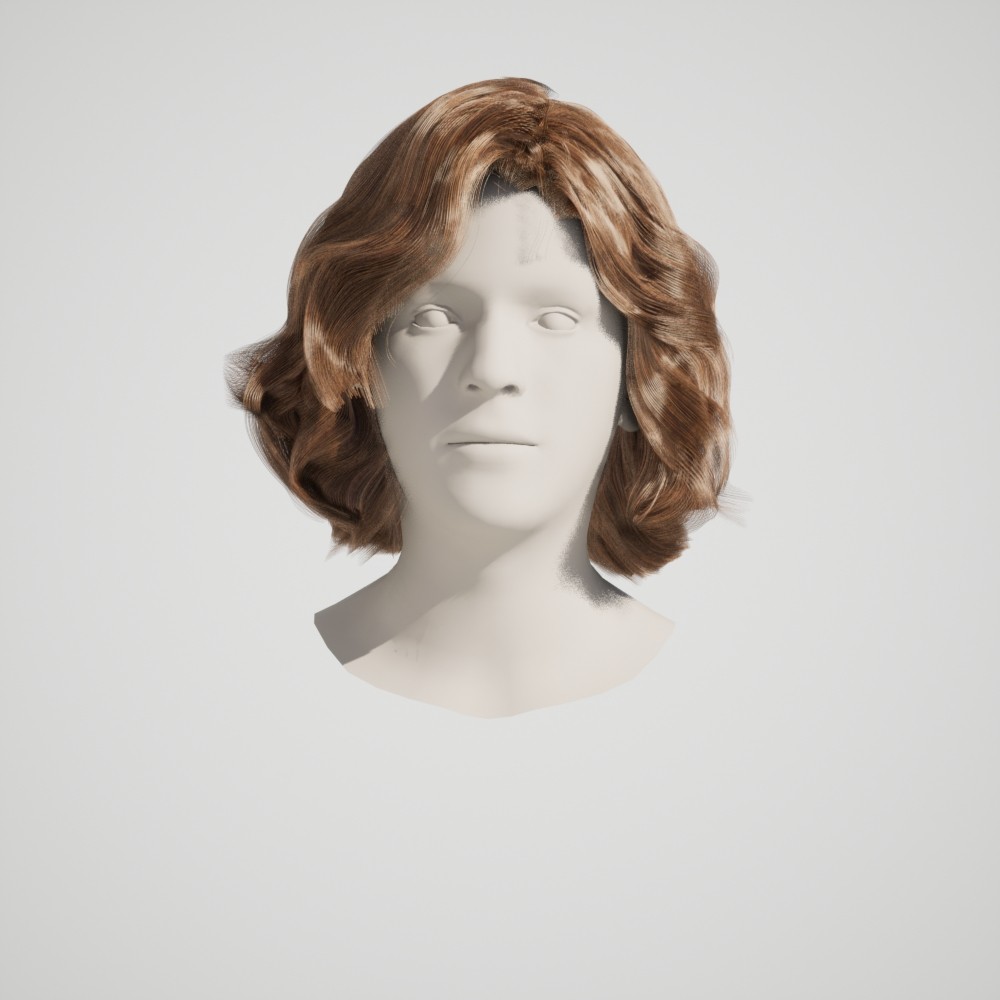}} 
        &
        \includegraphics[width=0.07\textwidth]{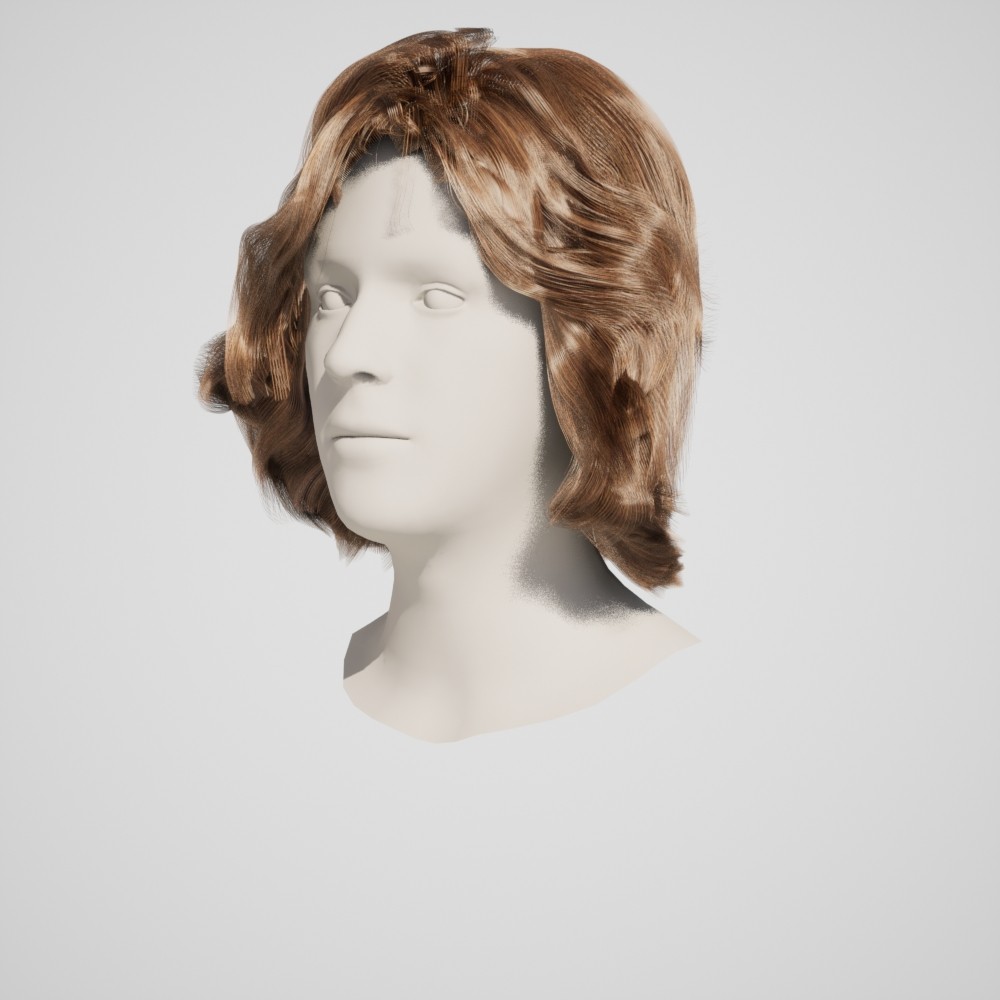} 
        &
        \multirow{2}{*}[0.481in]{\includegraphics[trim={125 210 125 40},clip,width=0.14\textwidth]{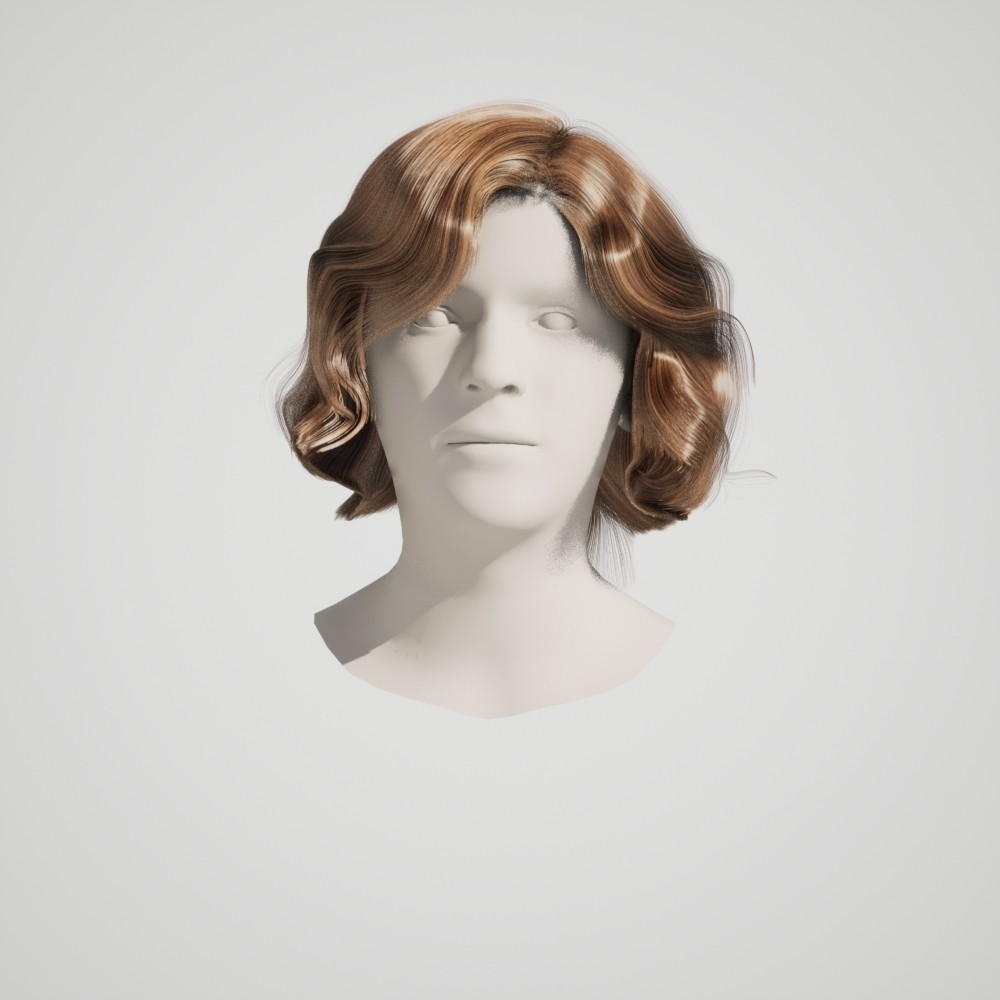}} 
        &
        \includegraphics[width=0.07\textwidth]{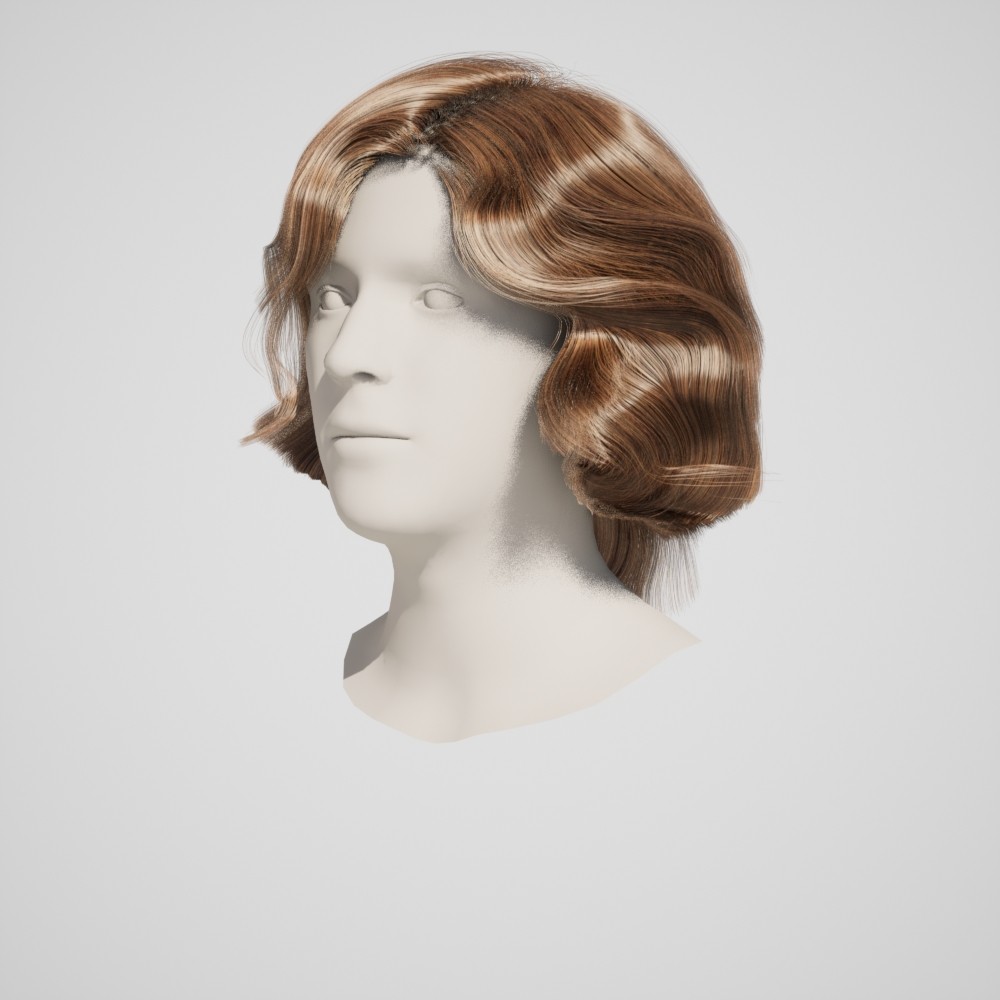} 
        &
        \multirow{2}{*}[0.481in]{\includegraphics[trim={125 210 125 40},clip,width=0.14\textwidth]{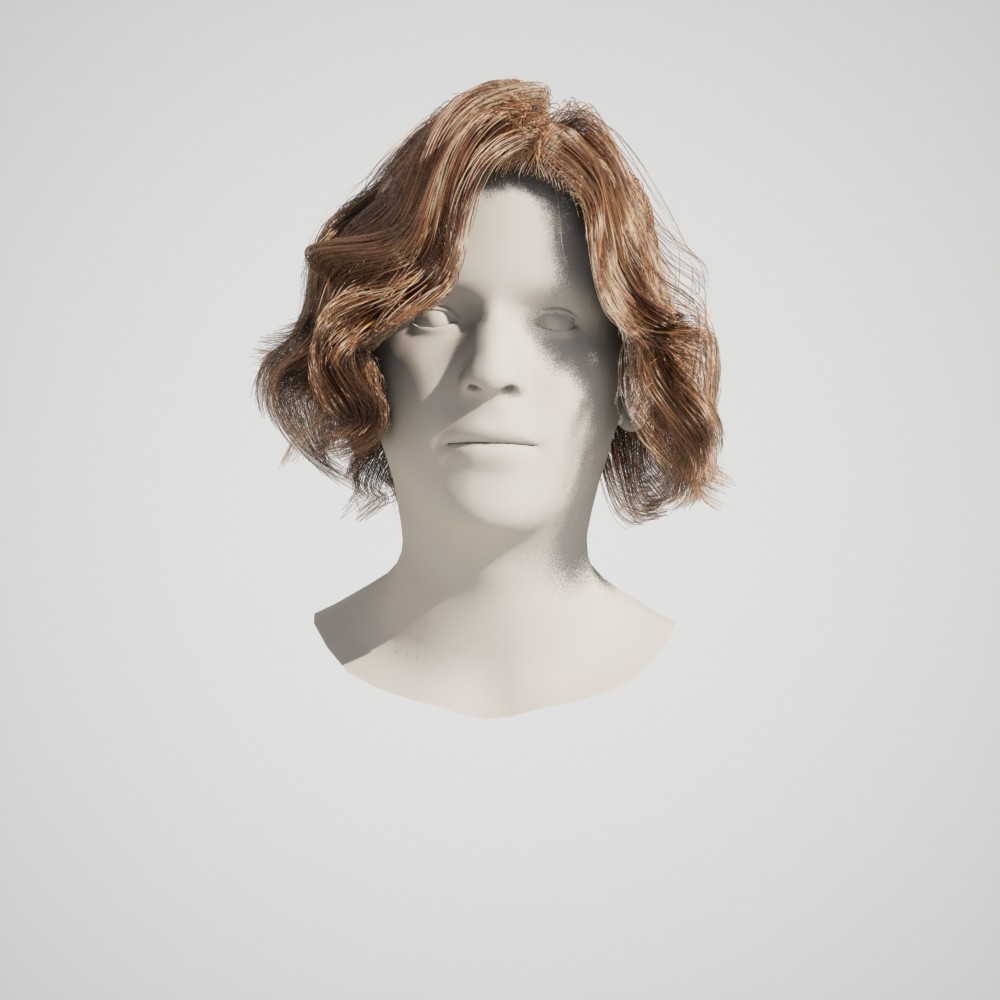}} 
        &
        \includegraphics[width=0.07\textwidth]{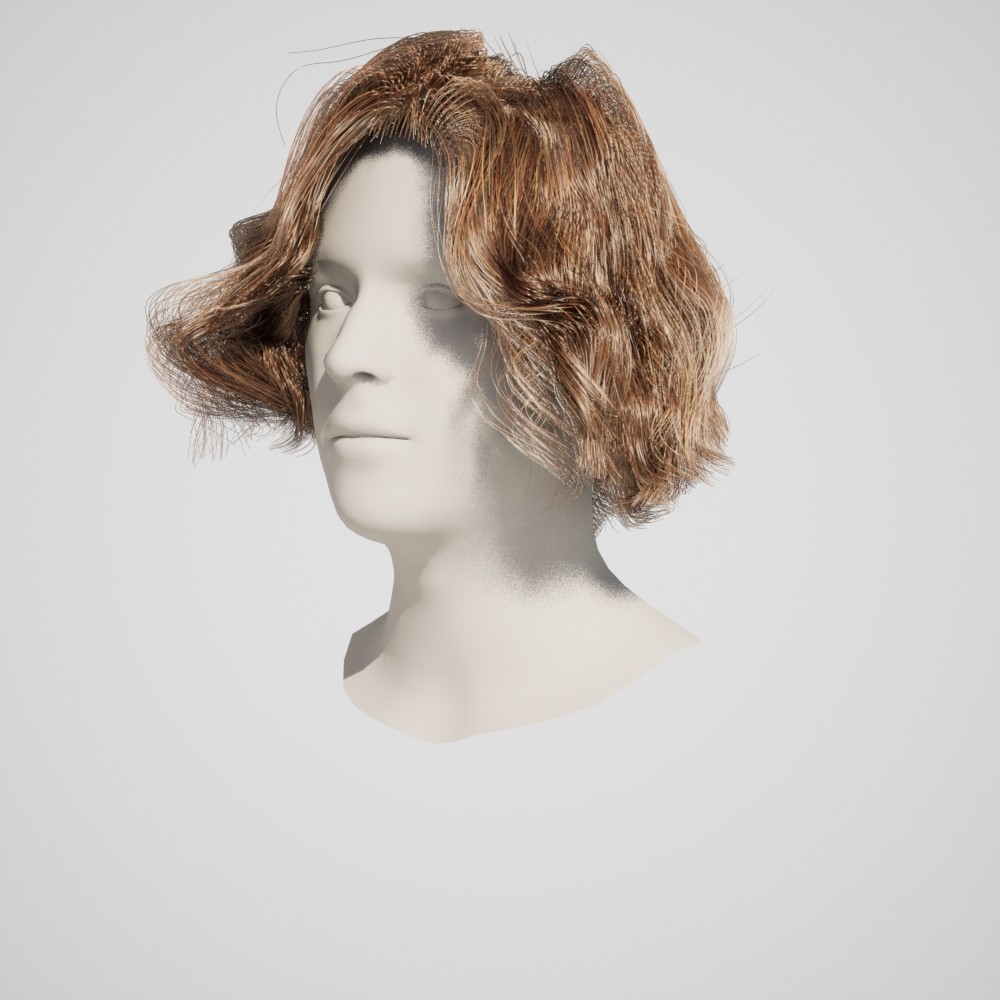} 
        &

        \multirow{2}{*}[0.481in]{\includegraphics[trim={125 210 125 40},clip,width=0.14\textwidth]{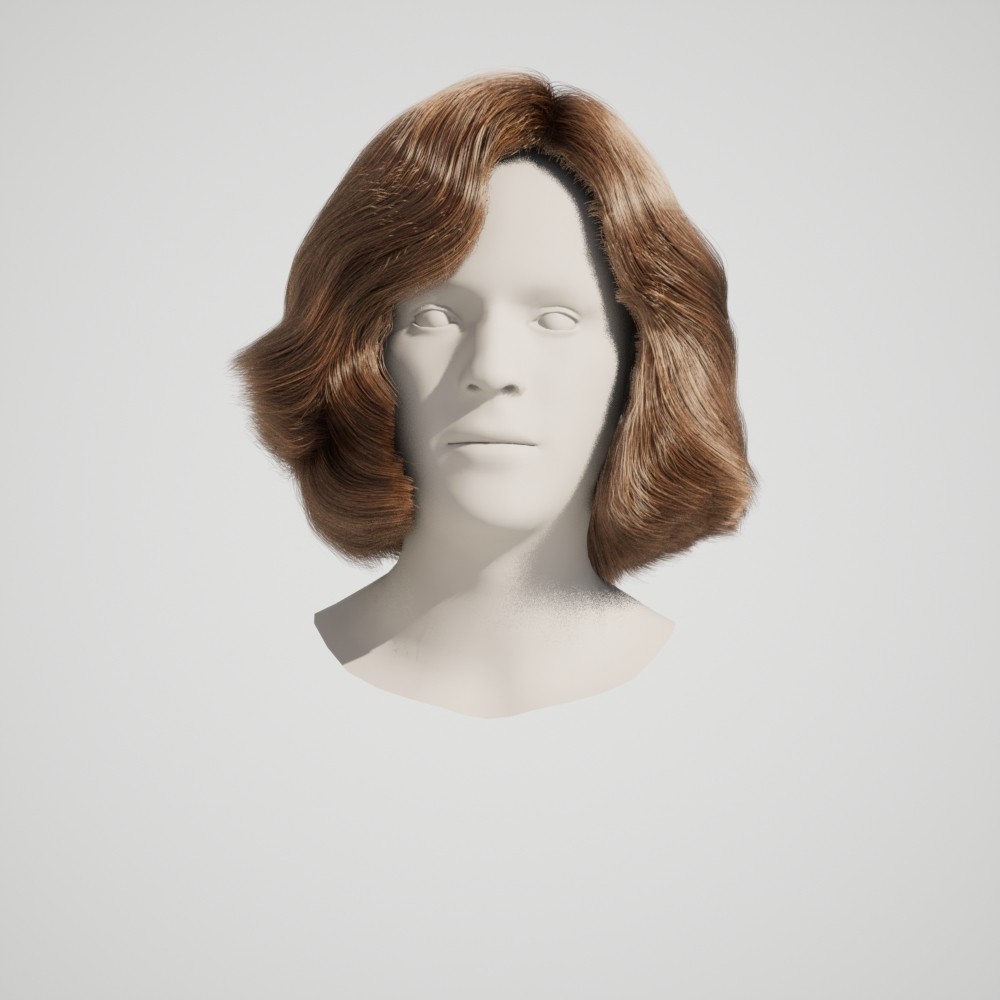}} 
        &
        \includegraphics[width=0.07\textwidth]{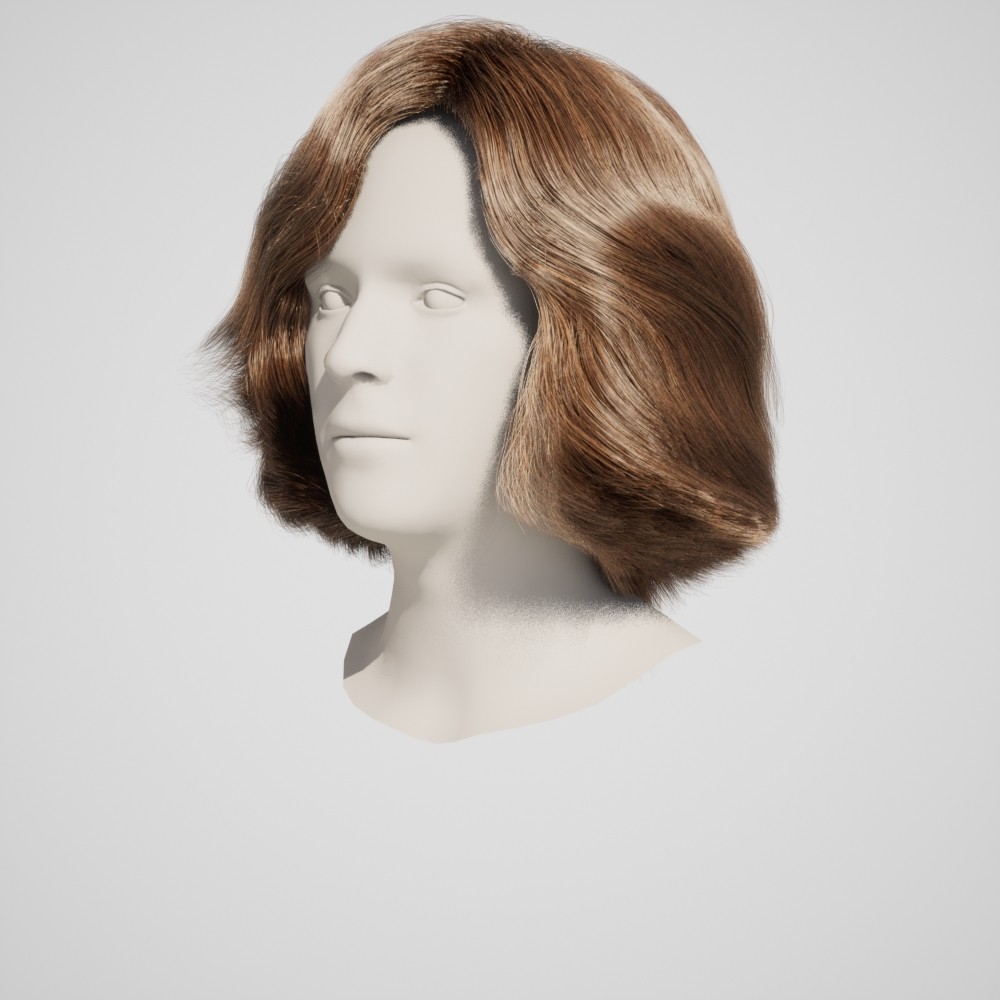}

        \\

        &
        &
        \includegraphics[width=0.07\textwidth]{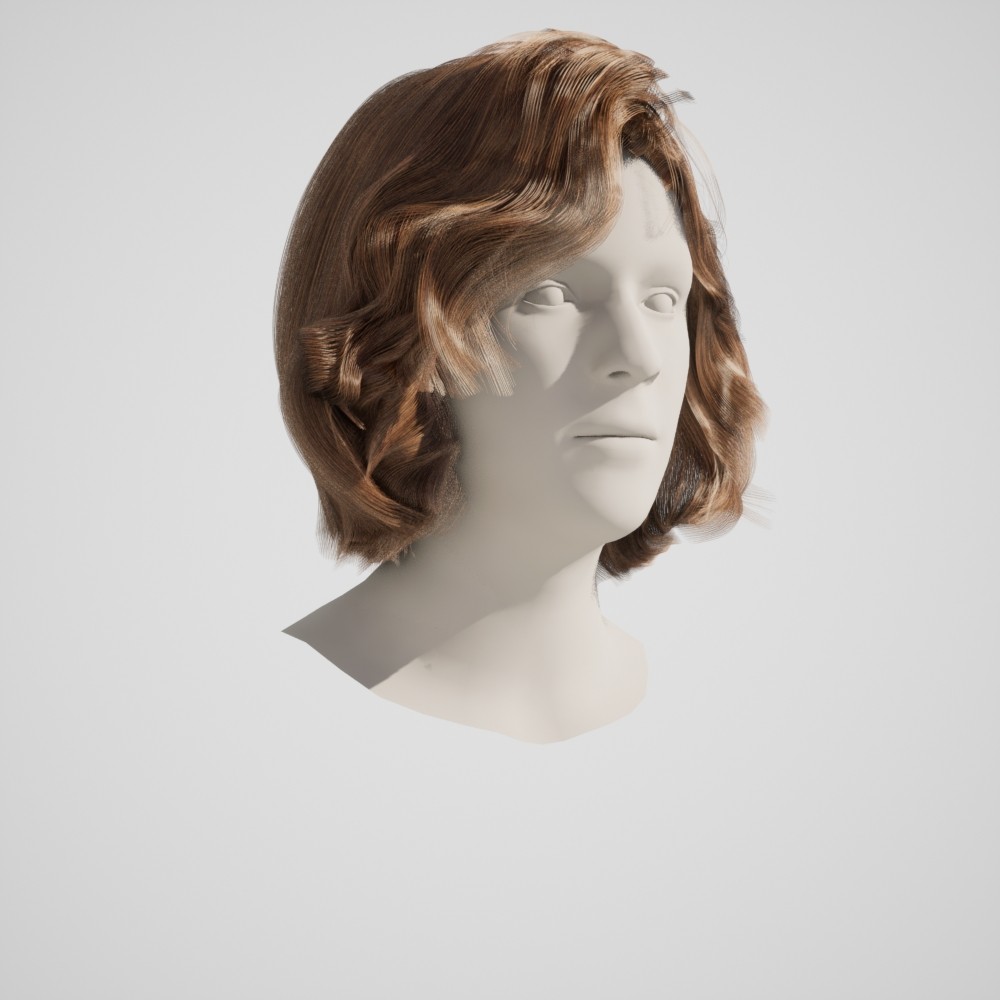}
        &
        &
        \includegraphics[width=0.07\textwidth]{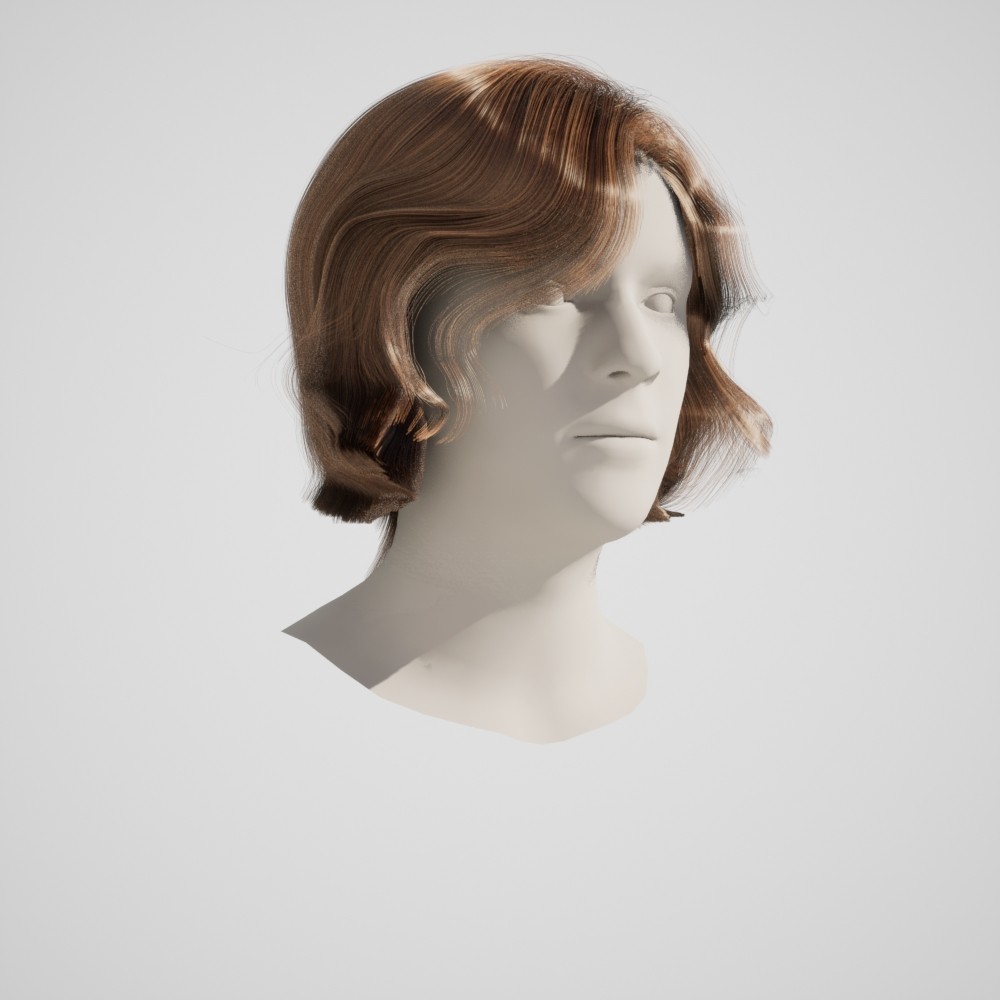}
        &
        &
        \includegraphics[width=0.07\textwidth]{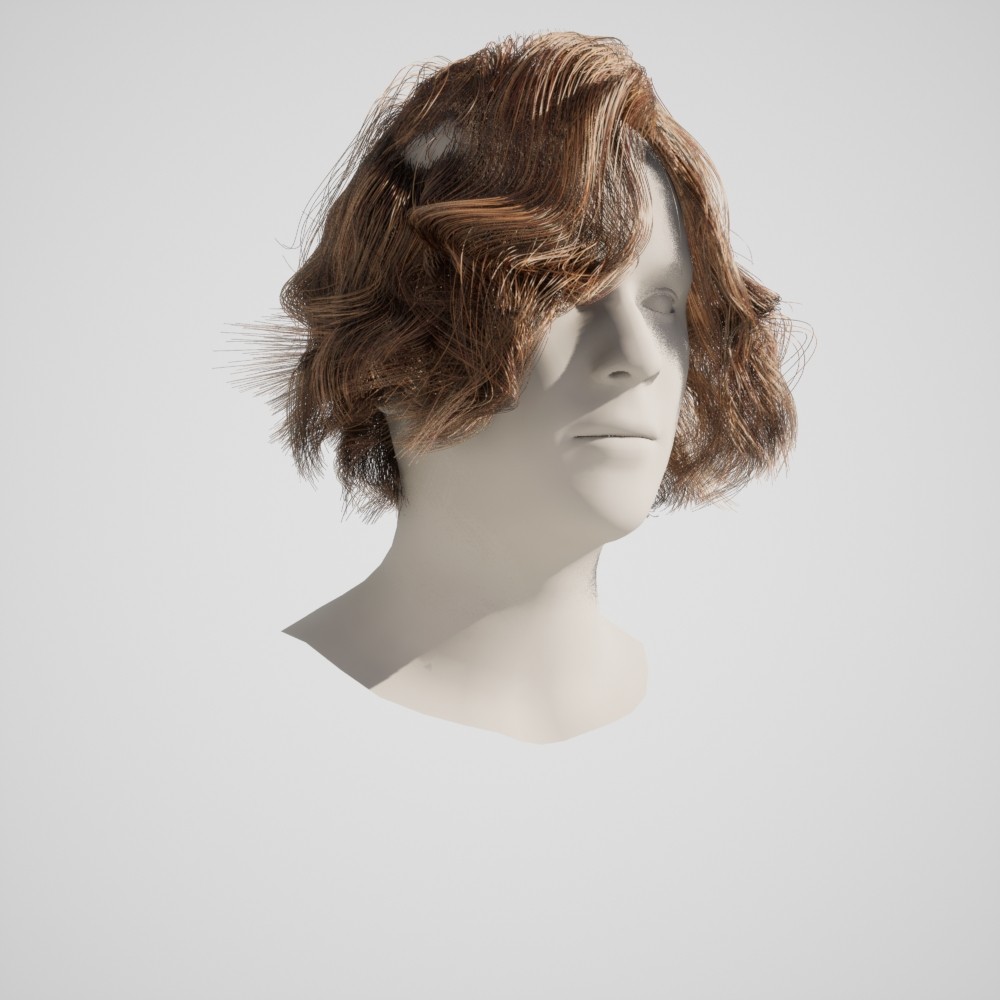}

        & 
        &
        \includegraphics[width=0.07\textwidth]{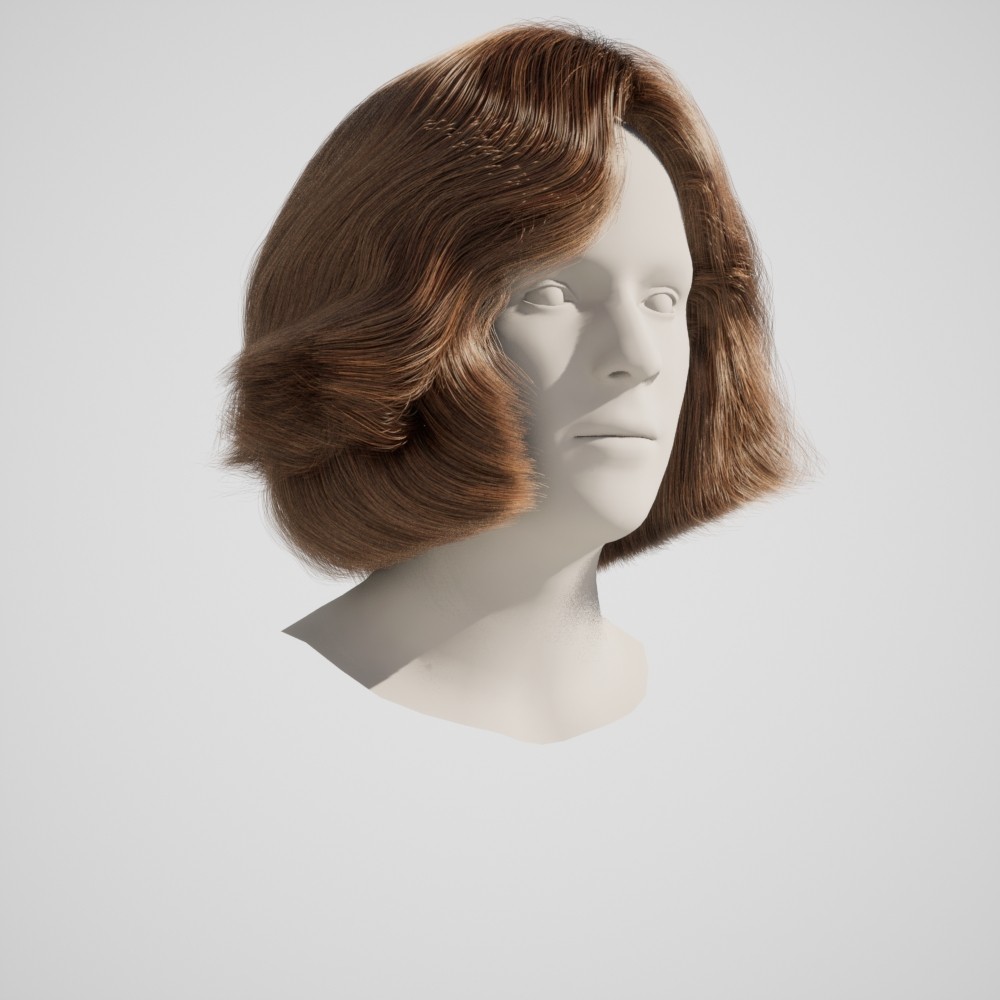} 

        \\

        \multirow{2}{*}[0.481in]{\includegraphics[width=0.14\textwidth]{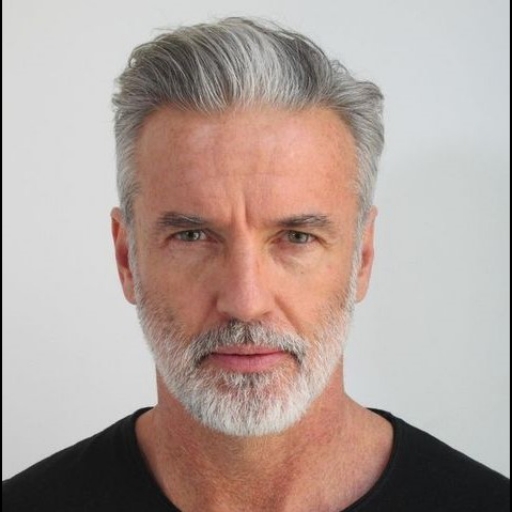}} 
            &
        \multirow{2}{*}[0.481in]{\includegraphics[trim={160 280 160 40},clip,width=0.14\textwidth]{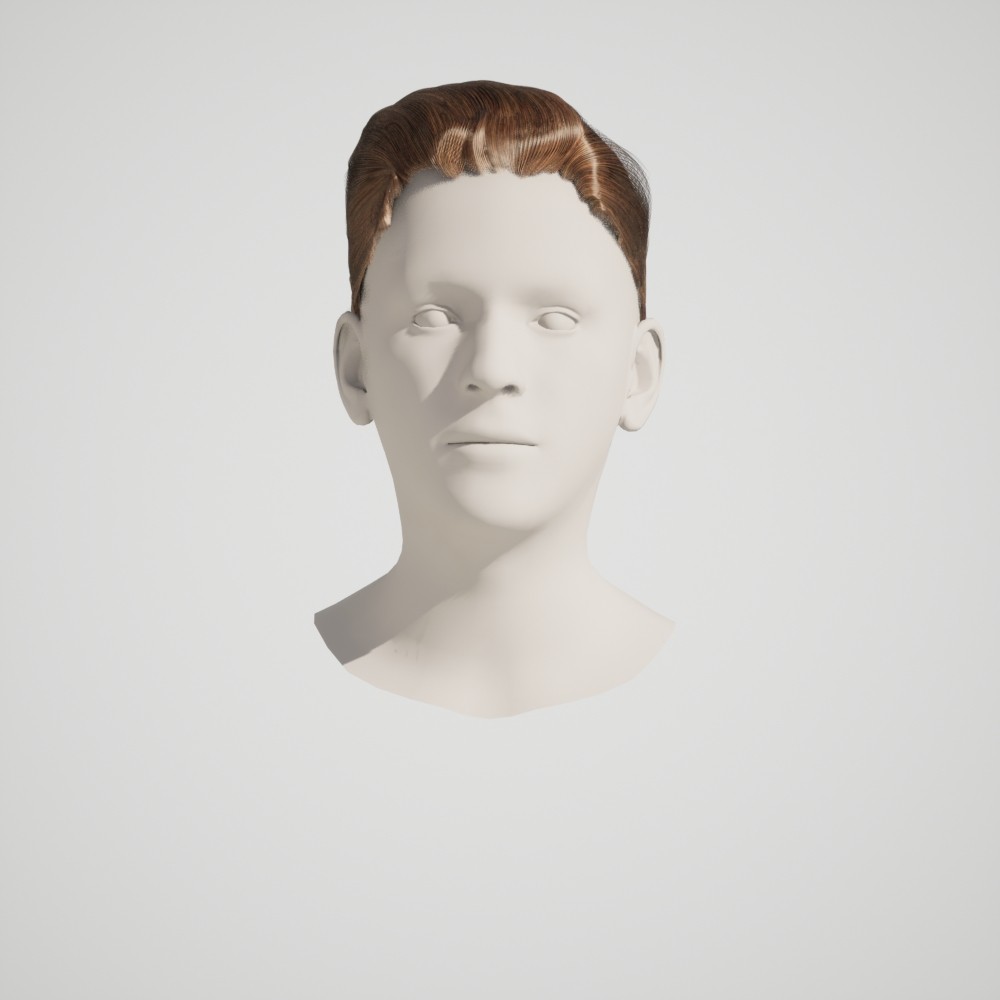}} 
        &
        \includegraphics[trim={100 160 100 40},clip,width=0.07\textwidth]{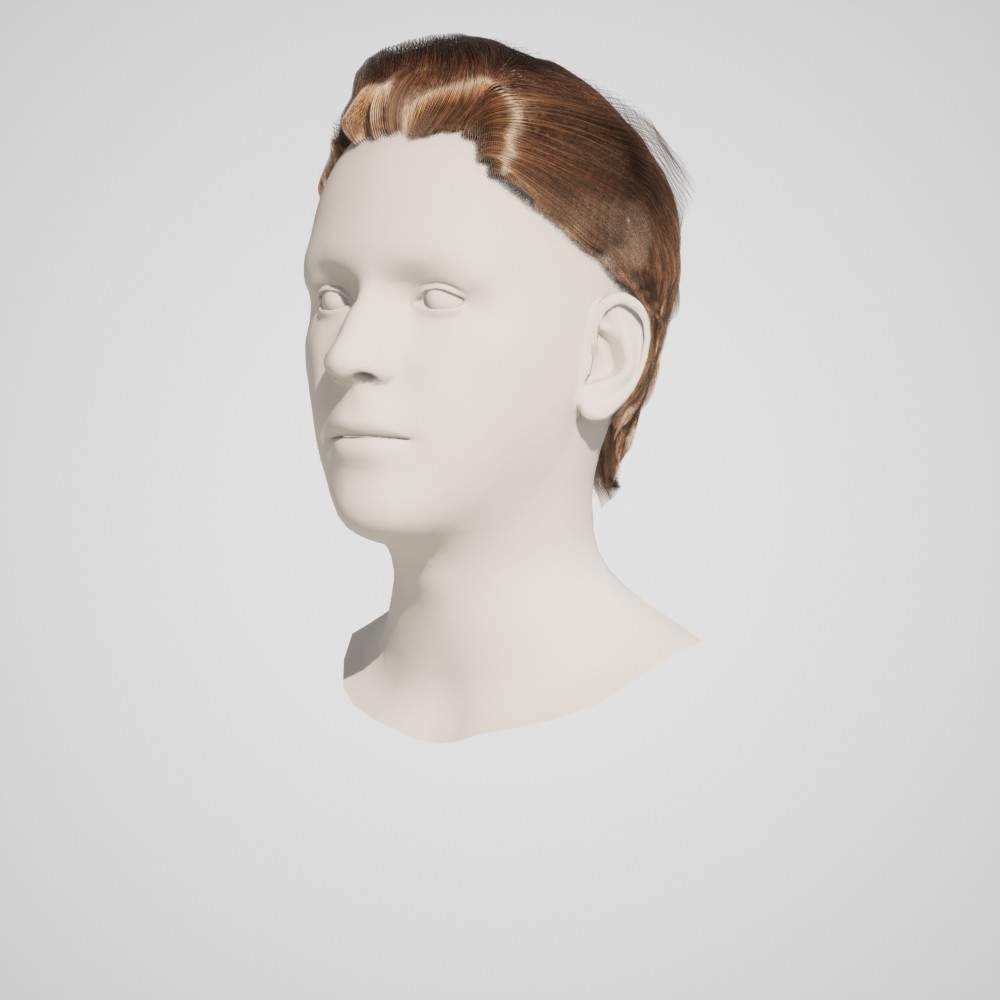} 
        &
        \multirow{2}{*}[0.481in]{\includegraphics[trim={160 280 160 40},clip,width=0.14\textwidth]{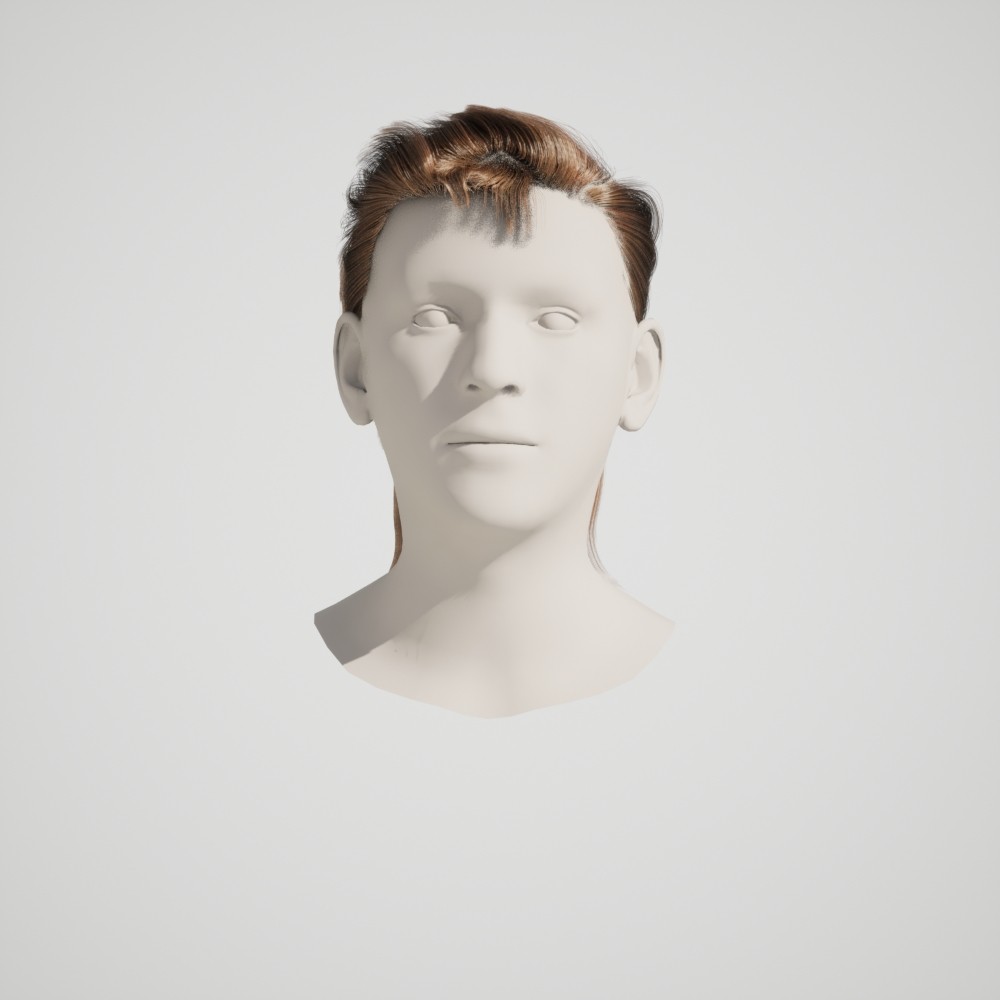}} 
        &
        \includegraphics[trim={100 160 100 40},clip,width=0.07\textwidth]{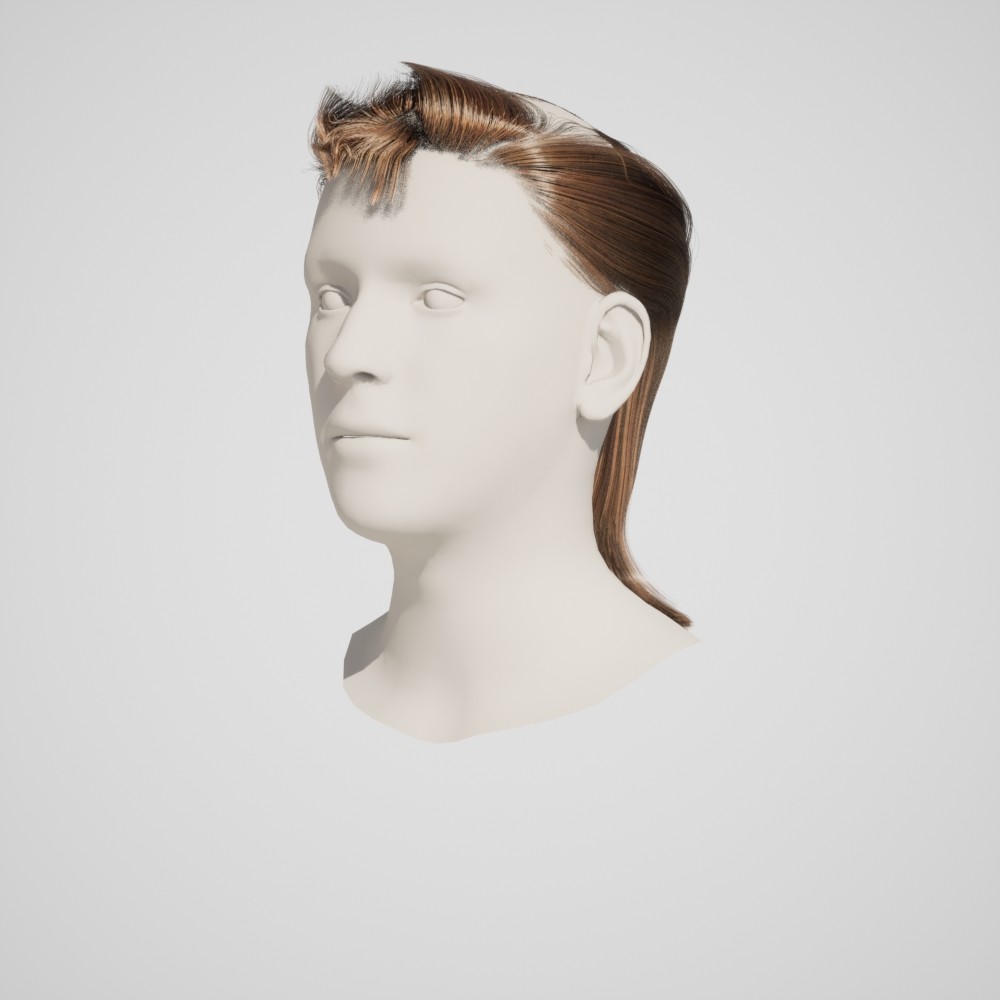} 
        &
        \multirow{2}{*}[0.481in]{\includegraphics[trim={160 280 160 40},clip,width=0.14\textwidth]{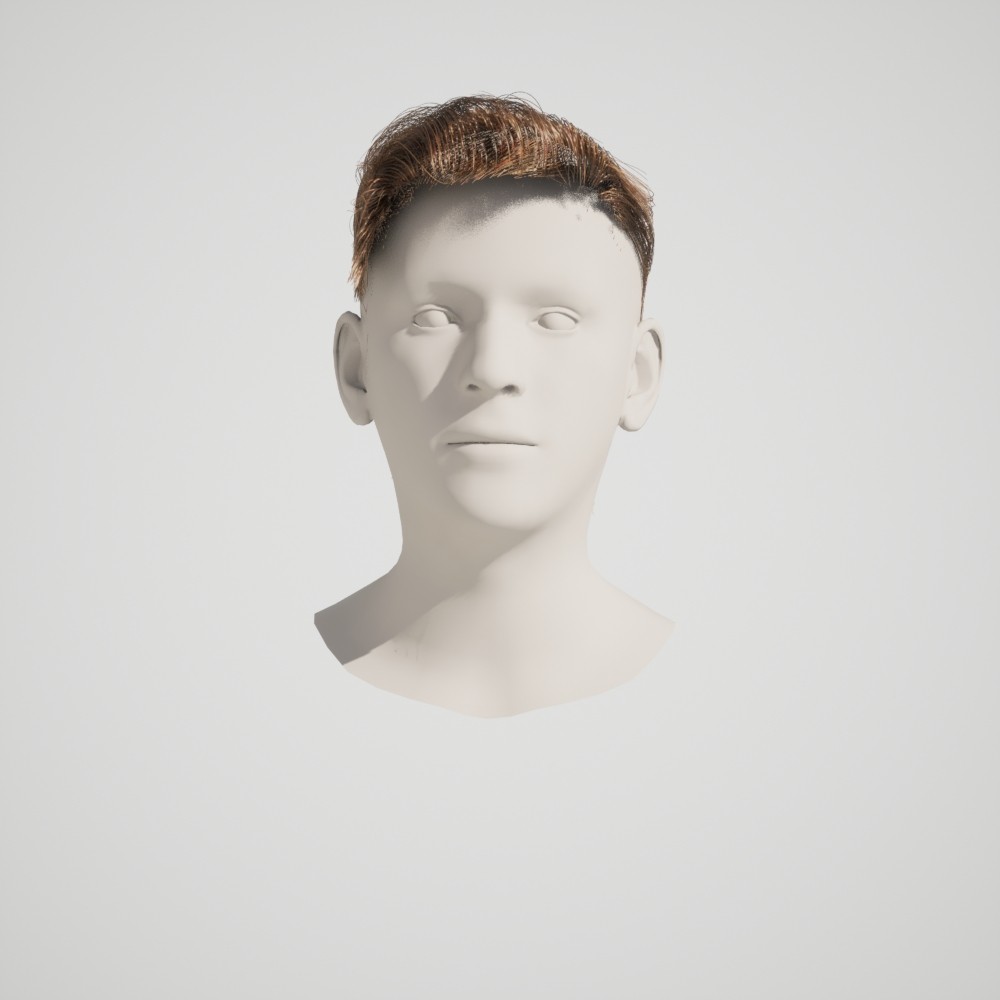}} 
        &
        \includegraphics[trim={100 160 100 40},clip,width=0.07\textwidth]{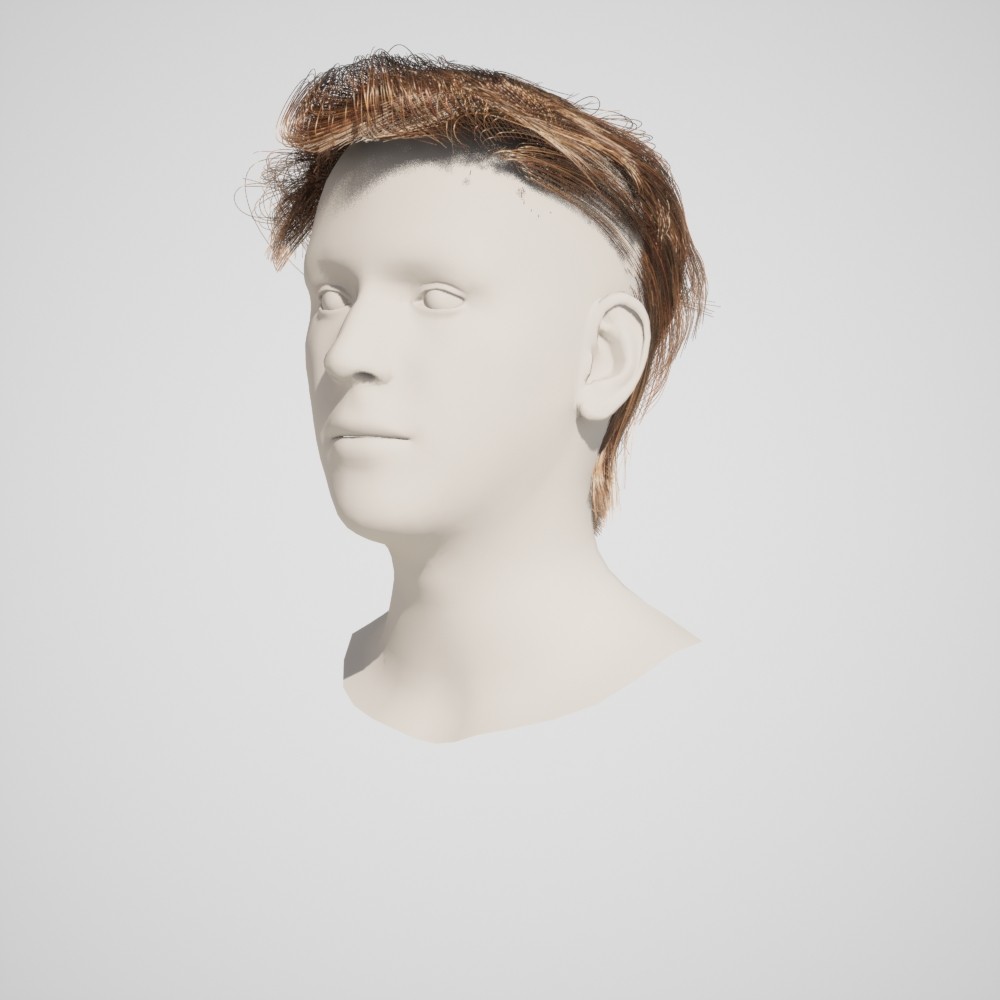} 
        &

        \multirow{2}{*}[0.481in]{\includegraphics[trim={160 280 160 40},clip,width=0.14\textwidth]{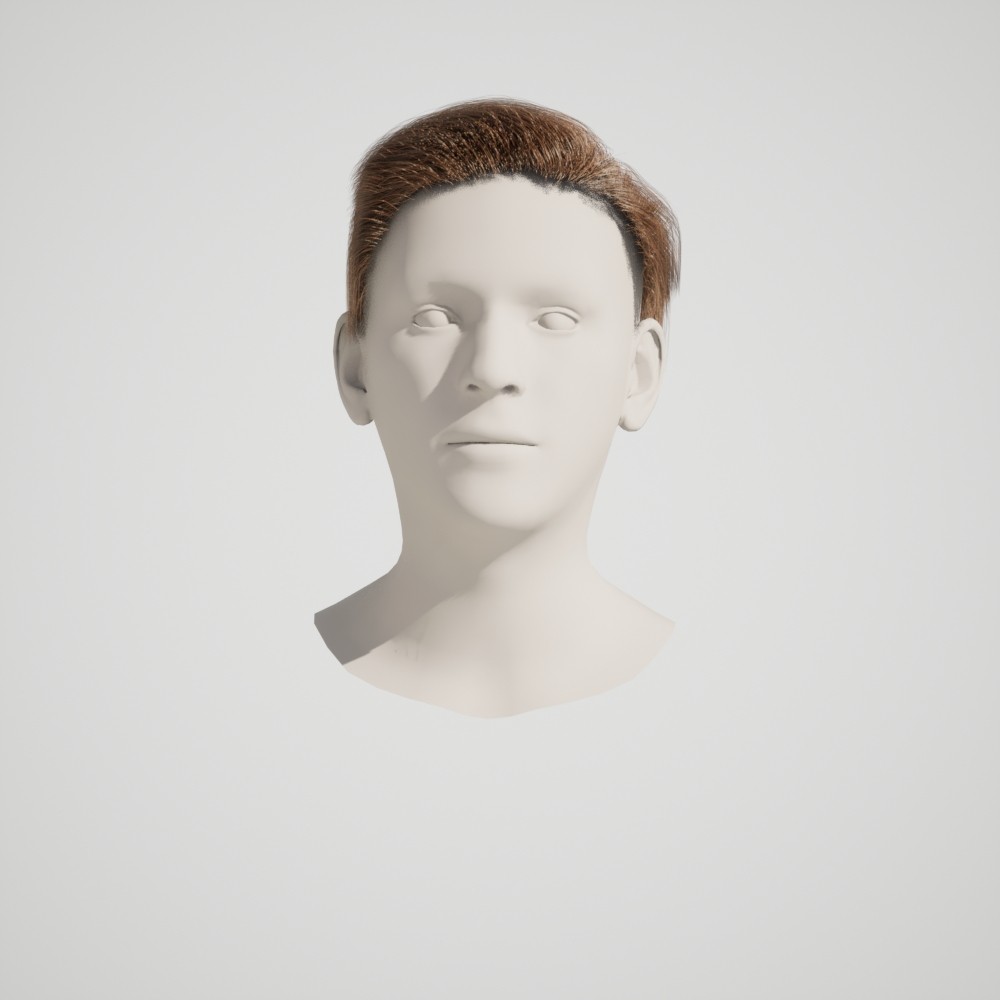}} 
        &
        \includegraphics[trim={100 160 100 40},clip,width=0.07\textwidth]{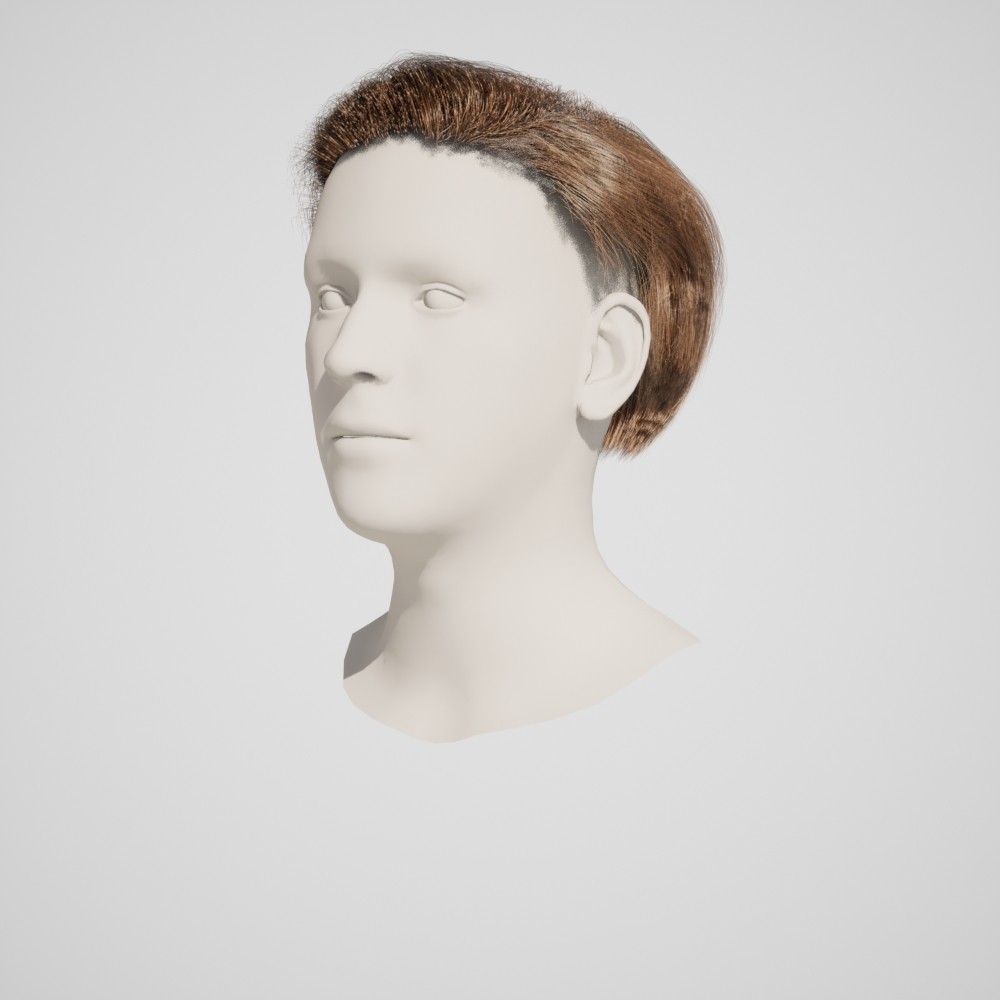}

        \\

        &
        &
        \includegraphics[trim={100 160 100 40},clip,width=0.07\textwidth]{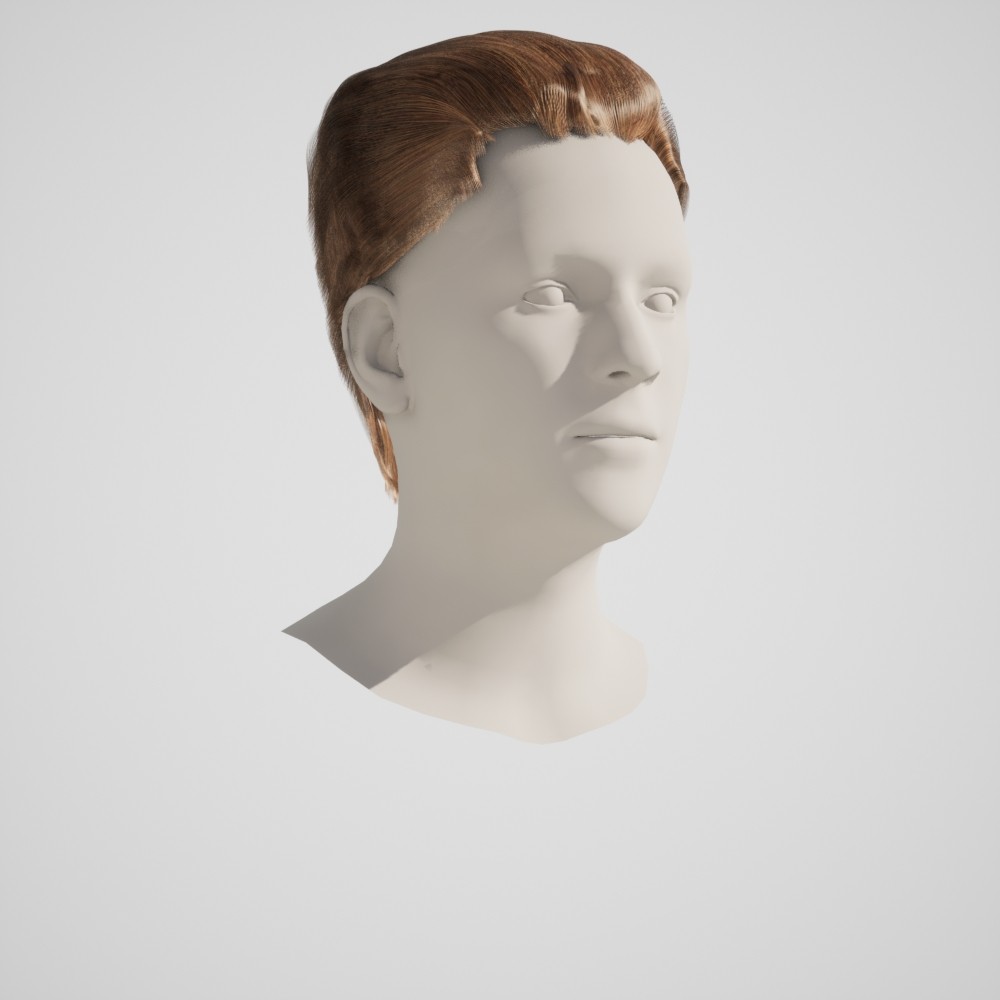}
        &
        &
        \includegraphics[trim={100 160 100 40},clip,width=0.07\textwidth]{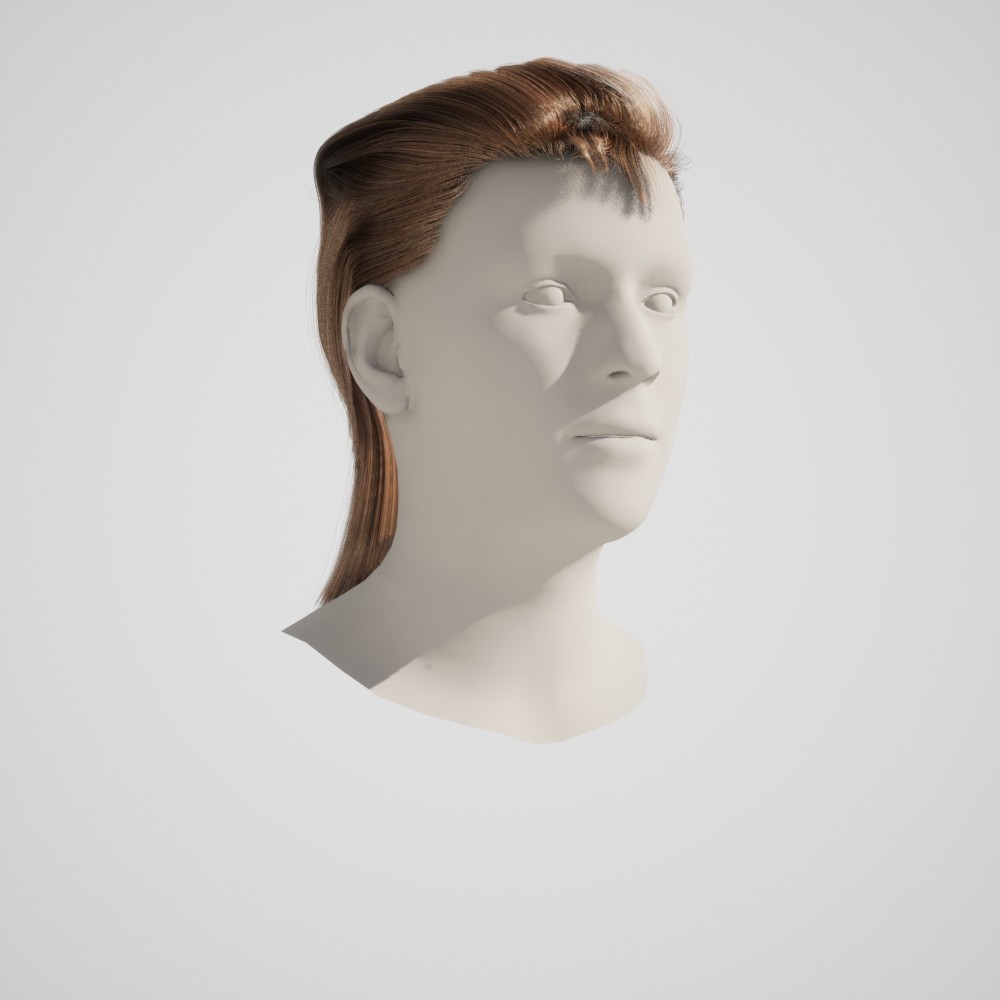}
        &
        &
        \includegraphics[trim={100 160 100 40},clip,width=0.07\textwidth]{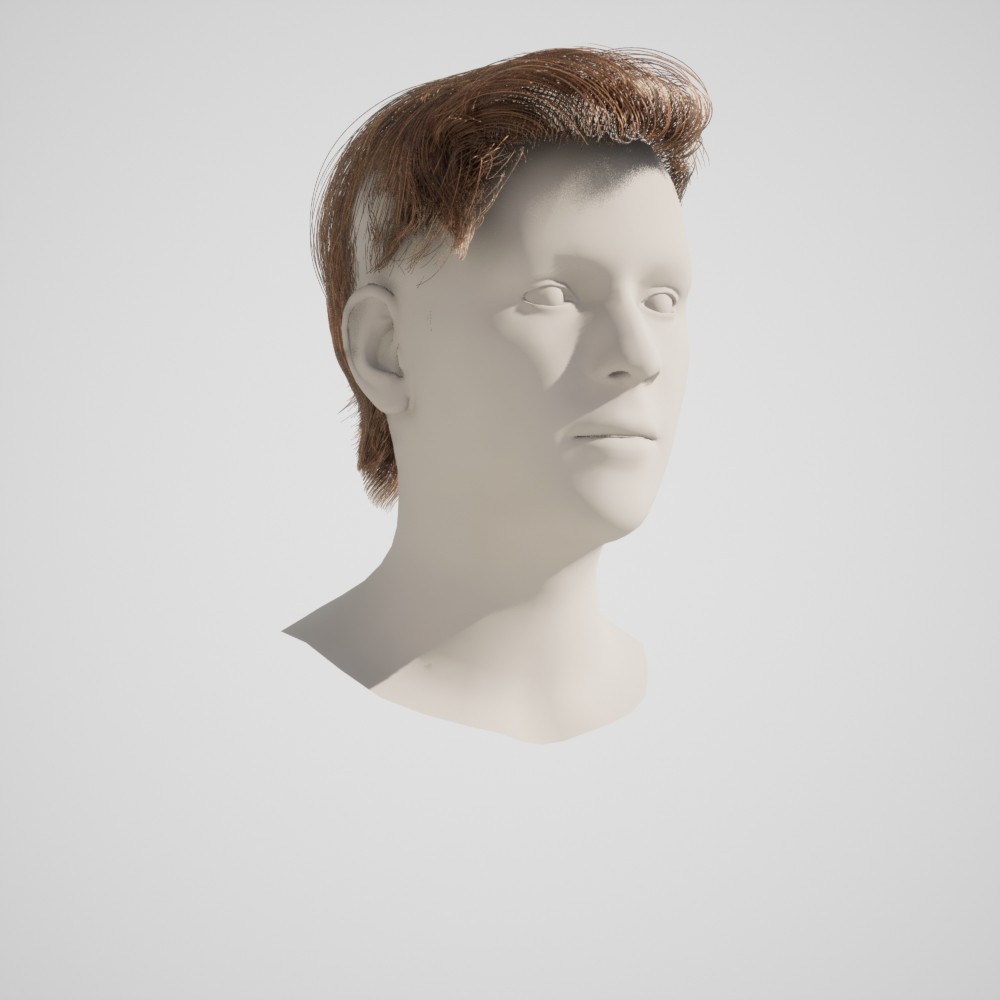}

        & 
        &
        \includegraphics[trim={100 160 100 40},clip,width=0.07\textwidth]{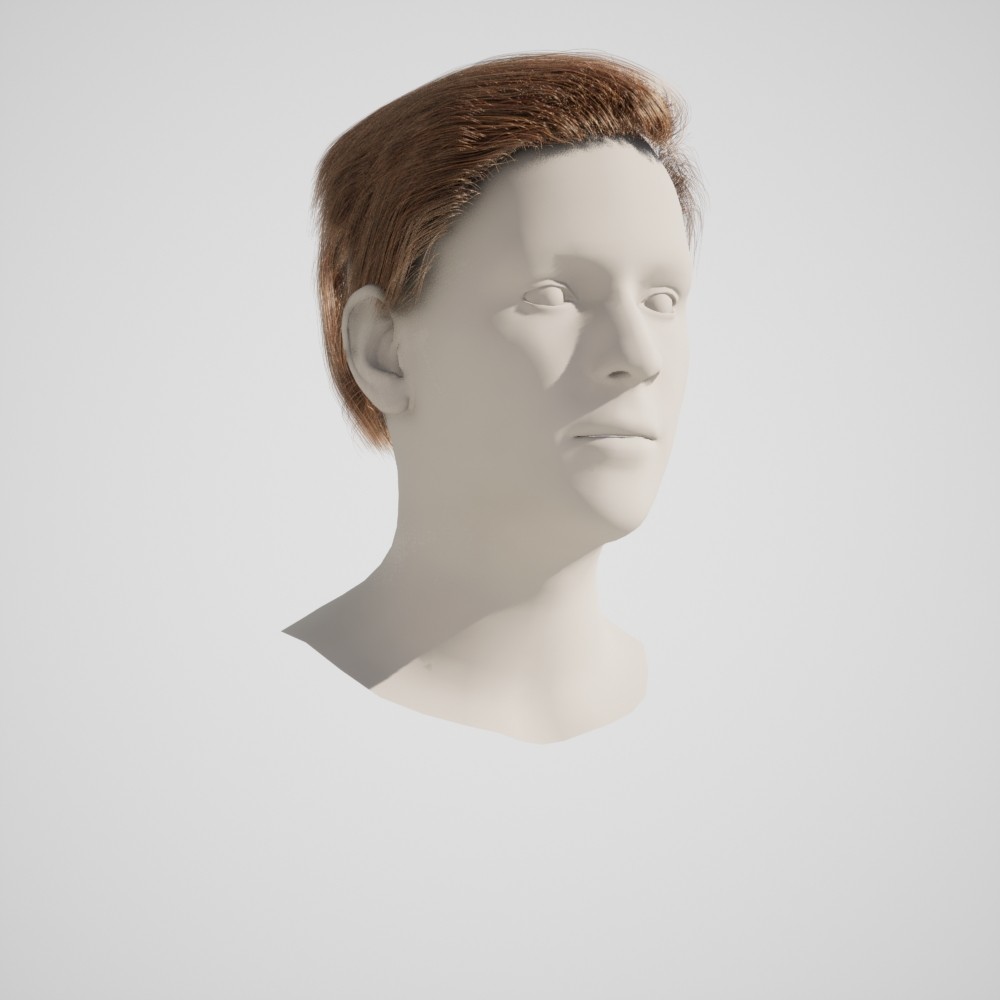}

        \\

        \multirow{2}{*}[0.481in]{\includegraphics[width=0.14\textwidth]{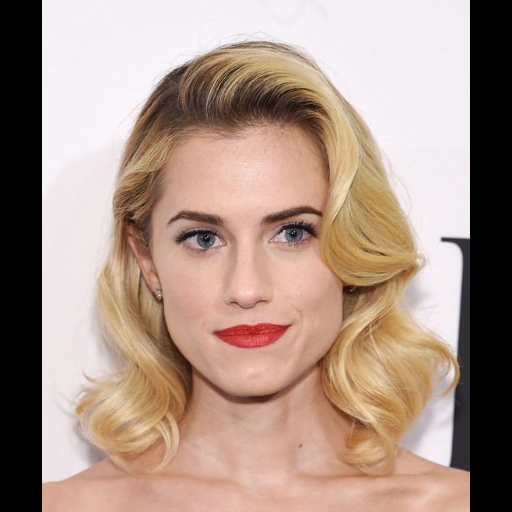}} 
            &
        \multirow{2}{*}[0.481in]{\includegraphics[trim={120 200 120 40},clip,width=0.14\textwidth]{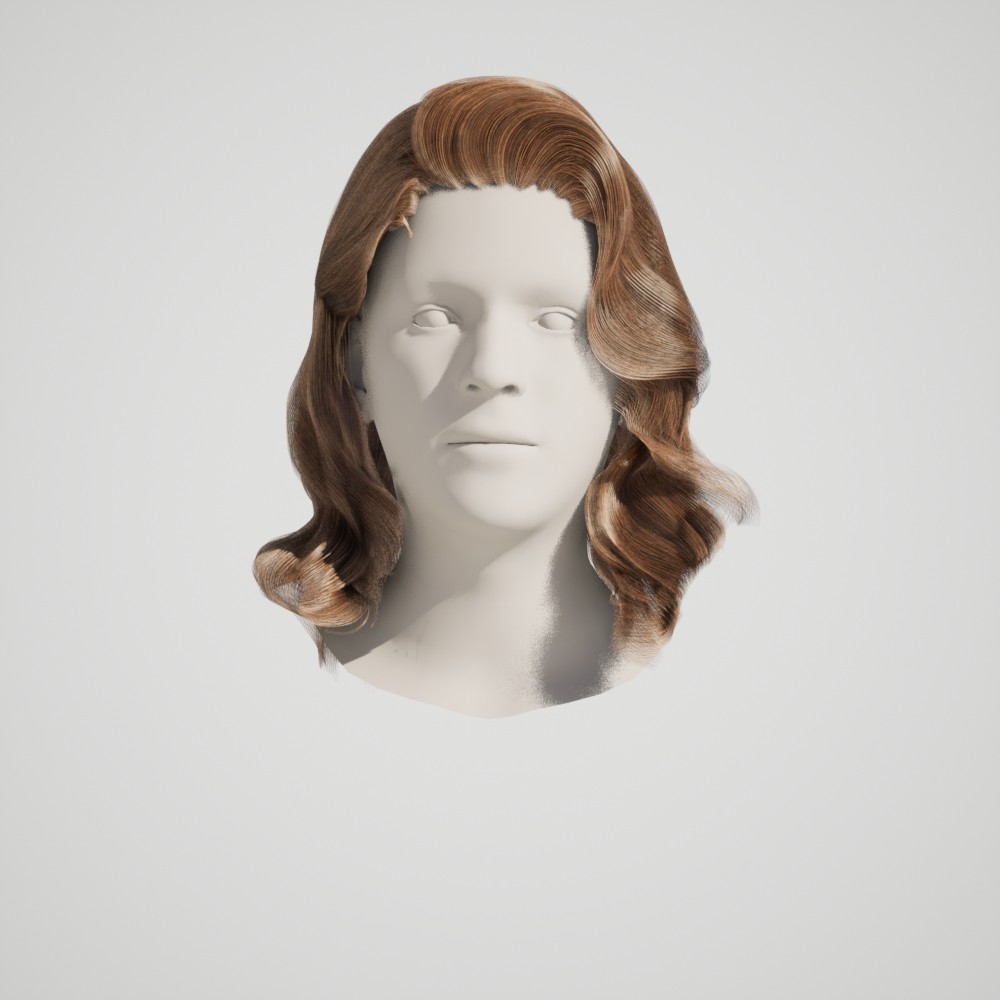}} 
        &
        \includegraphics[width=0.07\textwidth]{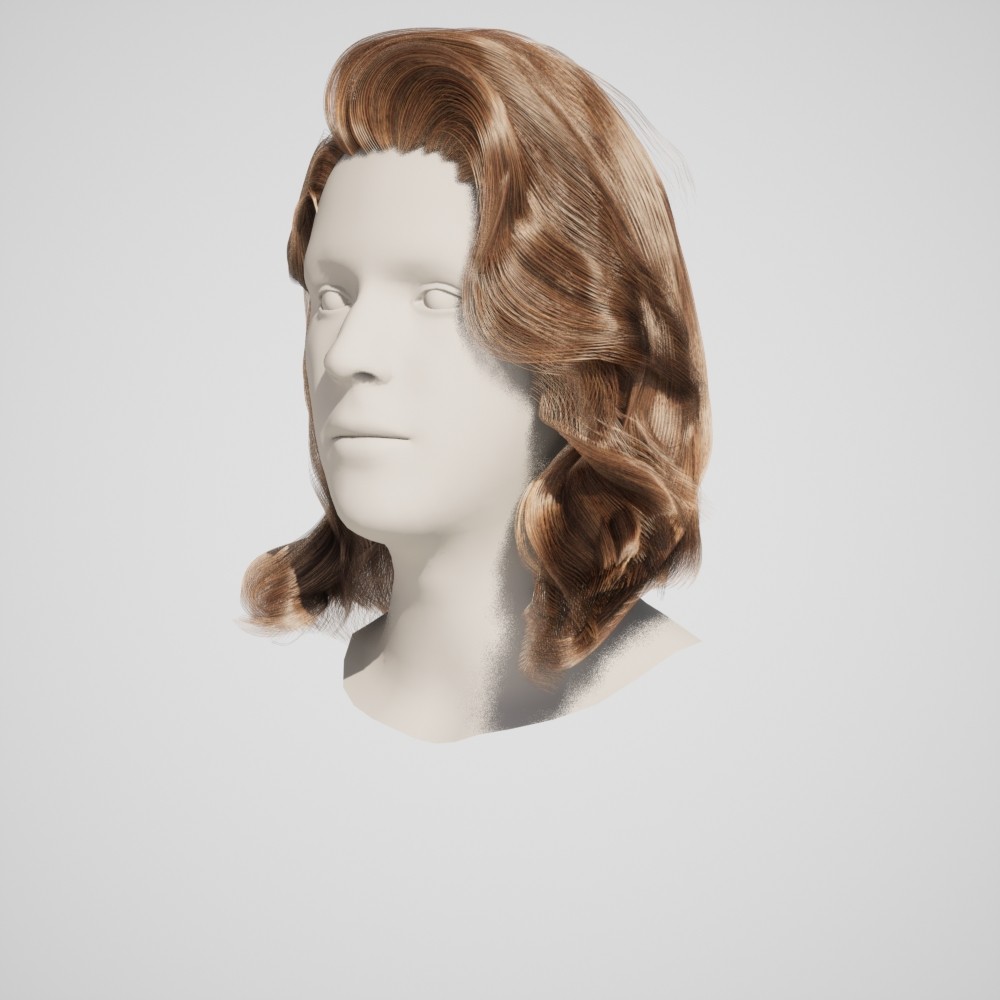} 
        &
        \multirow{2}{*}[0.481in]{\includegraphics[trim={120 200 120 40},clip,width=0.14\textwidth]{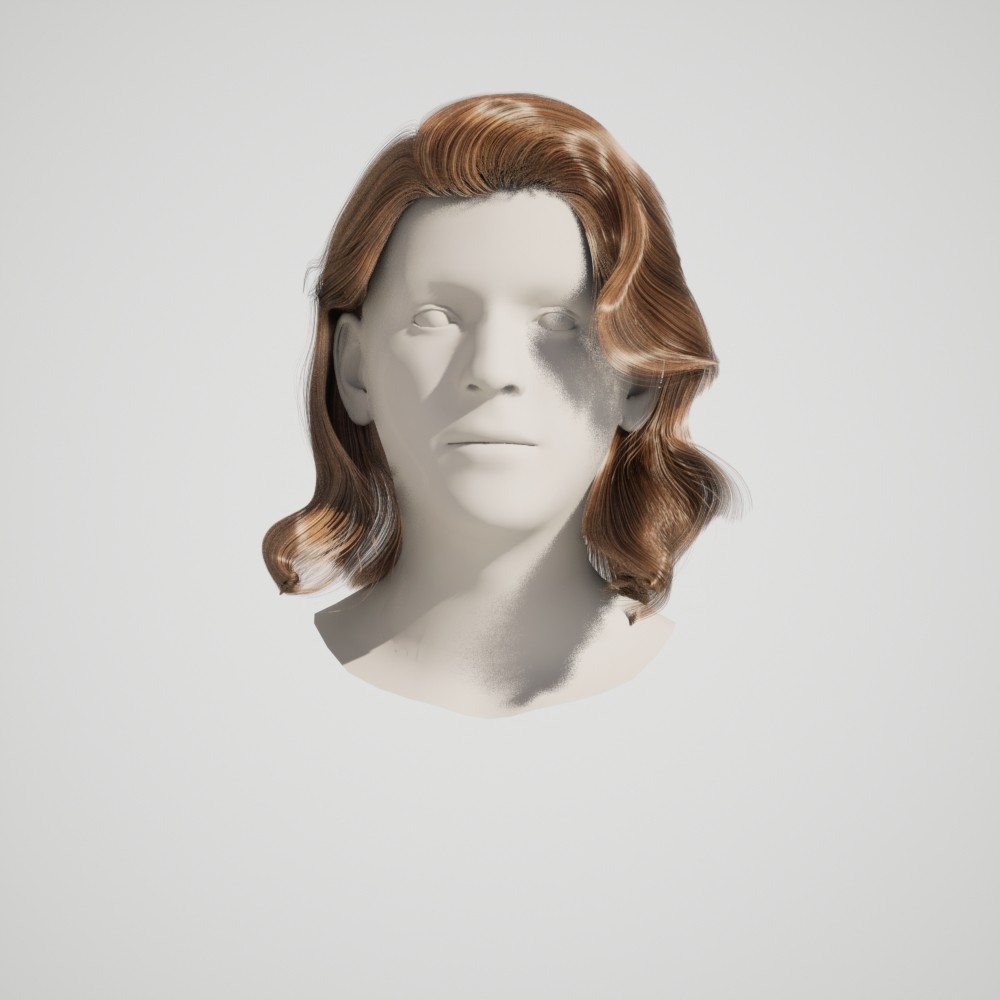}} 
        &
        \includegraphics[width=0.07\textwidth]{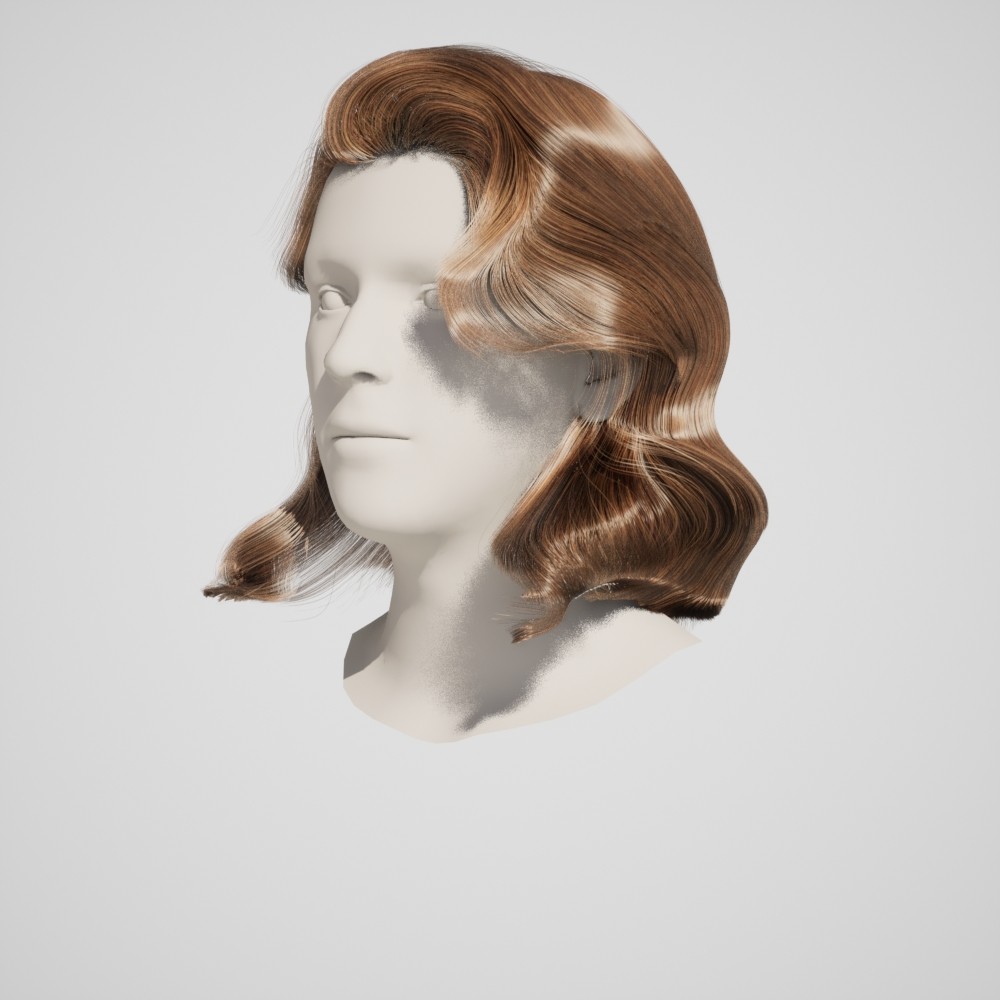} 
        &
        \multirow{2}{*}[0.481in]{\includegraphics[trim={120 200 120 40},clip,width=0.14\textwidth]{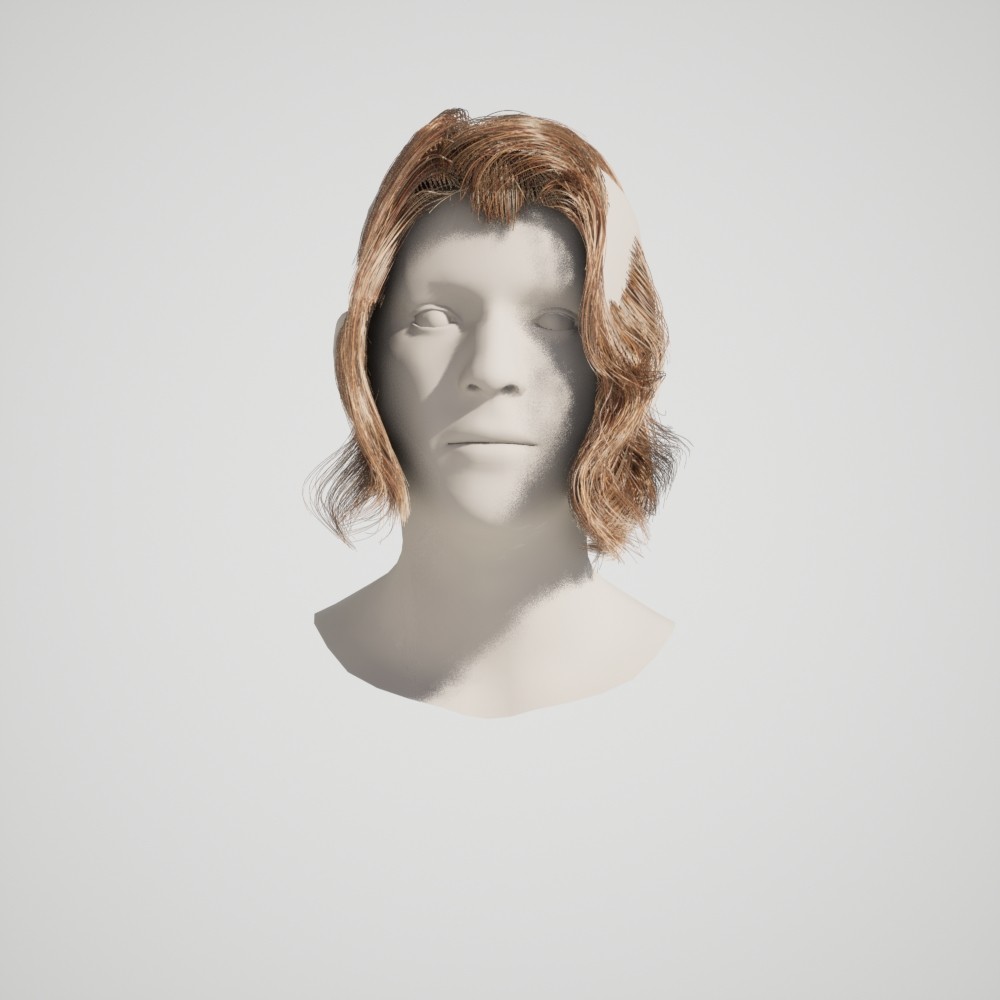}} 
        &
        \includegraphics[width=0.07\textwidth]{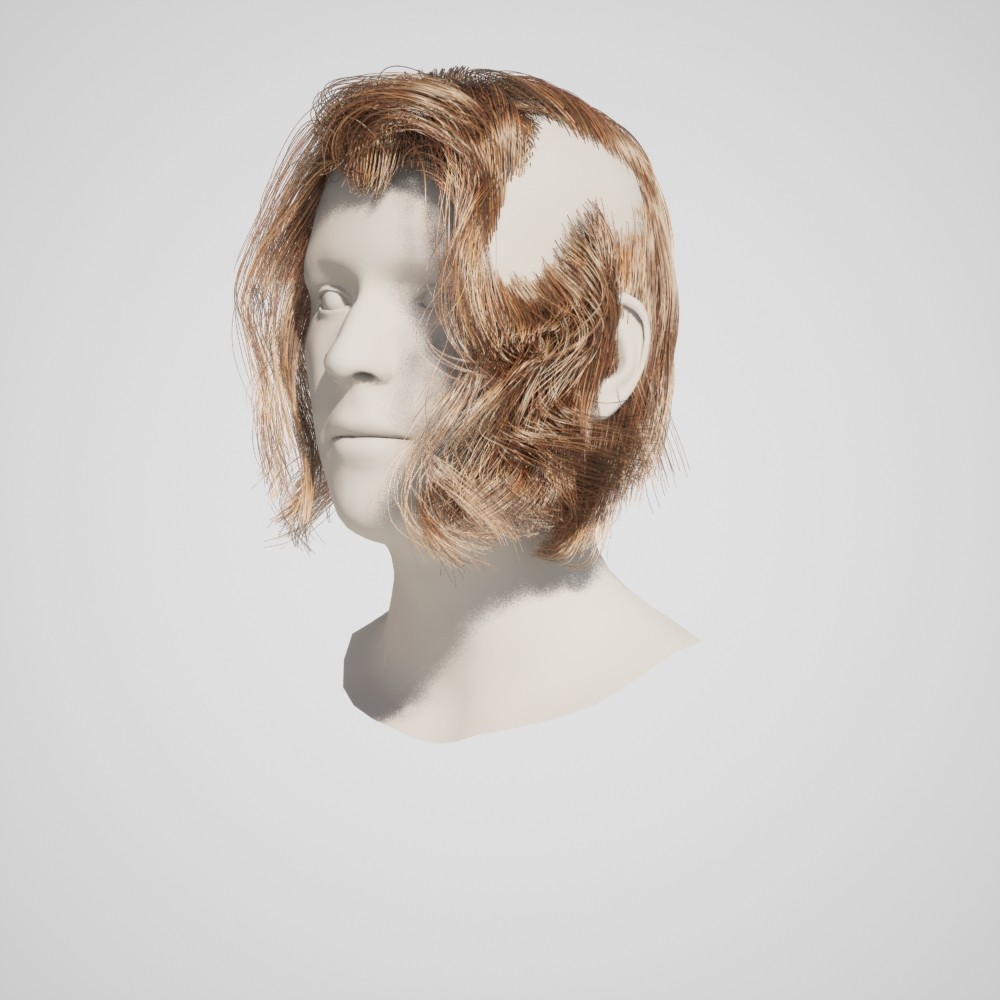} 
        &

        \multirow{2}{*}[0.481in]{\includegraphics[trim={120 200 120 40},clip,width=0.14\textwidth]{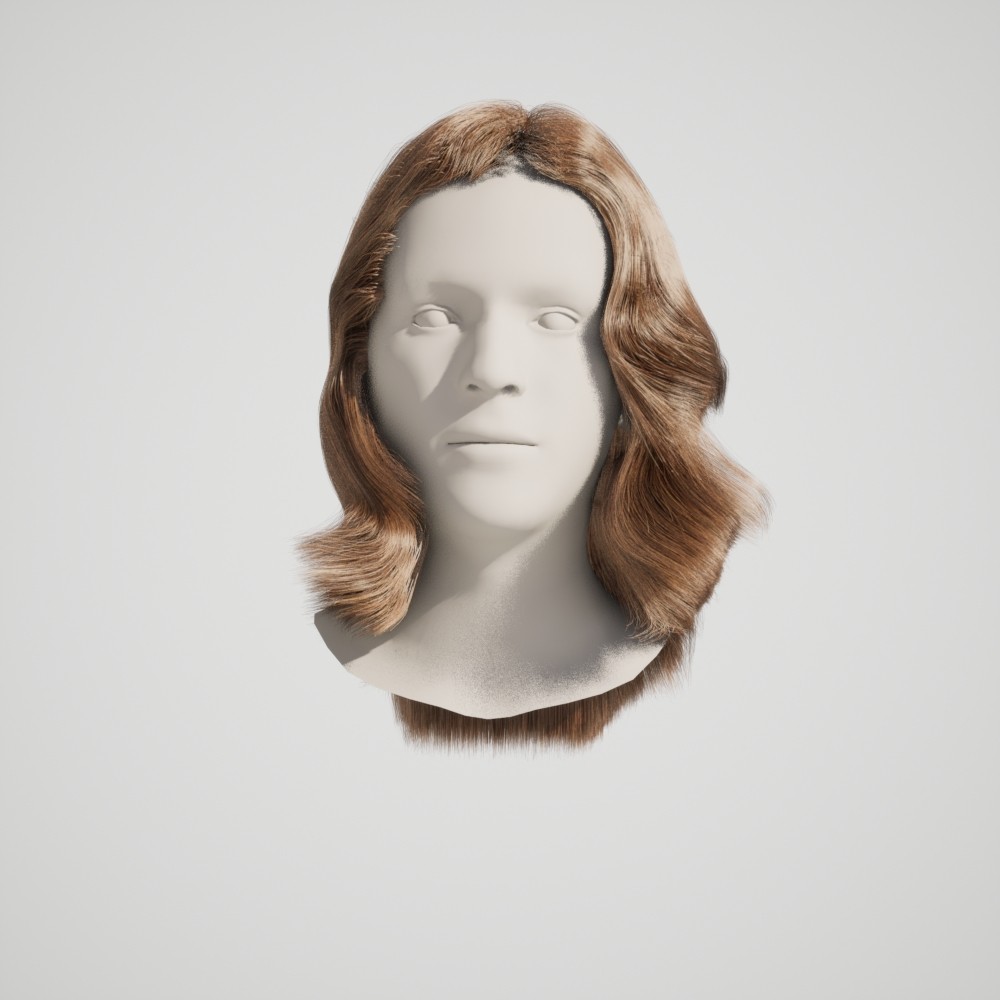}} 
        &
        \includegraphics[width=0.07\textwidth]{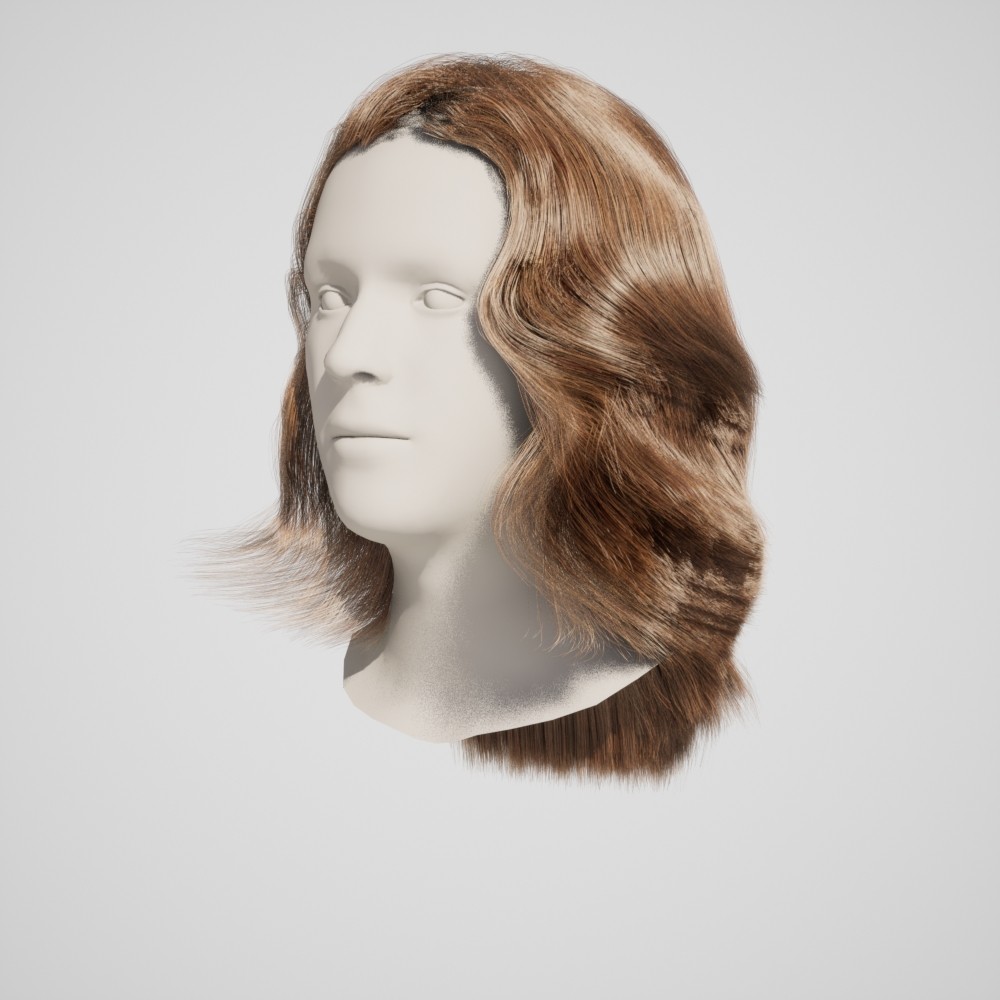}

        \\

        &
        &
        \includegraphics[width=0.07\textwidth]{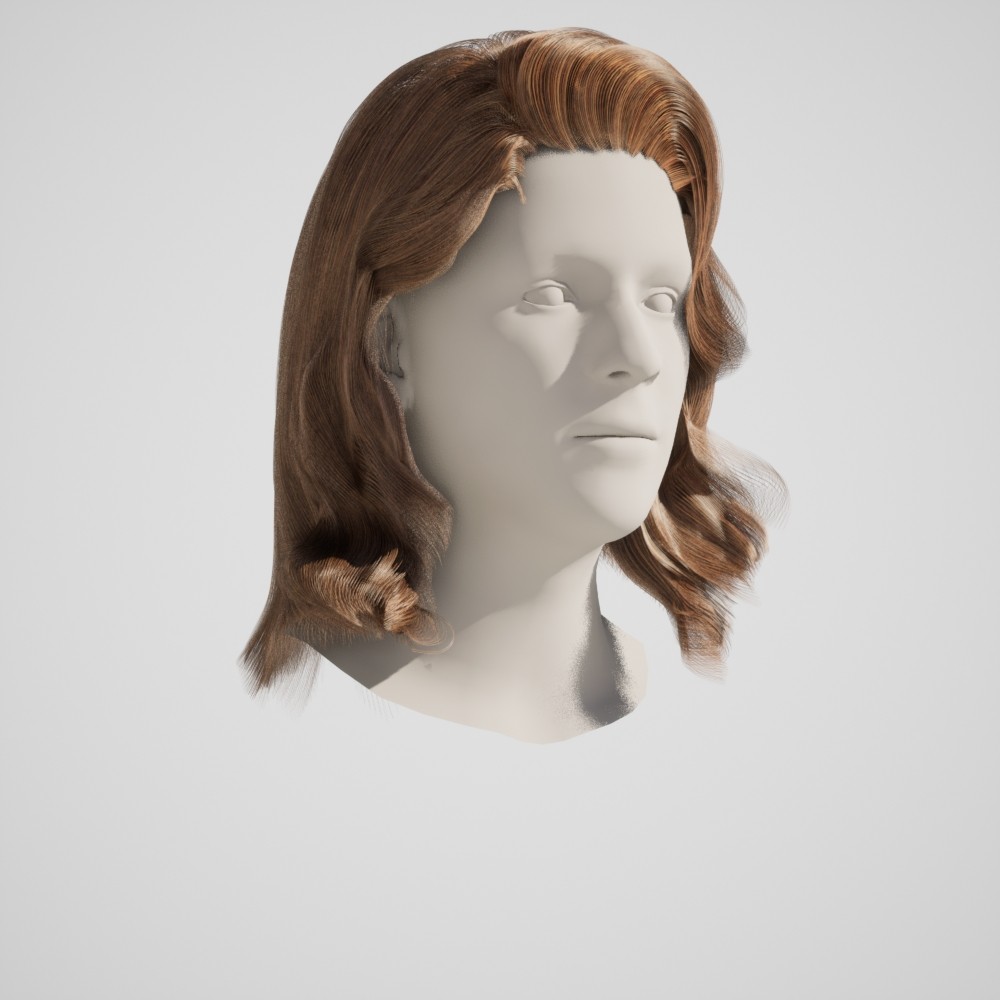}
        &
        &
        \includegraphics[width=0.07\textwidth]{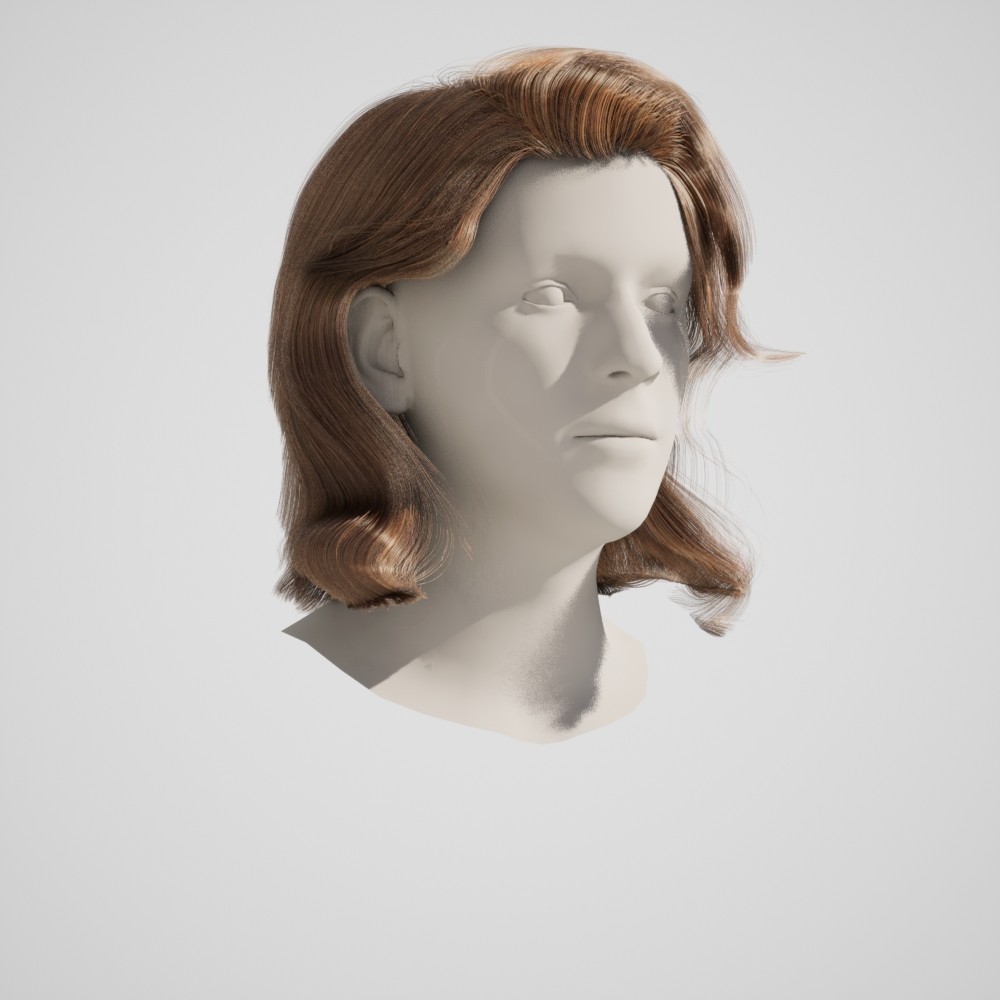}
        &
        &
        \includegraphics[width=0.07\textwidth]{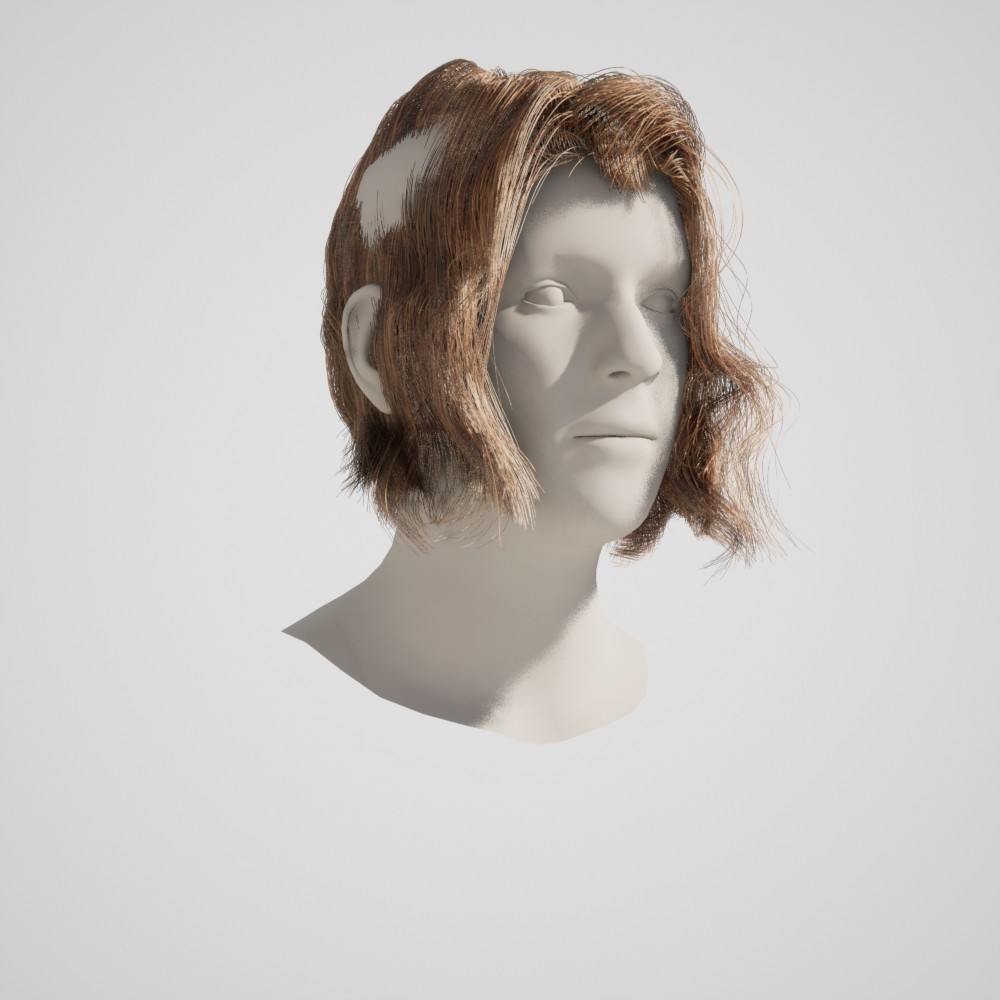}

        & 
        &
        \includegraphics[width=0.07\textwidth]{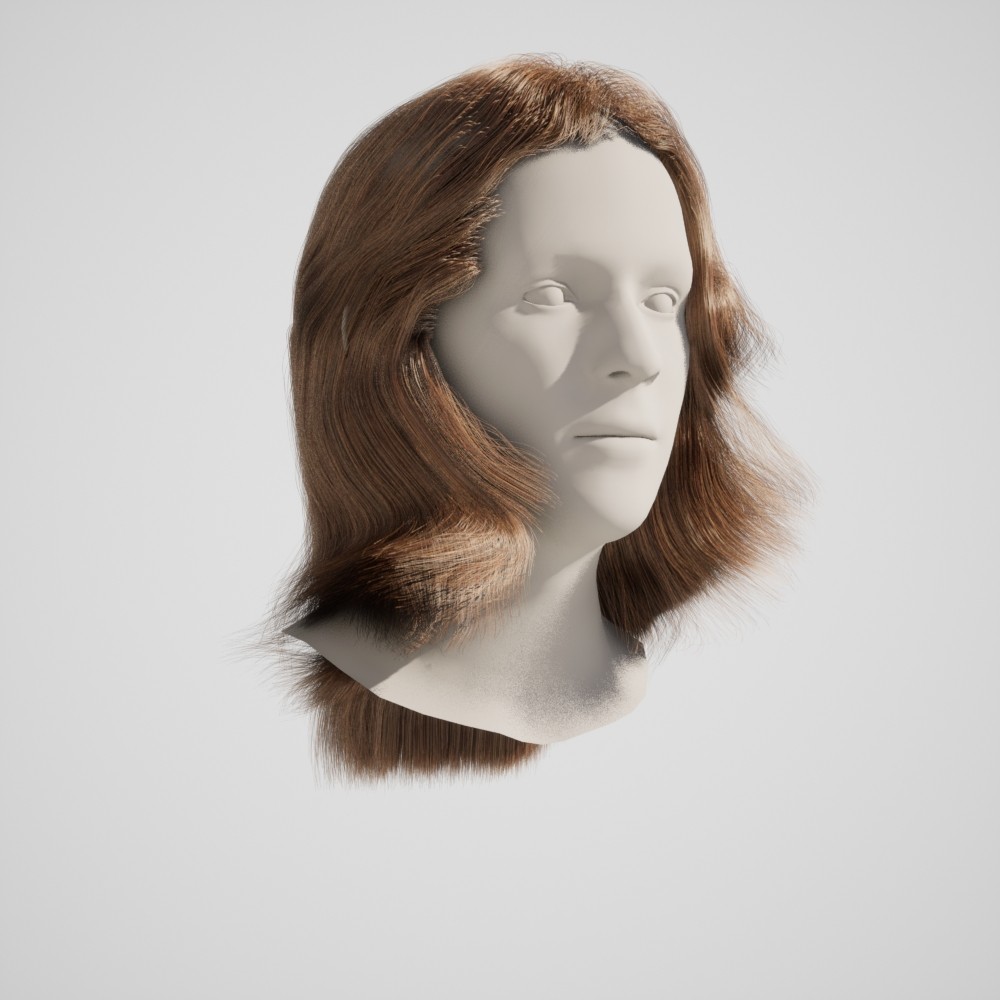} 

                \\

        \multirow{2}{*}[0.481in]{\includegraphics[width=0.14\textwidth]{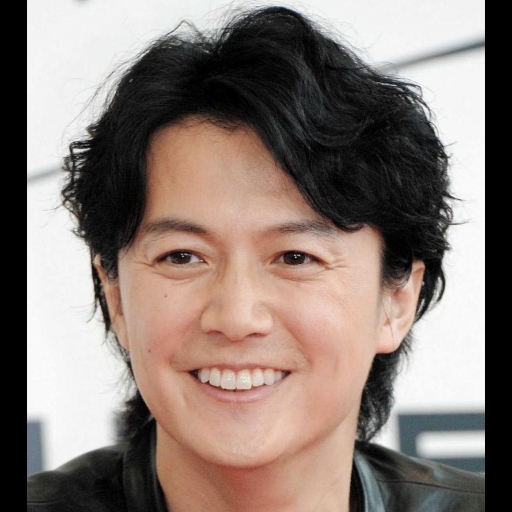}} 
            &
        \multirow{2}{*}[0.481in]{\includegraphics[trim={160 280 160 40},clip,width=0.14\textwidth]{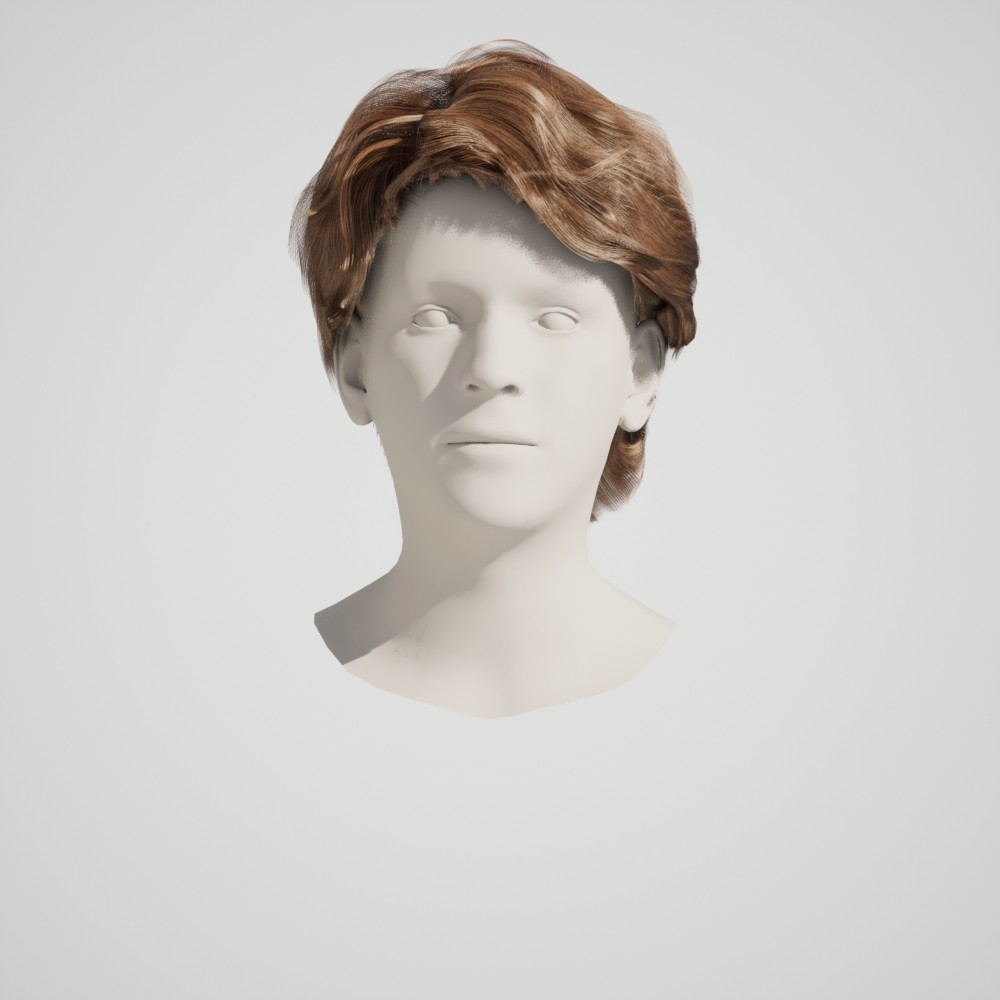}} 
        &
        \includegraphics[trim={100 200 100 0},clip,width=0.07\textwidth]{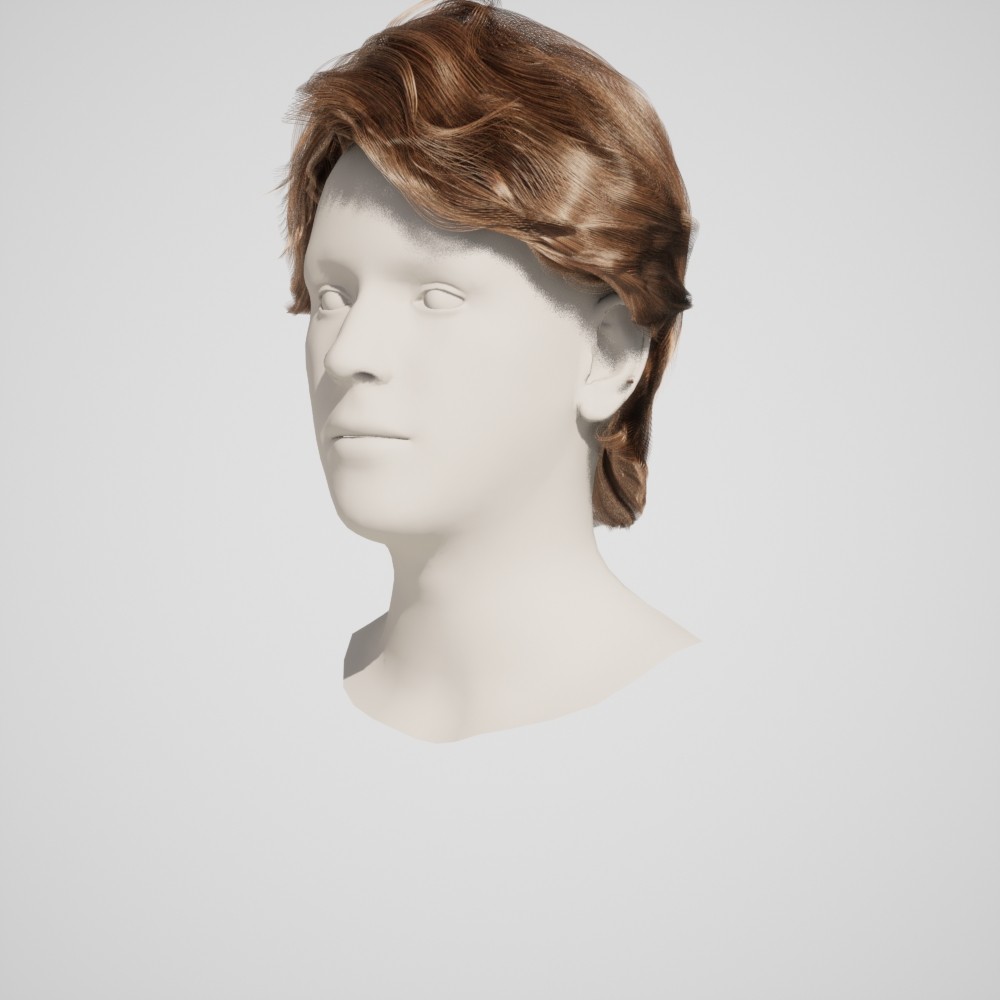} 
        &
        \multirow{2}{*}[0.481in]{\includegraphics[trim={160 280 160 40},clip,width=0.14\textwidth]{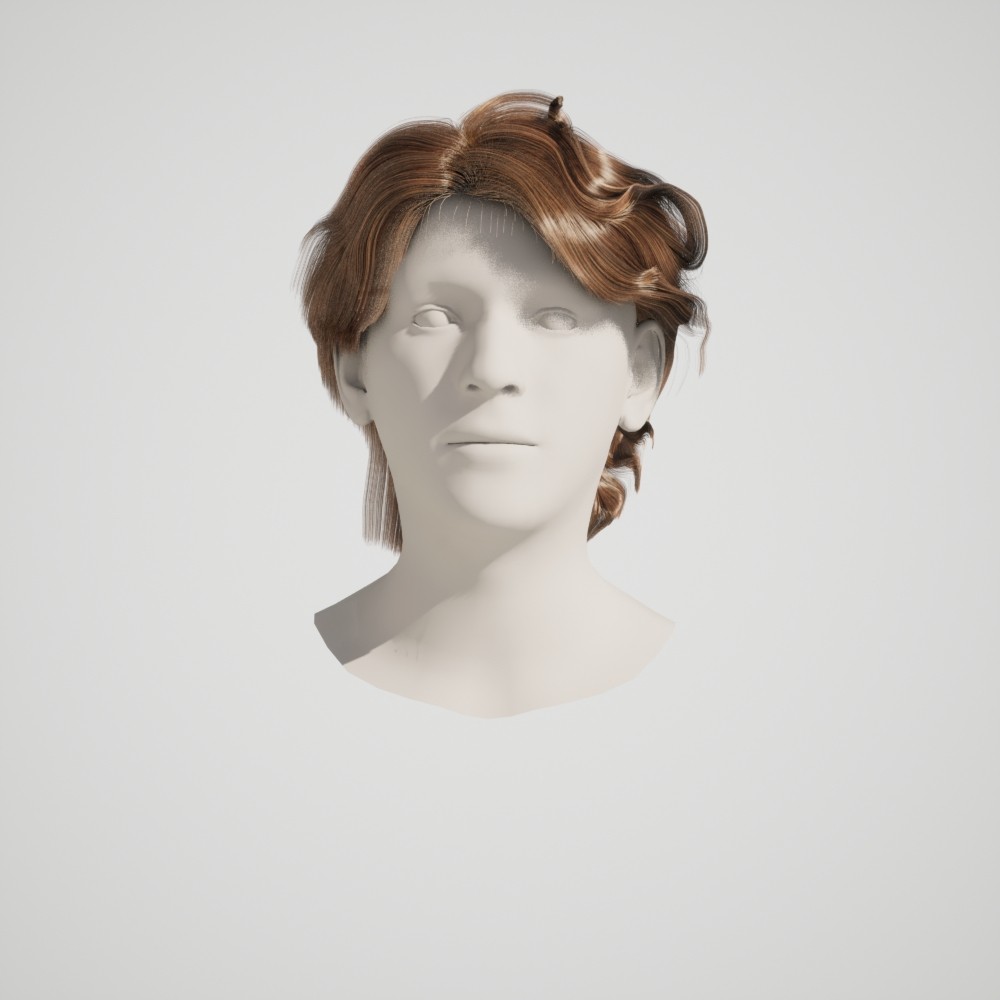}} 
        &
        \includegraphics[trim={100 200 100 0},clip,width=0.07\textwidth]{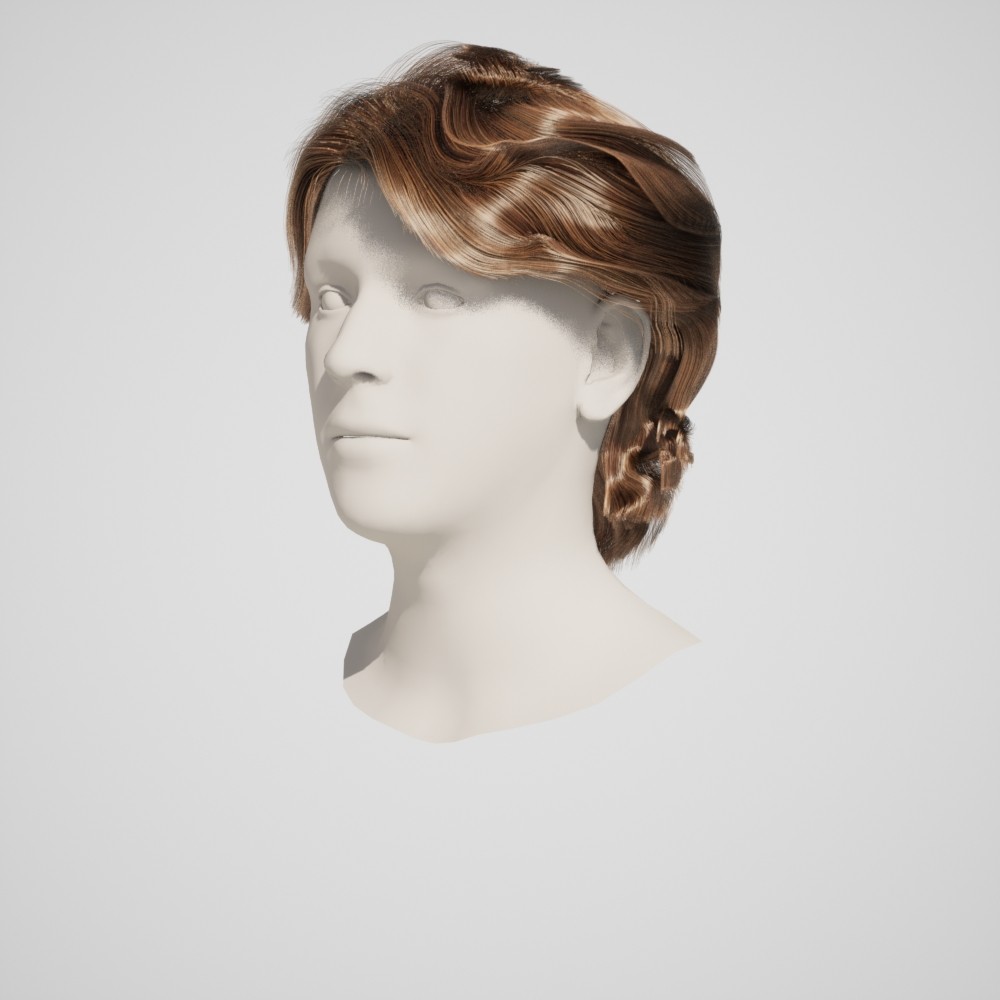} 
        &
        \multirow{2}{*}[0.481in]{\includegraphics[trim={160 280 160 40},clip,width=0.14\textwidth]{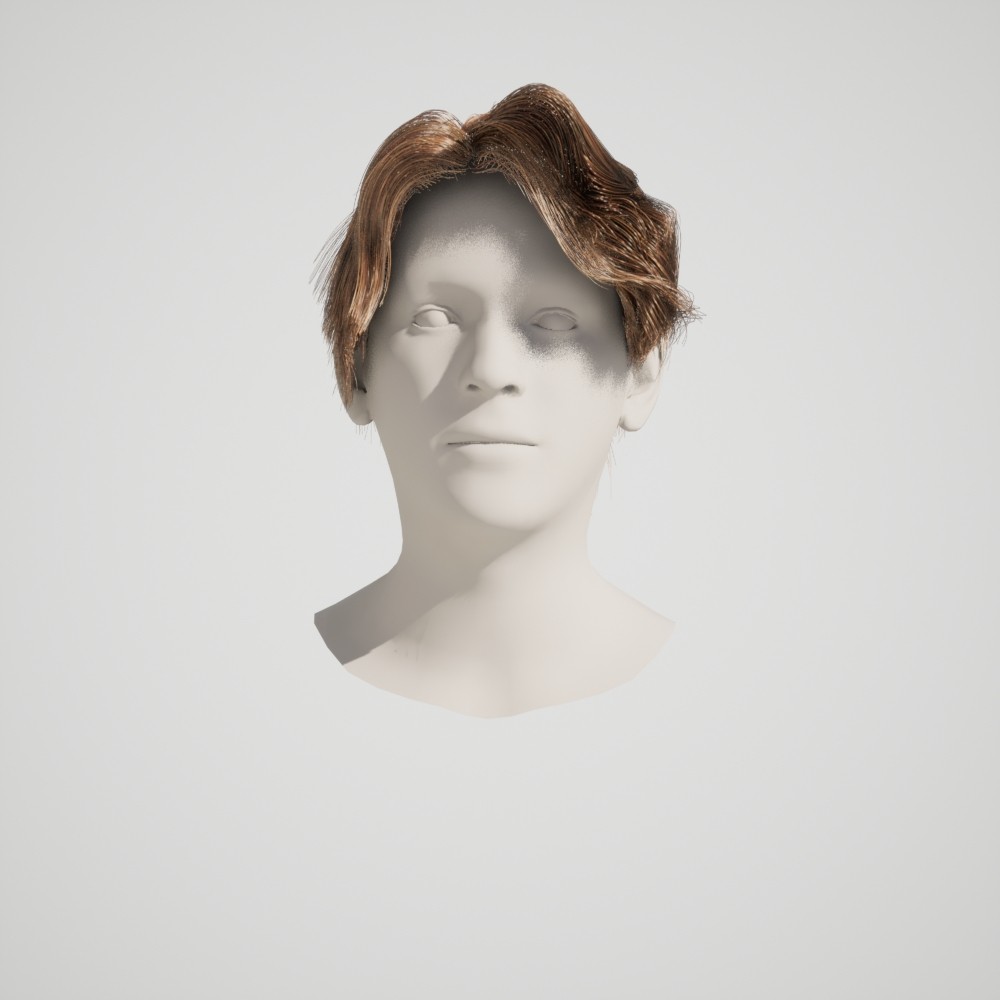}} 
        &
        \includegraphics[trim={100 200 100 0},clip,width=0.07\textwidth]{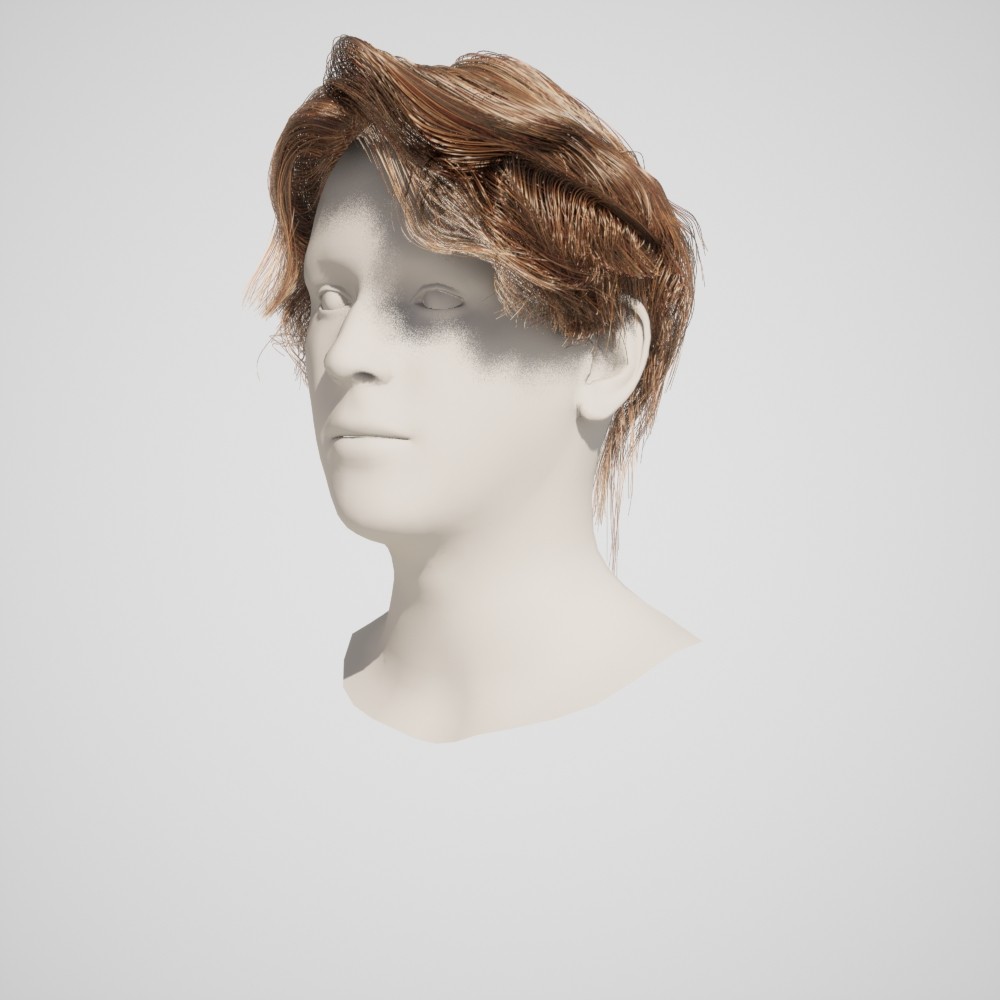} 
        &

        \multirow{2}{*}[0.481in]{\includegraphics[trim={160 280 160 40},clip,width=0.14\textwidth]{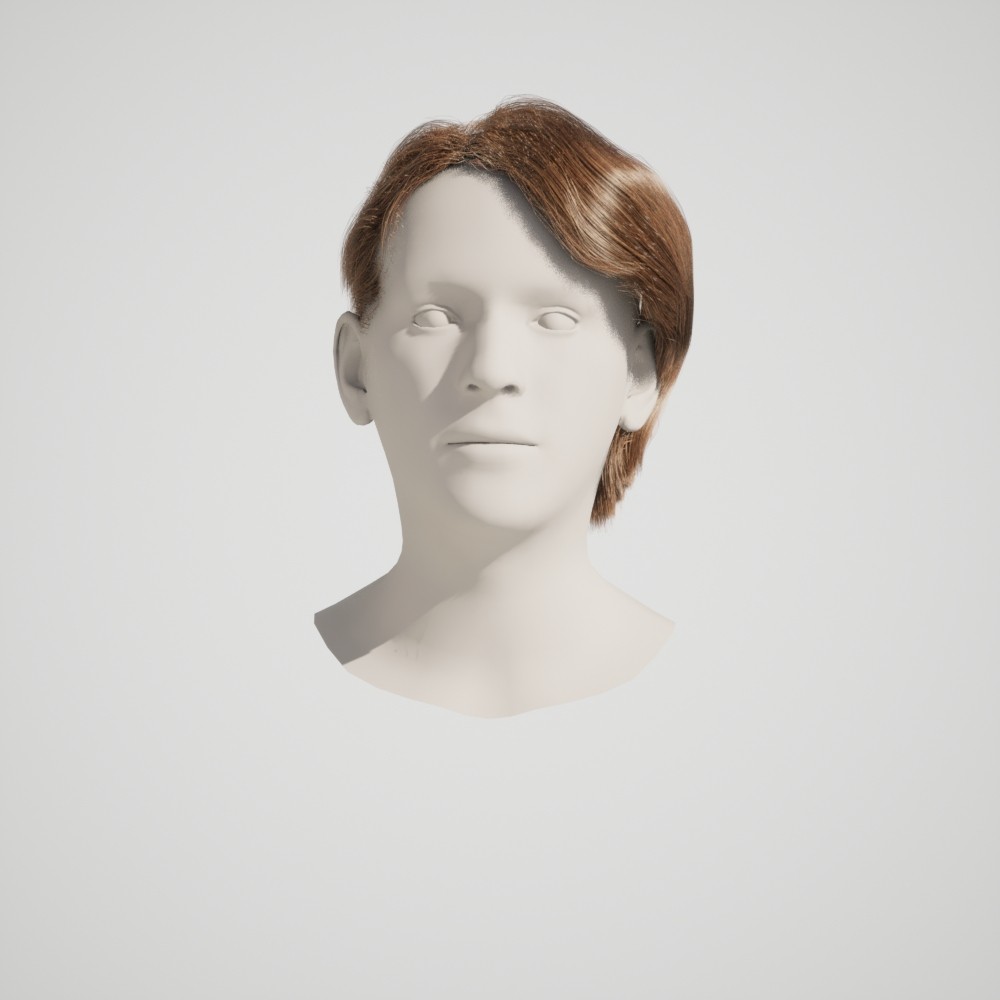}}
        &
        \includegraphics[trim={100 200 100 0},clip,width=0.07\textwidth]{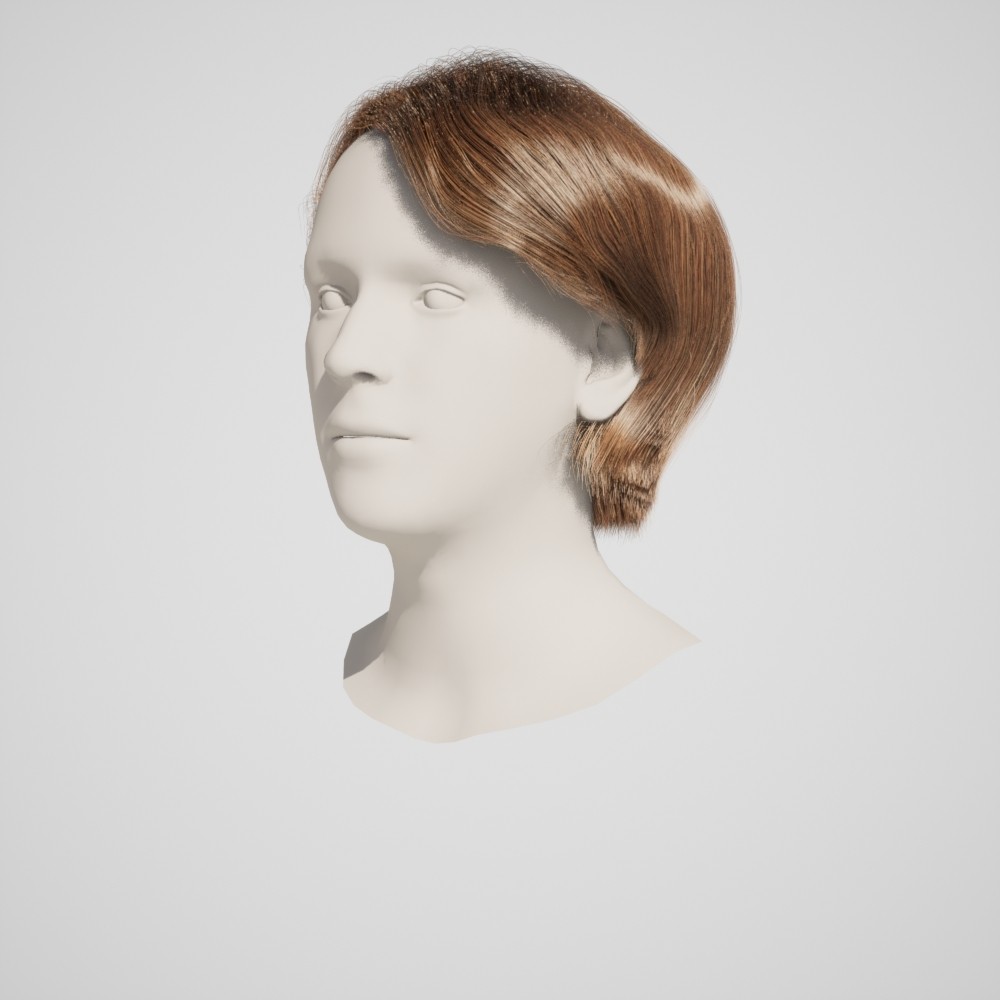}

        \\

        &
        &
        \includegraphics[trim={100 200 100 0},clip,width=0.07\textwidth]{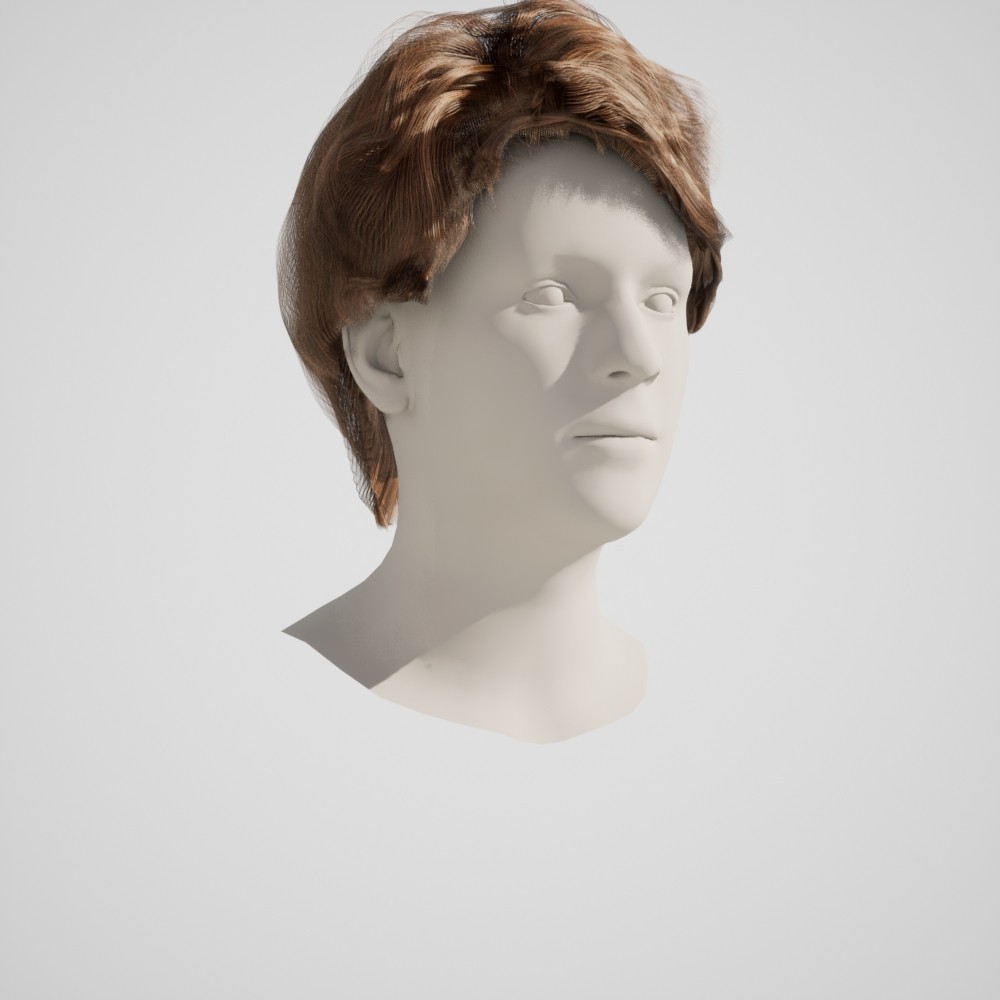}
        &
        &
        \includegraphics[trim={100 200 100 0},clip,width=0.07\textwidth]{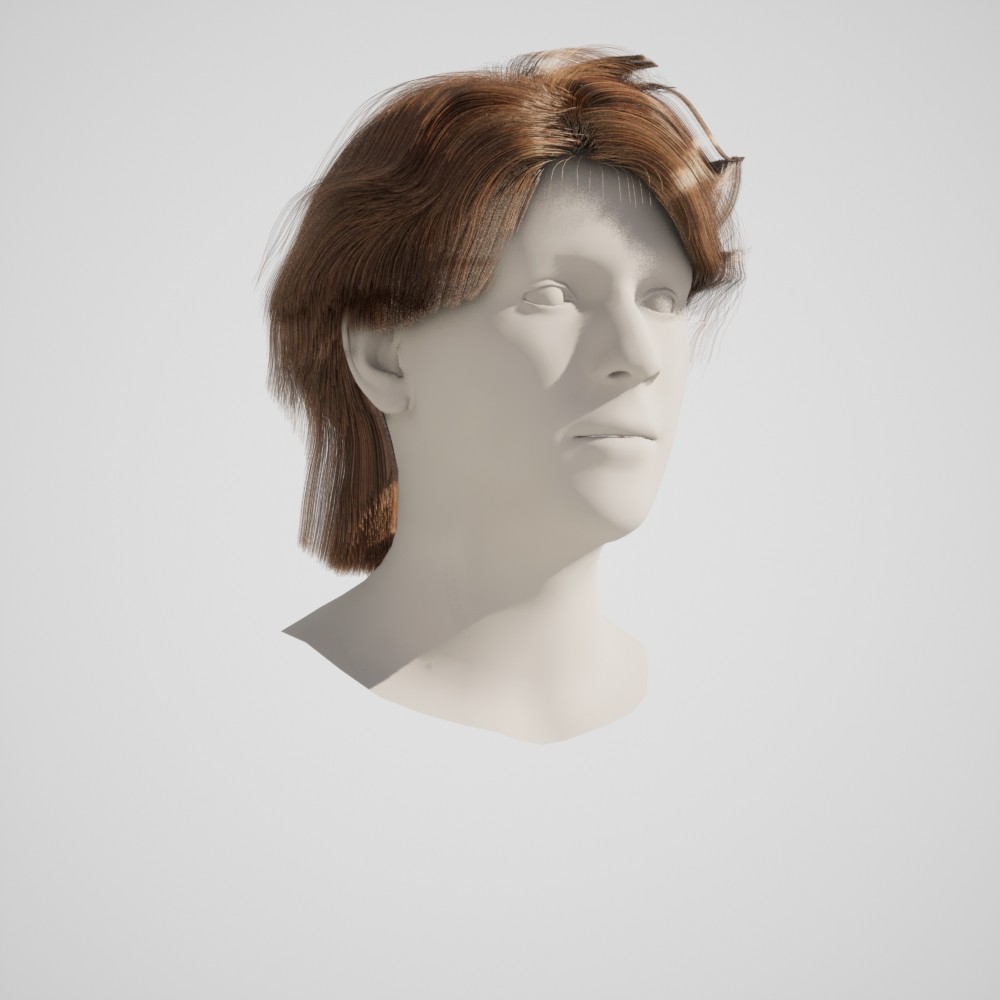}
        &
        &
        \includegraphics[trim={100 200 100 0},clip,width=0.07\textwidth]{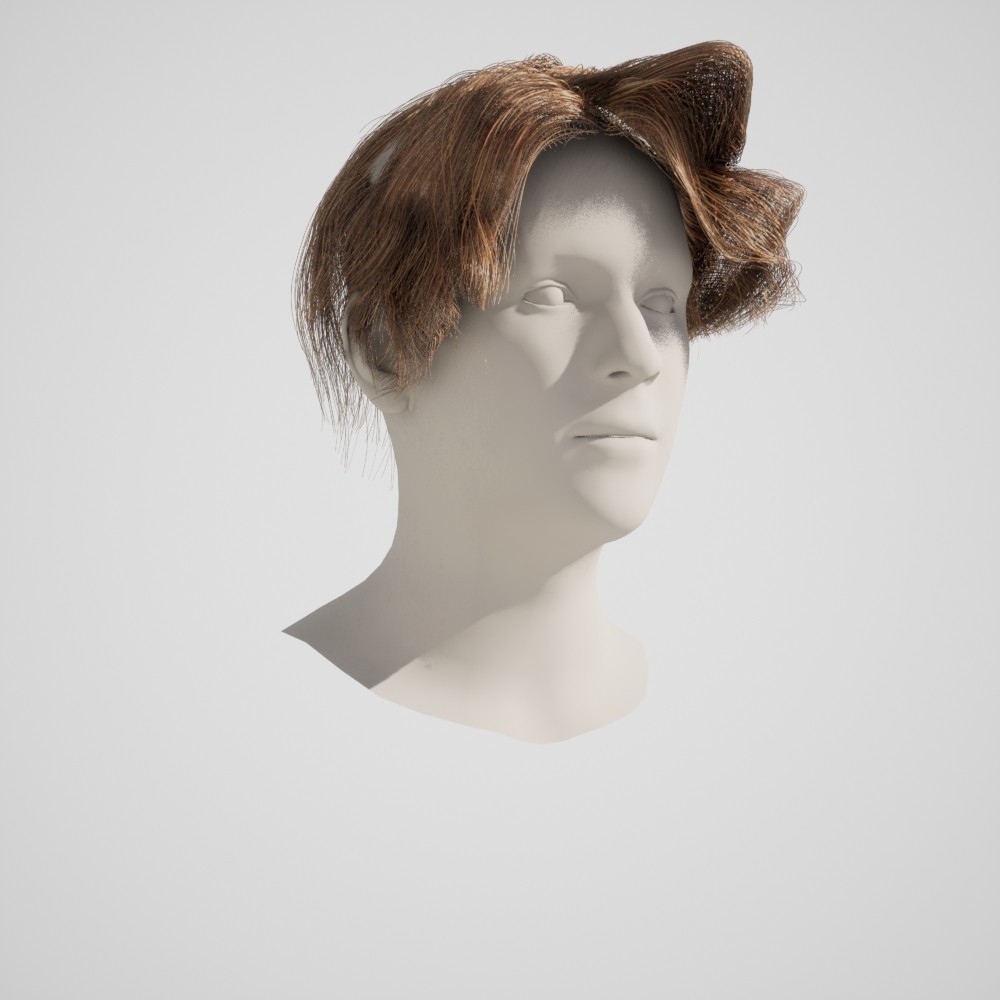}

        & 
        &
        \includegraphics[trim={100 200 100 0},clip,width=0.07\textwidth]{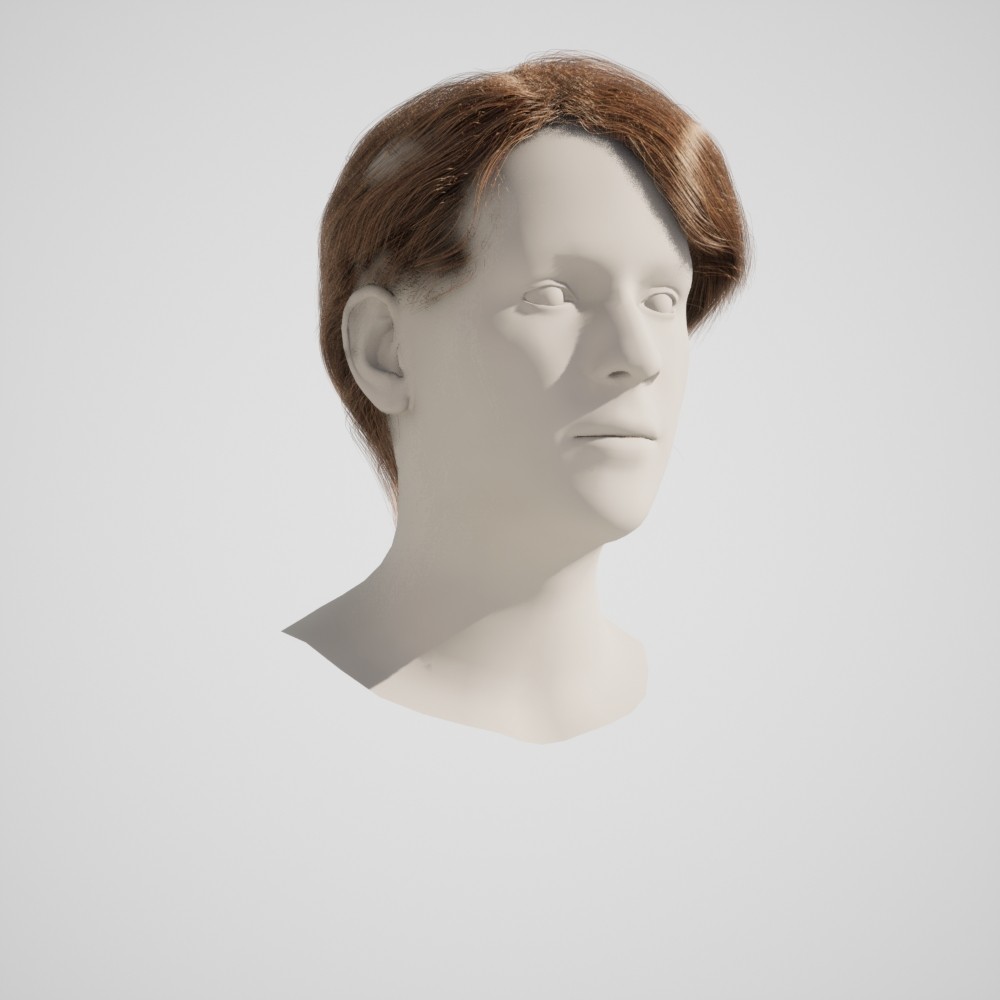} 

        \\

        \\
        \addlinespace

\multicolumn{1}{c}{\begin{small}\textbf{Image}\end{small}}  &
\multicolumn{2}{c}{\begin{small}\textbf{Ours}\end{small}} & 
\multicolumn{2}{c}{\begin{small}\textbf{Hairstep~\cite{hairstep}}\end{small}} & 
\multicolumn{2}{c}{\begin{small}\textbf{PERM~\cite{perm}}\end{small}} & 
\multicolumn{2}{c}{\begin{small}\textbf{NeuralHDHair~\cite{neuralhd}}\end{small}}
\end{tabular}
\vspace{-0.1cm}
\caption{\textbf{Comparison results with baseline.} NeuralHDHair and Hairstep produce smooth geometry, while Hairstep also has artifacts from the back view of short hairstyles (see row 4). PERM reconstructs simple hairstyles well (see row 1), but struggles with more complicated geometry.  Our method recovers more detailed hairstyles than others. Digital zoom-in is recommended.}
\label{fig:geom_compare}

\vspace{-0.2cm}
\end{figure*}

\begin{table}[t]
    \centering
    \vspace{0.2cm}
    \resizebox{\linewidth}{!}{
    \begin{tabular}{l|cc|cc}
        & \multicolumn{2}{c}{Synthetic data} & \multicolumn{2}{c}{Real data} \\ 
          & chamfer\_pts $\downarrow$ & chamfer\_angle $\downarrow$ &  $\mathcal{L}_\text{undir}$ $\downarrow$ & IoU $\uparrow$  \\

\hline
NeuralHDHair~\cite{neuralhd}    &  0.000251 &  0.627 & - & - \\
Hairstep~\cite{hairstep} &  0.001279 &  0.605 &  0.12& 0.8 \\
PERM       &  - &  - & 0.128 & 0.76\\
Our       &  0.000246 &  0.502 & 0.07& 0.97\\
\hline
    \end{tabular}
    }
    \caption{Quantitative comparison on real and synthetic data.}
    \label{tab:ablation_synthetic_dataset}
\end{table}

\section{Experiments}

\subsection{Comparisons to state of the art}

We compare our method with four different types of methods
(1) regression-based (HairNet~\cite{hairnet}), which aims to directly predict global hairstyle;
(2) hair-growing methods, which first learn global and local features of hairstyle and then use hair growing procedures to obtain strand-based geometry, (e.g.~NeuralHDHair~\cite{neuralhd}, Hairstep~\cite{hairstep}); (3) an optimization method in hairstyle prior space, PERM~\cite{perm}; and (4) a retrieval-based method, Hairmony~\cite{hairnet}.
PERM~\cite{perm} is the closest to ours in terms of training setup with the main difference being that they use a high-cost rendering procedure to retrieve the best initialization, which requires hours, while we use a lightweight regressor.
We compare all methods qualitatively and compute metrics on synthetic data for Hairstep~\cite{hairstep} and NeuralHDHair~\cite{neuralhd}, with additional metrics on real data for Hairstep~\cite{hairstep} and PERM~\cite{perm}.

\smallskip \noindent
\textbf{Quantitative comparison.}
First, we perform quantitative comparisons on synthetic data, see examples in the supplementary.
We use ten hairstyles and render them from a hemisphere with high-resolution textures to reduce the domain gap between synthetic renders and real images.
We launch Hairstep~\cite{hairstep}, NeuralHDHair~\cite{neuralhd}, and our method on the obtained data, align results in the same space, and calculate the chamfer distance on points and directions between reconstructed and ground-truth 3D hairstyles; 
see Table~\ref{tab:ablation_synthetic_dataset} for results.
Note that the metrics are evaluated on 10,000 strands sampled from the predicted and ground-truth hairstyles.
The high chamfer distance for Hairstep~\cite{hairstep} happens because of inaccurate length predictions, especially on the back of the hairstyle.
Our method has the lowest chamfer distance both for points and directions.

Second, we compare our model with Hairstep~\cite{hairstep} and PERM~\cite{perm} on 50 real images from a test subset of the Hairstep dataset with ground-truth 2D direction maps. We render the reconstructed hairstyles for both methods from a chosen viewpoint using OpenGL~\cite{opengl}. We evaluate the silhouette and orientation map quality based on the ground-truth reference. For the silhouette, we compute the average IoU, for orientations we use the minimal angular difference $\mathcal{L}_\text{undir}$. 
Table~\ref{tab:ablation_synthetic_dataset} shows that our method outperforms baselines both in silhouette coverage and orientation estimation.

\definecolor{tabfirst}{rgb}{1, 0.7, 0.7} %
\definecolor{tabsecond}{rgb}{1, 0.85, 0.7} %
\definecolor{tabthird}{rgb}{1, 1, 0.7} %

\begin{table}[tbh]
    \centering
    \resizebox{\linewidth}{!}{
    \begin{tabular}{l|ccccc}
         & chamfer\_pts $\downarrow$ & chamfer\_angle $\downarrow$ & angle error $\downarrow$ & mask $\downarrow$ & $\mathcal{L}_\text{undir}$ $\downarrow$ \\
     \hline    
coarse branch &0.00026 &  0.110 &                      15.81 &  0.517 &  0.735 \\
w/o pca       & 0.00026 &0.109 &  15.75 &                      0.598 & 0.738 \\
w/o curv      &  0.00022 & 0.110 & 15.61 &  0.545 &                      0.739 \\
\hline
    \end{tabular}
    }
    \vspace{-0.1cm}
    \caption{Ablation study of training losses of the coarse branch.}
    \label{tab:ablation_loss}
    \vspace{-0.15cm}
\end{table}

\begin{figure}[t]
\centering
  \includegraphics[width=0.95\linewidth]{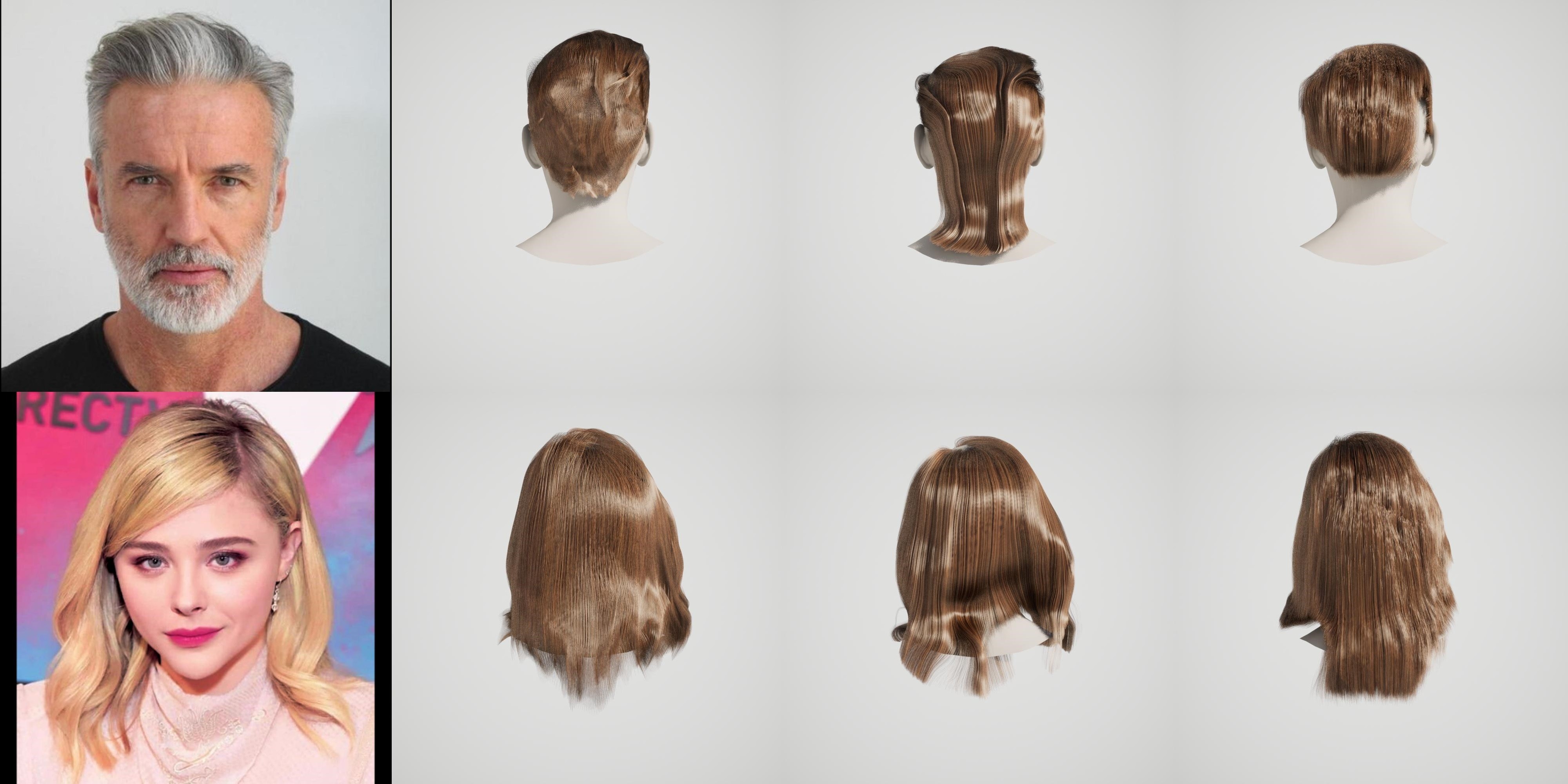}

\makebox[0.25\linewidth]{Image}%
\makebox[0.25\linewidth]{Ours}%
\makebox[0.2\linewidth]{Hairstep}%
\makebox[0.25\linewidth]{NeuralHDHair}%
  \vspace{-0.15cm}
  \caption{\textbf{Back view comparison} with Hairstep~\cite{hairstep} and NeuralHDHair~\cite{neuralhd}. The strands produced by our method on the back of the head are cleaner and more consistent with the input image.
  }
  \vspace{-0.2cm}
  \label{fig:back_view_comparison}
\end{figure}

\begin{figure}
  \centering
  \includegraphics[width=0.95\linewidth]{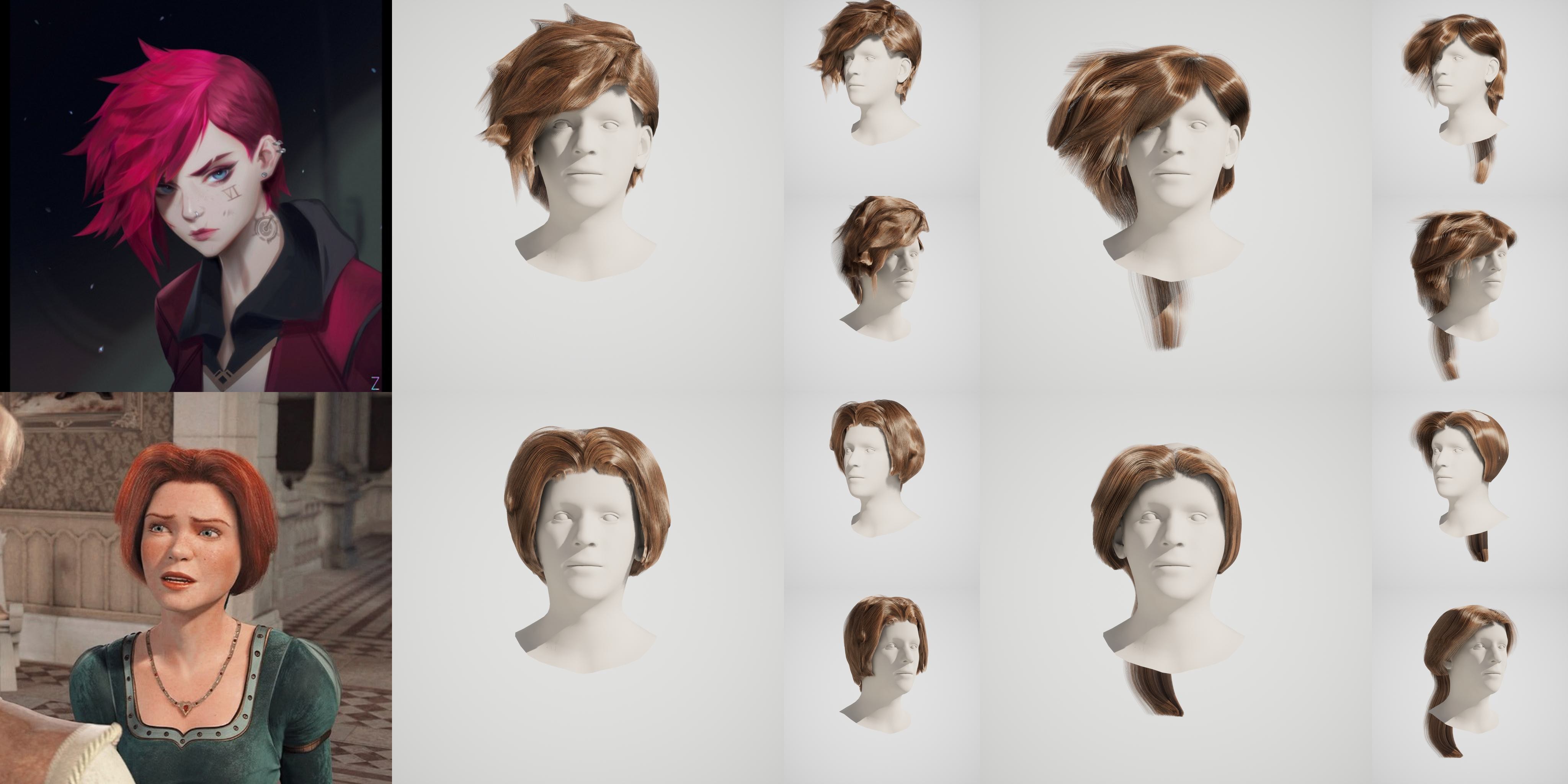}

  \caption{\textbf{Out-of-distribution comparison} of our method (second column) with Hairstep~\cite{hairstep} (last column).}
    \label{fig:cartoon_comparison_main}
    \vspace{-0.3cm}
\end{figure}

\begin{figure}
    \begin{center}
    \includegraphics[width=0.1\textwidth]{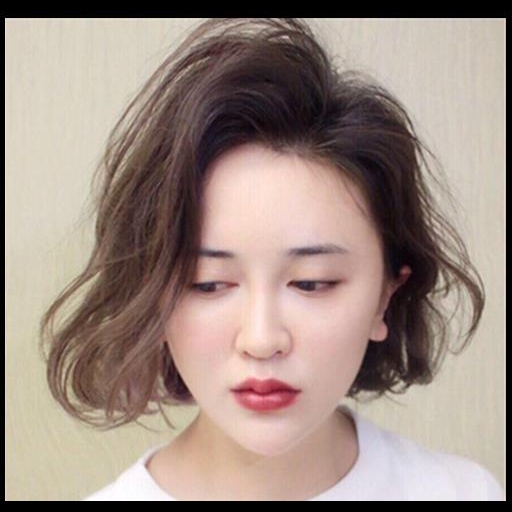}  
    \includegraphics[width=0.1\textwidth]{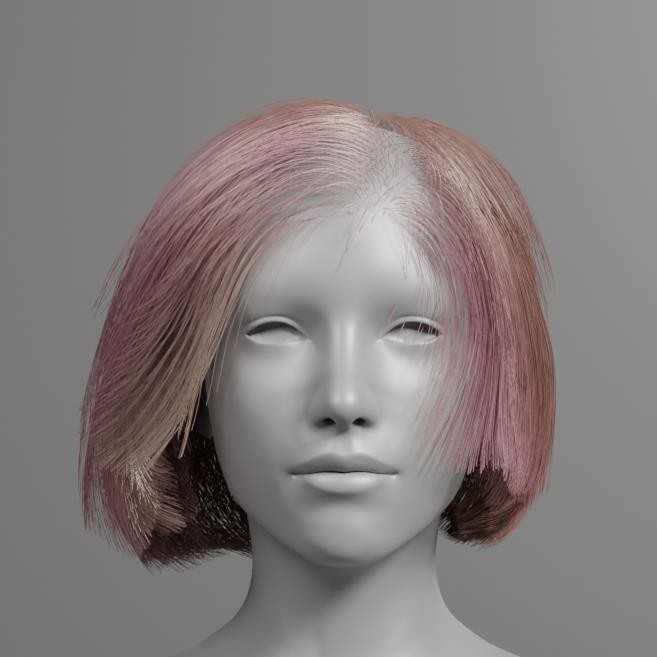}  
    \includegraphics[trim={180 360 180 0},clip,width=0.1\textwidth]{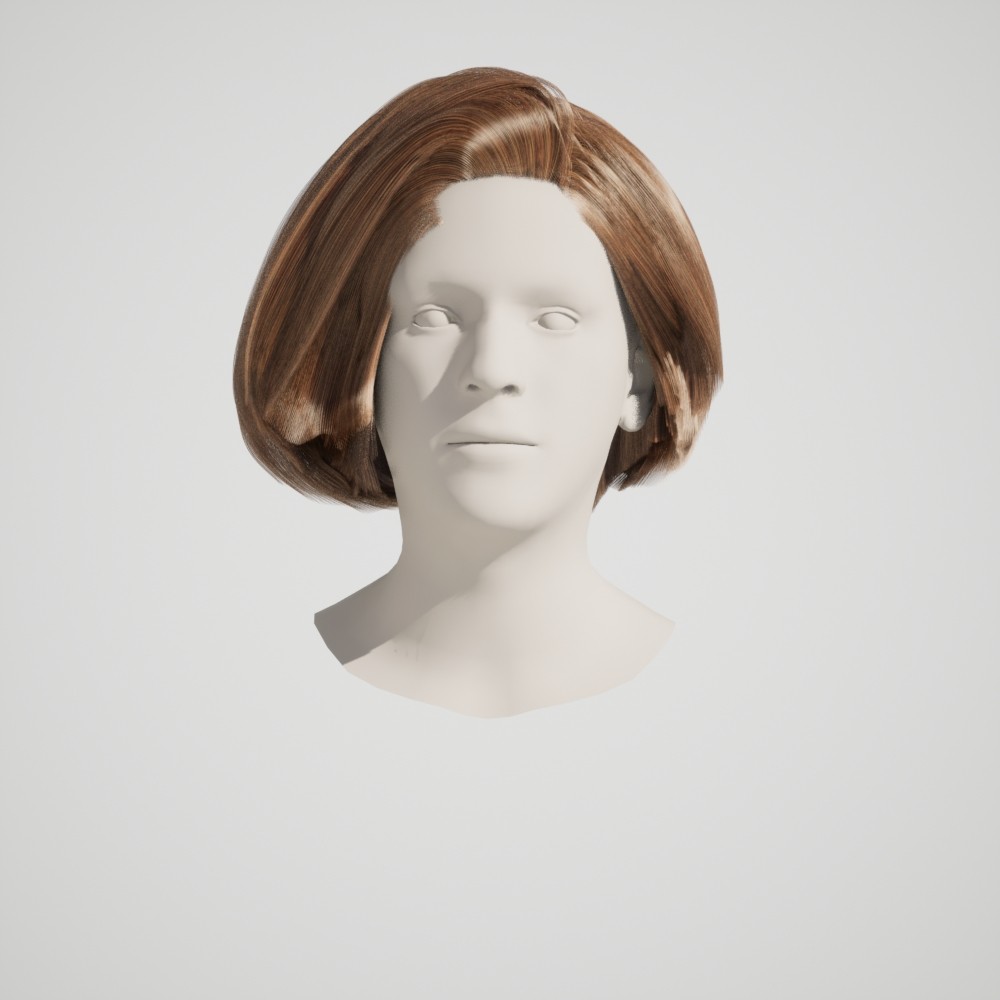}  
    \includegraphics[trim={259 518 259 0},clip,width=0.1\textwidth]{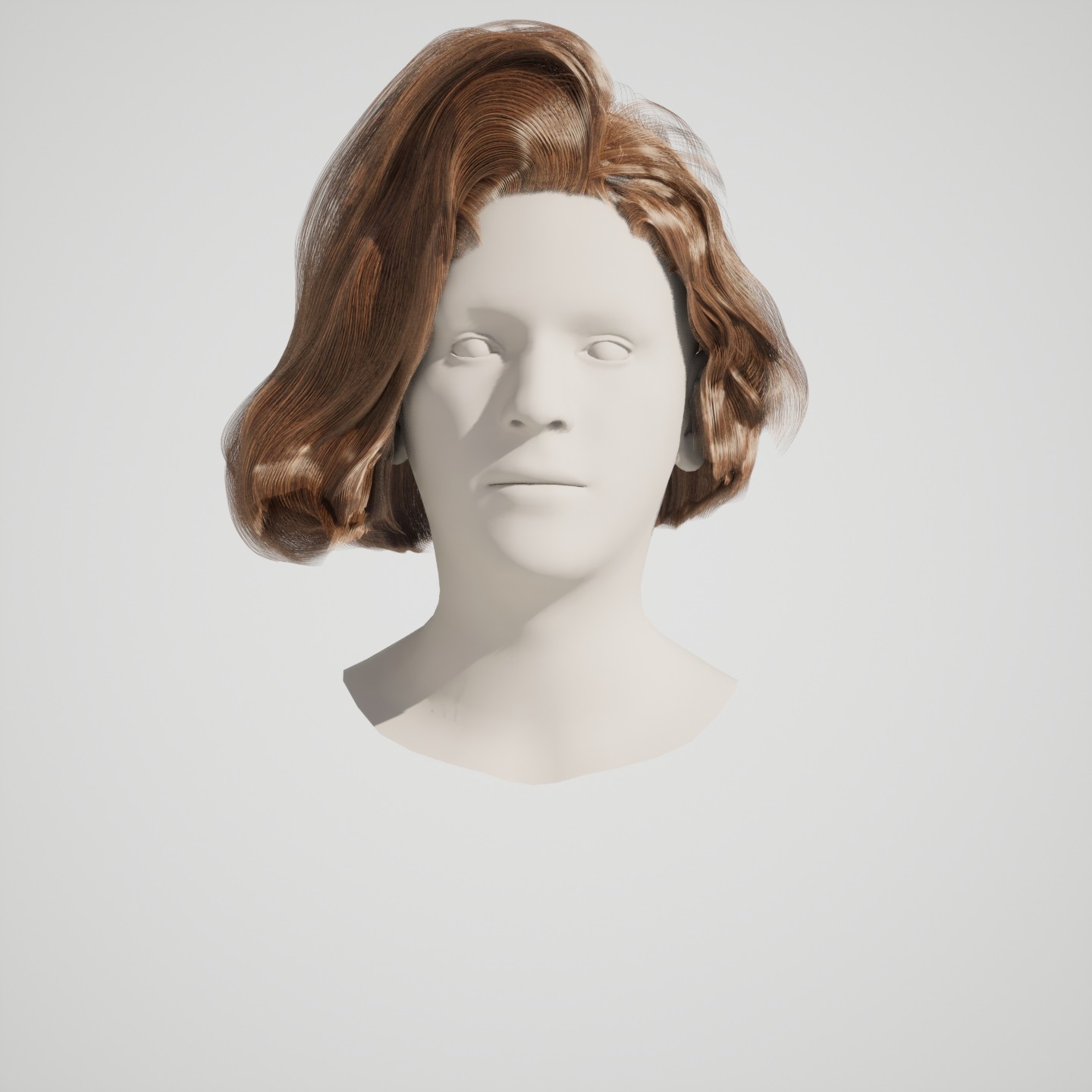}  \\
    
    \includegraphics[width=0.1\textwidth]{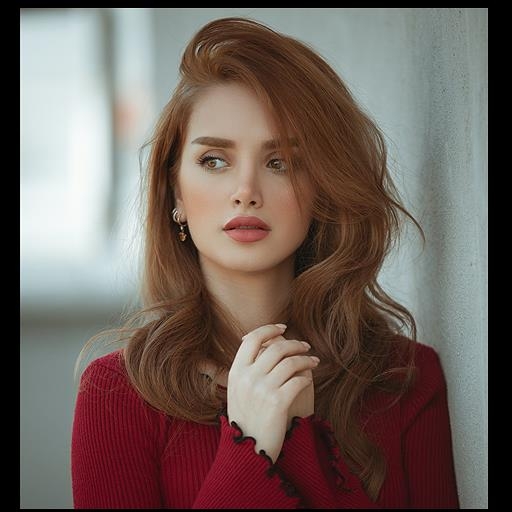}  
    \includegraphics[width=0.1\textwidth]{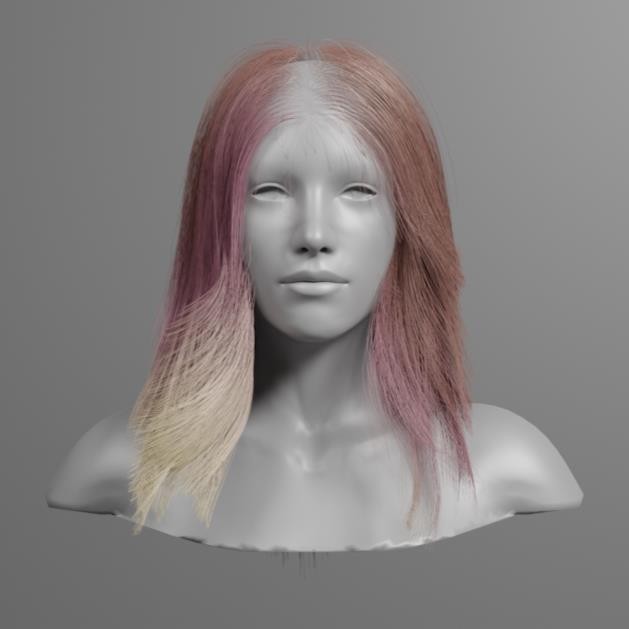}
    \includegraphics[trim={62 83 62 42},clip,width=0.1\textwidth]{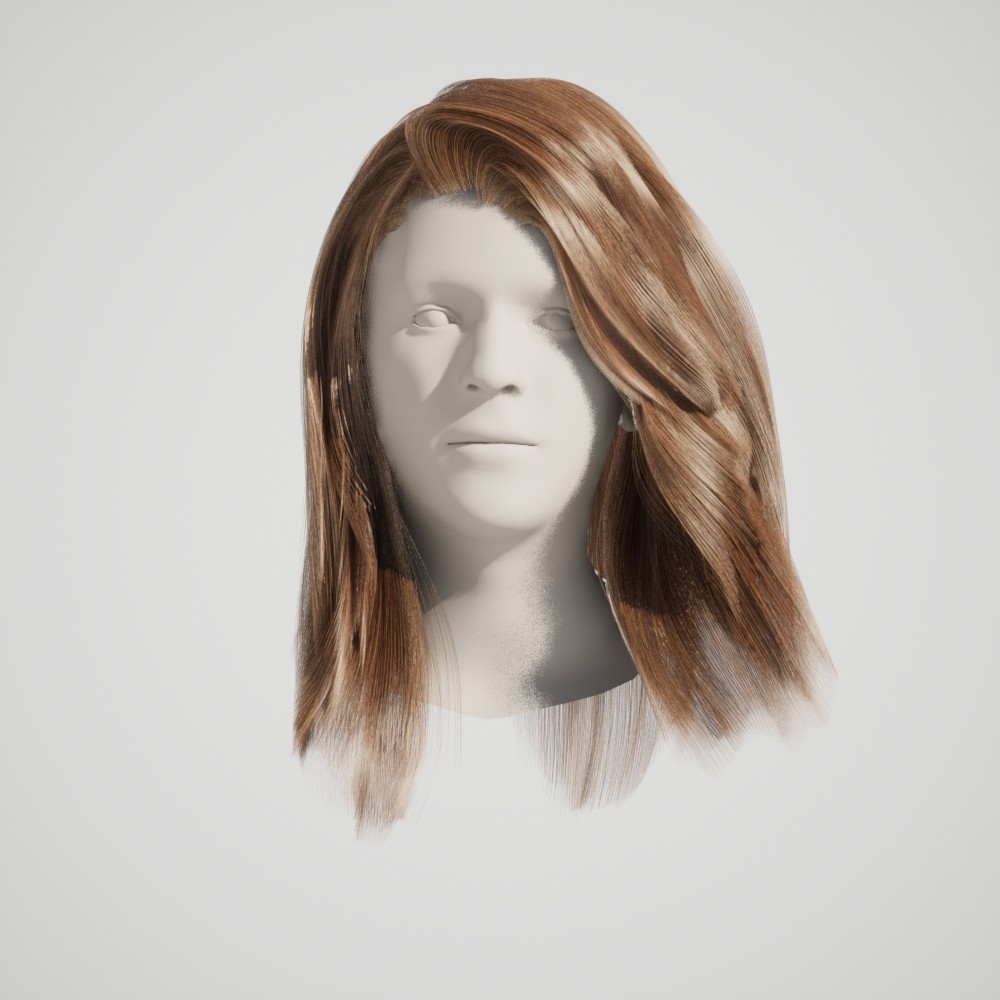}  
    \includegraphics[trim={90 120 90 60},clip,width=0.1\textwidth]{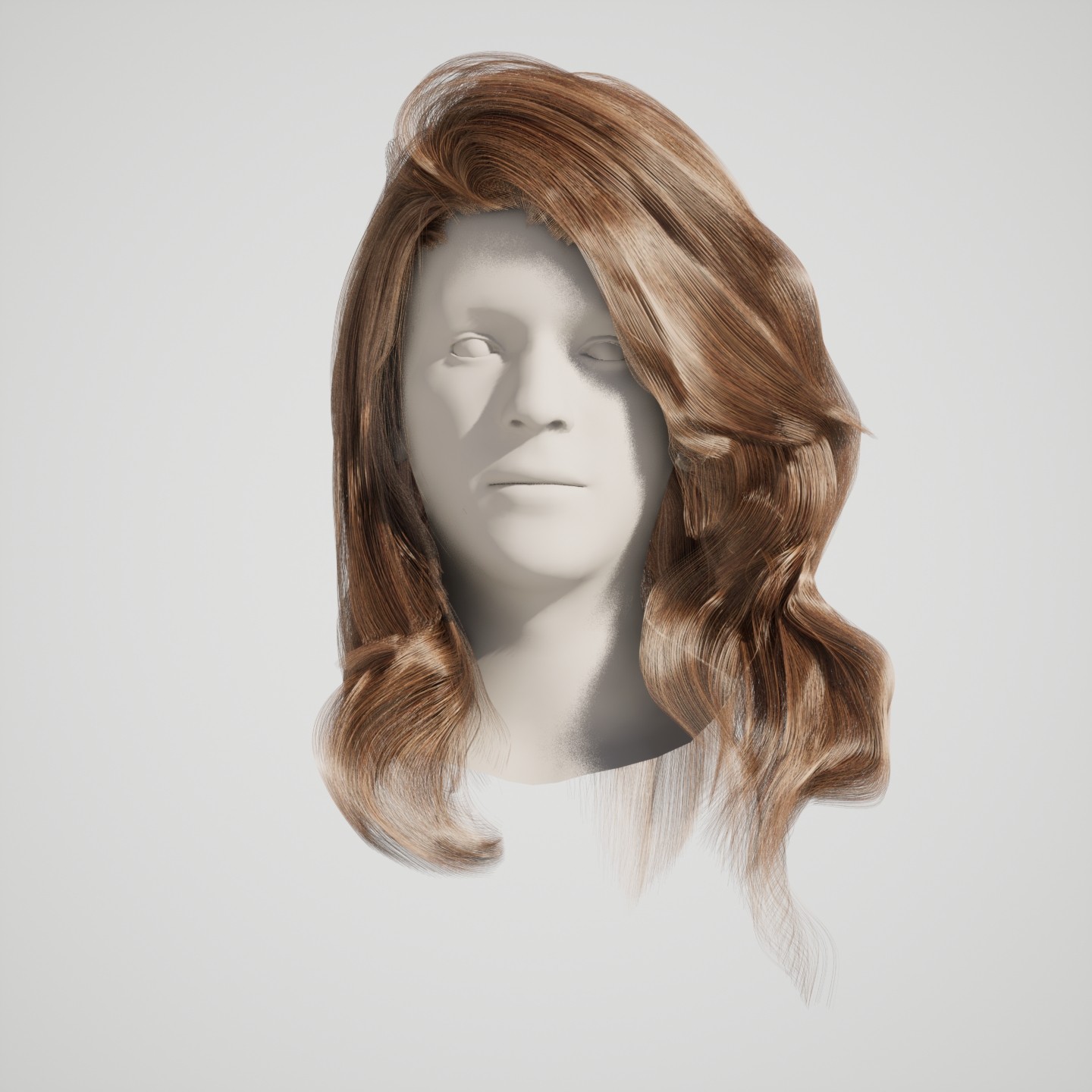}  \\

    \makebox[0.22\linewidth]{Image}%
    \makebox[0.22\linewidth]{HairNet}%
    \makebox[0.22\linewidth]{$\text{Ours}^{\text{w/o opt}}$}%
    \makebox[0.22\linewidth]{$\text{Ours}$}%
  
    \end{center}

    \vspace{-0.5cm}
    \caption{Comparison of our method with HairNet~\cite{hairnet}.}
    \label{fig:comaprison_hairnet}
     \vspace{-0.3cm}
\end{figure}

\newpage 
\noindent
\textbf{Qualitative comparison.}
Figure \ref{fig:geom_compare} compares our reconstruction method with other top-performing approaches on real images.
Hair-growing methods, NeuralHDHair~\cite{neuralhd} and Hairstep~\cite{hairstep} lack details due to the inherent smoothness of their growing procedure.
Hairstep also generates unrealistic geometry in the back, especially for short hairstyles (see Fig.~\ref{fig:geom_compare}, row 4, side view, Fig.~\ref{fig:back_view_comparison}, and Fig.~\ref{fig:cartoon_comparison_main}). 
PERM~\cite{perm} produces significant reconstruction errors, especially for wavy hairstyles, which may be due to poor initialization or an inefficient rendering procedure.
Our optimization-based method produces more detailed and accurate results.
Fig.~\ref{fig:cartoon_comparison_main} illustrates that our method outperforms Hairstep on out-of-distribution samples.
We also show a separate comparison between our method with and without optimization (see ``$\text{Ours}^{\text{w/o opt}}$''), with the regression-based approach HairNet~\cite{hairnet} (see Fig.~\ref{fig:comaprison_hairnet}), that struggles to obtain wavy structures. For comparison with the retrieval-based method Hairmony~\cite{hairmony}, please refer to the supplementary.

\subsection{Ablation studies}

We evaluate our losses, the influence of the depth map signal on reconstruction quality, as well as the training setup, using the following metrics:
a chamfer distance with normals between predicted and ground-truth hairstyle, and the mean angular error in degrees on a test subset of 50 hairstyles from the PERM~\cite{perm} dataset.
Additionally, we evaluate the view-related losses using a metric based on silhouette preservation using an $L_{2}$ distance and an undirected orientation metric $\mathcal{L}_\text{undir}$ using a subset of 33 real images:
\begin{equation}
    \mathcal{L}_\text{undir} = \min \big\{ \text{d}(b_p, \hat{b}_p), \text{d}(b_p, \hat{b}_p) \pm \pi \big\},
    \label{eq:undir}
\end{equation}
where $\text{d}$ denotes an angular distance between the direction vectors, and $\hat{b}_p$ is generated using oriented Gabor filters.

\smallskip \noindent
\textbf{Losses.} 
We ablate the contribution of each loss during the coarse stage training using a leave-one-out strategy.
We evaluate the quality using various metrics on both synthetic and real datasets (see Table~\ref{tab:ablation_loss}).
Omitting the loss on PCA components (``w/o pca'') yields comparable results on synthetic data, but leads to a notable drop in performance on real data.
Interestingly, removing the curvature loss (``w/o curv'') improves metrics for synthetic data, yet results in a significant decline in quality for real-world inputs. 
For the importance of the depth map as an input signal and direction loss, please refer to the supplementary.

\smallskip \noindent
\textbf{Mixing strategy.}
We analyze the impact of the mixing strategy (training on both synthetic and real data, at the same time) for the fine stage reconstruction model (Table~\ref {tab:tab_mixing_strategy}, ``w/o Mixing'' and ``hybrid training'').
When we add the mixing strategy, we see some degradation in terms of performance on synthetic data, but a significant improvement on real data.
We also experiment with using the rendering loss, but continue training only on synthetic data (see ``$\text{Prior}^{\text{syn only}}$'').
In addition, we show the importance of the mixing strategy for the final hairstyle reconstruction stage.

\begin{figure}
  \centering
  \includegraphics[width=\linewidth]{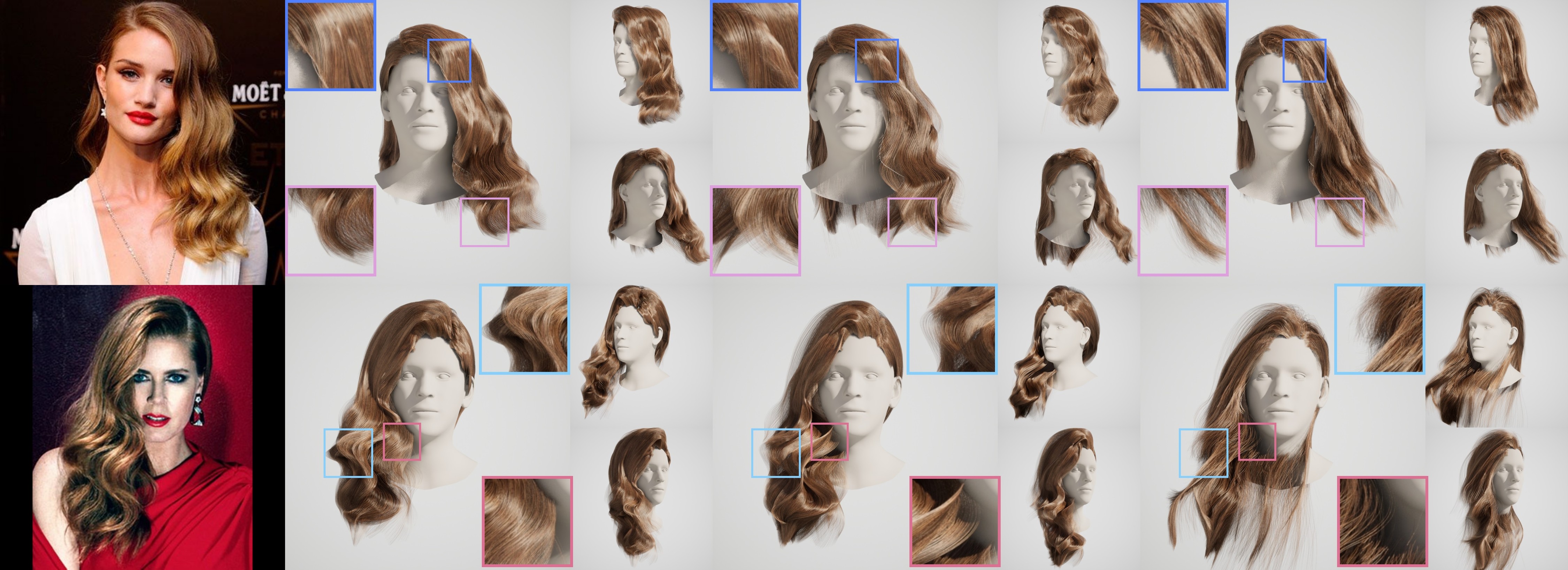}
  Image \hspace{1cm} Ours \hspace{1cm} w/o Mixing \hspace{1cm} w/o prior
  \vspace{-0.2cm}
  \caption{\textbf{Hair reconstruction stage.} The mixing strategy and prior space are important for single-view inversion.}
   \label{fig:mixing_prior_ablation}
   \vspace{-0.1cm}
\end{figure}

In Fig.~\ref{fig:mixing_prior_ablation}, we show results of optimization in the hairstyle prior space trained using the mixing strategy and without it (see ``Ours'' and ``w/o Mixing'').
Without the mixing strategy, there are more penetrations with the head, and the hairstyles look less realistic.

\smallskip \noindent
\textbf{Coarse-to-fine strategy.}
We evaluate the importance of training in separate stages by training a single-stage model that predicts all 64 PCA coefficients simultaneously.
To ensure training stability, we employ an unfreezing strategy, adding one more component to the optimization every 1,000 steps.
Additionally, we increase the model's capacity by adding several layers to the decoder architecture.
This single-stage model (``w/o coarse-to-fine'') is trained exclusively on synthetic data and is compared to the fine-stage model trained without the mixing strategy (``w/o Mixing'').
As shown in Table~\ref{tab:tab_mixing_strategy}, training in a single stage results in reduced quality for both synthetic and real data.

\smallskip \noindent
\textbf{Importance of prior during inversion.}
We show the importance of our prior model during inversion, comparing it with direct inversion in 3D space, see Fig.~\ref{fig:mixing_prior_ablation} (``Ours'' and ``w/o prior'').
To do that, we predict hairstyle initialization using our Hairstyle prior model and then, similar to GaussianHaircut~\cite{GaussianHaircut}, we optimize 3D directions to minimize the $\mathcal{L}_{real}$ loss.
Direct optimization in 3D space without a prior model leads to poor quality, especially for wavy hairstyles.

~

\definecolor{tabfirst}{rgb}{1, 0.7, 0.7} %
\definecolor{tabsecond}{rgb}{1, 0.85, 0.7} %
\definecolor{tabthird}{rgb}{1, 1, 0.7} %

\begin{table}[t]
    \centering
    \resizebox{\linewidth}{!}{
    \begin{tabular}{l|ccccc}
         & chamfer\_pts $\downarrow$ & chamfer\_angle $\downarrow$ & angle error $\downarrow$ & mask $\downarrow$ & $\mathcal{L}_\text{undir}$ $\downarrow$ \\

\hline
hybrid training             & 0.00030 &                      0.143 &18.03 &0.405 & 0.695 \\
w/o coarse-to-fine & 0.00032 & 0.135 &17.87 &                      0.578 & 0.732 \\
w/o Mixing   & 0.00032 & 0.128 & 17.02 &0.547 &                      0.735 \\
$\text{Prior}^{\text{syn only}}$ & 0.00025 & 0.136 & 17.63 & 0.519 & 0.733 \\

\hline
    \end{tabular}
    }
    \vspace{-0.1in}
    \caption{Ablation study w.r.t.~our mixing strategy, coarse-to-fine prediction of PCA components. %
    }
    \label{tab:tab_mixing_strategy}
\end{table}

\subsection{Applications}

\begin{figure}
  \includegraphics[width=\linewidth]{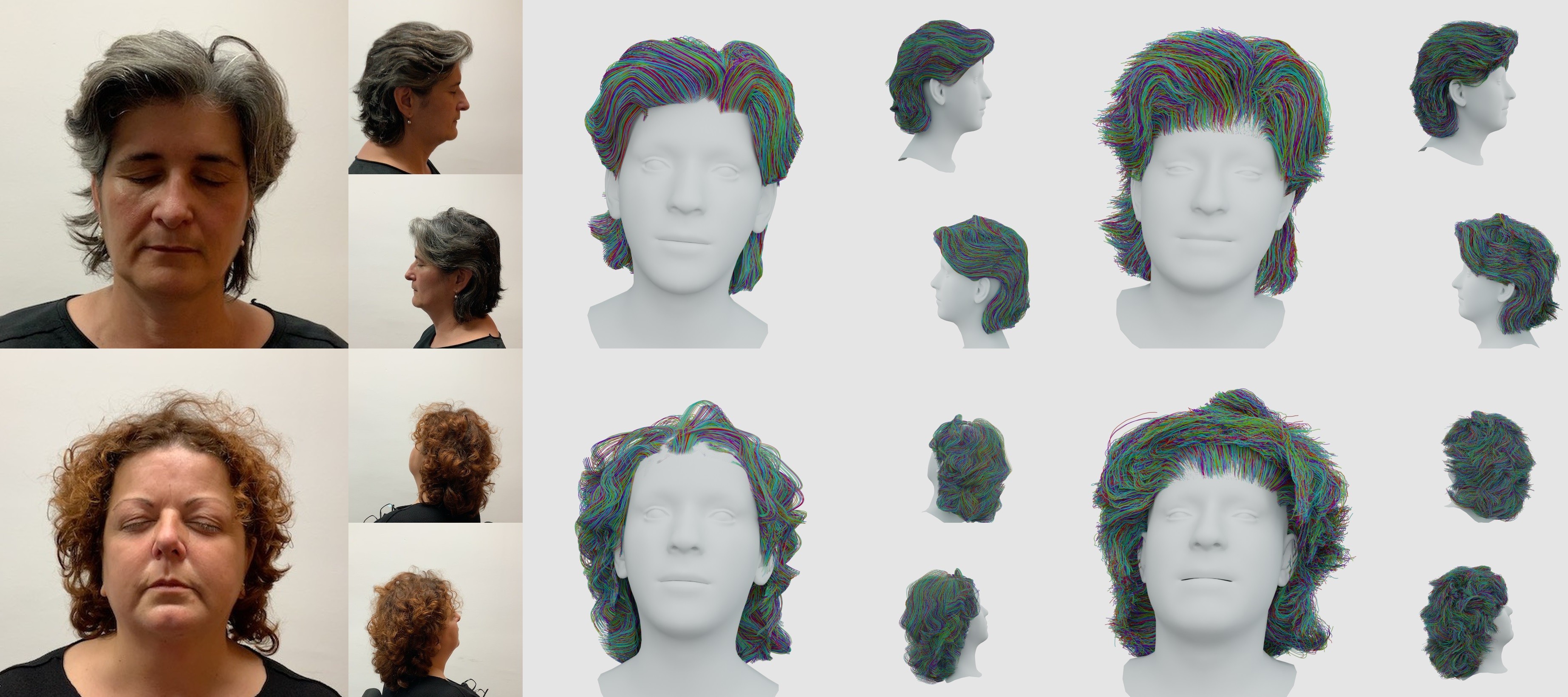}

  \makebox[0.33\linewidth]{Image}%
\makebox[0.33\linewidth]{Ours}%
\makebox[0.33\linewidth]{Gaussian Haircut}%
  
  \vspace{-0.2cm}
  \caption{\textbf{Multi-view Reconstruction.} Comparison of our method with Gaussian Haircut~\cite{GaussianHaircut} on 8 views.}
    \label{fig:multiview}
\end{figure}

\textbf{Multi-view reconstruction.} Our approach can be adapted to multi-view hair reconstruction, see suppl. doc.
Figure \ref{fig:multiview} shows the results of our method compared to Gaussian Haircut~\cite{GaussianHaircut} on several scenes using 8 views from H3DS dataset~\cite{h3ds}. 
While our method produces smoother results, it has more accurate geometry. Also, Gaussian Haircut~\cite{GaussianHaircut} is optimized for 10 hours on a single A100, while our method takes around 45 minutes, making it more efficient in the multi-view scenario.

\begin{figure}
    \centerline{    \includegraphics[width=\linewidth]{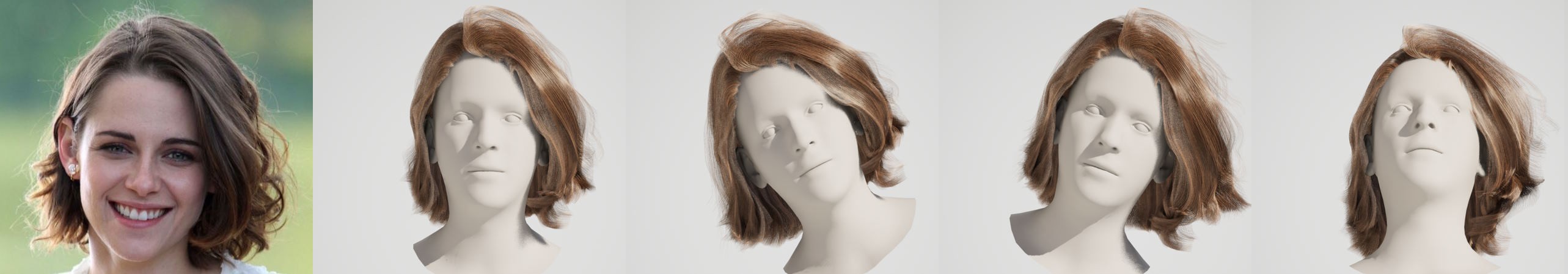}}
    \vspace{-0.1in}
    \caption{Reconstructed hairstyles animated using Unreal Engine.}
    \label{fig:sim}
\end{figure}

\begin{figure}
\centerline{  \includegraphics[width=\linewidth]{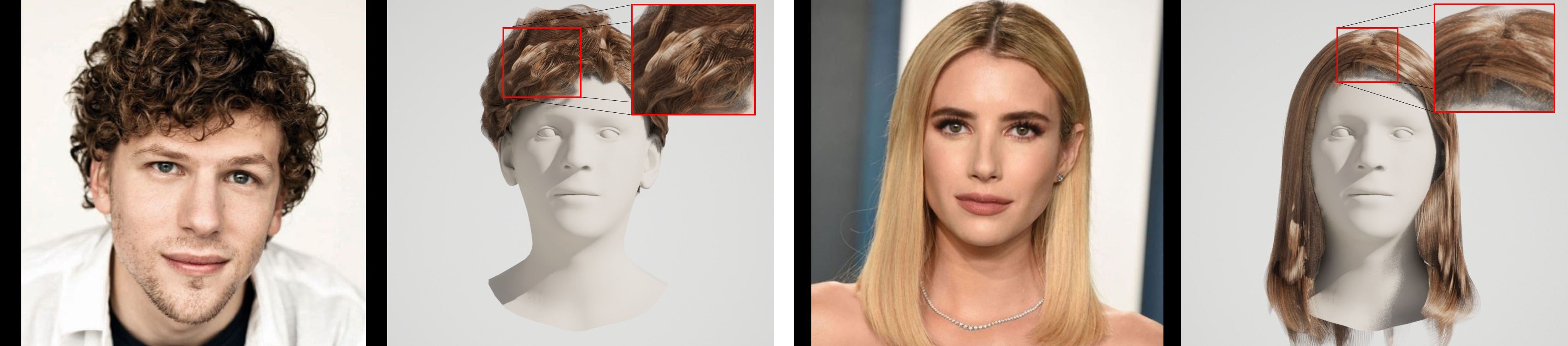}}
  \vspace{-0.1in}
  \caption{\textbf{Limitations.} Our method struggles with curly hairstyles (left example) due to a limited dataset and sometimes has problems with a partition (right example). }
    \label{fig:limitations}
\end{figure}

\smallskip \noindent\textbf{Hair Simulations.} Obtained hairstyles are fully compatible with CG engines and can be used for realistic rendering and simulations, see Fig.~\ref{fig:sim} and the supplementary video.

\subsection{Limitations and future work}

Our model depends on off-the-shelf estimators for camera, hair silhouette, and direction maps, making it sensitive to their errors.
Furthermore, our approach faces challenges in reconstructing highly curly hairstyles (Fig.~\ref{fig:limitations}, left) due to the limited hairstyles in the synthetic dataset. Expanding the dataset to include a broader range of diverse and complex hairstyles could improve these results.

We also observe difficulties where the hair parts 
(Fig.~\ref{fig:limitations}, right); these arise from fixed root positions and inaccuracies in the baldness masks. A joint optimization approach that dynamically adapts root positions and densification could address this limitation.

\section{Conclusion}

In this paper, we present a method for strand-based hair reconstruction from a single input photograph. 
To achieve this, we learn a strand-based hairstyle prior model using a dataset of synthetic and real data by introducing 3D geometry-based and rendering-based training objectives.
We leverage this model for single-view as well as multi-view hairstyle reconstruction, obtaining superior performance over competitors.
We show out-of-distribution examples and how our reconstructions can be used in simulations, which highlights the versatility of our method.%

\section*{Acknowledgements}
Vanessa Sklyarova and Malte Prinzler were supported by the Max Planck ETH Center for Learning Systems. Egor Zakharov's work was funded by the ``AI-PERCEIVE'' ERC Consolidator Grant, 2021. Justus Thies is supported by the ERC Starting Grant 101162081 ``LeMo'' and the DFG Excellence Strategy— EXC-3057. 
The authors would like to thank Yi Zhou for running PERM on the provided data and Keyu Wu for executing NeuralHDHair. We also thank Denys Nartsev, Arina Kuznetcova, and Tomasz Niewiadomski for their help during the project and Benjamin Pellkofer for IT support.

\paragraph{Disclosure.} 
While MJB is a co-founder and Chief Scientist at Meshcapade, his research in this project was performed solely at, and funded solely by, the Max Planck Society.

{
    \small
    \bibliographystyle{ieeenat_fullname}
    \bibliography{main}
}
\clearpage
\setcounter{section}{0}
\renewcommand{\thesection}{S\arabic{section}}
\section{Background}
To use 3D Gaussian Splatting for soft-rasterization of hair, we force Gaussians to lie on line segments.
We define the mean and covariance matrix of each individual Gaussian as:
\begin{equation}
    \mu_{ij} = \frac{1}{2} \big( p_{ij} + p_{i,j+1} \big),\quad C_{ij} = E_{ij} D_{ij} \big ( E_{ij} D_{ij} \big)^T.
\end{equation}
Here, $E_{ij} = \{ b_{ij}, t_{ij}, n_{ij} \}$ is a TBN basis associated with the strand curve, $b_{ij} = v_{ij} / |v_{ij}|, v_{ij} = p_{i,j+1} - p_{ij}$, where $v_{ij}$ denotes the segment vector and $b_{ij}$ its normalized direction vector.
$D_{ij}$ is defined as $D_{ij} = \text{diag} ( d_{ij}, \epsilon, \epsilon )$, where $d_{ij}$ is set to be proportional to the length of $v_{ij}$ and $\epsilon$ denotes a small value.
Such parametrization allows effective propagation of photometric information into hairstyle geometry.

\section{Training details}

\subsection{Strands parametrization}

For basis calculation, we launch the Incremental PCA method on all hairstyles from the PERM~\cite{perm} dataset.
We use 200 points for each strand to provide more degrees of freedom for hairstyles.
We found that PCA method can effectively compress each strand of size $200~\times~3$ into $64$ dimensions.

\subsection{Optimization details}

We use 4 NVIDIA A100 GPUs to train our entire method, which takes a total of 5 days and 19 hours.

For the \textbf{coarse stage}, we optimize the model for $420,000$ iterations (around 73 hours) with an effective batch size of $32$ using AdamW~\cite{adamw} with a weight decay of $0.001$ and a learning rate of $0.0001$.
We use the following weights: 
$\lambda_\text{PCA}=0.1$, $\lambda_\text{dir}=0.1$, $\lambda_\text{curv}=1$, and $\lambda_\text{mask}=0.0001$.

For the \textbf{fine stage}, we first fine-tune the fine branch with the ground-truth PCA map for the first 10 components for 200,000 iterations or around 45 hours.
We initialize the Encoder and Decoder architectures using the pretrained weights from the coarse branch.
For optimization, we use the following weights:
$\lambda_\text{PCA}=10$, $\lambda_\text{dir}=0.1$, $\lambda_\text{curv}=0.1$, and $\lambda_\text{mask}=0.0001$.
We also calculate visibility weights for points and apply a $3~\times$ weight for points, direction, and curvature losses.

Then, we finetune the coarse and fine branches together for 80,000 iterations on synthetic data (around 8-9 hours).
For optimization, we use the following weights:
$\lambda_\text{PCA}=1$, $\lambda_\text{dir}=0.1$, $\lambda_\text{curv}=0.1$, and $\lambda_\text{mask}=0.0001$.

Finally, we optimize the model \textbf{on mixed dataset} with real and synthetic data for 16,000 iterations (around 12 hours).
To balance the gradient propagation from real and synthetic data, we weight the photometric losses with weight $r=0.5$.
We use $L_{1}$ distance to calculate the depth loss.
We use the following parameters:
$\lambda_\text{PCA}=0.1$, $\lambda_\text{dir}=0.1$, $\lambda_\text{curv}=0.1$, $\lambda_\text{mask}=0.0001$, width for each gaussian = 0.005, $\lambda_{depth}=0.01, \lambda_{pen}=0.1, \lambda_{seg}=10, \lambda_{dirmap}=5$. 

During the \textbf{inversion stage}, we optimize the hairstyle for 400 iterations on a single A100, which takes around 10 minutes.
We use the following parameters: width for each gaussian is set to $0.00035$, upsampling from the texture size $64\times 64$ to $256\times 256$,  $\lambda_{seg}=1, \lambda_{dirmap}=0.8, \lambda_{pen}=0.3, \lambda_{depth}=0.01$.

\begin{figure}
\centering
  \includegraphics[width=\linewidth]{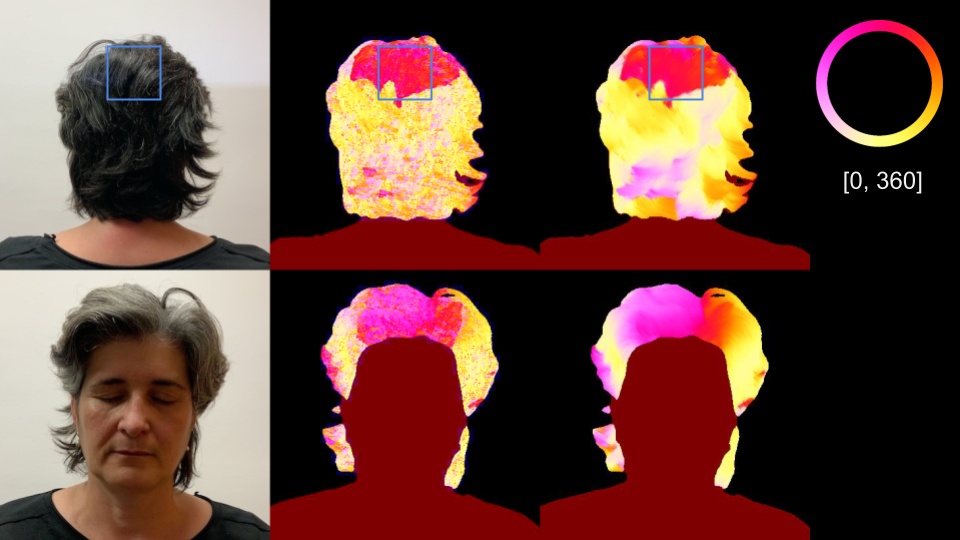}
  \makebox[0.3\linewidth]{Image}%
    \makebox[0.3\linewidth]{Gabor+Dirmap}%
    \makebox[0.3\linewidth]{Dirmap}%
    \makebox[0.1\linewidth]{}%
  \caption{Combination of Gabor map with Direction map from Hairstep~\cite{hairstep}. Although the model predicts correct directions for frontal views, it produces incorrect results for back views.
  }
    \label{fig:combination_gabor_dirmap}
\end{figure}

\subsection{Preprocessing}

We align all synthetic hairstyles to the same bust model.
To simplify the training, we ensure that real and synthetic data share the same scale range.
This is achieved by applying an affine transformation to the input image, using facial keypoints estimated from both the real image and the rendered bust model.
For real and synthetic data, we extract depth using Depth Pro~\cite{depth-pro}.
We normalize the depth map values within the hair silhouette by following these steps:
(i) we erode the hair segmentation mask to reduce boundary artifacts;
(ii) we clip the depth values using the 2nd and 98th quantiles;
and (iii) perform min-max normalization.

\subsection{Interpolation}
To obtain interpolated hairstyles from our guiding strands during the inversion stage, we utilize an upsampling method with a weighted combination of nearest and bilinear interpolation following HAAR~\cite{haar}, adapted to operate in 3D space.
This results in around 10,000 strands during optimization.
For visualization purposes, we repeat this procedure and increase up to 30,000 strands.

\section{Evaluation details}

\begin{figure}
    \begin{center}
    \includegraphics[trim=120 150 120 90,clip,width=0.09\textwidth]{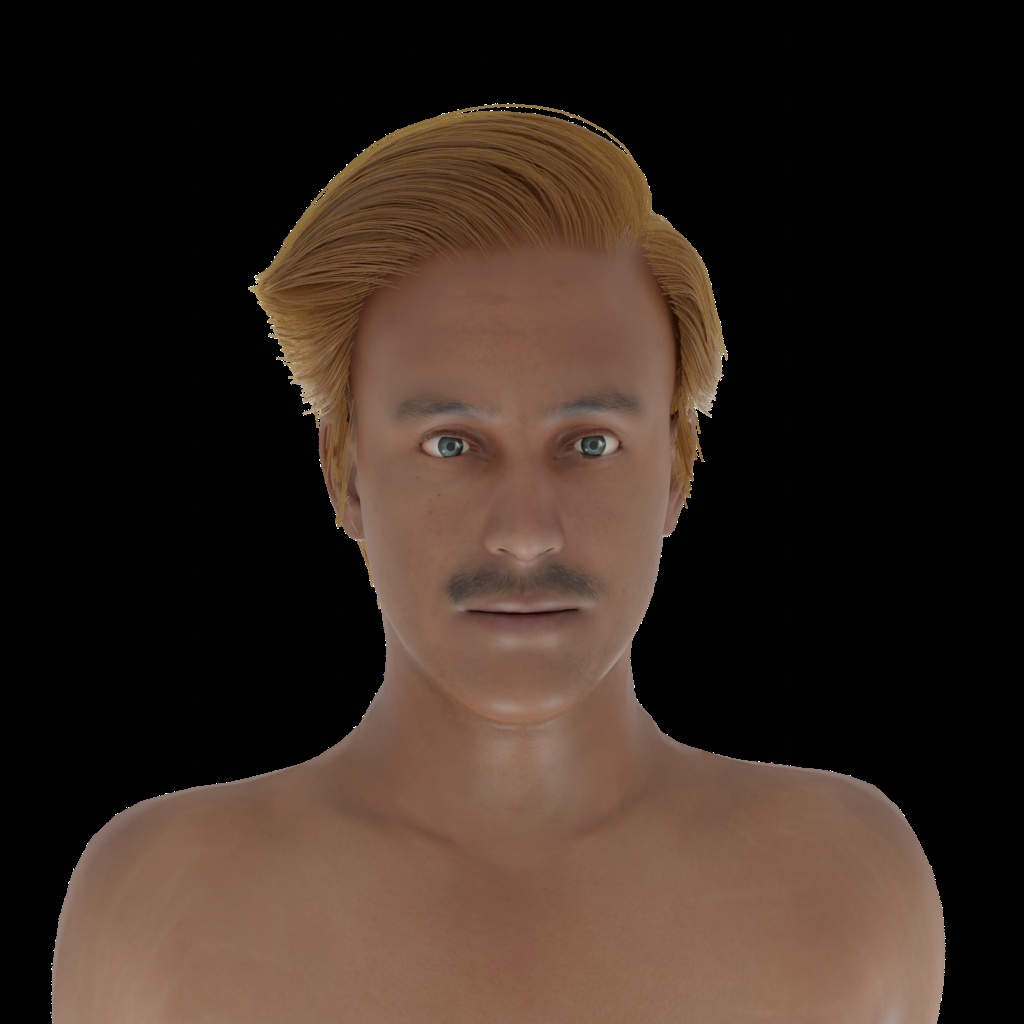}
    \includegraphics[trim=120 150 120 90,clip,width=0.09\textwidth]{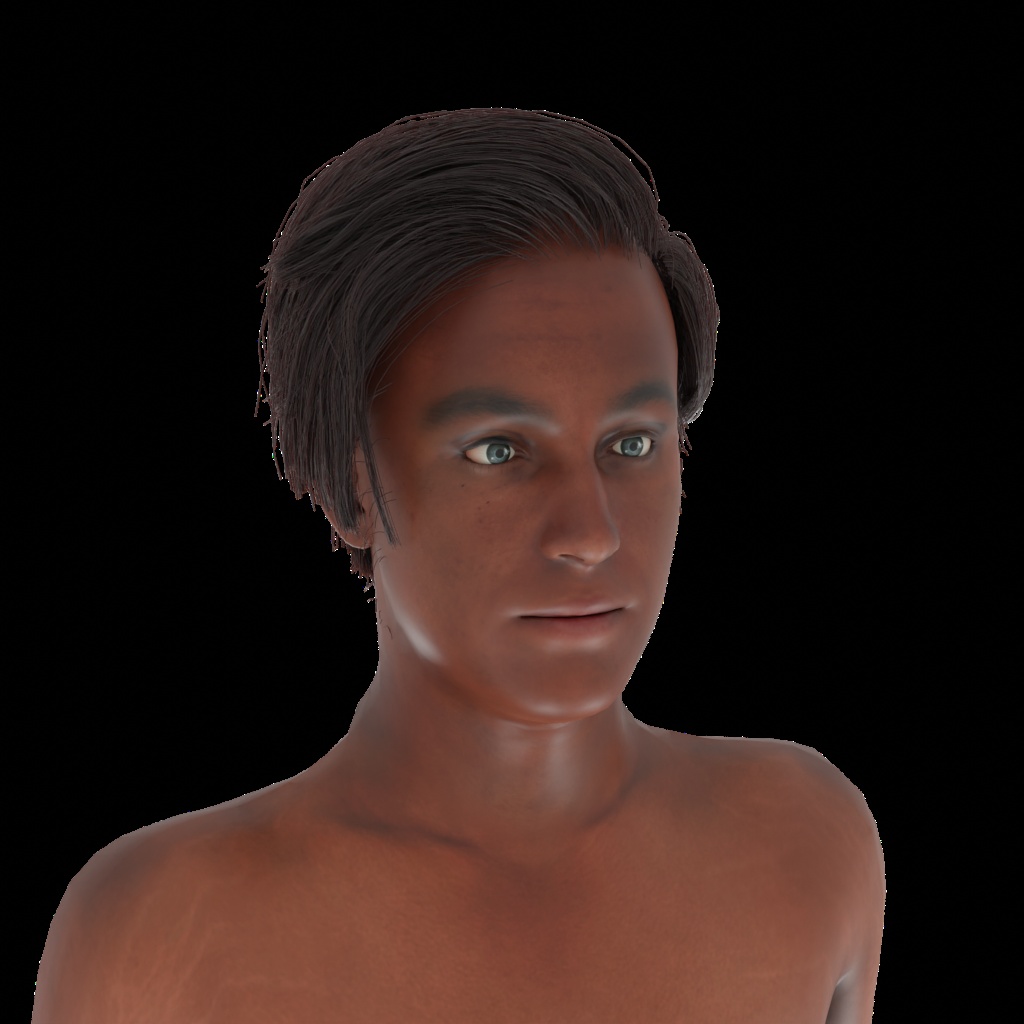} 
    \includegraphics[trim=120 160 120 80,clip,width=0.09\textwidth]{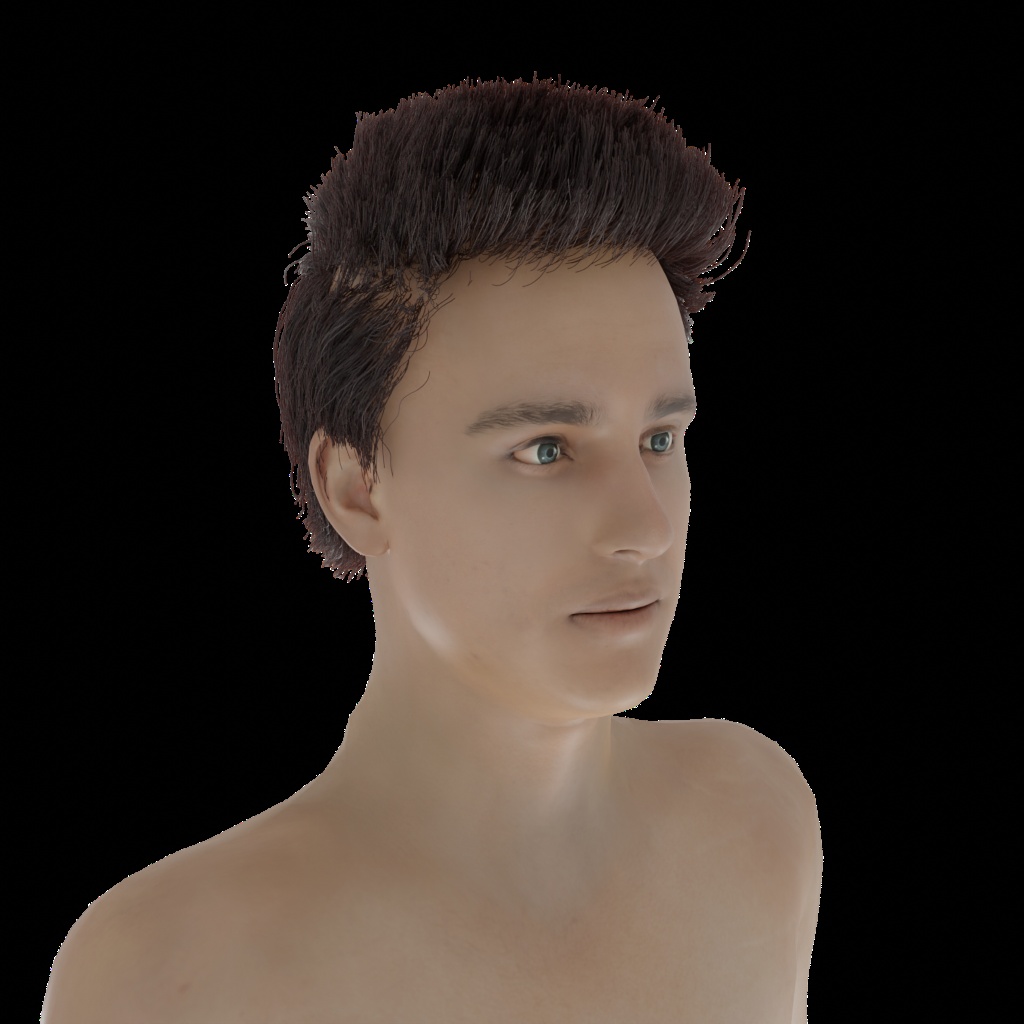}
    \includegraphics[trim=120 150 120 90,clip,width=0.09\textwidth]{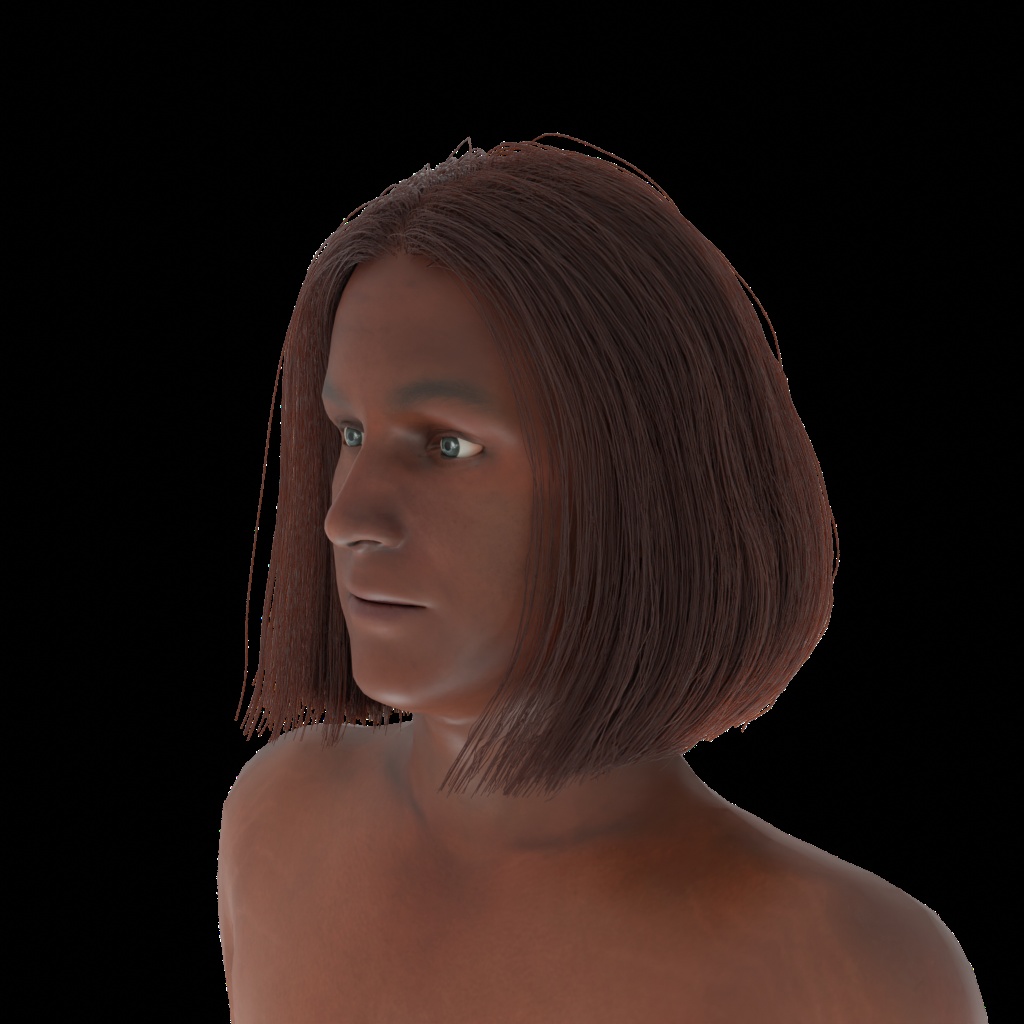} 
     \includegraphics[trim=120 150 120 90,clip,width=0.09\textwidth]{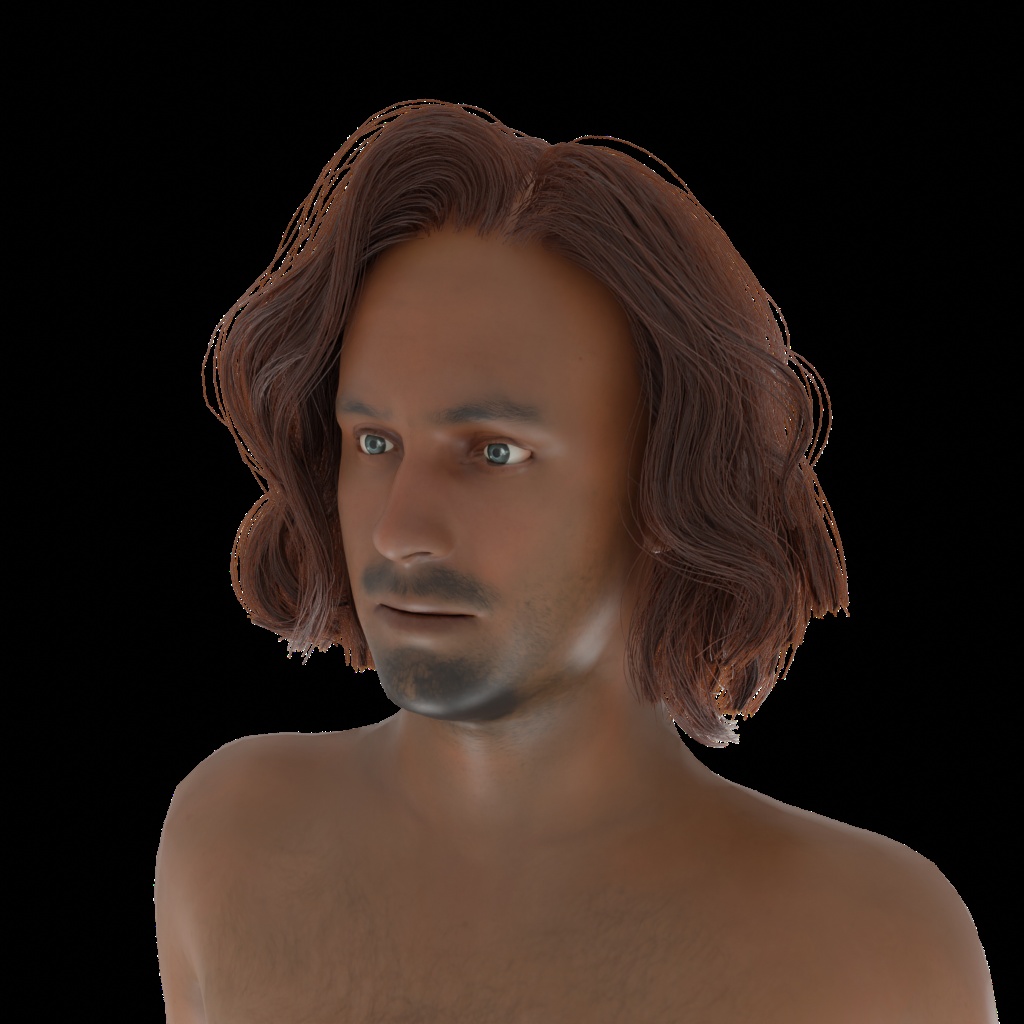}\\ 
    \includegraphics[trim=120 150 120 90,clip,width=0.09\textwidth]{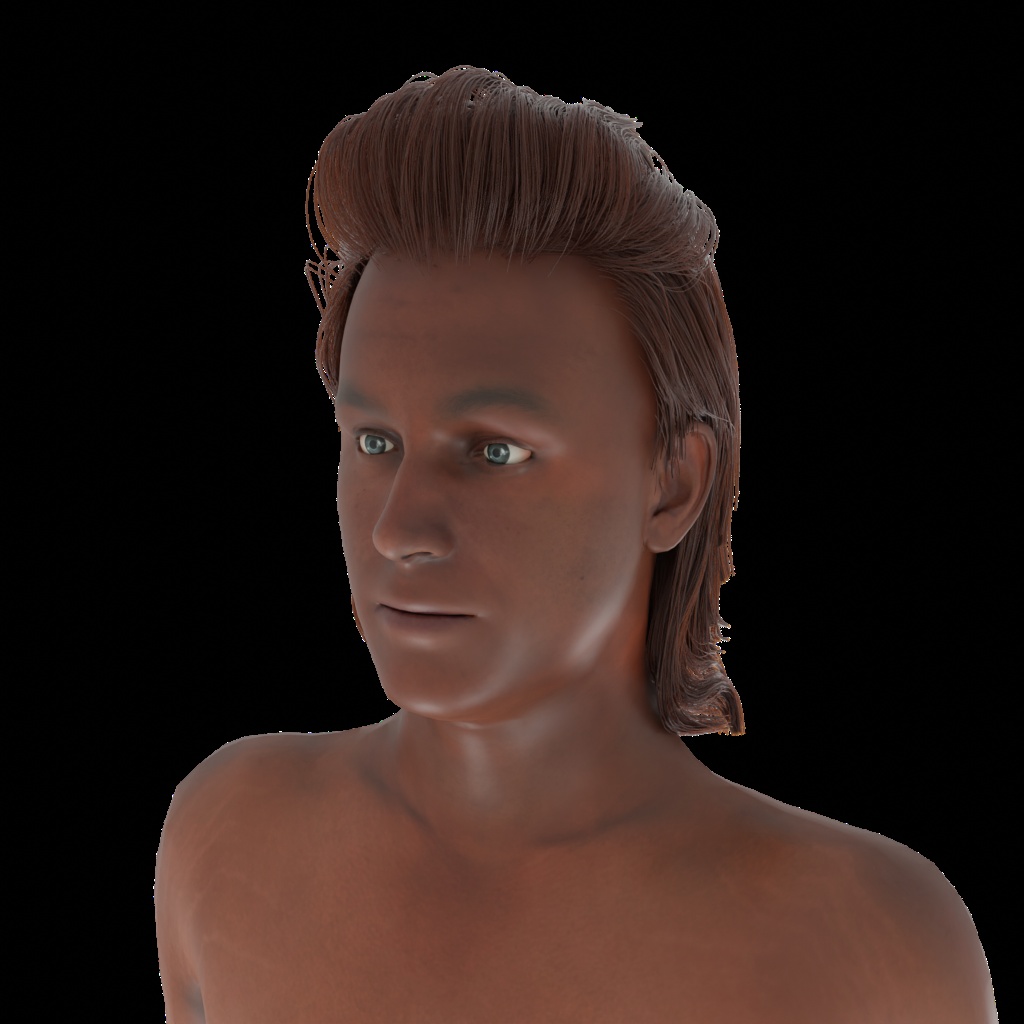}
    \includegraphics[trim=120 150 120 90,clip,width=0.09\textwidth]{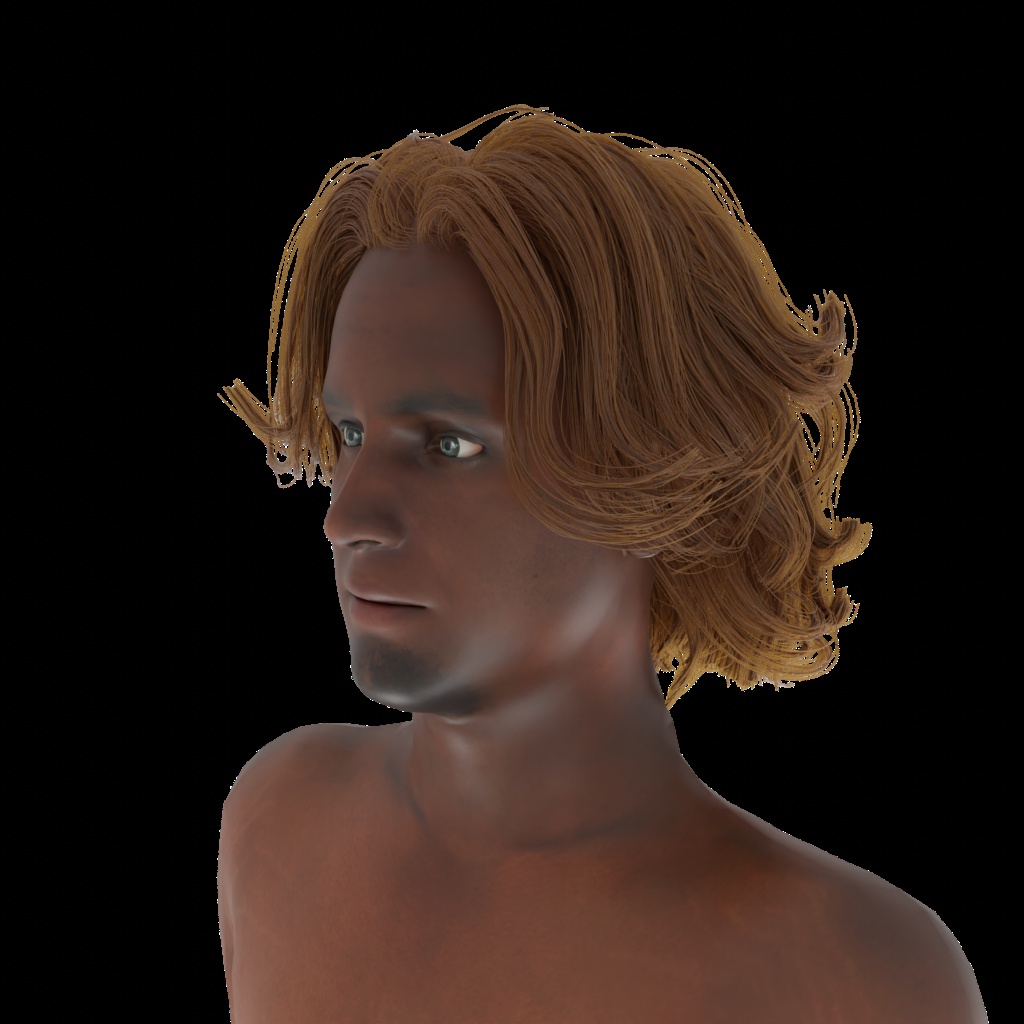}  
    \includegraphics[trim=120 150 120 90,clip,width=0.09\textwidth]{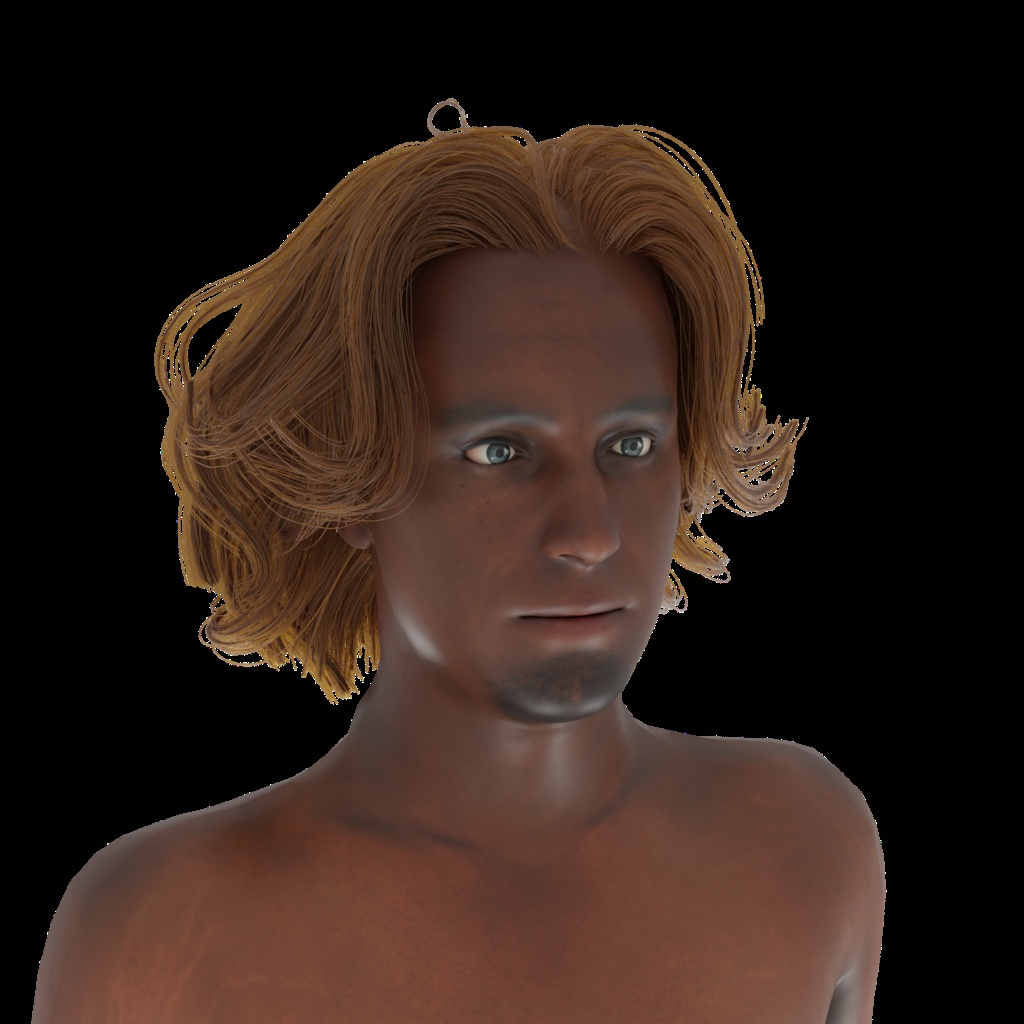} 
    \includegraphics[trim=120 150 120 90,clip,width=0.09\textwidth]{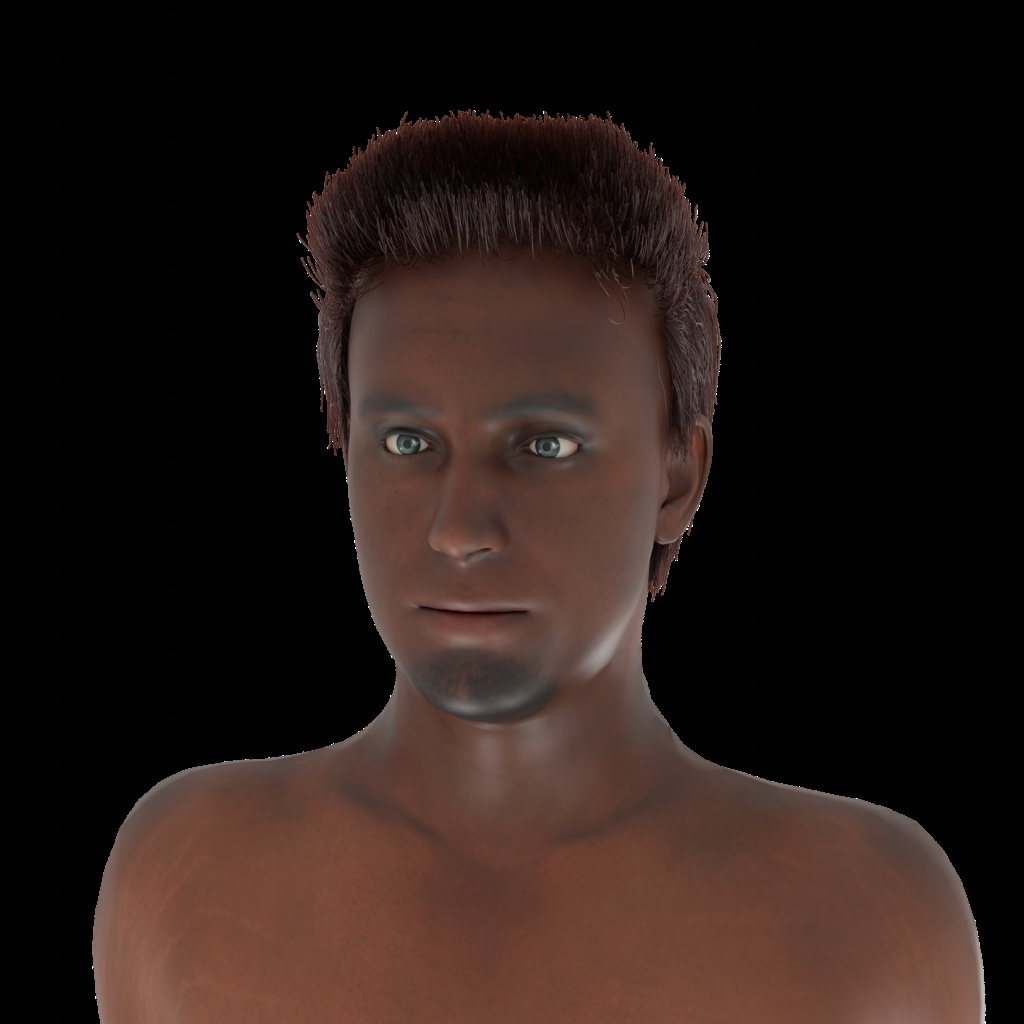} 
     \includegraphics[trim=120 150 120 90,clip,width=0.09\textwidth]{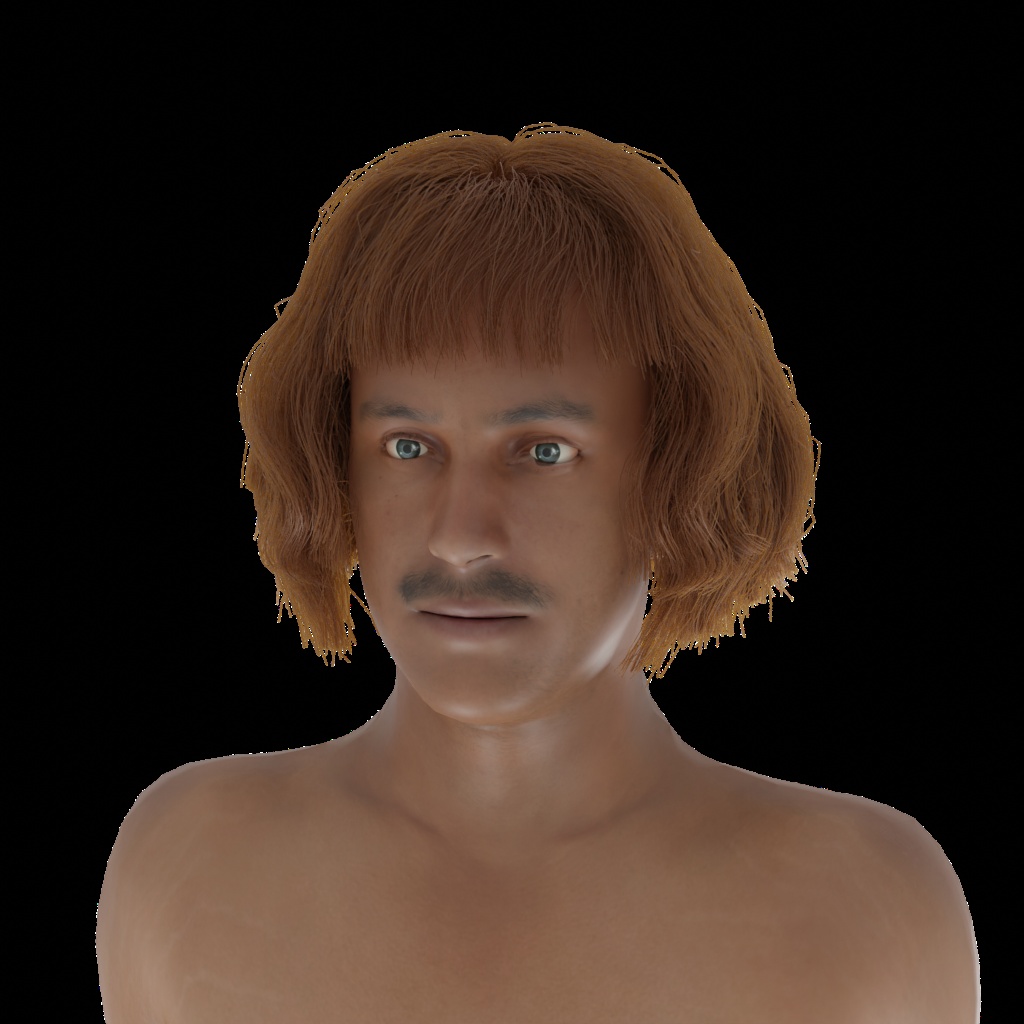}\\  
    \end{center}
    \vspace{-0.25cm}
    \caption{\textbf{Synthetic dataset} is generated by rendering ground-truth hairstyles in Blender.}\label{fig:hairstyles}
      \vspace{-0.5cm}
\end{figure}

\paragraph{Baselines.}
We compare our method with Hairstep~\cite{hairstep} using the original publicly available code.
Results for NeuralHDHair~\cite{neuralhd} and PERM~\cite{perm} were provided by the respective authors.
For Hairnet~\cite{hairnet}, we use images extracted from the Hairstep~\cite{hairstep} paper, and do not compare on more samples, as the performance of this model is significantly worse than other existing methods.
Finally, in Figure~\ref{fig:comaprison_hairmony}, we compare with Hairmony~\cite{hairmony} on several images and use the corresponding results extracted from their paper.

All baselines use augmented versions of the USC-HairSalon~\cite{hairsalon} dataset.
The PERM dataset is constructed similarly, so the synthetic datasets used here and in prior work are comparable.
Retraining all models under identical conditions is not feasible since prior works did not release training code. 

\paragraph{Quantitative comparison.}
For quantitative comparison of our method with the baselines, we render synthetic hairstyles using Blender~\cite{blender} (see images in Figure~\ref{fig:hairstyles}).

\begin{figure}
\centering
\begin{center}
    \centering
    \includegraphics[width=0.48\textwidth]{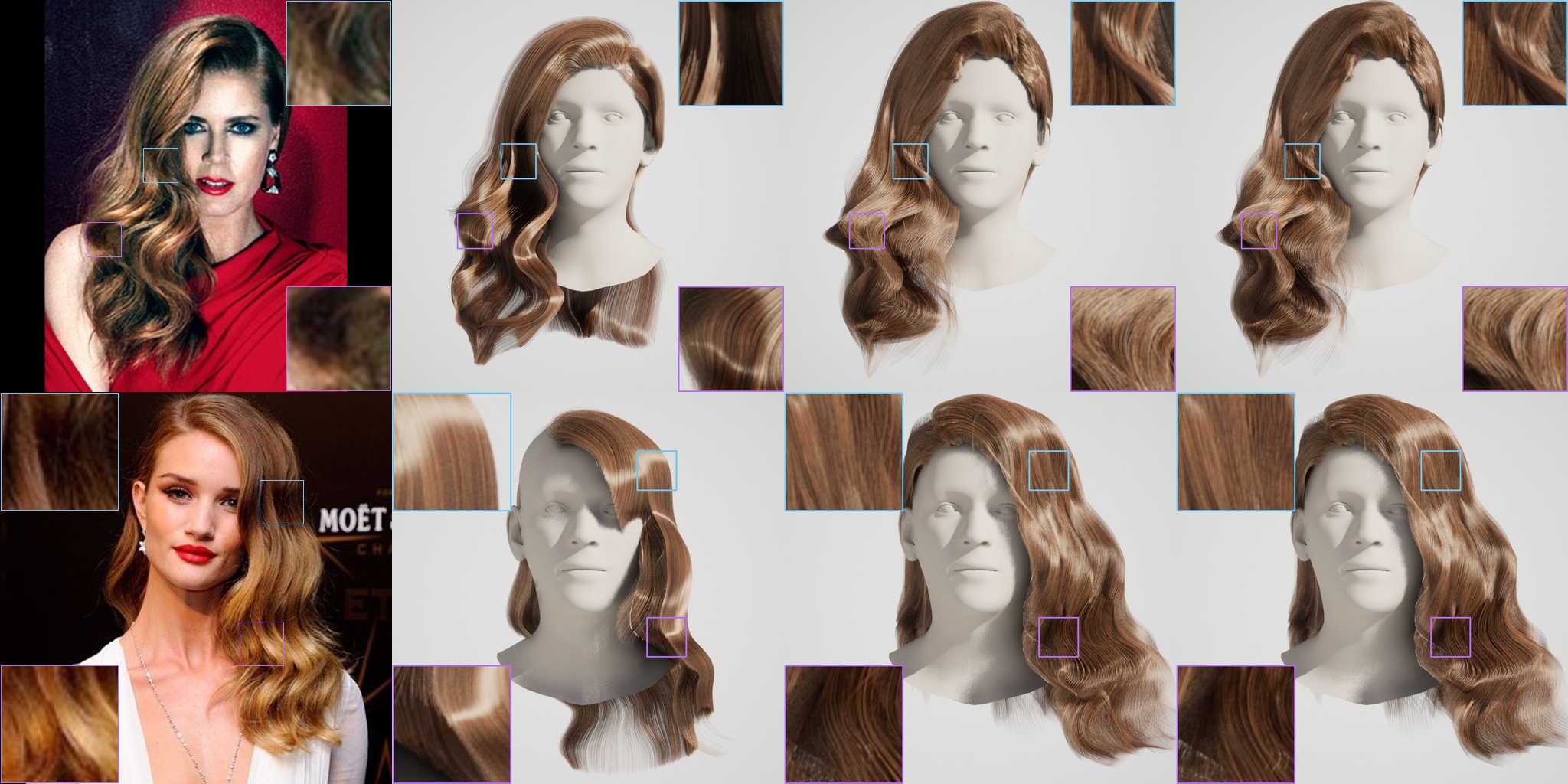}   
    \makebox[0.12\textwidth]{Image}%
    \makebox[0.12\textwidth]{Hairstep}%
    \makebox[0.12\textwidth]{$\text{Ours}^{\text{same cost}}$}%
    \makebox[0.12\textwidth]{$\text{Ours}$}

    \vspace{-0.2cm}
    
\end{center}%
\caption{\textbf{Comparison with Hairstep} under the same computational cost.}
\label{fig:ablation_additional_hairstep}
\end{figure}

\paragraph{Inference time.}
Perm~\cite{perm} is an optimization-based method based on a retrieval procedure that could take around 4 hours.
Hairstep~\cite{hairstep} takes around 3-5 minutes to reconstruct 30,000 strands on an NVIDIA RTX3090.
Our method employs 400 optimization iterations, requiring approximately 10 minutes on an A100, yet it achieves superior strand quality compared to Hairstep~\cite{hairstep} within the same time frame, see ``$\text{Ours}^{\text{same cost}}$'' in Figure~\ref{fig:ablation_additional_hairstep} for comparison under the same computation budget.

\paragraph{Visualization.} To show rendering results, we use Unreal Engine~\cite{unrealengine}. For NeuralHDHair~\cite{neuralhd} we have the 60,000 strands, while for Hairstep~\cite{hairstep}
and our around 30,000. For PERM~\cite{perm} original number of strands from the authors was used, as additional interpolation may lead to worse results.

\section{Applications}
\subsection{Multi-view optimization} 

In this section, we show more results of integrating our prior model in the multi-view reconstruction scenario.
Rather than relying solely on the direction map, we enhance it with Gabor orientation maps to incorporate finer details.
Since the direction map estimator from Hairstep~\cite{hairstep} struggles with accurate predictions from side and back views (compare Figure~\ref{fig:combination_gabor_dirmap}, first row) as well as loose high-frequency details from frontal, we propose using an augmented Gabor direction map (see ``Gabor+Dirmap'' in Figure~\ref{fig:combination_gabor_dirmap}) for near-frontal views and an undirected Gabor map for other viewpoints.
The optimization process follows the same weighting scheme as for the single-view scenario, with the only difference that for the orientation loss we compute either $\mathcal{L}_\text{dir}$ or $\mathcal{L}_\text{undir}$ depending on the view.

During optimization, we input a \textbf{near-frontal} image into the encoder to extract more accurate features, while supervising from other views using the same Gaussian Splatting-based procedure with soft-rasterization of hair strands.
At each iteration, we randomly sample a single view for supervision.
We also experimented with rendering and applying weighted loss from all views simultaneously, but this slowed down the training.

\begin{figure*}
\includegraphics[width=0.98\textwidth]{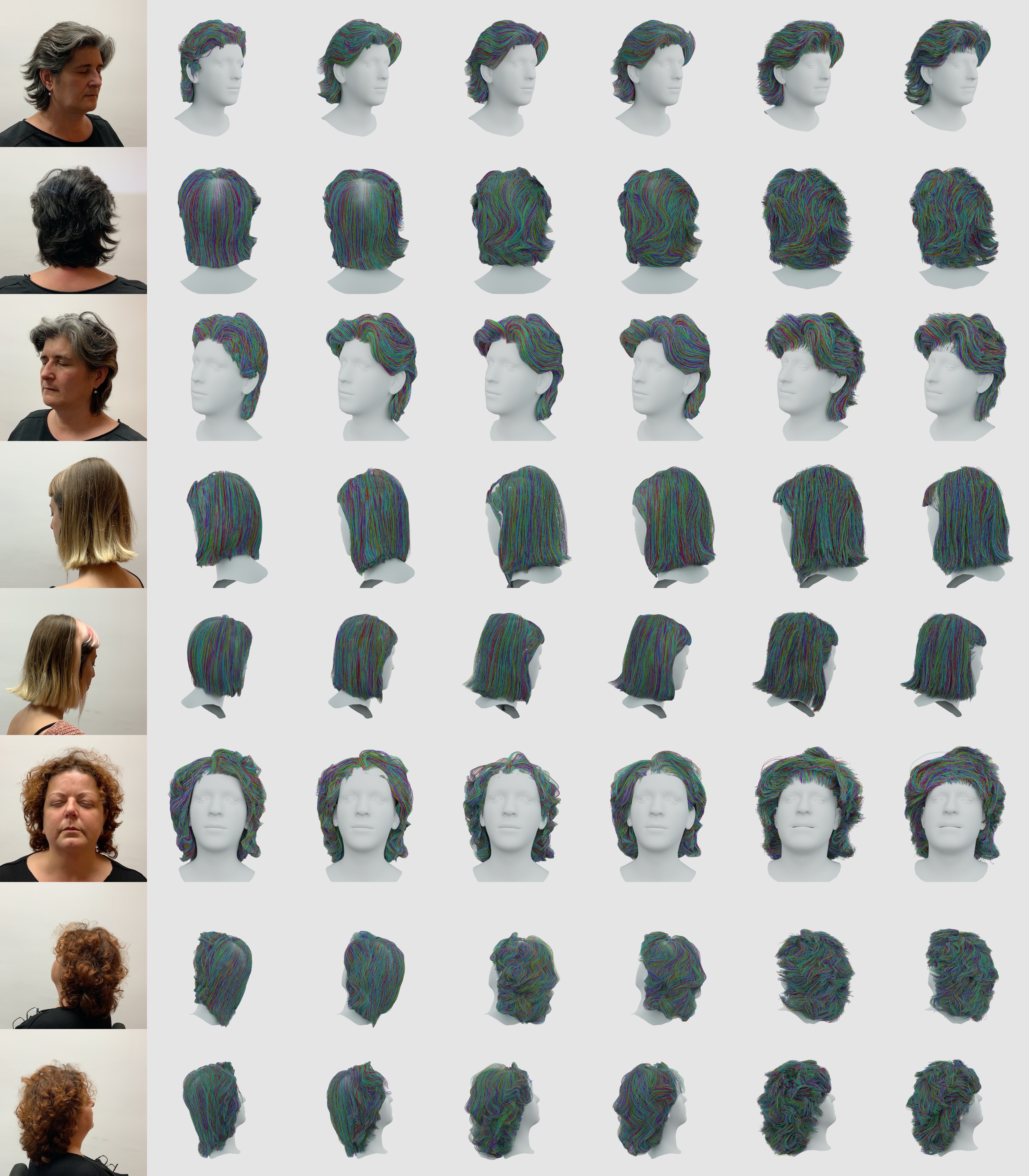} 
    \vspace{0.1cm}
\makebox[0.14\textwidth]{Image}%
\makebox[0.14\textwidth]{Ours (1 view)}%
\makebox[0.14\textwidth]{Ours (3 views)}
\makebox[0.14\textwidth]{Ours (8 views)}%
\makebox[0.14\textwidth]{Ours (32 views)}%
\makebox[0.14\textwidth]{GH (8 views)}
\makebox[0.14\textwidth]{GH (32 views)}%

    \caption{\textbf{Extended qualitative comparison using real-world multi-view scenes~\cite{h3ds} with Gaussian Haircut (GH)~\cite{GaussianHaircut}}. We compared in a scenario with 1, 3, 8, and 32 views available. Note, GH fails in scenarios with 1 and 3 views. Digital zoom-in is recommended.}
    \label{fig:mv_suppmat_comparison}
\end{figure*}

~
In Figure~\ref{fig:mv_suppmat_comparison}, we show reconstruction results of our method using different numbers of views, such as $1$, $3$, $8$, and $32$ on H3DS dataset~\cite{h3ds}.
We compare our approach to Gaussian Haircut (GH)~\cite{GaussianHaircut} for the $8$- and $32$-view cases, as GH fails to produce reasonable results with only $1$ or $3$ views.
For a fair comparison, we lunch GH using its official repository and default configuration settings.
~
Our optimization process involves $400$ steps for the $1$ view case, $800$ steps for $3$ views, and $1600$ steps for $8$ and $32$ views.
Running all $1600$ steps takes approximately $45$ minutes, significantly faster than GH, which takes around $10$ hours. 
~

In Figure~\ref{fig:mv_suppmat_comparison}, GH has some freezing issues with strands that happen because of direct optimization for directions in 3D space.
Also, it has some issues with the scalp mask, which may be resolved with an improved scalp estimator algorithm.
Our method has some smoothing in the results because of the use of interpolation. This could be potentially resolved by using a learnable neural interpolation scheme during the optimization stage.
We notice an improved curly geometry for our method (see Figure~\ref{fig:mv_suppmat_comparison}, last 3 rows).
While the hairstyle in GH appears more detailed, some strands exhibit noisy structures.
Potentially, both approaches could be combined to achieve more detailed hair while staying within the hairstyle prior to prevent unrealistic structures.

\subsection{Simulations}

For simulation results, we use Unreal Engine~\cite{unrealengine}. 
To do that, we convert hairstyles into Alembic format and import as a groom into Unreal.
For simulation, we use around $30,000$ strands.

\begin{figure}
    \begin{center}
    \includegraphics[width=0.15\textwidth]{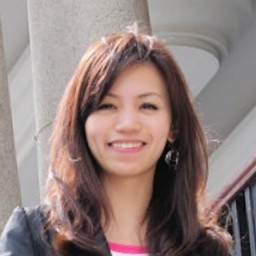}  
    \includegraphics[width=0.15\textwidth]{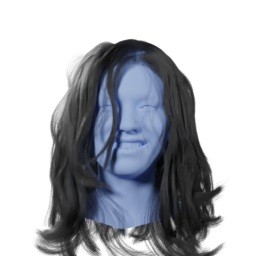}  
    \includegraphics[trim={120 240 120 0},clip,width=0.15\textwidth]{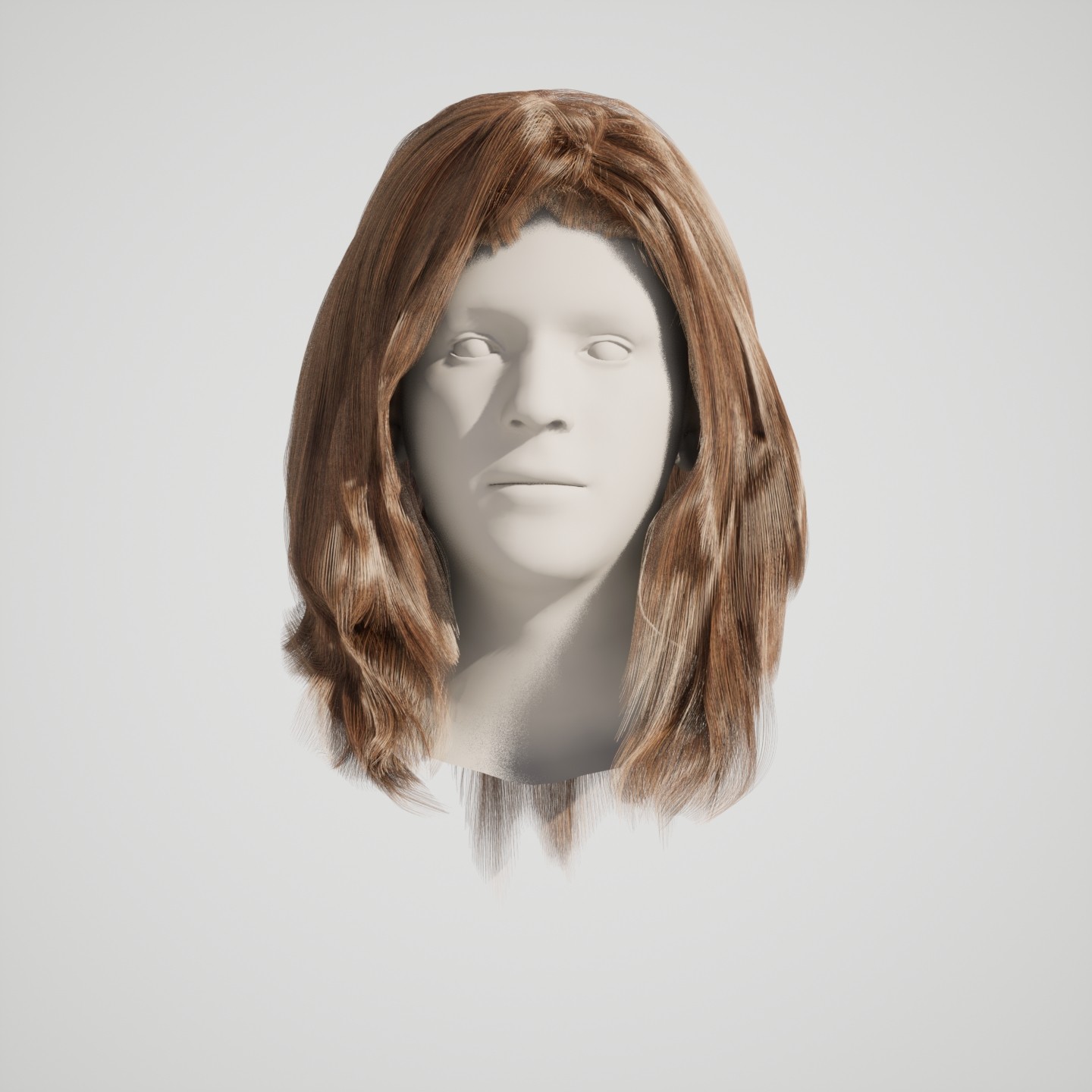}  \\
    
    \includegraphics[width=0.15\textwidth]{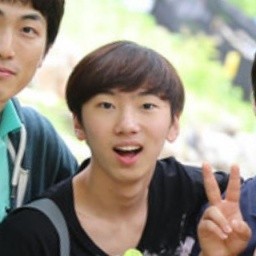}  
    \includegraphics[width=0.15\textwidth]{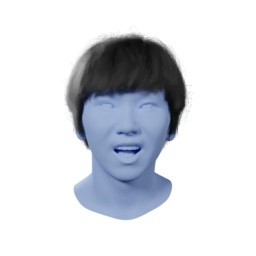}
    \includegraphics[trim={120 240 120 0},clip,width=0.15\textwidth]{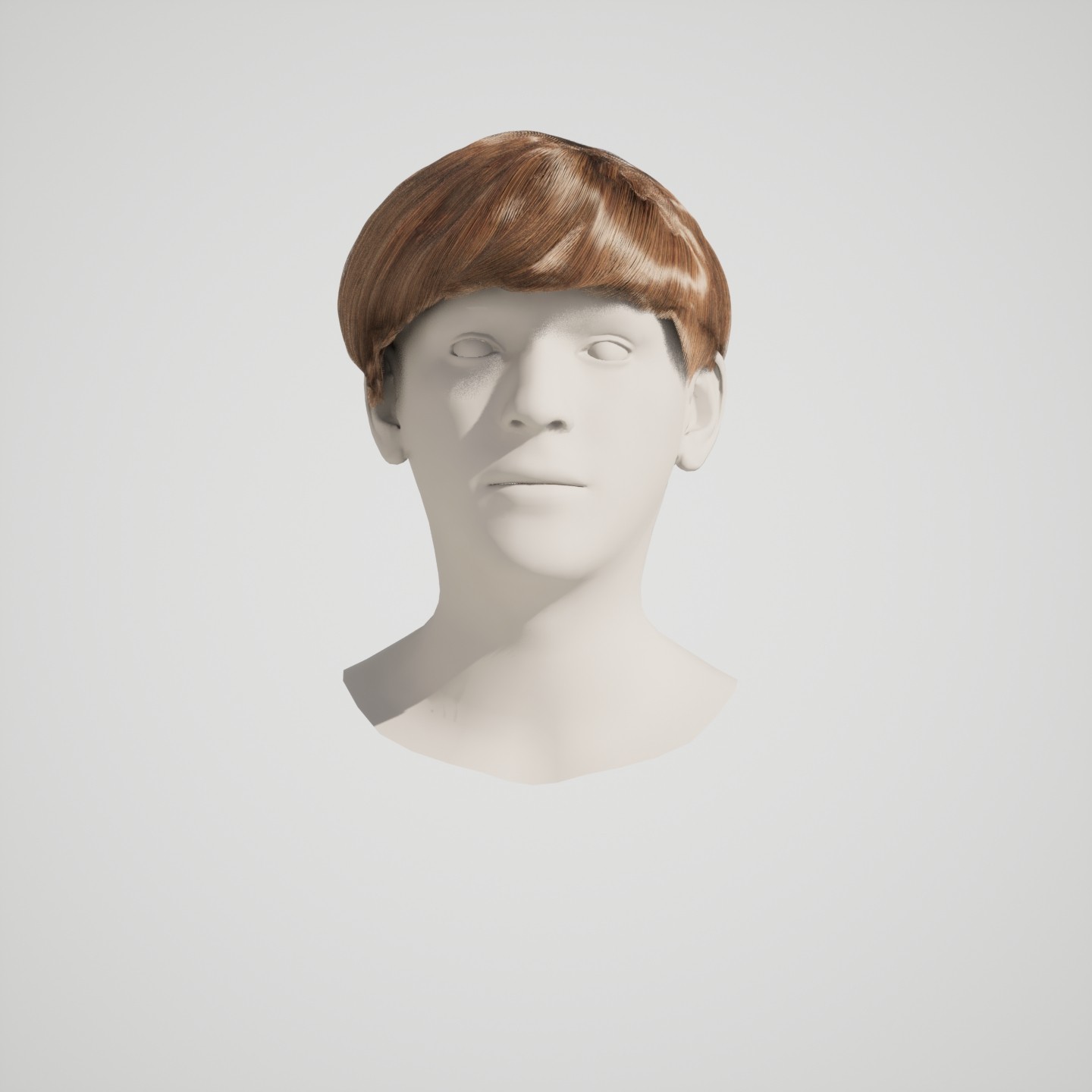}  \\
  
    \end{center}
    \vspace{-0.4cm}
    \makebox[0.3\linewidth]{Image}%
    \makebox[0.36\linewidth]{Hairmony~\cite{hairmony}}%
    \makebox[0.35\linewidth]{Ours}%
    
    \caption{\textbf{Comparison} of our model with retrieval-based method Hairmony~\cite{hairmony} .}\label{fig:comaprison_hairmony}
\end{figure}

\begin{figure}[ht]
    \centering
    \begin{subfigure}{0.16\textwidth}
        \includegraphics[trim=140 140 50 50,clip,width=\textwidth]{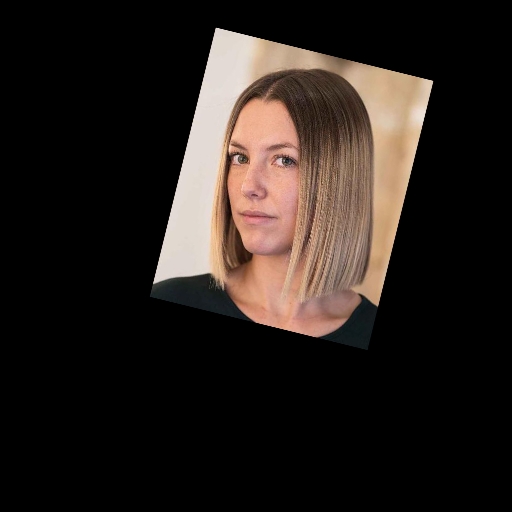}
        \caption*{Image}
    \end{subfigure}
    \begin{subfigure}{0.16\textwidth}
        \includegraphics[trim=140 140 50 50,clip,width=\textwidth]{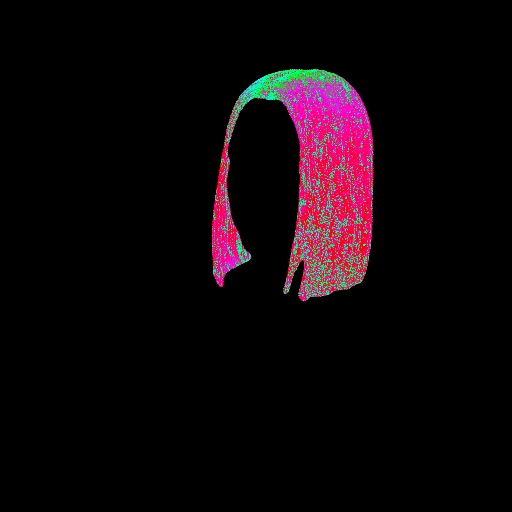}
        \caption*{Ground-truth}
    \end{subfigure}

    \begin{subfigure}{0.16\textwidth}
        \includegraphics[trim=140 140 50 50,clip,width=\textwidth]{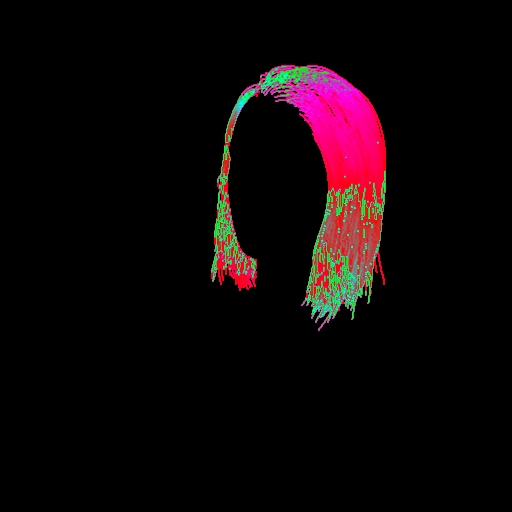}
        \caption*{w Mixing}
    \end{subfigure}
    \begin{subfigure}{0.16\textwidth}
        \includegraphics[trim=140 140 50 50,clip,width=\textwidth]{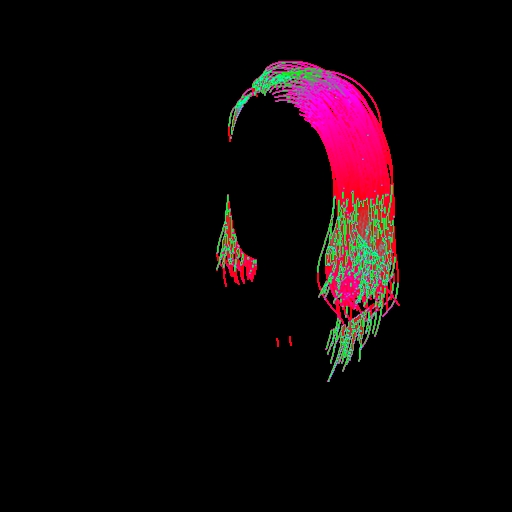}
        \caption*{w/o Mixing}
    \end{subfigure}

    \vspace{-0.1cm}
    \caption{\textbf{Mixing strategy.} The performance of the Hairstyle prior model with and without mixing strategy. The model trained on real images can better regress the hair silhouette and orientations. The color in the image corresponds to the direction in the orientation maps.}
    \label{fig:mixing_strategy_color_image_regressor}
\end{figure}

\begin{figure*}
    \setlength{\tabcolsep}{0pt} %
    \renewcommand{\arraystretch}{0.0} %

\resizebox{\textwidth}{!}{
    \begin{tabular}{@{}ccccccccc}
        \multirow{2}{*}[0.481in]{\includegraphics[width=0.14\textwidth]{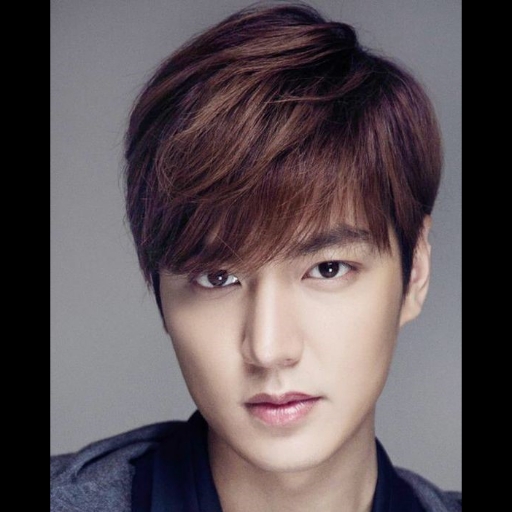}} 
            &
        \multirow{2}{*}[0.481in]{\includegraphics[trim={125 200 125 50},clip,width=0.14\textwidth]{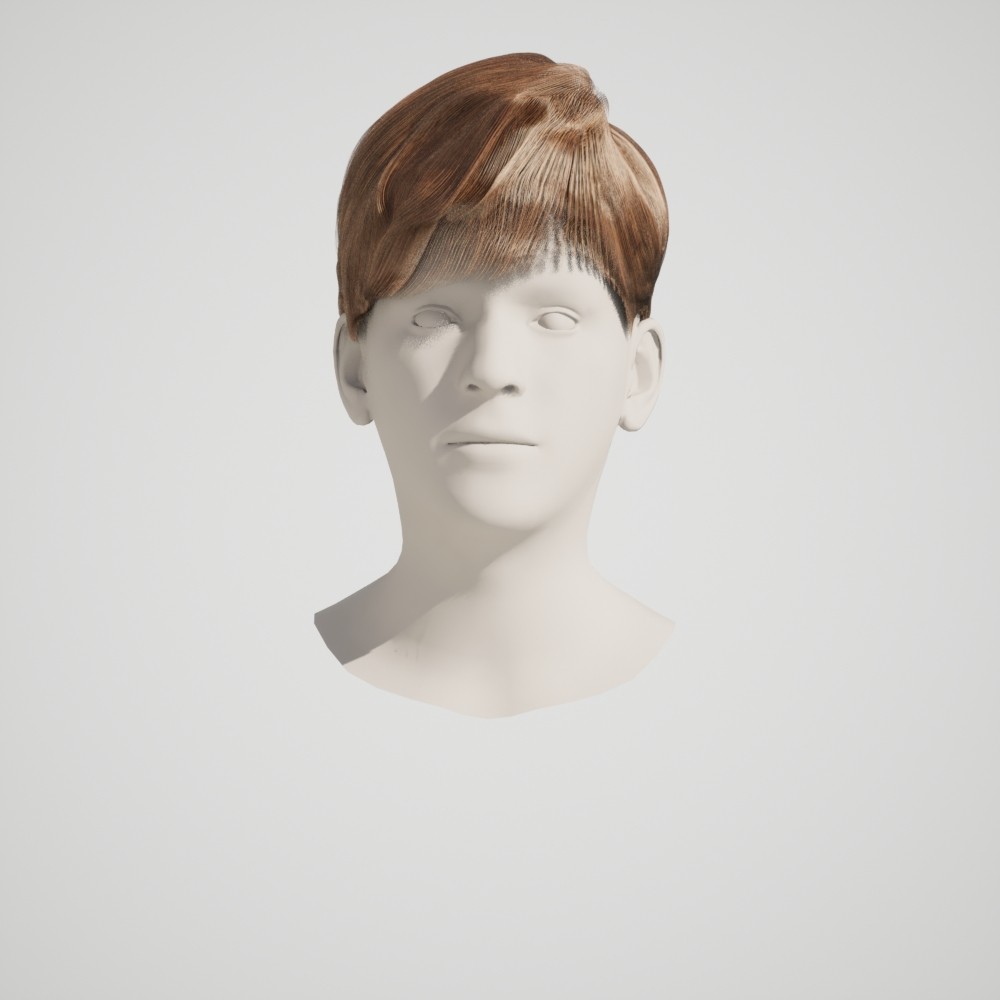}} 
        &
        \includegraphics[width=0.07\textwidth]{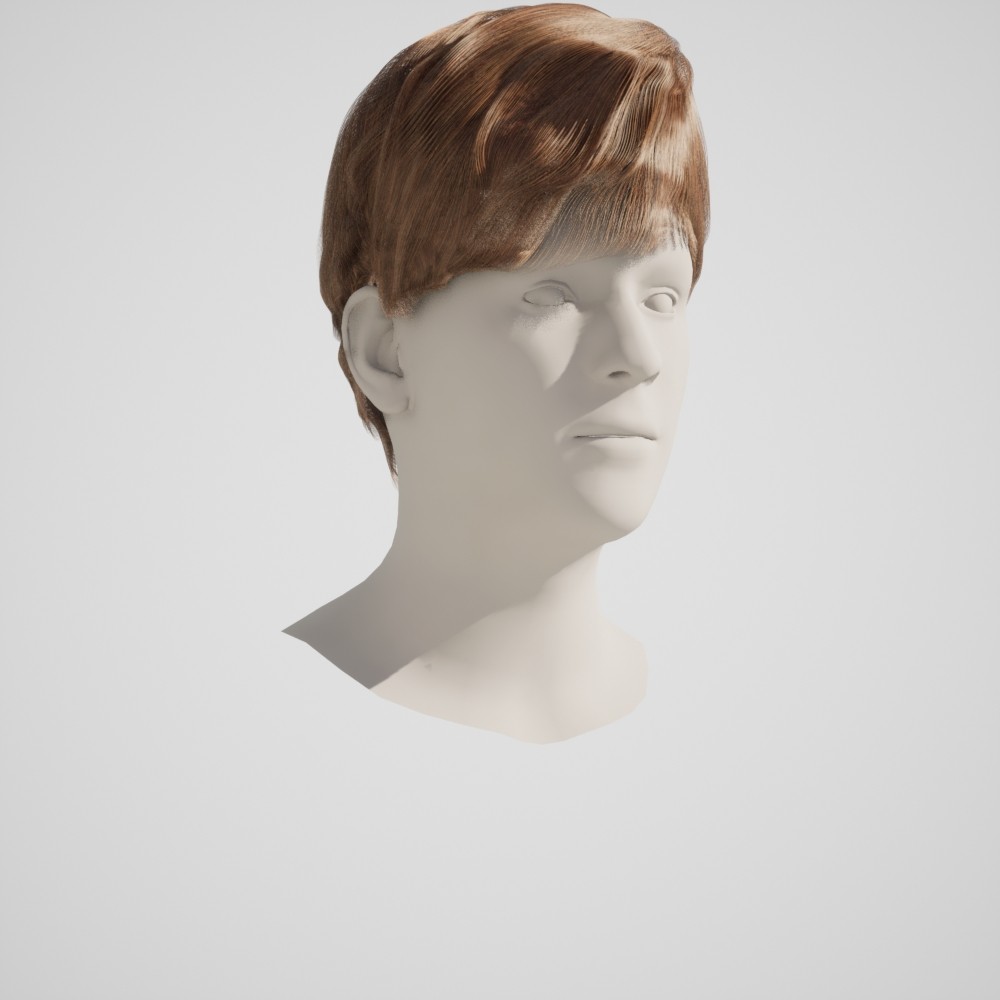} 
        &
        \multirow{2}{*}[0.481in]{\includegraphics[trim={125 200 125 50},clip,width=0.14\textwidth]{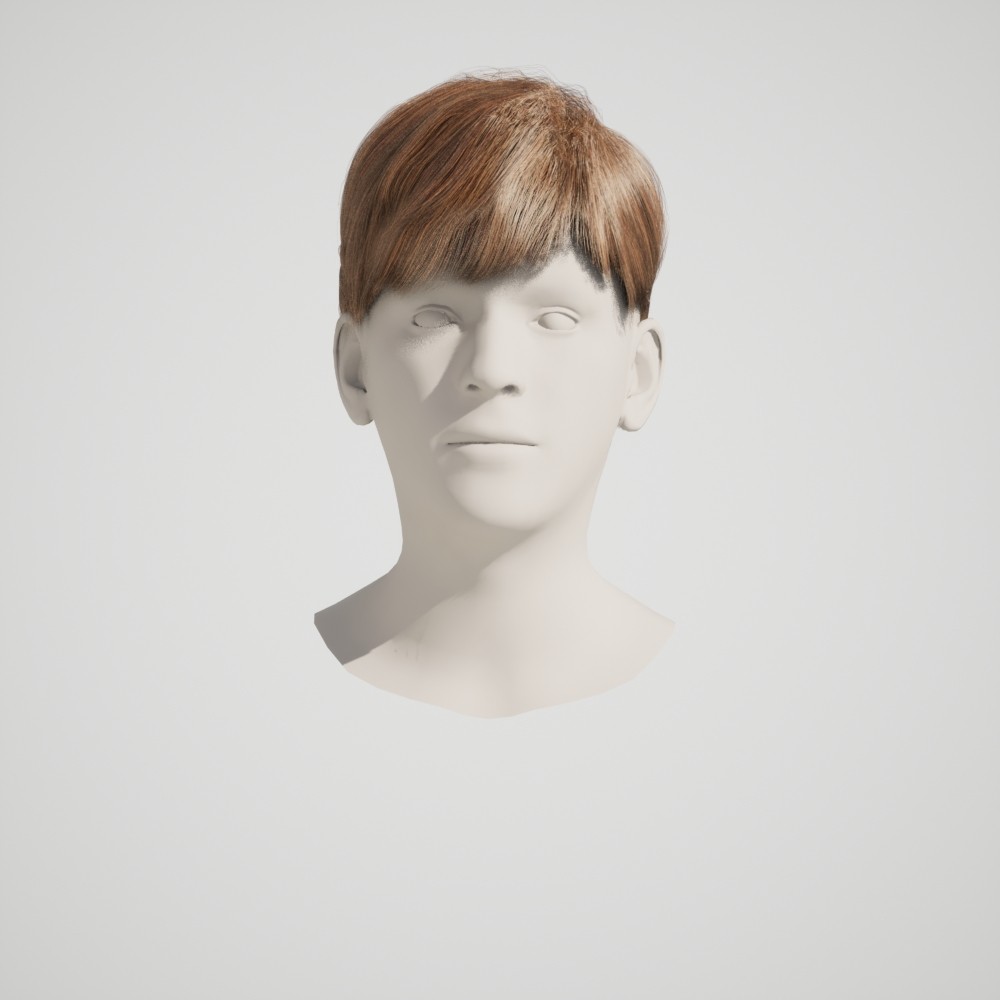}} 
        &
        \includegraphics[width=0.07\textwidth]{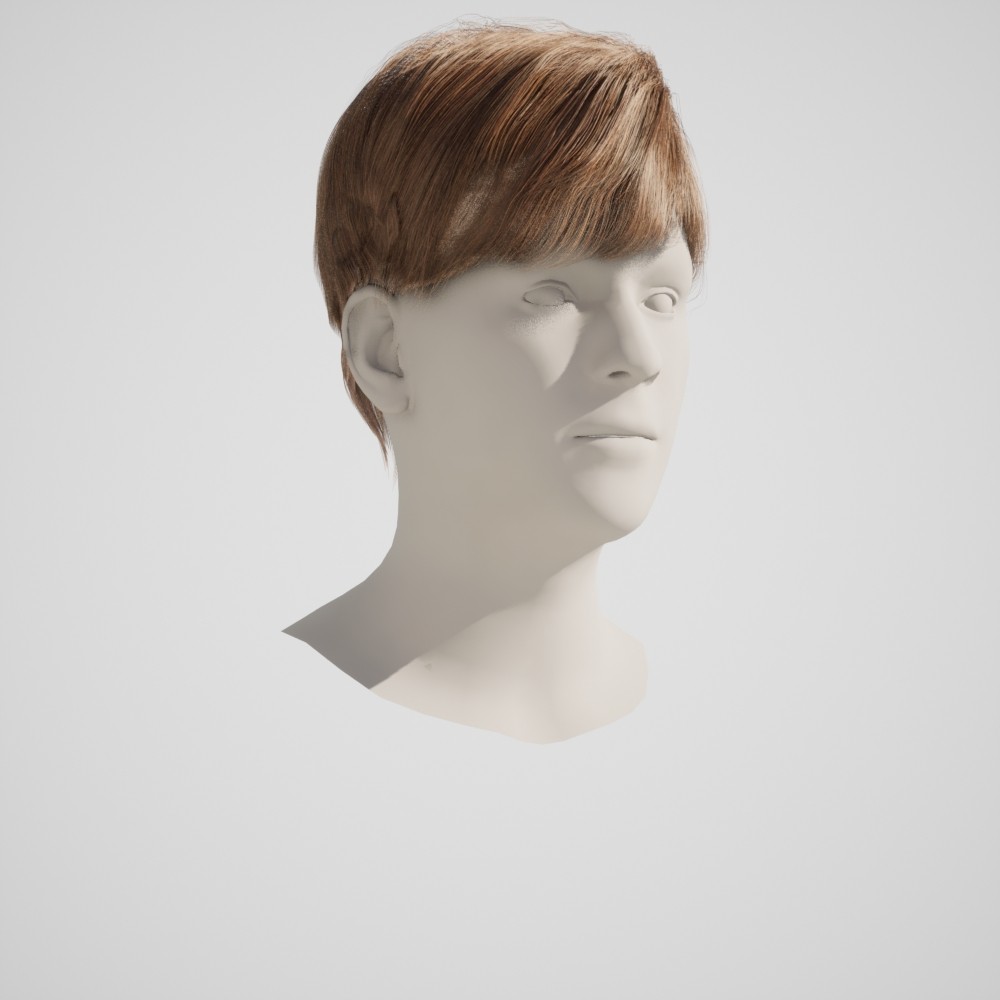} 
        &
        \multirow{2}{*}[0.481in]{\includegraphics[trim={125 200 125 50},clip,width=0.14\textwidth]{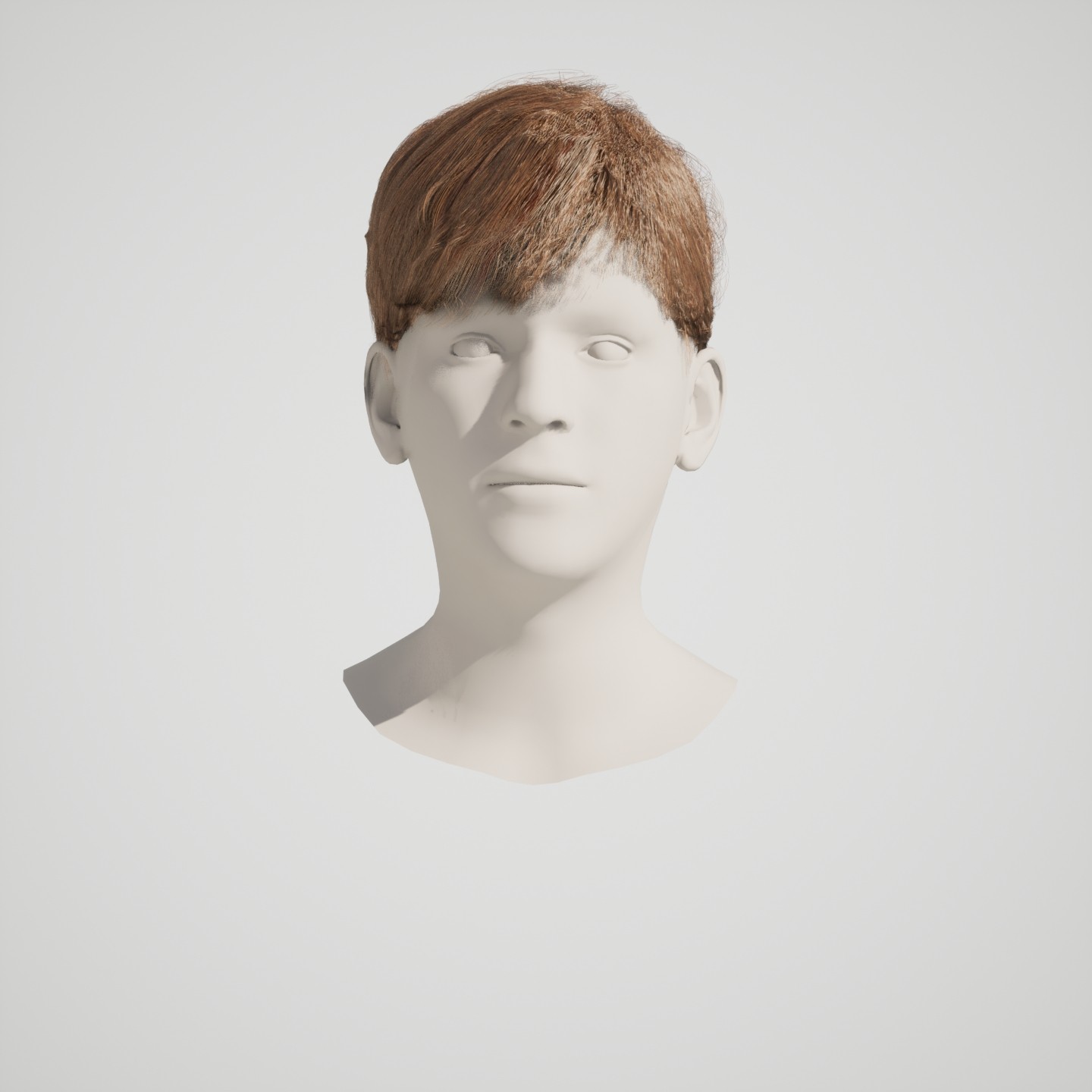}} 
        &
        \includegraphics[width=0.07\textwidth]{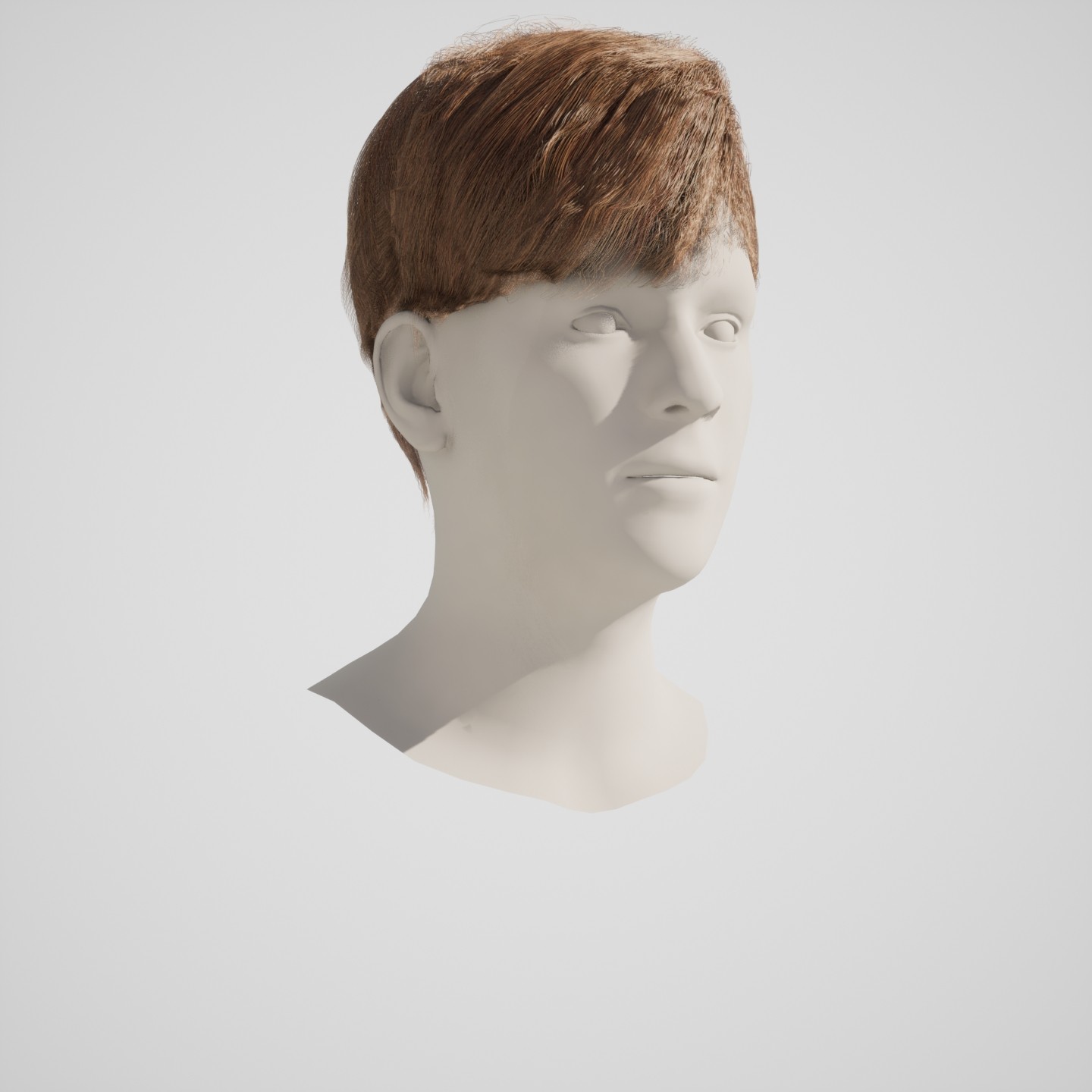} 
       &
        \multirow{2}{*}[0.481in]{\includegraphics[trim={125 200 125 50},clip,width=0.14\textwidth]{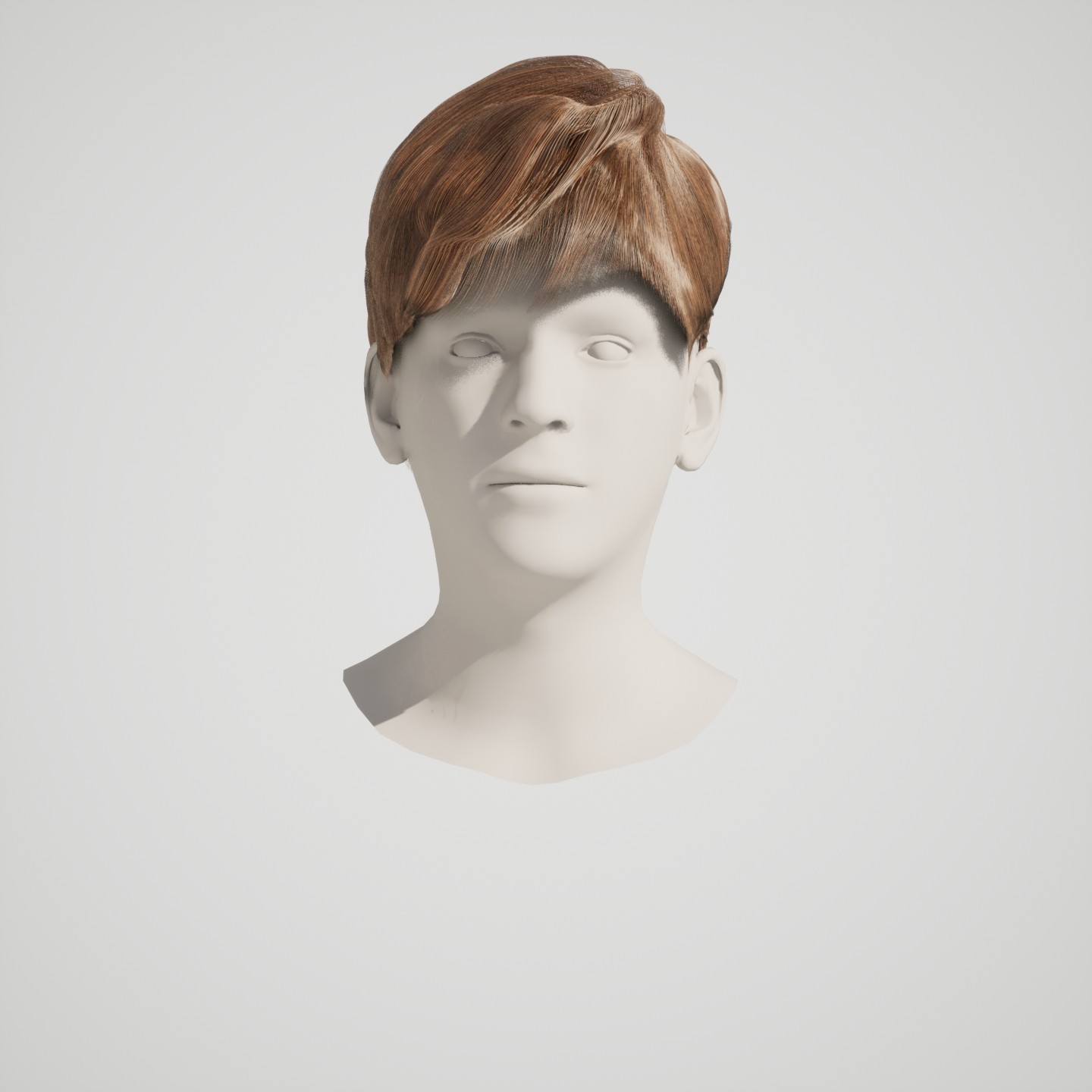}} 
        &
        \includegraphics[width=0.07\textwidth]{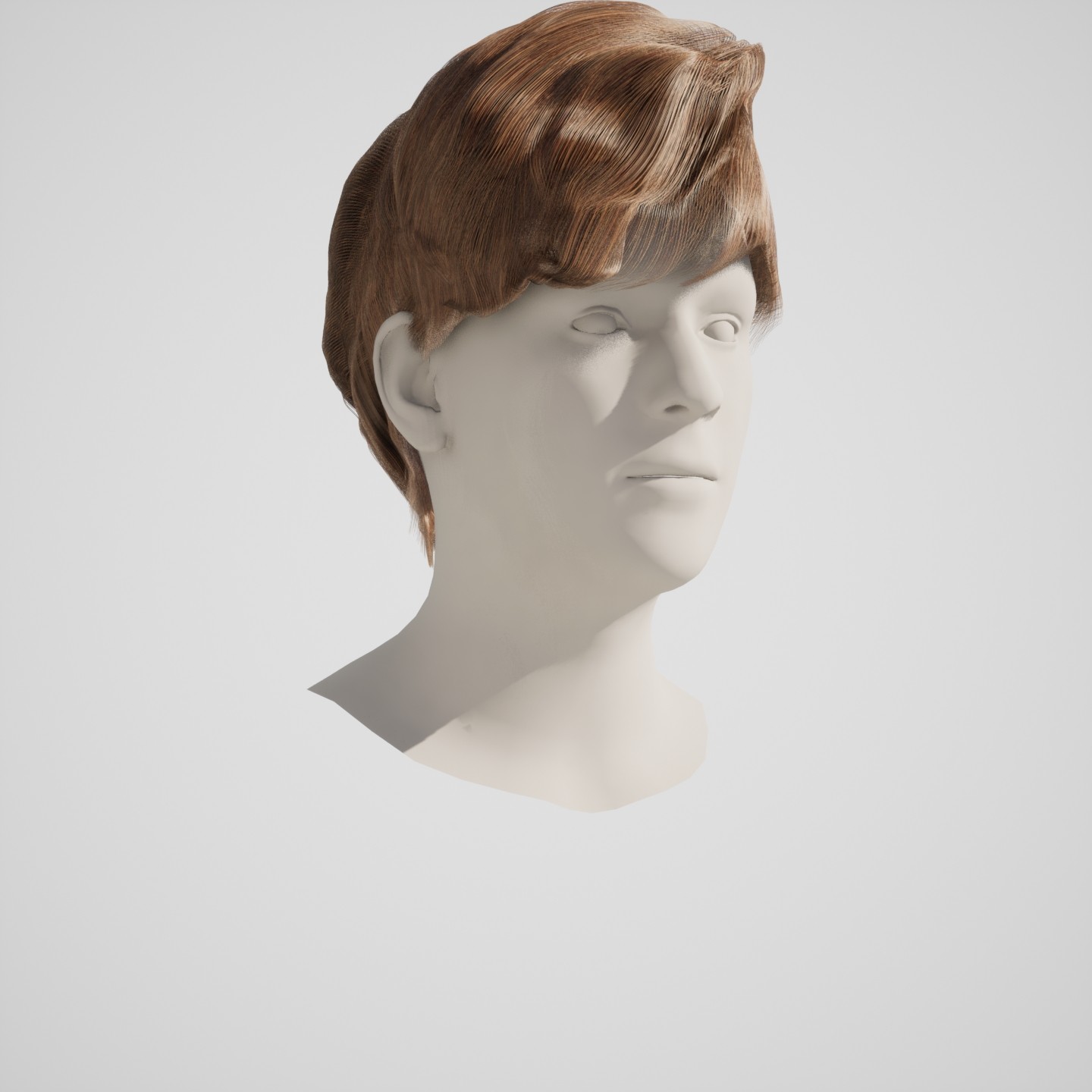}

        \\

        &
        &
        \includegraphics[width=0.07\textwidth]{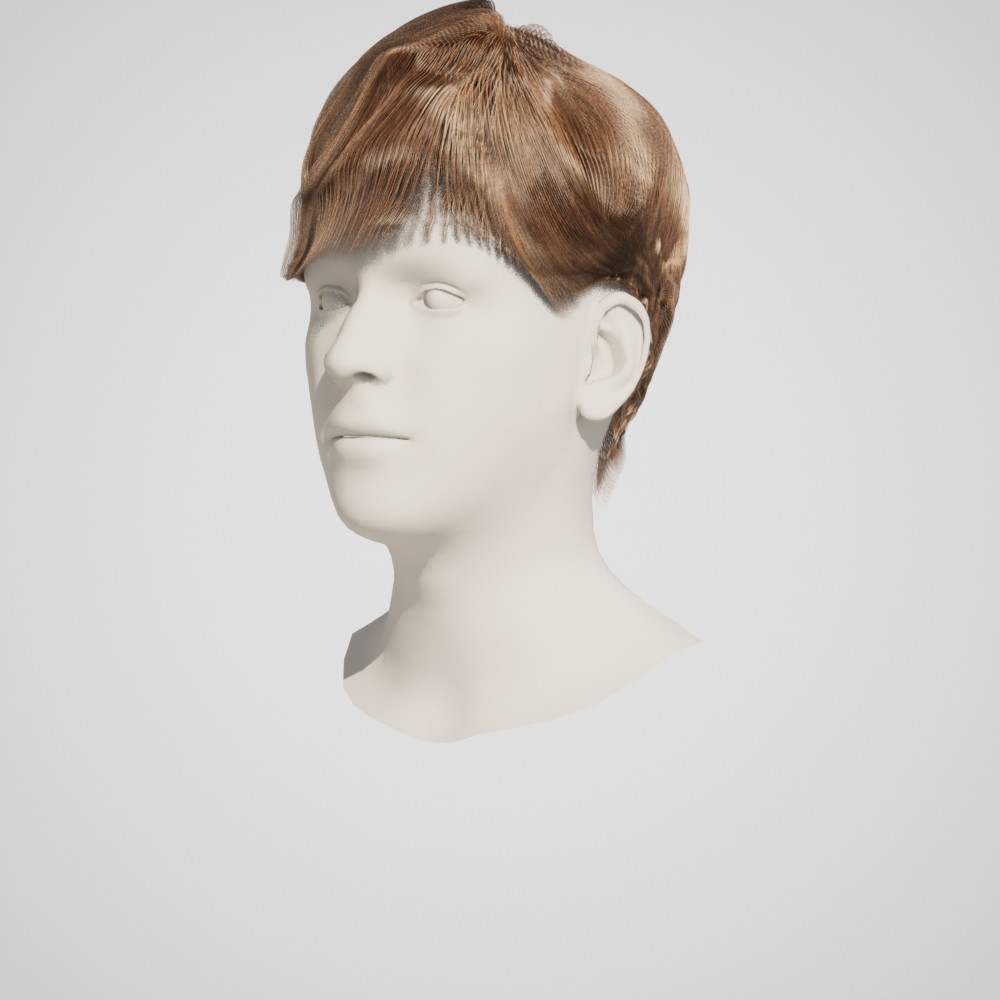} 
        &
        &
        \includegraphics[width=0.07\textwidth]{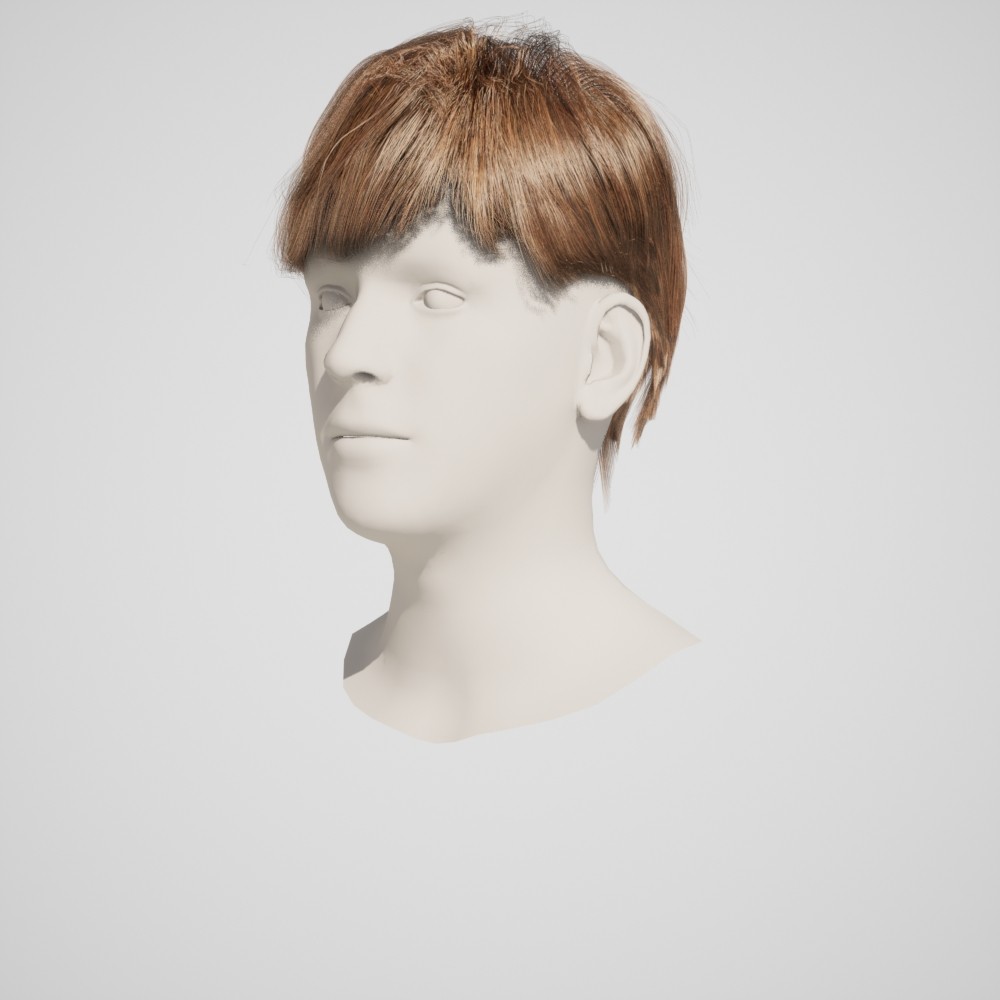} 
        & 
        &
       \includegraphics[width=0.07\textwidth]{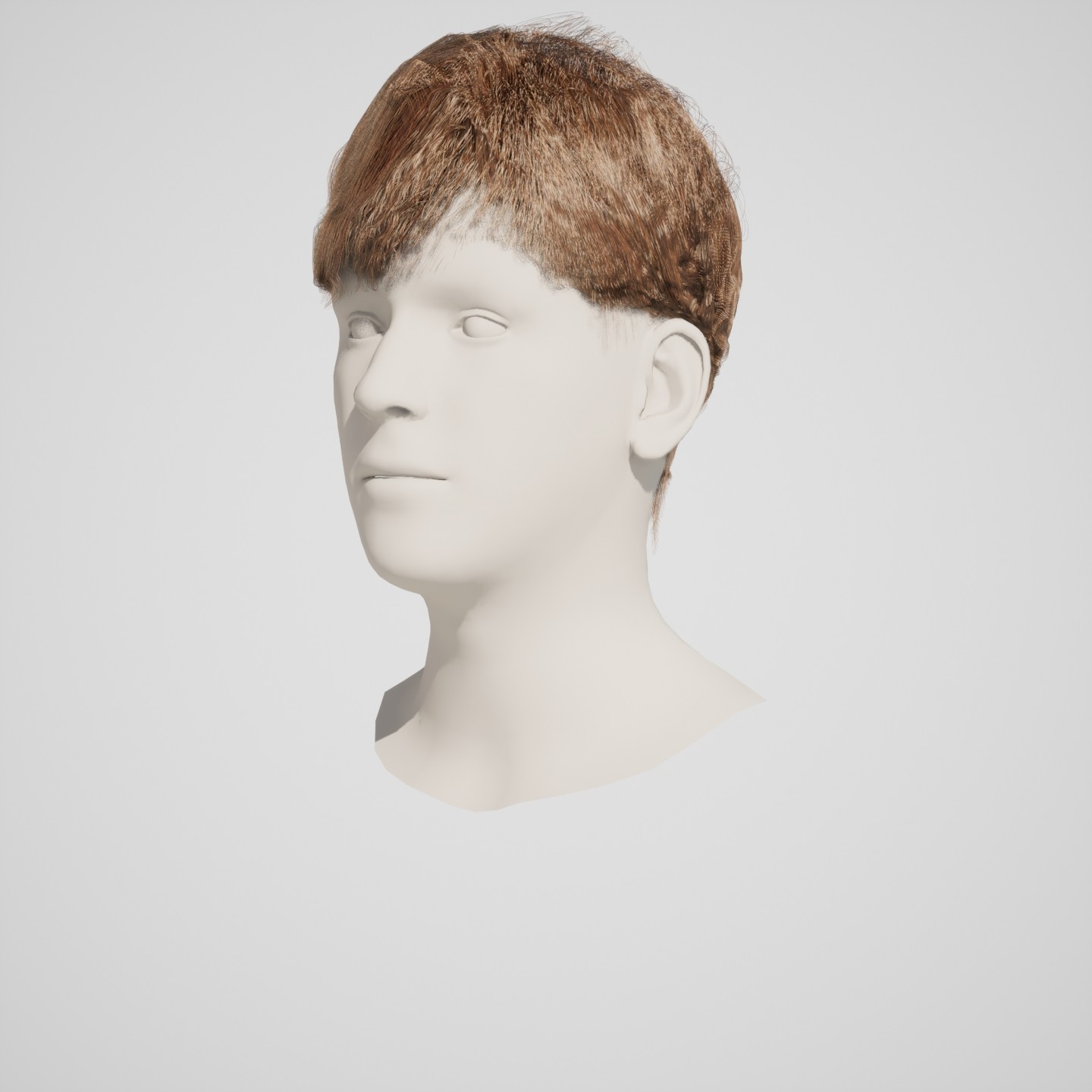}  

        & 
        &
        \includegraphics[width=0.07\textwidth]{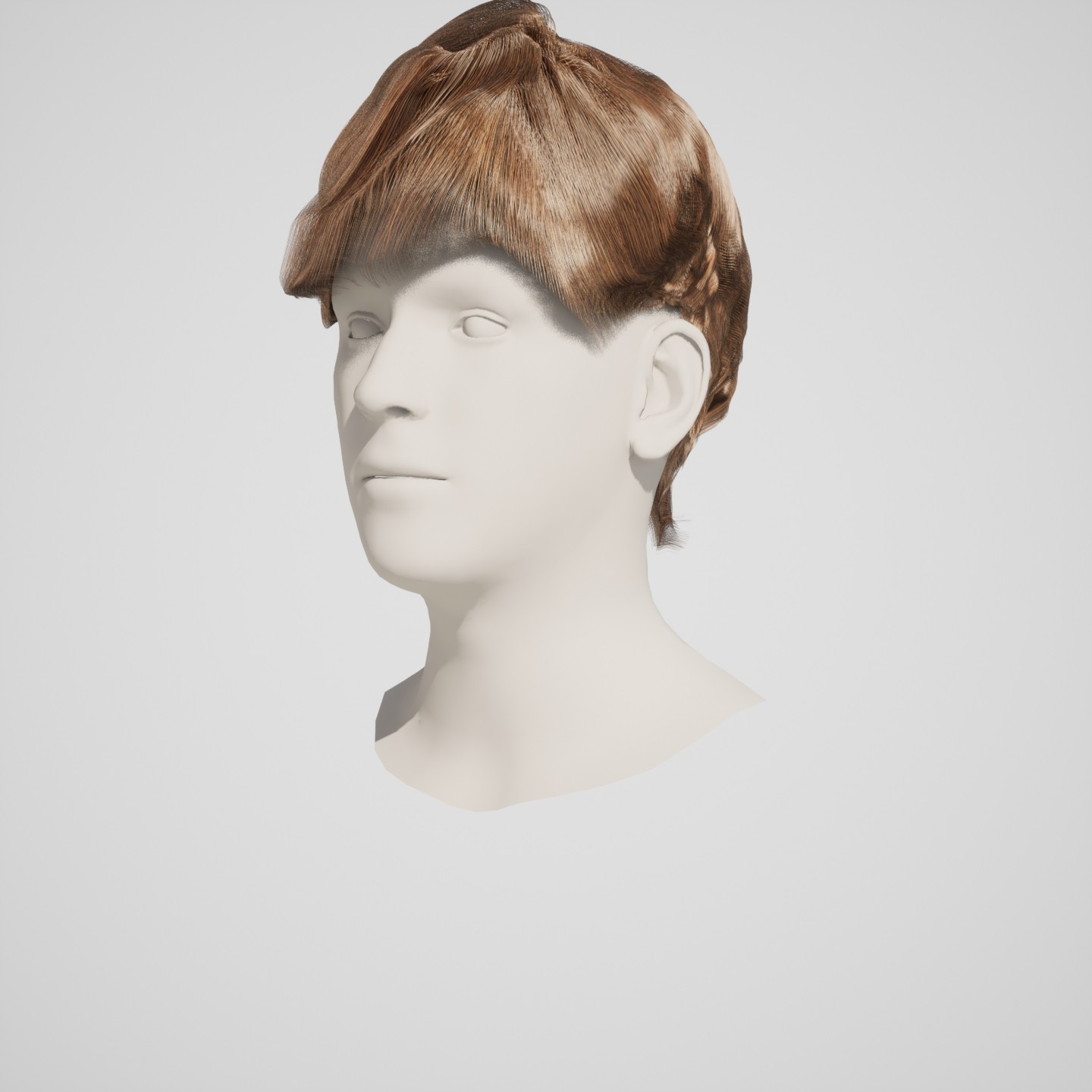}

        \\

                \multirow{2}{*}[0.481in]{\includegraphics[width=0.14\textwidth]{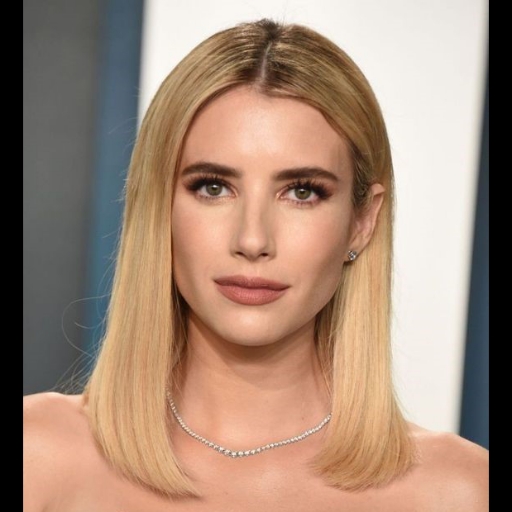}} 
            &
        \multirow{2}{*}[0.481in]{\includegraphics[trim={125 200 125 50},clip,width=0.14\textwidth]{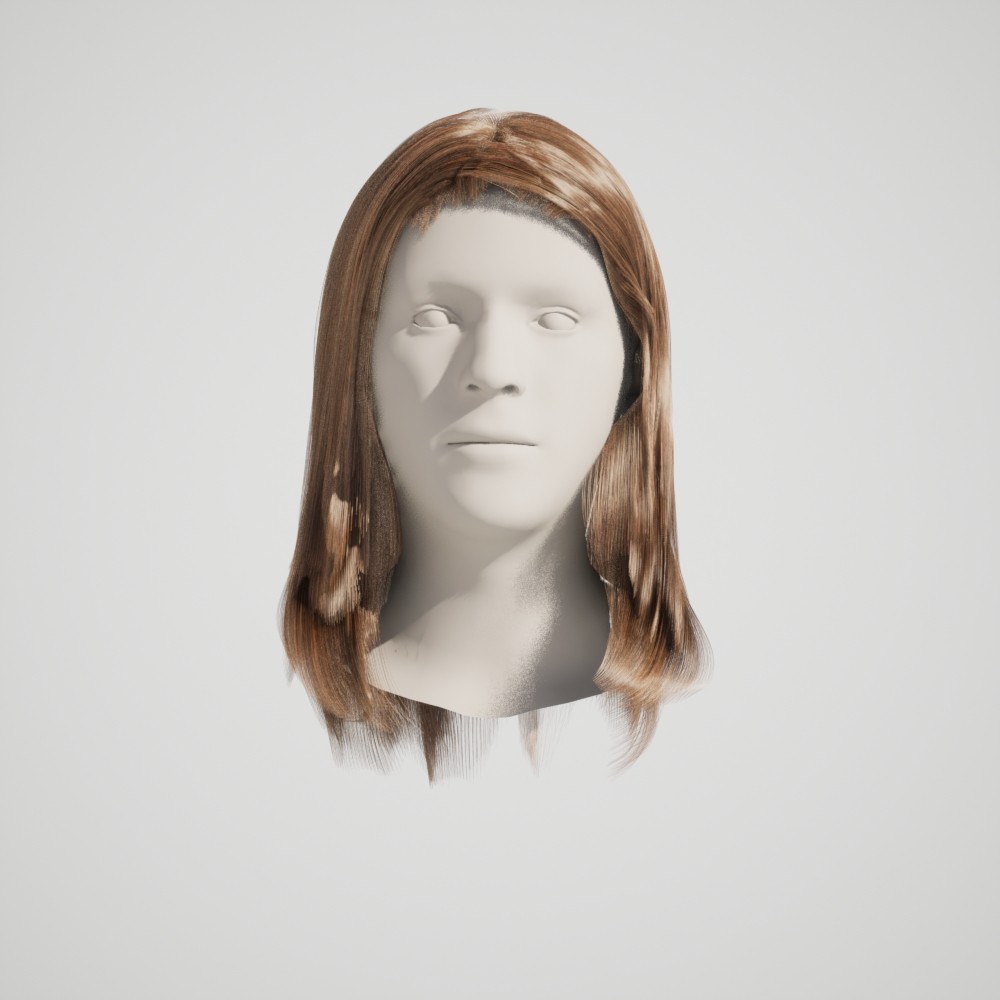}} 
        &
        \includegraphics[width=0.07\textwidth]{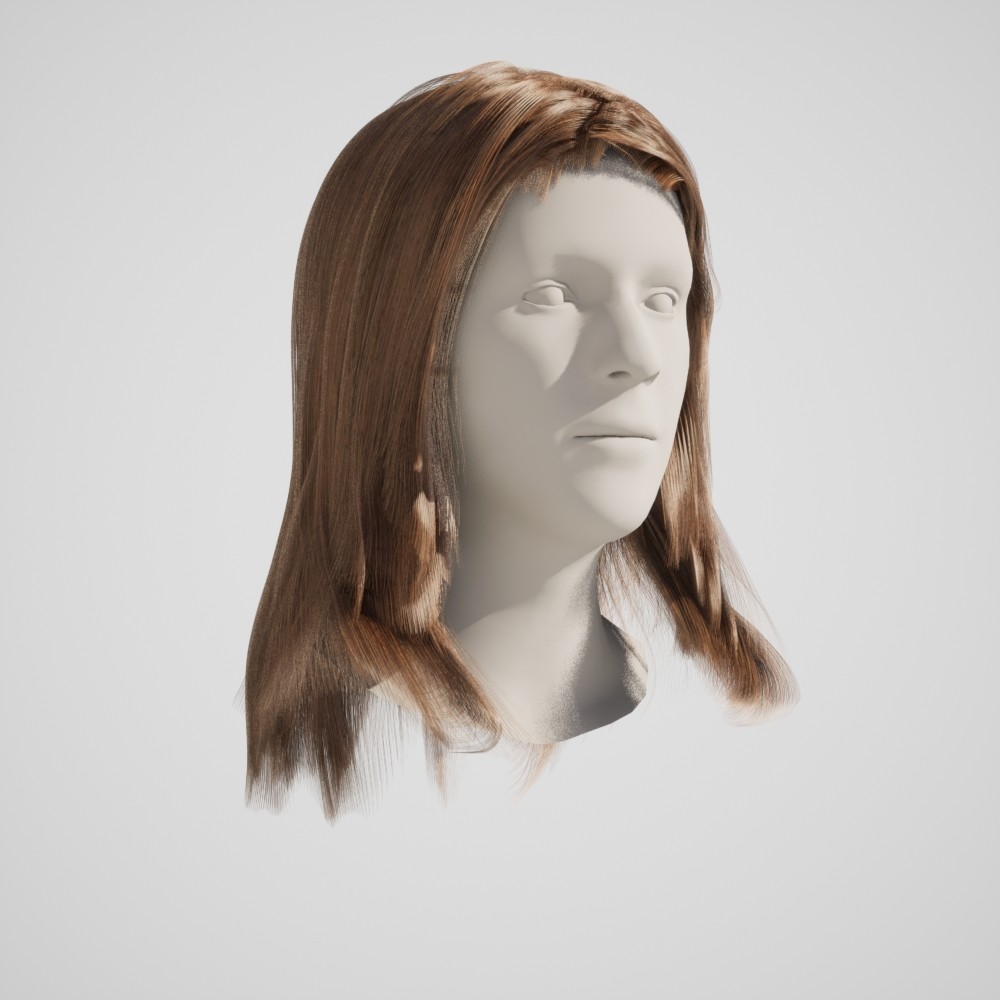} 
        &
        \multirow{2}{*}[0.481in]{\includegraphics[trim={125 200 125 50},clip,width=0.14\textwidth]{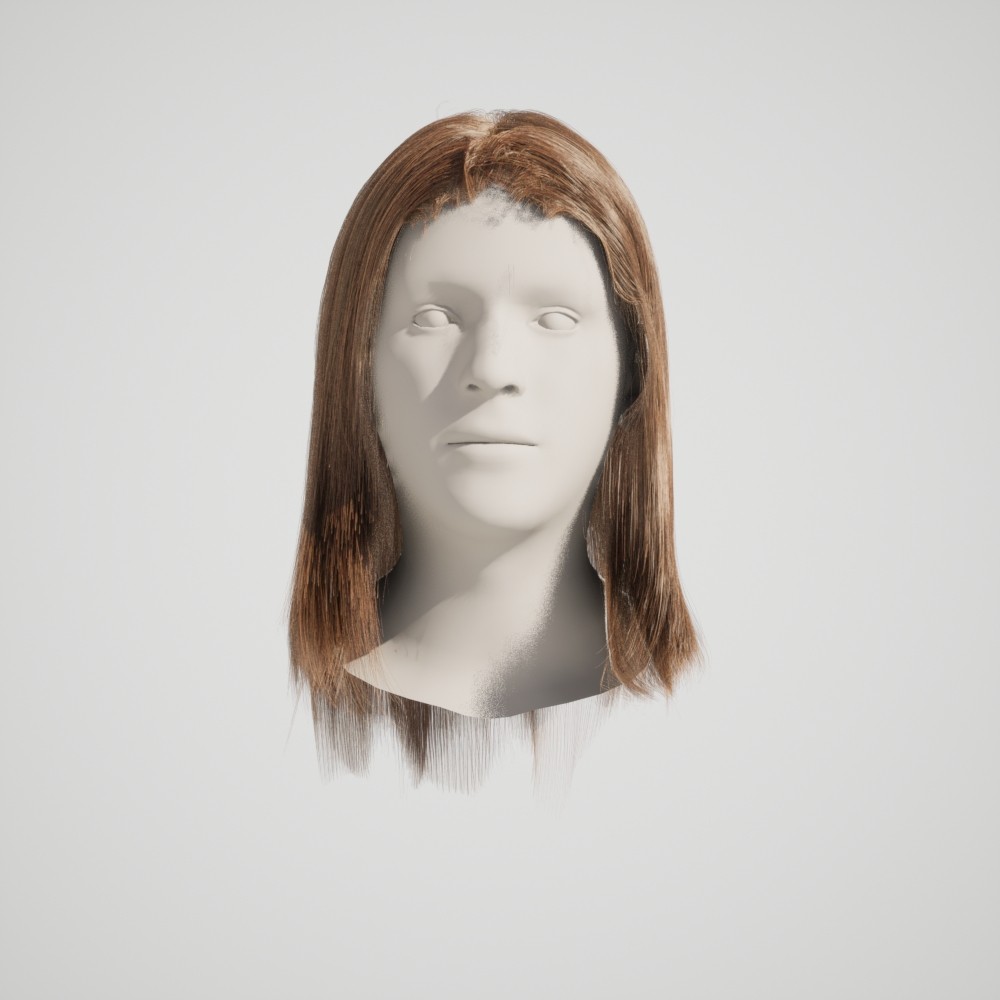}} 
        &
        \includegraphics[width=0.07\textwidth]{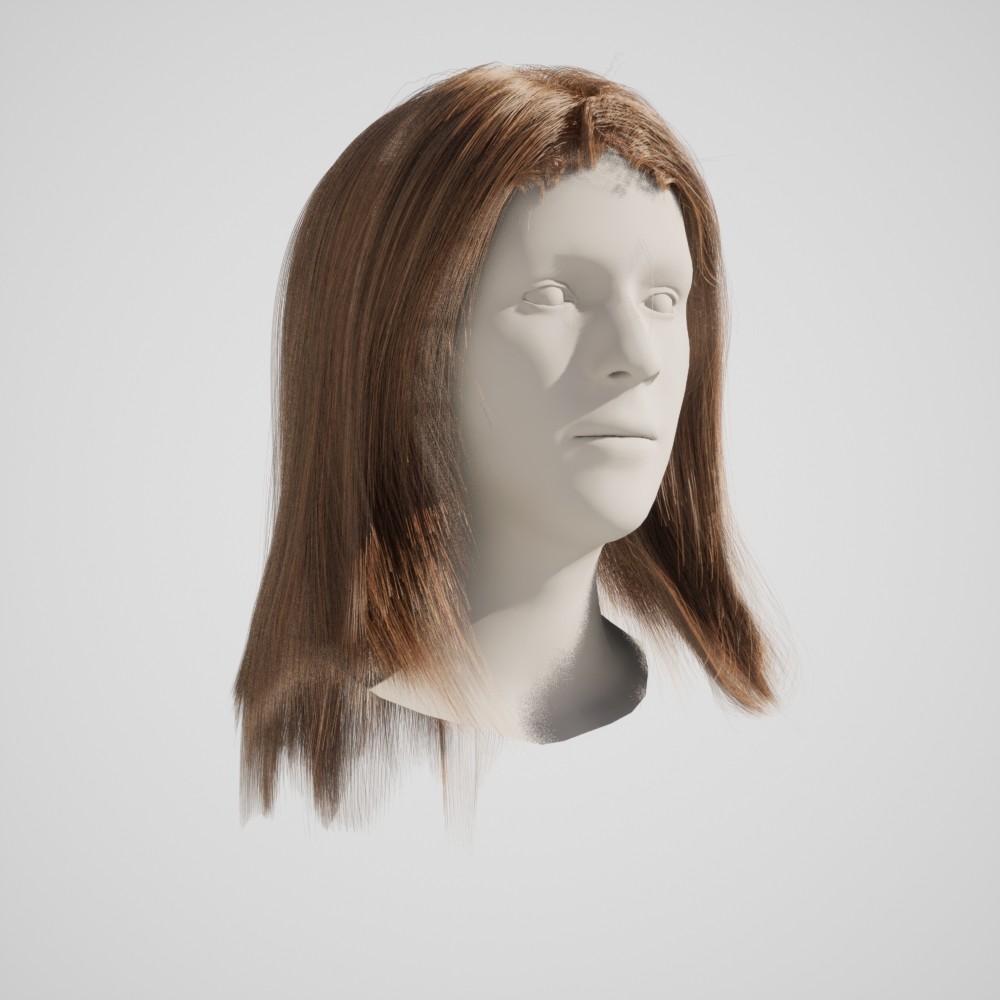} 
        &
        \multirow{2}{*}[0.481in]{\includegraphics[trim={125 200 125 50},clip,width=0.14\textwidth]{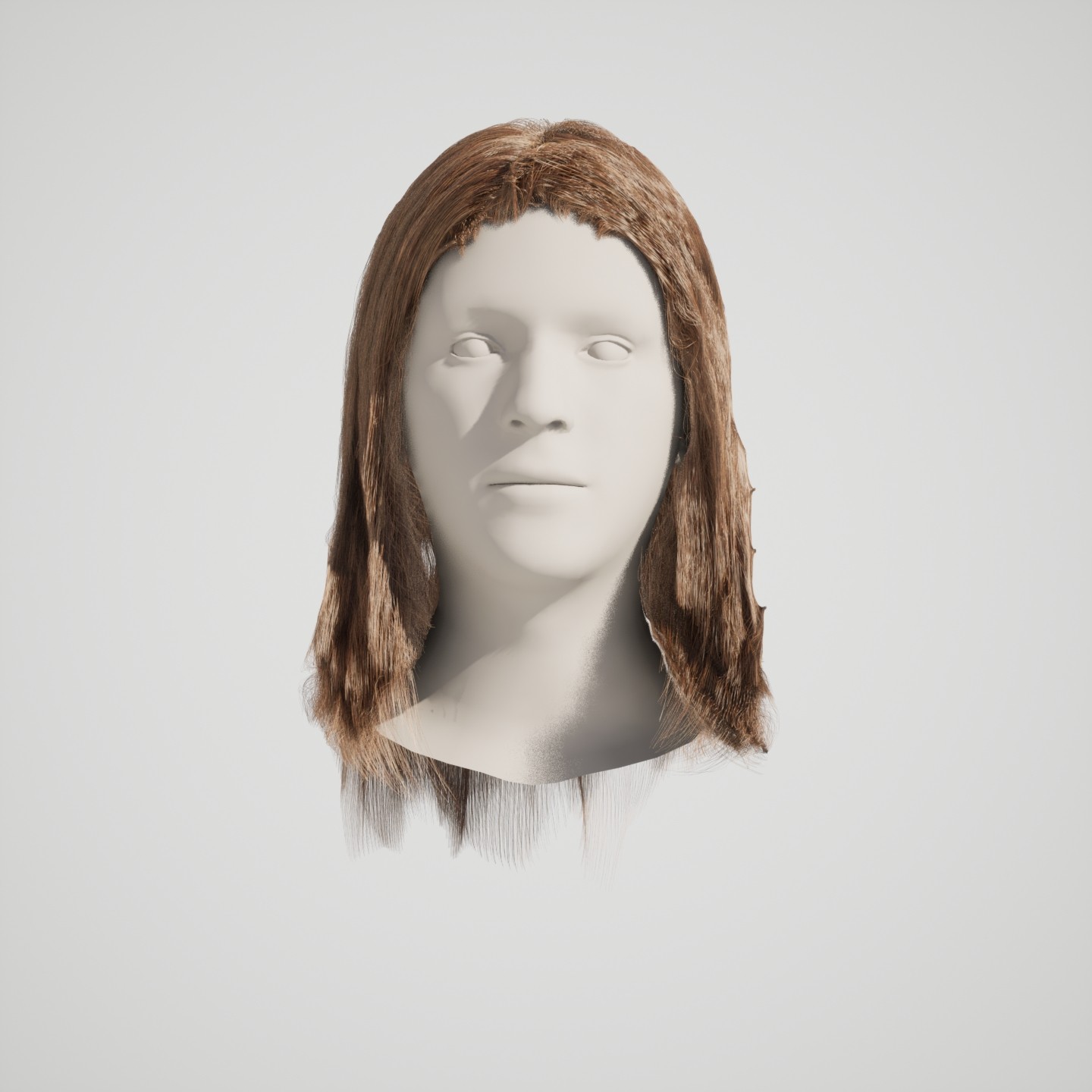}} 
        &
        \includegraphics[width=0.07\textwidth]{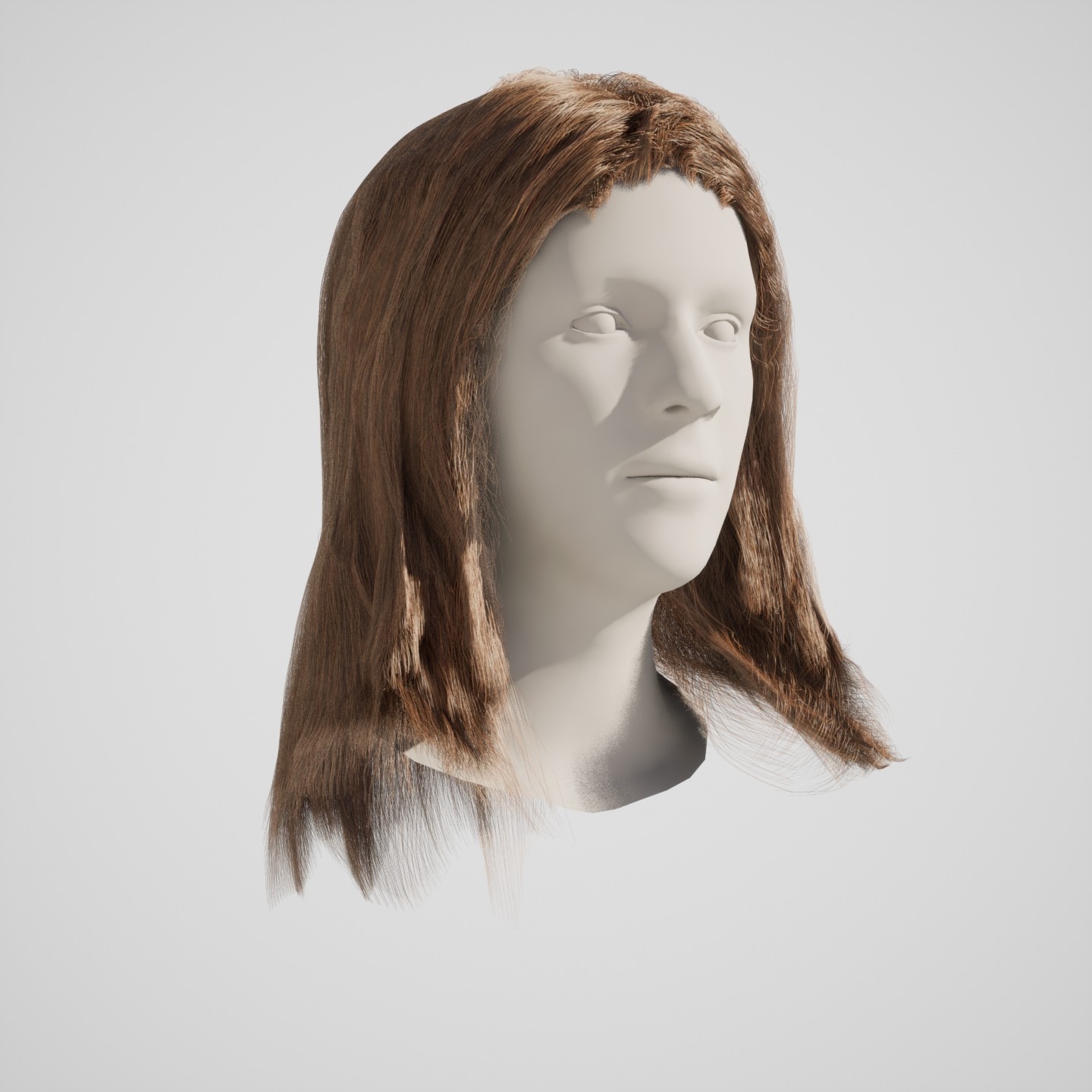} 
       &
        \multirow{2}{*}[0.481in]{\includegraphics[trim={125 200 125 50},clip,width=0.14\textwidth]{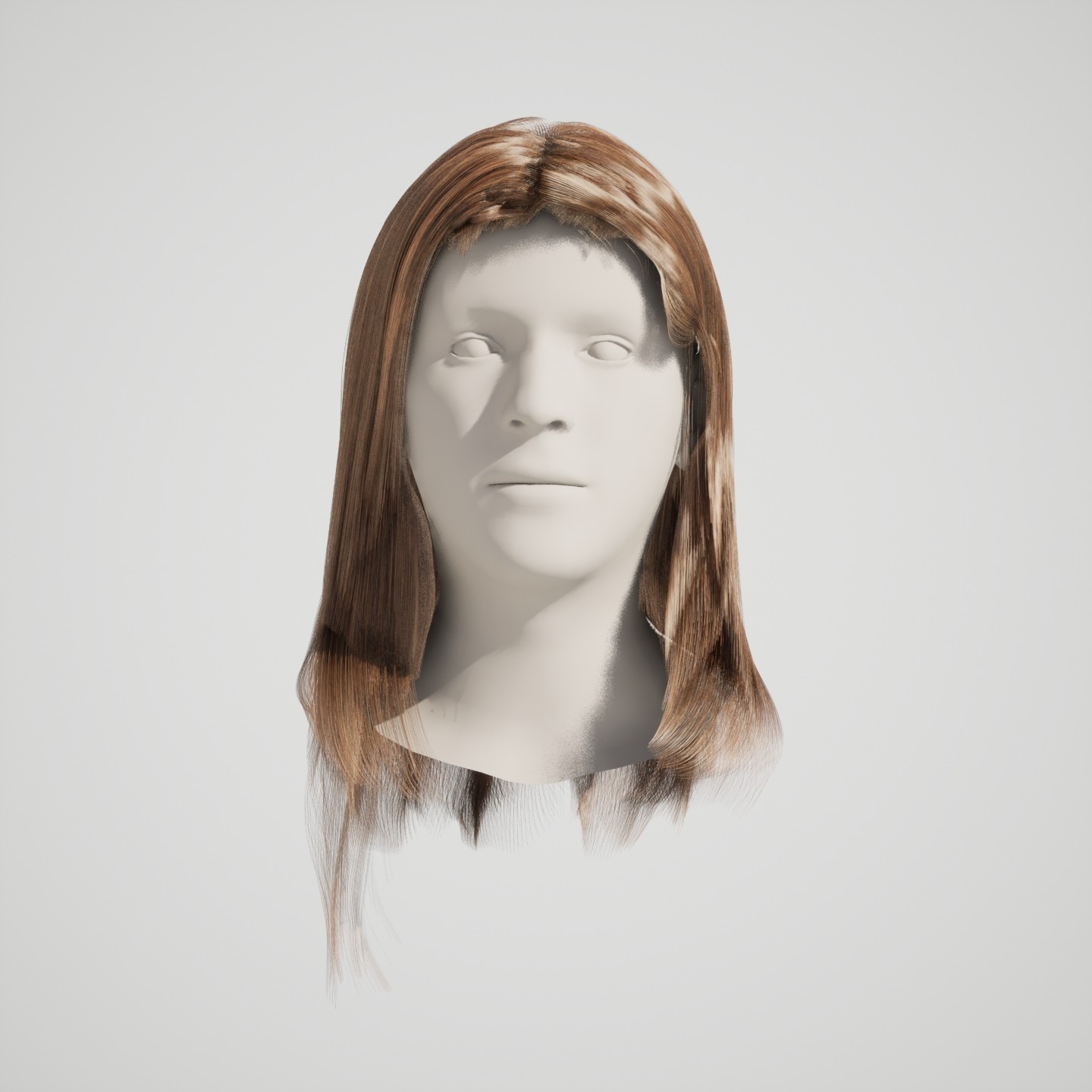}} 
        &
        \includegraphics[width=0.07\textwidth]{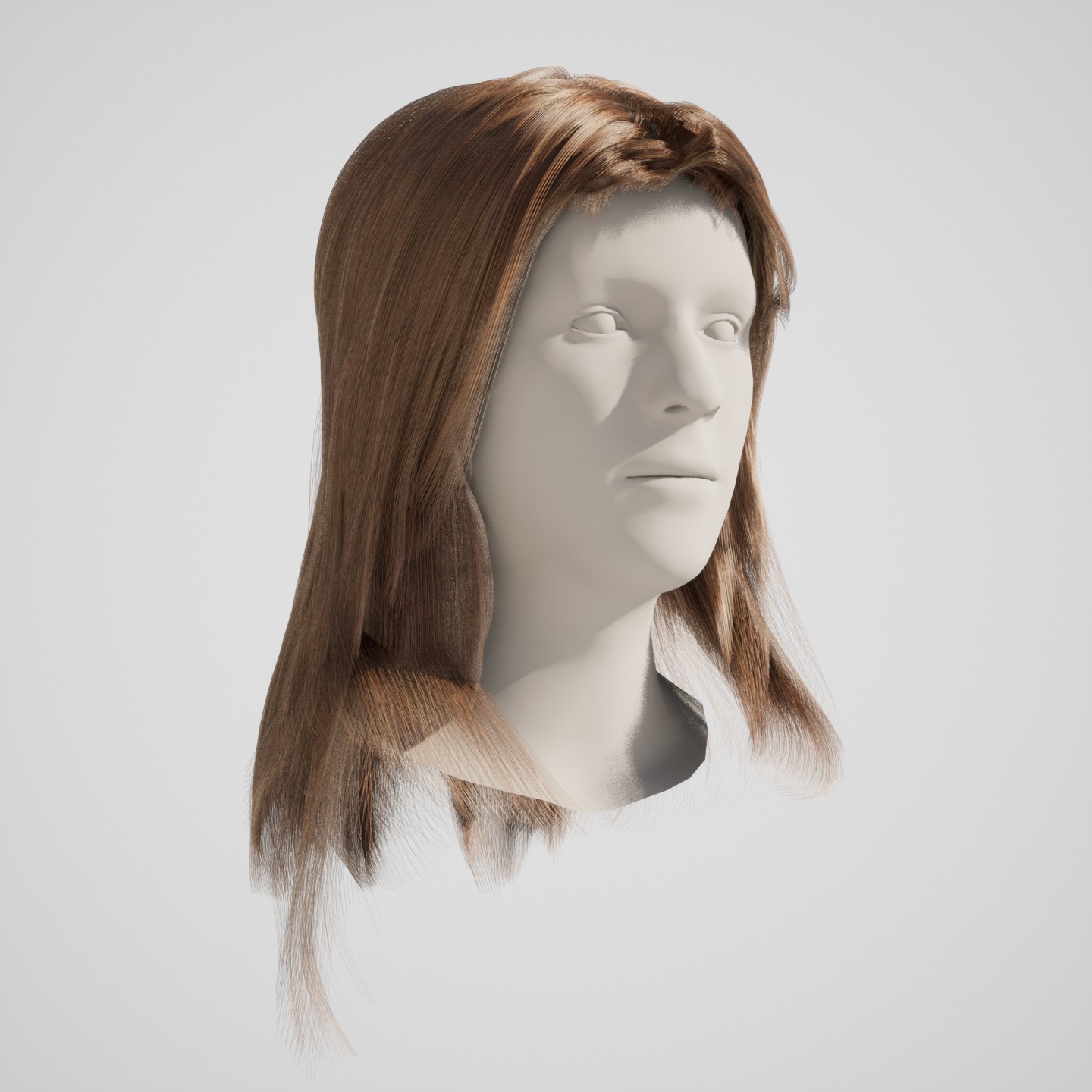}

        \\

        &
        &
        \includegraphics[width=0.07\textwidth]{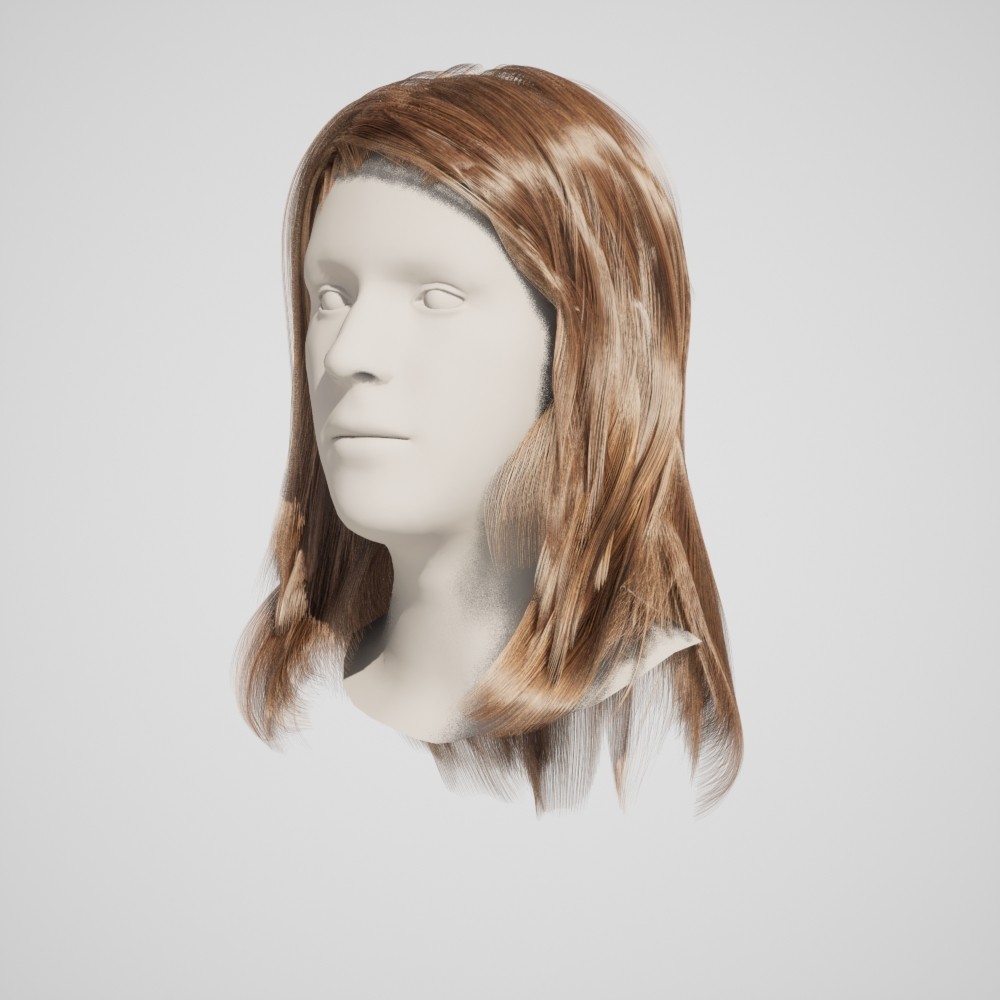} 
        &
        &
        \includegraphics[width=0.07\textwidth]{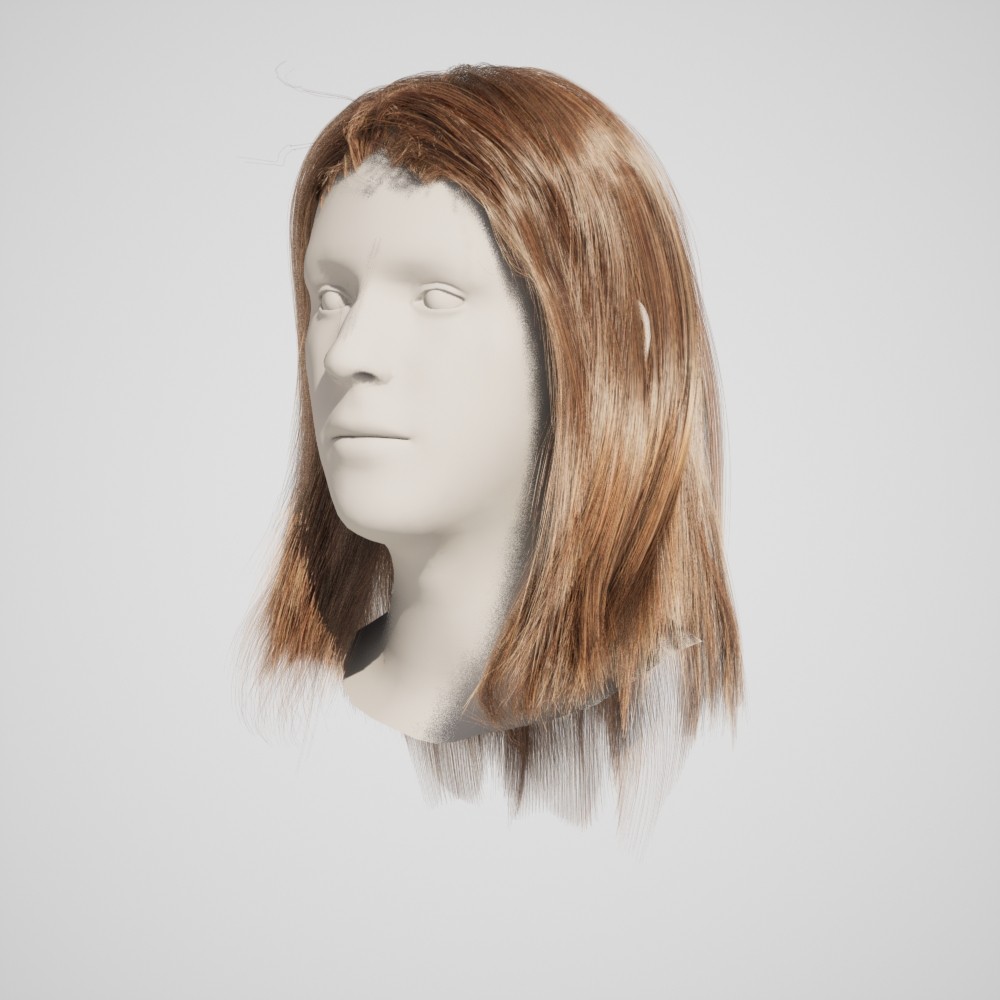} 
        & 
        &
       \includegraphics[width=0.07\textwidth]{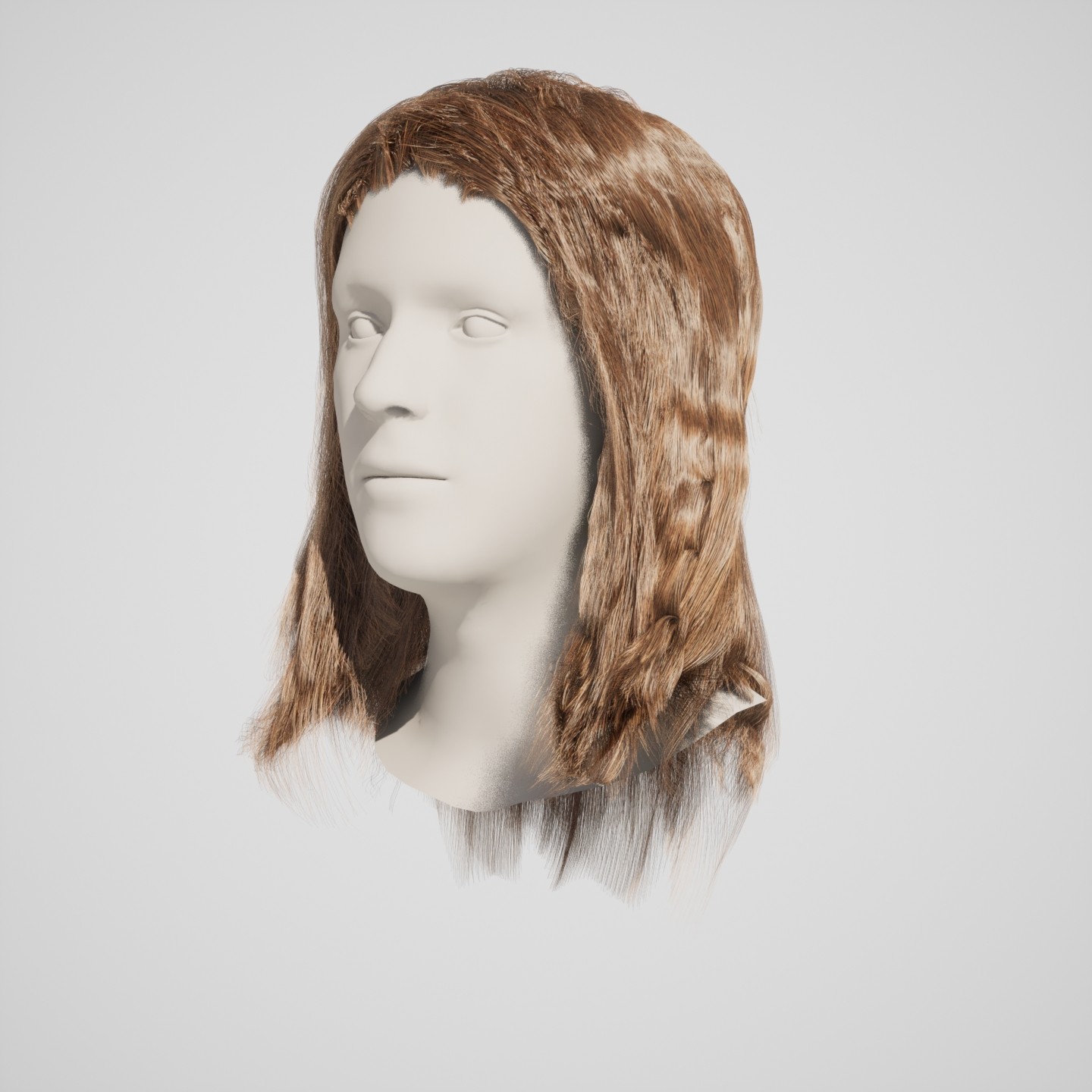}  

        & 
        &
        \includegraphics[width=0.07\textwidth]{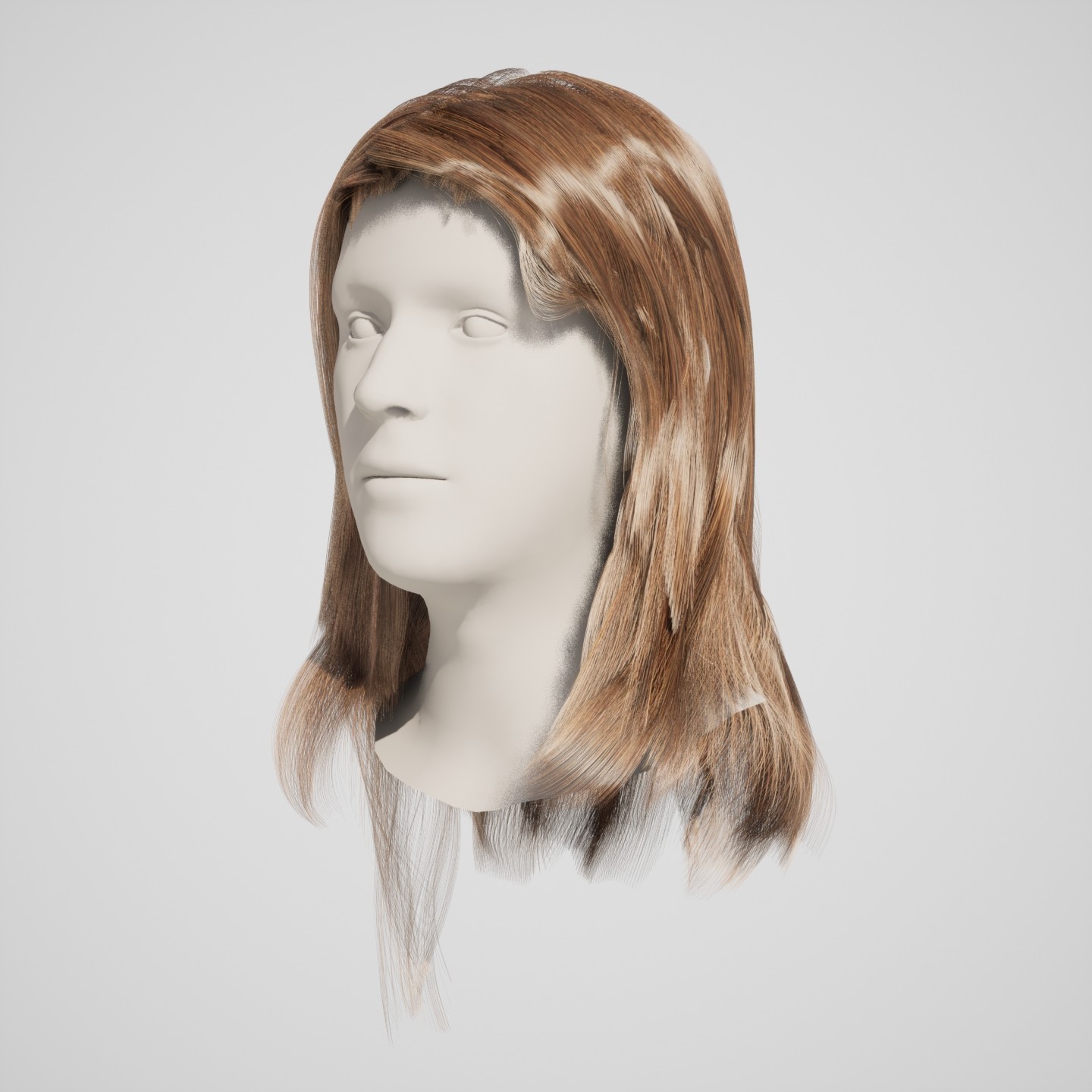}

        \\

        \addlinespace

\multicolumn{1}{c}{Image}  &
\multicolumn{2}{c}{Ours} & 
\multicolumn{2}{c}{w/o $prior_{3D}$} & 
\multicolumn{2}{c}{w/o $prior_{pca}$} &
\multicolumn{2}{c}{w/o Enc}
\end{tabular}
}
\caption{\textbf{Hair reconstruction stage.} We demonstrate the importance of optimizing in Hairstyle prior space of the model for single-view inversion (compare ``Ours'' to ``w/o $prior_{3D}$'' and ``w/o $prior_{pca}$''). Also, we show a scenario when only the decoder is optimized while the encoder is kept fixed (see ``w/o Enc'').}
\label{fig:suppmat_prior_ablation}

\end{figure*}

\section{Additional experiments} 

\subsection{Reconstruction results}

\begin{figure*}
\includegraphics[width=0.98\textwidth]{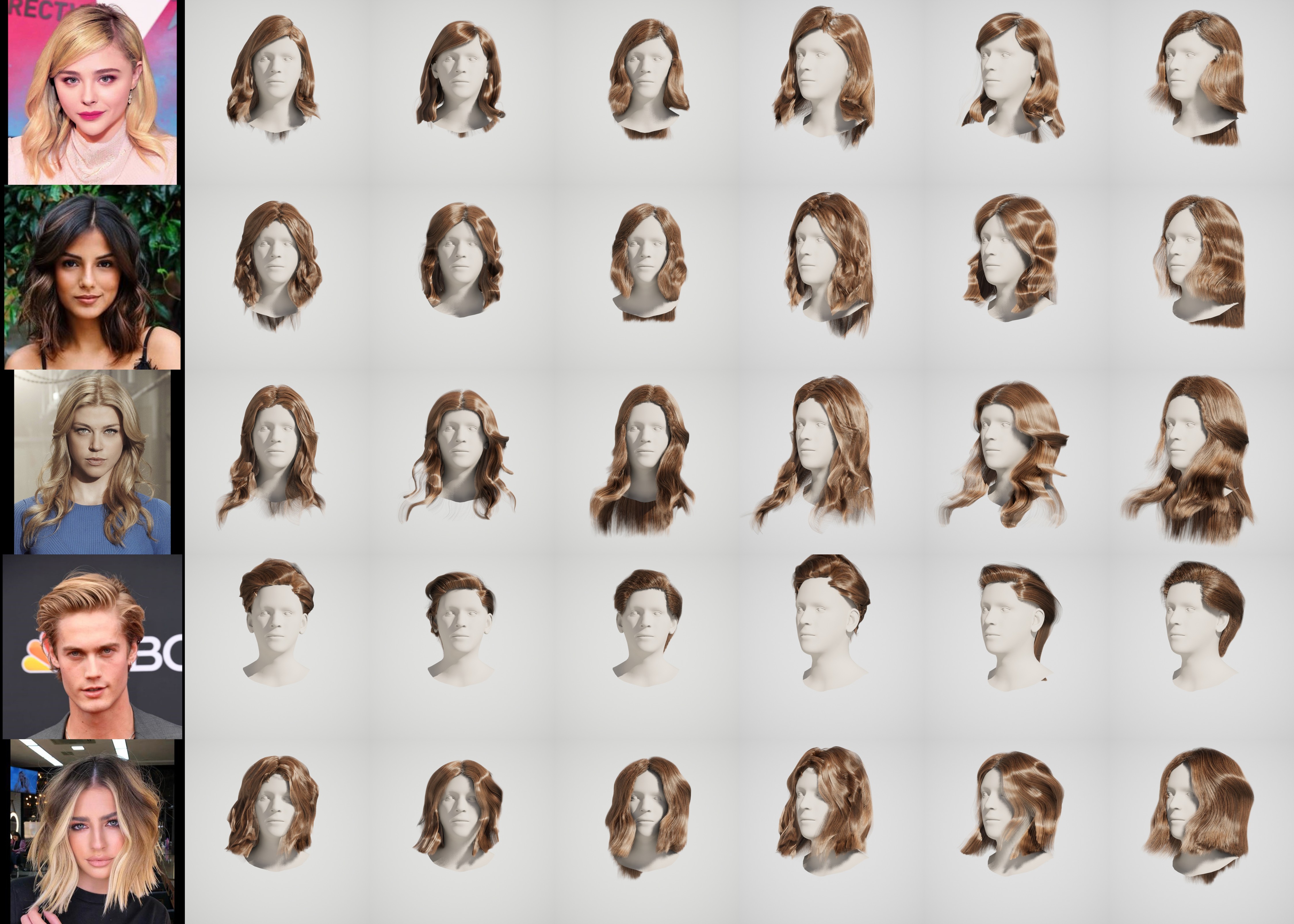} 
    \vspace{0.1cm}
\makebox[0.14\textwidth]{Image}%
\makebox[0.14\textwidth]{Ours}%
\makebox[0.14\textwidth]{Hairstep}%
\makebox[0.14\textwidth]{NeuralHDHair}%
\makebox[0.14\textwidth]{Ours}%
\makebox[0.14\textwidth]{Hairstep}%
\makebox[0.14\textwidth]{NeuralHDHair}%

    \caption{\textbf{Extended qualitative comparison with Hairstep~\cite{hairstep} and NeuralHDHair~\cite{neuralhd}.} Note that in NeuralHDHair the number of rendered strands is twice as compared to our method and Hairstep.}
    \label{fig:more_scenes_comp1}
\end{figure*}

\begin{figure*}
\includegraphics[width=0.98\textwidth]{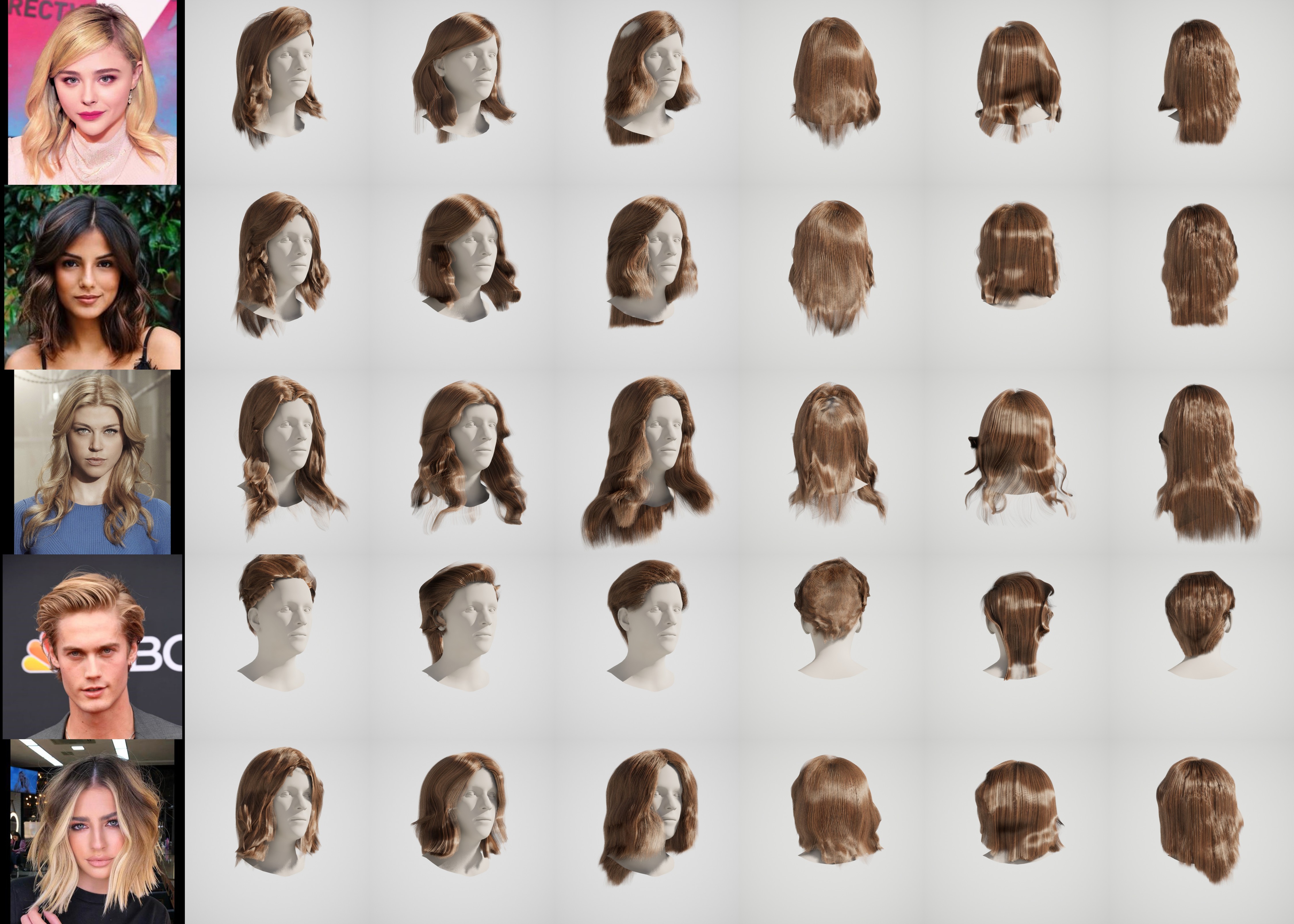} 
    \vspace{0.1cm}
\makebox[0.14\textwidth]{Image}%
\makebox[0.14\textwidth]{Ours}%
\makebox[0.14\textwidth]{Hairstep}%
\makebox[0.14\textwidth]{NeuralHDHair}%
\makebox[0.14\textwidth]{Ours}%
\makebox[0.14\textwidth]{Hairstep}%
\makebox[0.14\textwidth]{NeuralHDHair}%

    \caption{\textbf{Extended qualitative comparison with Hairstep~\cite{hairstep} and NeuralHDHair~\cite{neuralhd}.}}
    \label{fig:more_scenes_comp2}
\end{figure*}

\paragraph{More baselines.}
In Figure~\ref{fig:comaprison_hairmony}, we show a comparison with the retrieval-based method Hairmony~\cite{hairmony}.
While retrieval-based methods could predict general hair style, they do not contain any personalized details and are restricted by the diversity and quality of the dataset.
We show an extended comparison with Hairstep~\cite{hairstep} and NeuralHDHair~\cite{neuralhd} in Figures~\ref{fig:more_scenes_comp1}, ~\ref{fig:more_scenes_comp2}.

\begin{figure*}
\centering
  \includegraphics[width=0.65\textwidth]{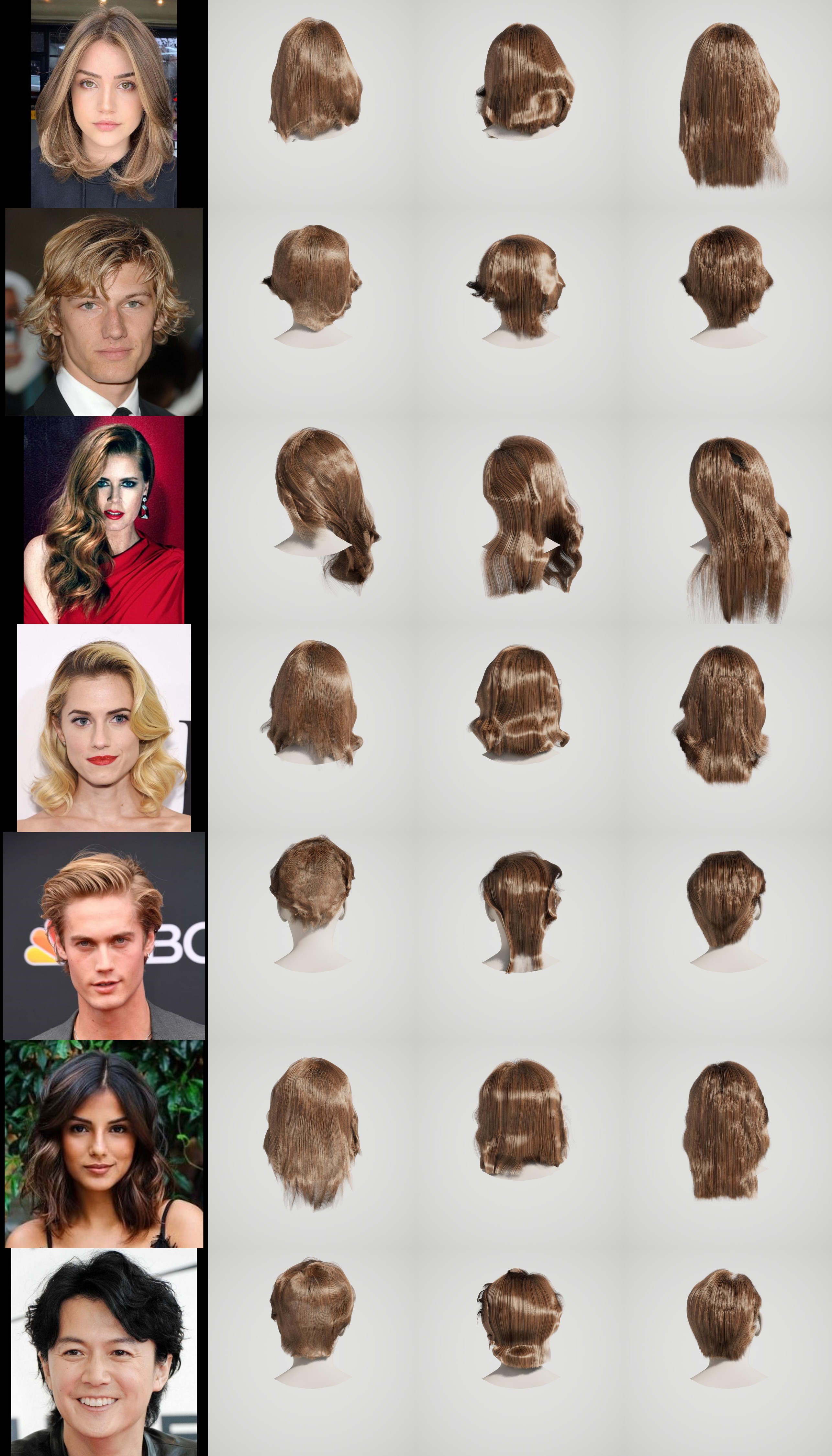}
    \makebox[0.15\textwidth]{Image}%
    \makebox[0.15\textwidth]{Ours}%
    \makebox[0.15\textwidth]{Hairstep}%
    \makebox[0.15\textwidth]{NeuralHDHair}

  \caption{\textbf{Back view comparison.} Comparison of back view geometry of reconstructions obtained by our method, Hairstep~\cite{hairstep} and NeuralHDHair~\cite{neuralhd}.
  }

    \label{fig:back_view_comparison_supplement}
\end{figure*}

\begin{figure*}
\centering
\includegraphics[width=\textwidth]{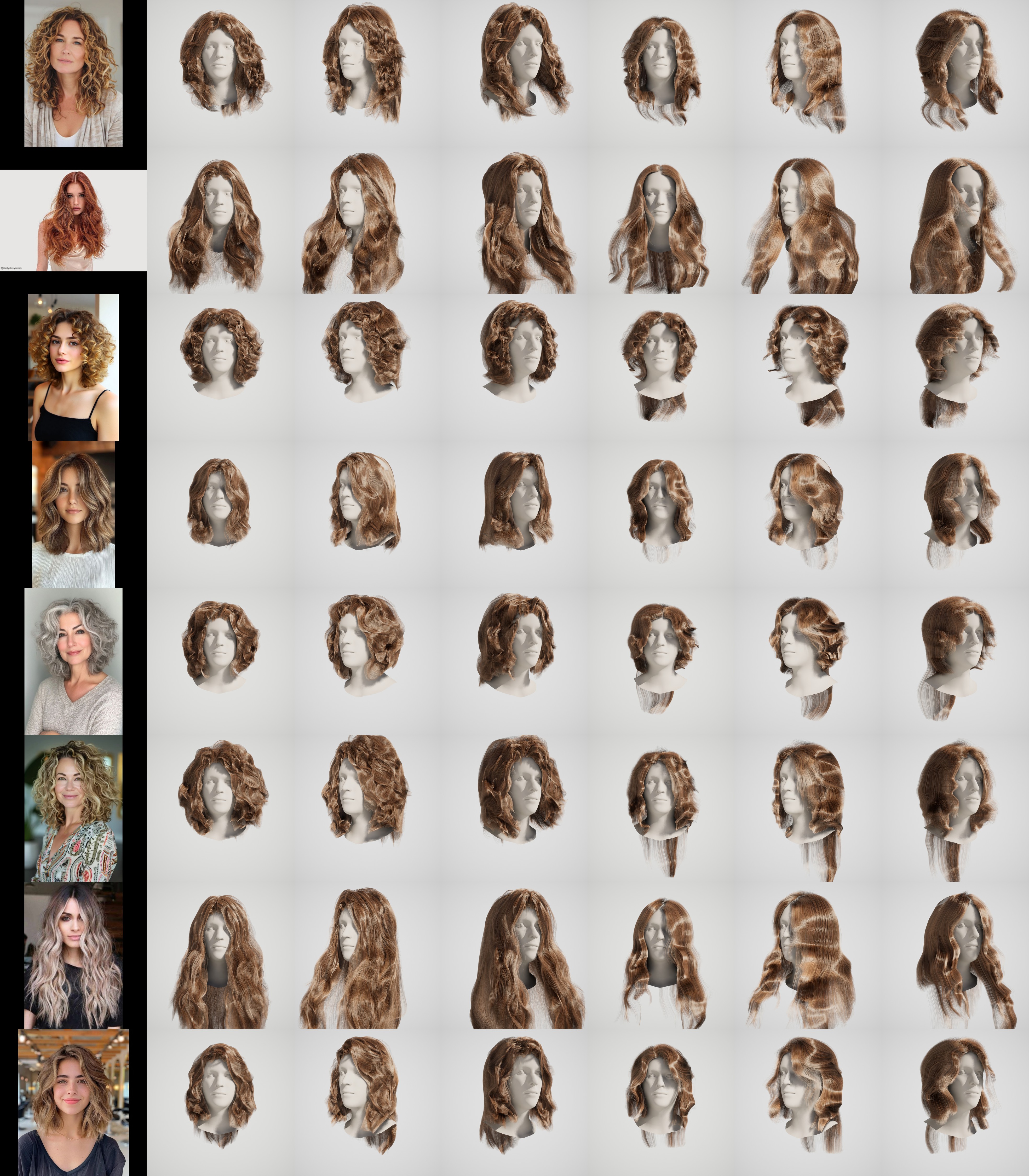} 
    \vspace{0.1cm}
     \makebox[0.14\textwidth]{Image}%
\makebox[0.428\textwidth]{Ours}%
\makebox[0.428\textwidth]{Hairstep}

    \caption{\textbf{Extended qualitative comparison of our method (columns 2–4) with Hairstep~\cite{hairstep} (last three columns) on wavy samples.} Our method can reconstruct curlier structures with more realistic back geometry.}
    \label{fig:wavy_suppmat_comparison}
\end{figure*}

\begin{figure*}
\centering
\includegraphics[width=0.89\textwidth]{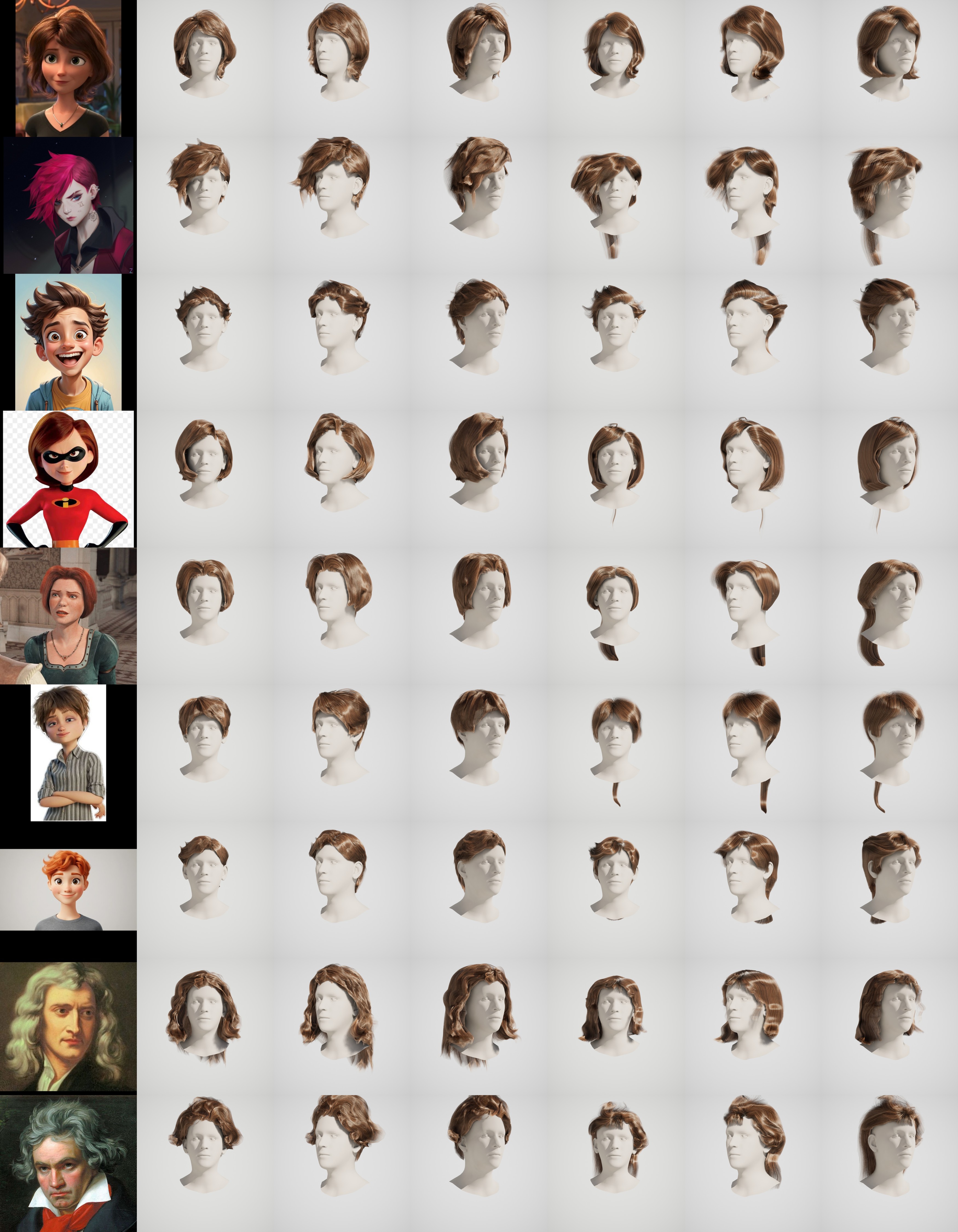} 
    \vspace{0.1cm}
     \makebox[0.24\textwidth]{Image}%
\makebox[0.38\textwidth]{Ours}%
\makebox[0.38\textwidth]{Hairstep}

    \caption{\textbf{Extended qualitative comparison of our method (columns 2–4) with Hairstep~\cite{hairstep} (last three columns)} on out-of-distribution samples.}
    \label{fig:cartoon_suppmat_comparison}
\end{figure*}

\paragraph{Back views.} 
We show additional visualization results for back views of our method, Hairstep~\cite{hairstep}, and NeuralHDHair~\cite{neuralhd} (see Figure~\ref{fig:back_view_comparison_supplement}).
While NeuralHDHair~\cite{neuralhd} produces accurate and smooth results for back views, the method lacks details from the frontal view.
Hairstep~\cite{hairstep} reconstructs unrealistic back views, especially for short hairstyles (see Figure \ref{fig:cartoon_suppmat_comparison}).
Our method generates more realistic results for all types of hair (see Figures~\ref{fig:wavy_suppmat_comparison}, \ref{fig:cartoon_suppmat_comparison}).

\paragraph{Wavy hairstyles.} 
In Figure~\ref{fig:wavy_suppmat_comparison}, we show a comparison with Hairstep~\cite{hairstep} on wavy hairstyles.
While our method has problems in the accurate estimation of curls from a single image, it outperforms Hairstep~\cite{hairstep} in terms of quality.

\begin{figure*}
\centering
\includegraphics[width=0.89\textwidth]{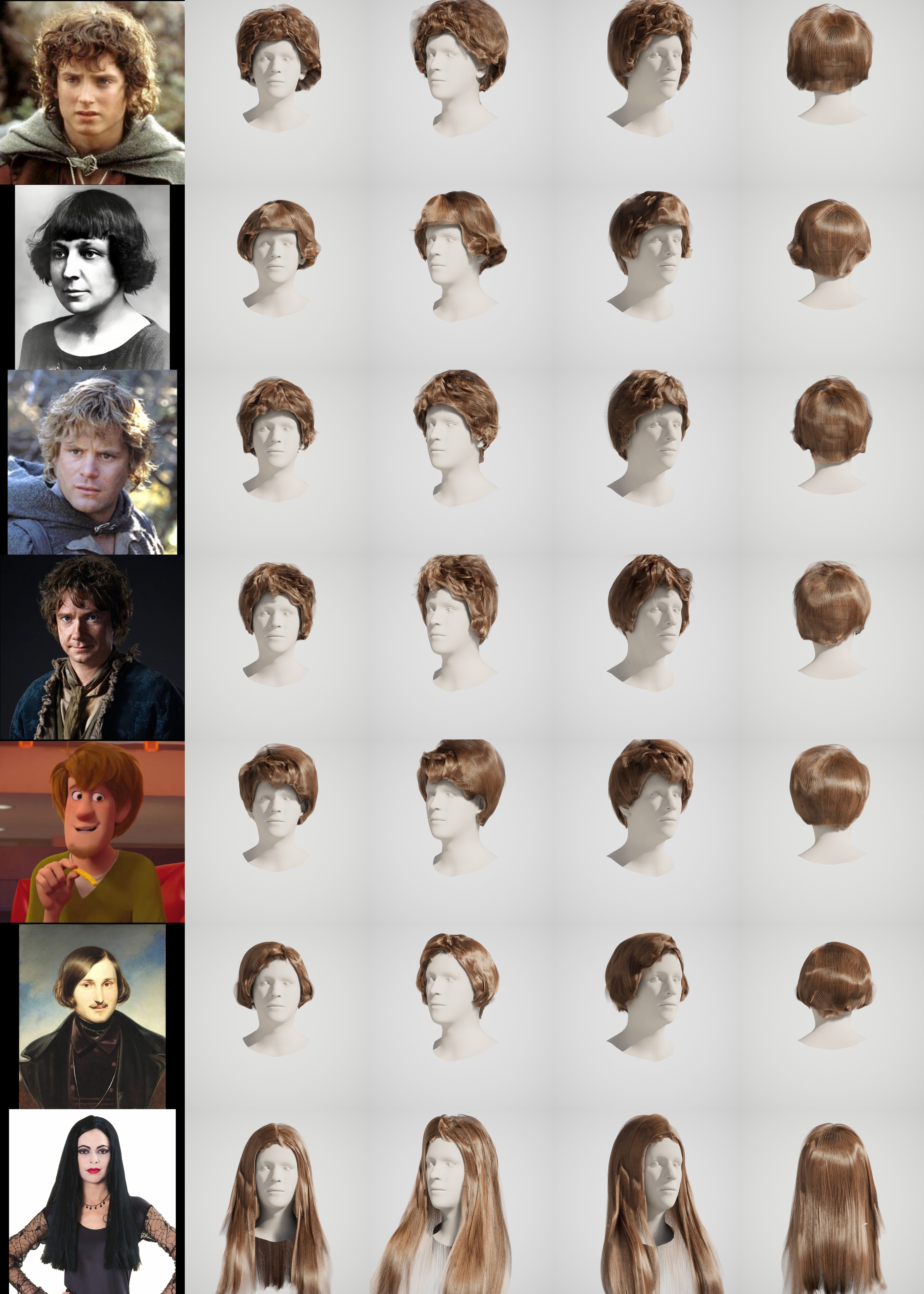} 
    \vspace{0.1cm}
     
    \caption{\textbf{More results} of our method on out-of-distribution data.}
    \label{fig:suppmat_movie1}
\end{figure*}

\begin{figure*}
\centering
\includegraphics[width=0.89\textwidth]{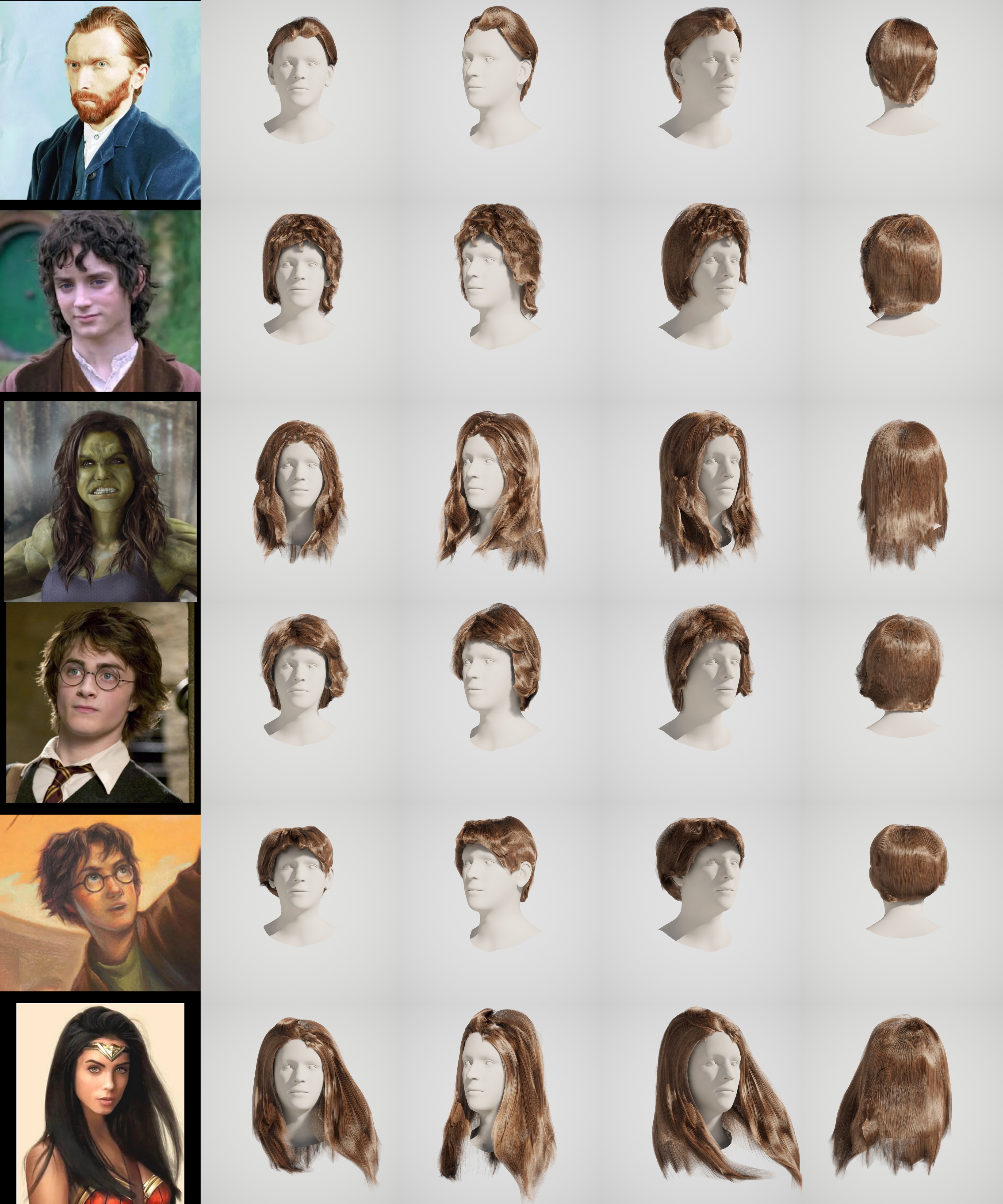} 
    \vspace{0.1cm}
     
    \caption{\textbf{More results} of our method on out-of-distribution data.}
    \label{fig:suppmat_movie2}
\end{figure*}

\paragraph{Out-of-distribution samples.} 
We present results of our method on out-of-distribution samples (Figures ~\ref{fig:cartoon_suppmat_comparison},~\ref{fig:suppmat_movie1},~\ref{fig:suppmat_movie2}), demonstrating its ability to reconstruct hairstyles for portraits, movie and cartoon characters, as well as drawings. 

\begin{figure*}
\centering
\includegraphics[width=0.75\textwidth]{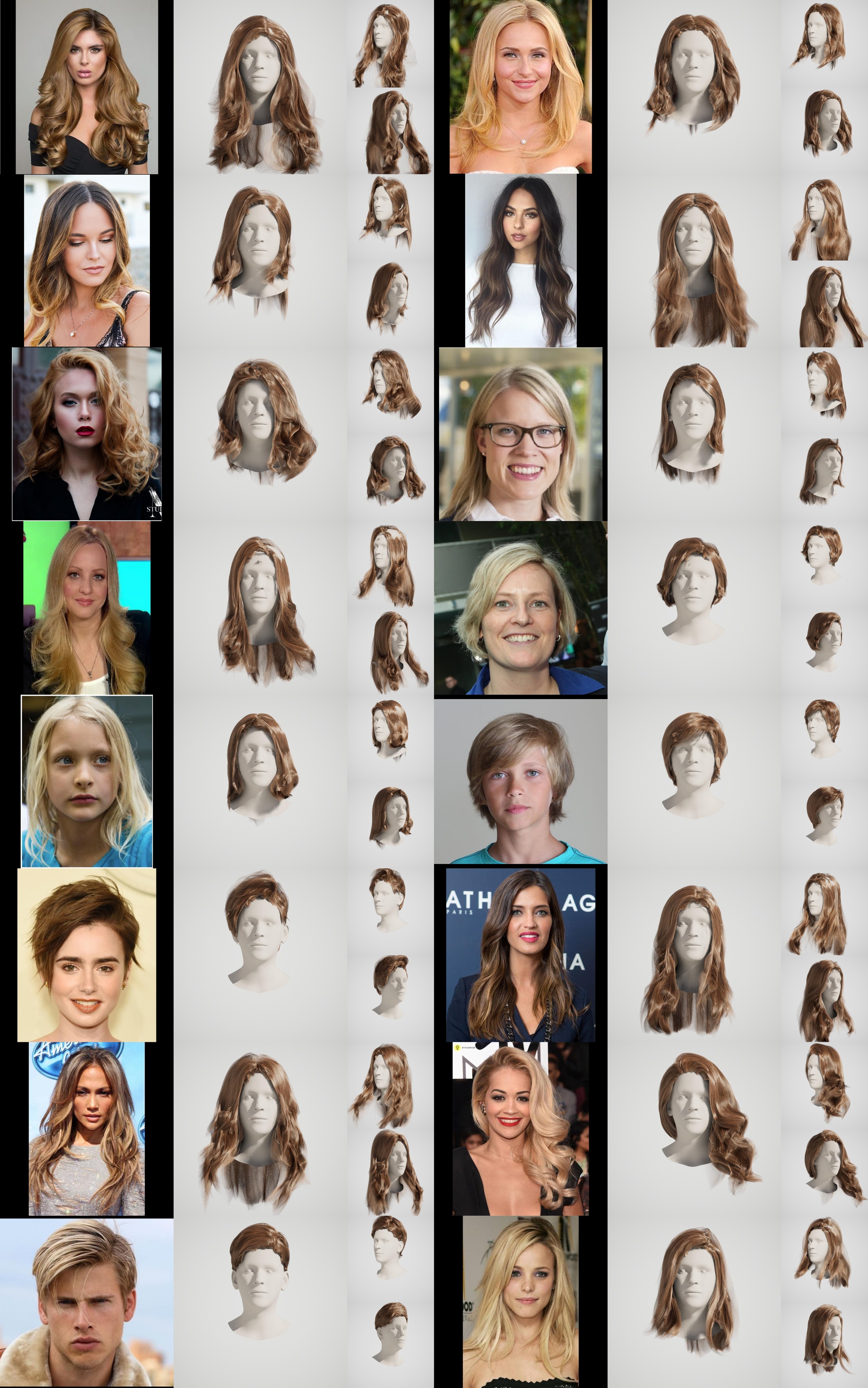} 
    \vspace{0.1cm}
    \caption{\textbf{Additional results} of our model.}
    \label{fig:new2}
\end{figure*}

\begin{figure*}
\centering
\includegraphics[width=0.9\textwidth]{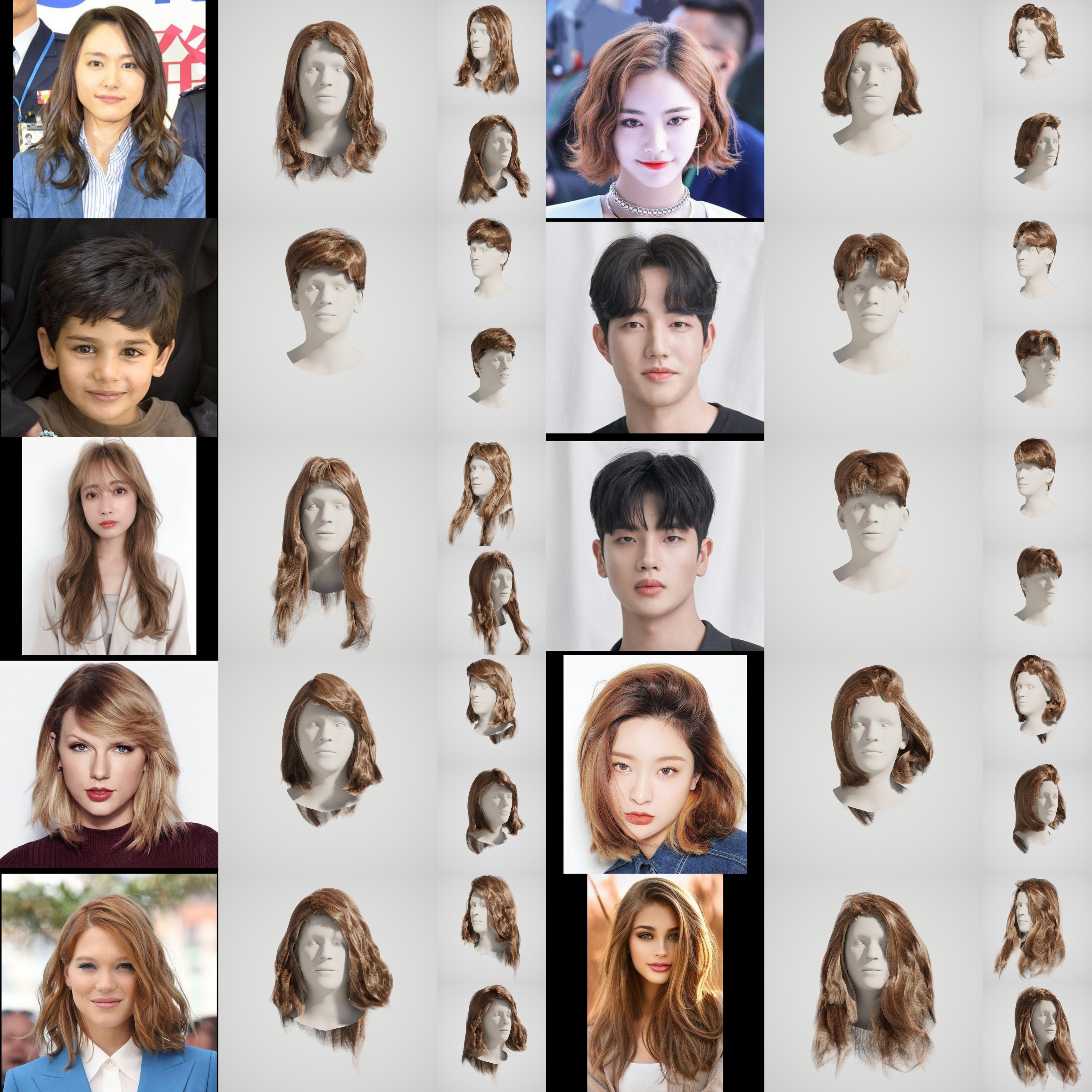} 
    \vspace{0.1cm}
    \caption{\textbf{Additional results} of our model.}
    \label{fig:new1}
\end{figure*}

\begin{figure*}
\centering
\includegraphics[width=0.9\textwidth]{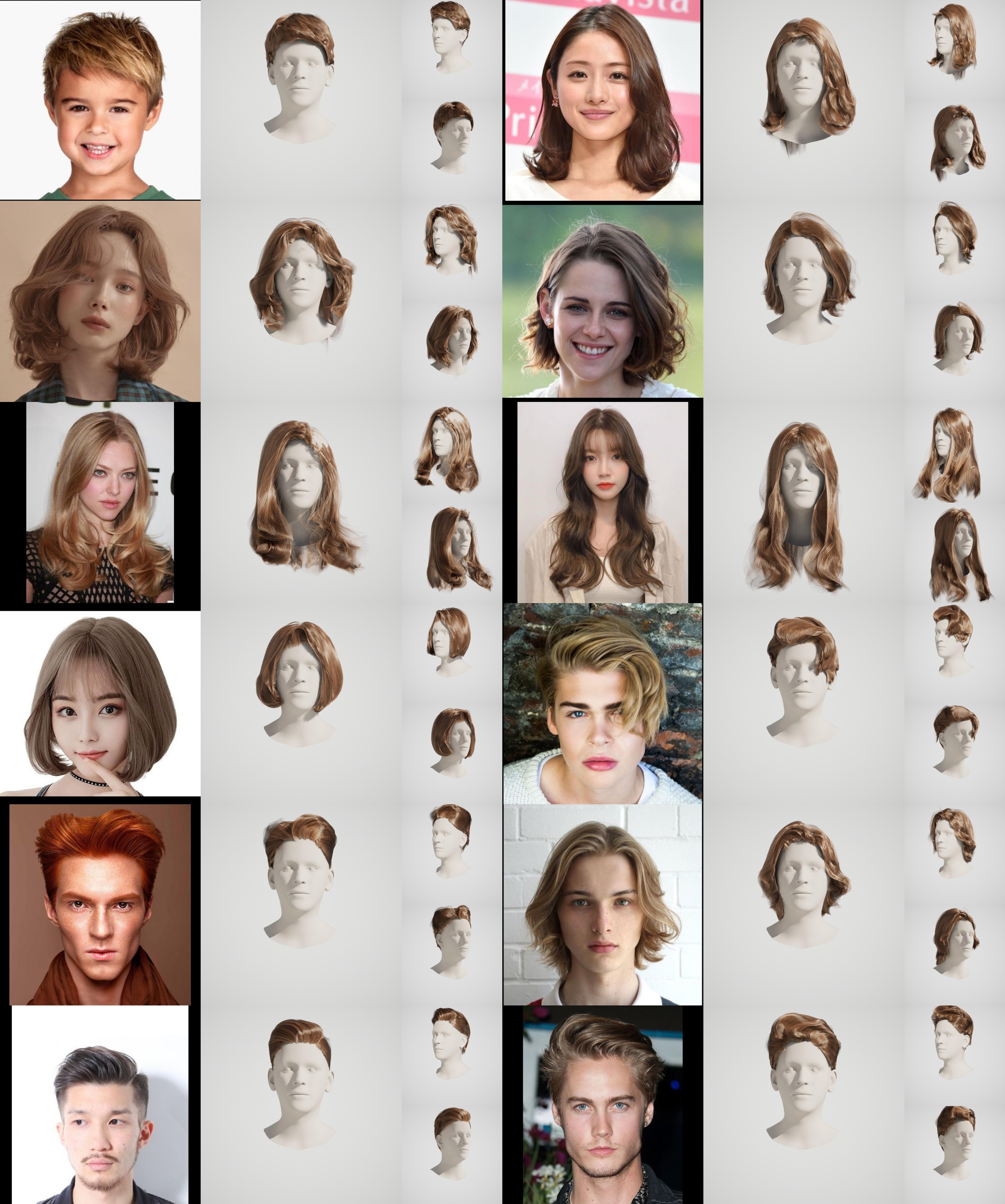} 
    \vspace{0.1cm}
    \caption{\textbf{Additional results} of our model.}
    \label{fig:new3}
\end{figure*}

\paragraph{More results.} 
Finally, we show our reconstruction results on more images (see Figures
~\ref{fig:new1}, \ref{fig:new2}, \ref{fig:new3}). 
Note that baldness artifacts or inaccuracies in hairstyles arise from failures in the camera or segmentation estimators. 
Additional camera fine-tuning and holistic reconstruction may resolve this problem.

\begin{figure*}
\centering
\begin{center}
    \centering
    \includegraphics[width=1\textwidth]{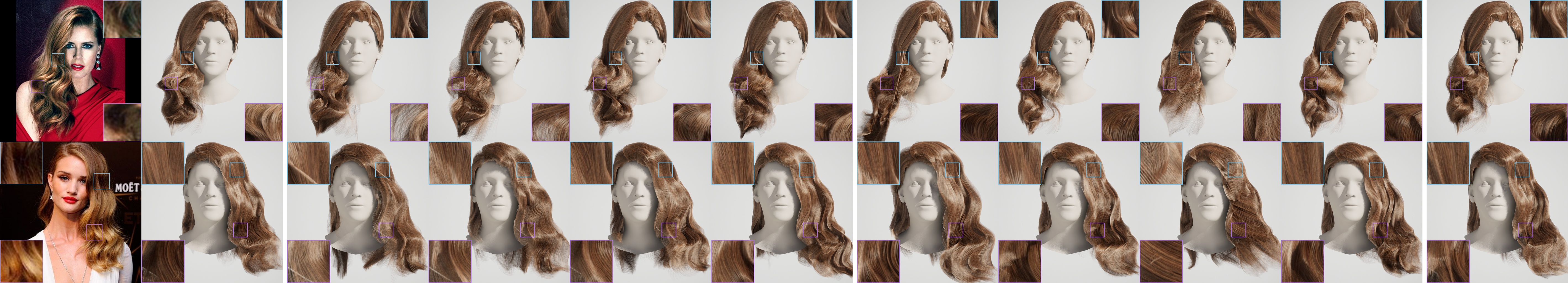}   

    \vspace{-0.1cm}
    \makebox[0.5\textwidth]{%
    \hspace{0.5cm}
    \scriptsize %
    \begin{tabular}{@{}*{11}{>{\centering\arraybackslash}p{0.07\textwidth}}@{}}
    Image &  $\text{Ours}$ & $\text{Prior}^{\text{syn only}}$ &
    $\text{Prior}^{\text{w/o silh}}$ & $\text{Prior}^{\text{w/o dir}}$ & $\text{Prior}^{\text{w/o penetr}}$   &  $\text{opt}^{\text{w/o silh}}$ & $\text{opt}^{\text{w/o depth}}$ & $\text{opt}^{\text{w/o dir}}$ & $\text{opt}^{\text{w/o penetr}}$
    & $\text{opt}^{\text{w/ blur depth}}$
    \end{tabular}
    }
    \vspace{-0.5cm}
    
\end{center}%
\caption{\textbf{Extended ablation on losses.} We show the importance of mixing strategy by training a prior model using rendering loss computed only on synthetic data (see $\text{Prior}^{\text{syn only}}$). Also, we provide an ablation on losses during hybrid model training with post optimization using \textit{all} losses, and their contribution during optimization with our final Hybrid model. Lastly, we analyze robustness results of our method with blurred depth as input.}
\label{fig:ablation_additional_losses}
\end{figure*}

\definecolor{tabfirst}{rgb}{1, 0.7, 0.7} %
\definecolor{tabsecond}{rgb}{1, 0.85, 0.7} %
\definecolor{tabthird}{rgb}{1, 1, 0.7} %

\begin{table}[tbh]
    \centering
    \resizebox{\linewidth}{!}{
    \begin{tabular}{l|ccccc}
         & chamfer\_pts $\downarrow$ & chamfer\_angle $\downarrow$ & angle error $\downarrow$ & mask $\downarrow$ & $\mathcal{L}_\text{undir}$ $\downarrow$ \\
     \hline    
coarse branch &0.00026 &  0.110 &                      15.81 &  0.517 &  0.735 \\
w/o depth input     &                      0.00031 &                      0.114 &                      16.33 &                      0.548 & 0.737 \\
w/o dir       &  0.00028 &                      0.112 &                      15.95 &                      0.553 & 0.737 \\
\hline
    \end{tabular}
    }
    \caption{\textbf{Extended ablation on losses and usage of depth representation} as input during the training of the coarse branch.}
    \label{tab:ablation_coarse_stage_suppmat}
\end{table}

\begin{table}[t]
    \centering
    \resizebox{0.95\linewidth}{!}{
    \begin{tabular}{l|ccccc}
         & chamfer\_pts $\downarrow$ & chamfer\_angle $\downarrow$ & angle error $\downarrow$ & mask $\downarrow$ & $\mathcal{L}_\text{undir}$ $\downarrow$ \\

\hline
hybrid training       & 0.00030 & 0.143 & 18.03 & 0.405 & 0.695 \\
$\text{Prior}^{\text{w/o dir}}$       & 0.00028 & 0.140 & 17.84 & 0.418 & 0.723 \\
$\text{Prior}^{\text{w/o penetr}}$        & 0.00030 & 0.139 & 17.60 & 0.412 & 0.689 \\
$\text{Prior}^{\text{w/o silh}}$         & 0.00033 & 0.141 & 17.96 & 0.571 & 0.710 \\
\hline
    \end{tabular}
    }
    \caption{\textbf{Contribution of losses} during training prior model.}
    \label{tab:ablation_additional}
\end{table}

\subsection{Extended ablation}
\begin{figure*}
\includegraphics[width=0.98\textwidth]{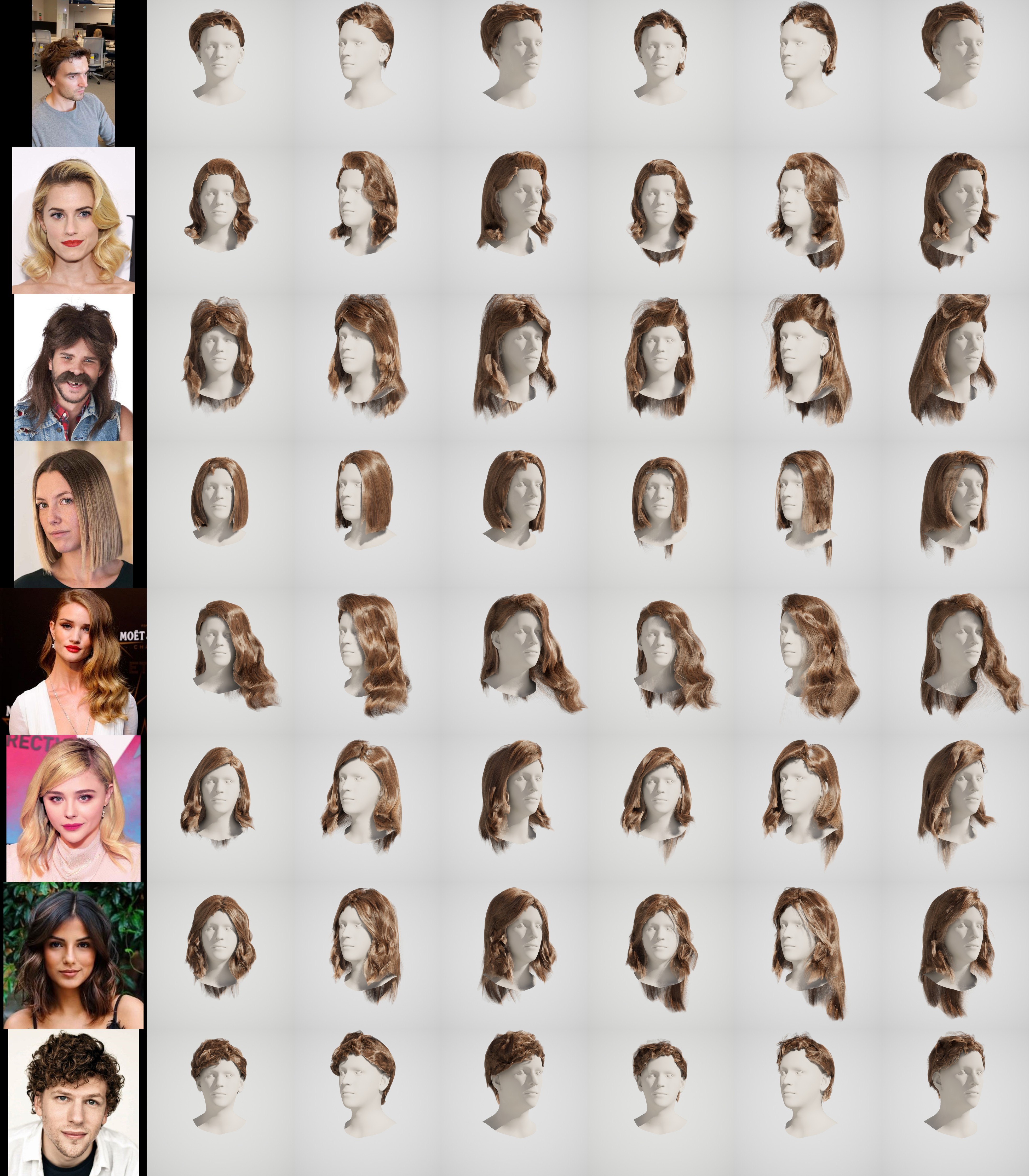} 
    \vspace{0.1cm}
\makebox[0.142\textwidth]{Image}%
\makebox[0.428\textwidth]{Ours}%
\makebox[0.428\textwidth]{w/o Mixing}%

    \caption{\textbf{Extended ablation on importance of training on synthetic and real data.} Results of ``Ours'' correspond to columns 2-4, while ``w/o Mixing'' to 5-7.}
    \label{fig:mixing_suppmat1}
\end{figure*}

\begin{figure*}
\includegraphics[width=0.98\textwidth]{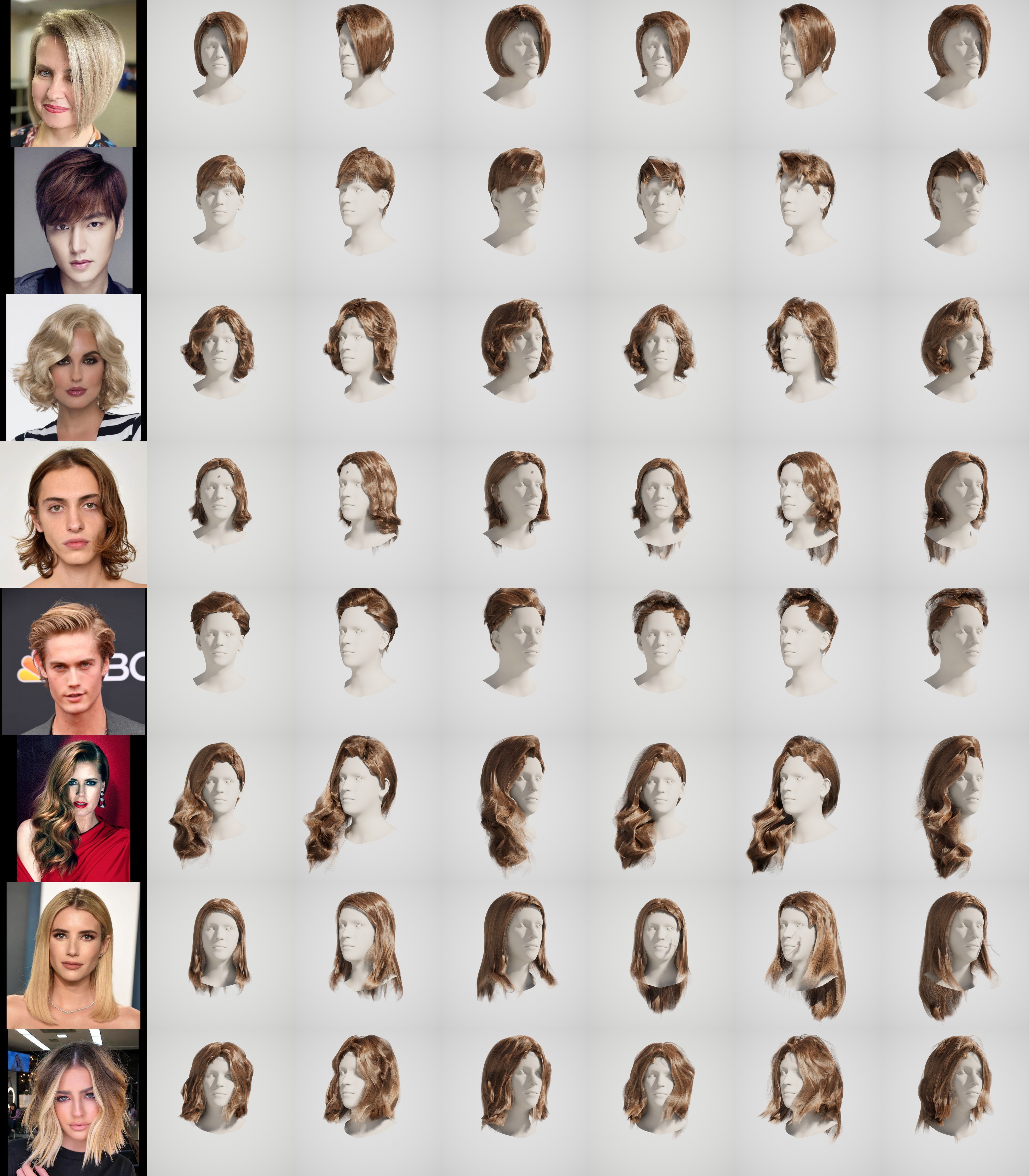} 
    \vspace{0.1cm}
\makebox[0.142\textwidth]{Image}%
\makebox[0.428\textwidth]{Ours}%
\makebox[0.428\textwidth]{w/o Mixing}%

       \caption{\textbf{Extended ablation on importance of training on synthetic and real data.} Results of ``Ours'' correspond to columns 2-4, while ``w/o Mixing'' to 5-7.}
    \label{fig:mixing_suppmat2}
\end{figure*}

We present an extended version of our ablation study in Figures~\ref{fig:suppmat_prior_ablation},
\ref{fig:ablation_additional_losses}, 
\ref{fig:mixing_suppmat1}, \ref{fig:mixing_suppmat2} and Tables~\ref{tab:ablation_coarse_stage_suppmat}, \ref{tab:ablation_additional}.
First, we show the ablation study on losses and input representation used during training the coarse branch, the Hybrid model, and during the inversion stage.
Then, we show more reconstruction results obtained using a Hairstyle prior that is trained on a mixture of synthetic and real images, and without it. To disentangle the contribution of rendering loss from the mixture strategy, we train our prior model with rendering loss computed only on synthetic data.
Also, we clarify the importance of optimization in the learned Hairstyle prior space compared to direct optimization of directions in 3D space or in PCA hair map.
Finally, we conduct an experiment with and without the optimization of the encoder during inversion.

\paragraph{Coarse stage.} 
In Table~\ref{tab:ablation_coarse_stage_suppmat}, we show an extended ablation of using depth as input and the usage of direction loss during training the coarse branch model. Without depth input or direction loss, we see the degradation of quality across all metrics.

\paragraph{Importance of losses.}
In Figure~\ref{fig:ablation_additional_losses}, we extend an ablation study on losses for (1) the optimization phase ($\text{opt}^{*}$ with our final Hybrid model), and (2) Hybrid model training ($\text{Prior}^{*}$; results are shown w/ opt which uses \emph{all} losses).
During both training and optimization, the orientation loss (dir) plays a critical role in improving fine-grained strand details (see close-ups).
The silhouette (silh) improves pixel alignment.
Removing depth maps worsens the results (see ``$\text{opt}^\text{w/o depth}$'').
Our method is also robust to depth corruption (see ``$\text{opt}^{\text{w/ blur depth}}$'', where we blurred the depth maps).
Lastly, penetration loss leads to better internal geometry.

In Table~\ref {tab:ablation_additional}, we calculate the metrics on synthetic and real data with omitted losses during training the Hybrid model. Surprisingly, omitting penetration loss improves metrics on synthetic data, but results in increased mask loss on real data. Excluding direction or silhouette loss produces bad results on real data.

\paragraph{Importance of Mixing strategy.} 
In Figure~\ref {fig:mixing_strategy_color_image_regressor}, we show results of our Hairstyle prior before (``w/o Mixing'') and after training (``w Mixing'') on a mix of synthetic and real data.
Model ``w Mixing'' can provide better hairstyle initialization in terms of hair silhouette and orientations for the inversion stage.
Note, here we rasterize orientations of obtained hairstyles using OpenGL~\cite{opengl} and color them for visualization purposes.

In Figures  ~\ref{fig:mixing_suppmat1}, ~\ref{fig:mixing_suppmat2}, we show more results of inversion in the hairstyle prior space trained with (see ``Ours'') and without mixing strategy (see ``w/o Mixing'').
The model that is not trained on real images results in more unrealistic structures and penetrations.  
To disentangle the contribution of mixing strategy from rendering loss, in Figure~\ref{fig:ablation_additional_losses} (see ``$\text{Prior}^{\text{syn only}}$''), we show inversion results in the space of a prior model trained with rendering loss computed only on synthetic data.

\paragraph{Importance of optimization in prior space.}
In the proposed ``Ours''  configuration, we jointly optimize both the encoder and decoder architectures.
The result of the optimization hairstyle in the 3D space after retrieving the coarse structure from the Hairstyle prior noted as ``w/o $prior_{3D}$''.
Compared to our optimization setup, we decrease the learning rate to $0.00001$ while doing $400$ steps for inversion.
Using more steps does not improve the quality of results.
While this approach fails for wavy hairstyles, it produces realistic results for simpler, straight hairstyles, see Figure~\ref{fig:suppmat_prior_ablation}.
Additionally, instead of optimizing the directions of the strands in the 3D space, as done in Gaussian Haircut~\cite{GaussianHaircut}, we optimize a PCA texture map, initialized from the hairstyle prior (see ``w/o $prior_{pca}$'').
Interestingly, this method introduces noisy artifacts in the generated strands. 

\paragraph{Optimization of Encoder.} 
We examine optimization within our prior space while keeping the encoders frozen (see Figure~\ref{fig:suppmat_prior_ablation}, ``w/o Enc'').
We find that this configuration underperforms compared to jointly optimizing both the encoder and decoder architectures.

\clearpage

\clearpage

\clearpage

\clearpage

\clearpage

\clearpage

\clearpage

\clearpage

\clearpage

\clearpage

\clearpage

\clearpage

\end{document}